%% file: main.tex
\documentclass[acmsmall]{acmart}

\AtBeginDocument{%
  \providecommand\BibTeX{{%
    \normalfont B\kern-0.5em{\scshape i\kern-0.25em b}\kern-0.8em\TeX}}}

\setcopyright{acmcopyright}
\copyrightyear{2021}
\acmJournal{TODAES}
\acmVolume{27}
\acmNumber{3}
\acmArticle{20}
\acmYear{2021}
\acmMonth{11}
\acmDOI{10.1145/3486618}

\input{packages}
\input{macros}

\begin{document}

\title{Enable Deep Learning on Mobile Devices: Methods, Systems, and Applications}

\author{Han Cai}
\authornote{All student authors have contributed equally to this work and are listed in the alphabetical order. Song Han is the corresponding author.}
\email{hancai@mit.edu}
\author{Ji Lin}
\authornotemark[1]
\email{jilin@mit.edu}
\author{Yujun Lin}
\authornotemark[1]
\email{yujunlin@mit.edu}
\author{Zhijian Liu}
\authornotemark[1]
\email{zhijian@mit.edu}
\author{Haotian Tang}
\authornotemark[1]
\email{kentang@mit.edu}
\author{Hanrui Wang}
\authornotemark[1]
\email{hanrui@mit.edu}
\author{Ligeng Zhu}
\authornotemark[1]
\email{ligeng@mit.edu}
\author{Song Han}
\email{songhan@mit.edu}
\affiliation{%
  \institution{Massachusetts Institute of Technology}
  \streetaddress{77 Massachusetts Avenue}
  \city{Cambridge}
  \state{MA}
  \country{USA}
  \postcode{02139}
}

\input{text/abstract}



\maketitle

\input{text/intro}
\input{text/compression}
\input{text/automl}
\input{text/training}

\input{text/domain/main}
\input{text/system}
\input{text/conclusion}

\bibliographystyle{acmj}
\bibliography{reference}

\end{document}

%% file: packages.tex
\usepackage{color,xcolor}
\usepackage{epsfig}
\usepackage{graphicx}

\usepackage{adjustbox}
\usepackage{array}
\usepackage{booktabs}
\usepackage{colortbl}
\usepackage{float,wrapfig}
\usepackage{hhline}
\usepackage{multirow}
\usepackage{subcaption} 

\usepackage{amsmath,amsthm}
\usepackage{bm}
\usepackage{nicefrac}
\usepackage{microtype}

\usepackage{changepage}
\usepackage{extramarks}
\usepackage{fancyhdr}
\usepackage{setspace}
\usepackage{soul}
\usepackage{xspace}

\usepackage{url}

\usepackage[ruled,vlined]{algorithm2e}
\usepackage{enumitem}

\usepackage{titlesec}
\usepackage{listings}
\usepackage{enumitem}
\usepackage{lipsum}

\usepackage{pifont}

\usepackage[perpage]{footmisc}
\usepackage{threeparttable}

%% file: macros.tex
\newcolumntype{L}[1]{>{\raggedright\let\newline\\\arraybackslash\hspace{0pt}}m{#1}}
\newcolumntype{C}[1]{>{\centering\let\newline\\\arraybackslash\hspace{0pt}}m{#1}}
\newcolumntype{R}[1]{>{\raggedleft\let\newline\\\arraybackslash\hspace{0pt}}m{#1}}

\newcommand{\sect}[1]{Section~\ref{#1}}

\newcommand{\fig}[1]{Figure~\ref{#1}}

\newcommand{\tab}[1]{Table~\ref{#1}}

\newcommand{\ignorethis}[1]{}

\DeclareMathOperator*{\argmin}{arg\,min}

\makeatletter
\DeclareRobustCommand\onedot{\futurelet\@let@token\@onedot}
\def\@onedot{\ifx\@let@token.\else.\null\fi\xspace}

\def\eg{\emph{e.g}\onedot} 
\def\ie{\emph{i.e}\onedot} 
 
\def\etc{\emph{etc}\onedot} \def\vs{\emph{vs}\onedot}
\def\wrt{w.r.t\onedot} 
\def\etal{\emph{et al}\onedot}
\makeatother

\definecolor{citecolor}{RGB}{34,139,34}
\definecolor{mydarkblue}{rgb}{0,0.08,1}
\definecolor{mydarkgreen}{rgb}{0.02,0.6,0.02}
\definecolor{mydarkred}{rgb}{0.8,0.02,0.02}
\definecolor{mydarkorange}{rgb}{0.40,0.2,0.02}
\definecolor{mypurple}{RGB}{111,0,255}
\definecolor{myred}{rgb}{1.0,0.0,0.0}
\definecolor{mygold}{rgb}{0.75,0.6,0.12}
\definecolor{mydarkgray}{rgb}{0.66, 0.66, 0.66}

\newcommand{\myparagraph}[1]{\vspace{5pt}\noindent\textbf{#1}}

\definecolor{qxkcolor}{RGB}{215,235,255}
\definecolor{softmaxcolor}{RGB}{230,235,255}
\definecolor{probxvcolor}{RGB}{255,255,235}

\usepackage{xcolor}

%% file: text/abstract.tex
\begin{abstract}

Deep neural networks (DNNs) have achieved unprecedented success in the field of artificial intelligence (AI), including computer vision, natural language processing and speech recognition. However, their superior performance comes at the considerable cost of computational complexity, which greatly hinders their applications in many resource-constrained devices, such as mobile phones and Internet of Things (IoT) devices. Therefore, methods and techniques that are able to lift the efficiency bottleneck while preserving the high accuracy of DNNs are in great demand in order to enable numerous edge AI applications.
This paper provides an overview of efficient deep learning methods, systems and applications. We start from introducing popular model compression methods, including pruning, factorization, quantization as well as compact model design. To reduce the large design cost of these manual solutions, we discuss the AutoML framework for each of them, such as neural architecture search (NAS) and automated pruning and quantization. We then cover efficient on-device training to enable user customization based on the local data on mobile devices. Apart from general acceleration techniques, we also showcase several task-specific accelerations for point cloud, video and natural language processing by exploiting their spatial sparsity and temporal/token redundancy. Finally, to support all these algorithmic advancements, we introduce the efficient deep learning system design from both software and hardware perspectives.

\end{abstract}

%% file: text/intro.tex
\section{Introduction}

Deep neural networks (DNNs) have revolutionized the field of artificial intelligence (AI) and have delivered impressive performance in computer vision \cite{krizhevsky2012imagenet,simonyan2015very,he2016deep}, natural language processing~\cite{sutskever2014sequence,vaswani2017attention,devlin2018bert,brown2020language} and speech recognition~\cite{hinton2012deep,deng2013recent,xiong2018microsoft}. They can be applied in various real-world scenarios, such as mobile phones~\cite{ota2017deep,ignatov2018ai,xu2019first}, self-driving cars~\cite{bojarski2016end,codevilla2018end,amini2019variational,liu2021efficient} and smart hospitals~\cite{haque2017towards,liu20183d,yeung2018bedside}. However, their superior performance comes at the cost of high computational complexity. For instance,
the state-of-the-art machine translation model~\cite{vaswani2017attention} requires more than 10G multiply-and-accumulates (MACs) to process a sentence of only 30 words;
the popular LiDAR perception model~\cite{choy20194d} needs more than 2000G MACs per second (\ie, 10 frames).

Such a high computational cost is far beyond the capabilities of most mobile devices, ranging from the vehicles to the mobile phones and the Internet of Things (IoT) devices, since their hardware resources are tightly constrained by the form factor, battery and heat dissipation. These computation workloads, however, cannot be delegated to the cloud server as they can be very sensitive to the latency (\eg, autonomous driving) and/or privacy (\eg, healthcare)~\cite{zhu2019deep,liu2020datamix}. Therefore, efficient deep learning is in great demand in order to lift the roadblock for mobile AI applications.

\input{figText/fig_teaser}

To accelerate the neural network inference, researchers have proposed a variety of model compression techniques, including pruning~\cite{han2015learning,liu2017learning,he2017channel}, low-rank factorization~\cite{xue2013restructuring,kim2015compression,zhang2016accelerating} and quantization~\cite{han2016deep,courbariaux2016binarized,jacob2018quantization}. Besides building upon existing large models, researchers have also explored designing efficient neural networks directly from scratch, including MobileNets~\cite{howard2017mobilenets,sandler2018mobilenetv2}, ShuffleNets~\cite{ma2018shufflenet,zhang2018shufflenet} and SqueezeNets~\cite{ma2018shufflenet,zhang2018shufflenet}. These solutions usually require considerable human efforts since there are a bunch of knobs that need to be tuned jointly to achieve the optimal performance: \eg, the pruning ratio and the quantization bitwidth of each layer. To this end, there have been many explorations to use automated machine learning (AutoML) to provide push-the-button solutions to free the human from the time-consuming design process, such as neural architecture search (NAS)~\cite{zoph2017neural,tan2019mnasnet,liu2019darts,cai2019proxylessnas,guo2020single}, automated pruning~\cite{he2018amc,yang2018netadapt,liu2019metapruning} and automated quantization~\cite{wang2019haq,wang2020apq,wang2020hardware}. However, the benefit of AutoML does not come for free as it will increase the carbon footprints significantly: \eg, Evolved Transformer~\cite{so2019evolved} produces the lifetime carbon dioxide emissions of five U.S. cars (in \fig{fig:carbon}). To achieve green and sustainable AI, researchers have proposed to efficiently search efficient neural architectures~\cite{cai2020once,wang2020hat}, which can achieve the same level of accuracy while reducing the carbon footprints by orders of magnitudes.

Besides the inference, neural network training can be very expensive as well, which hinders the on-device training and, consequently, the user customization on the mobile devices. To tackle this, researchers have proposed various memory-efficient training algorithms, such as gradient checkpointing~\cite{chen2016training}, activation pruning~\cite{dai2020sparsetrain} and low-bit quantization~\cite{zhou2018dorefa}. In most use cases, the mobile model only needs to be finetuned a little bit on the local user data to provide the specialization. Hence, another stream of research attempts to improve the efficiency of transfer learning~\cite{mudrakarta2019k,cai2020tinytl}. Lately, researchers have also introduced the federated learning~\cite{konevcny2016federated} to aggregate the users' trained models without compromising the privacy.

\input{figText/fig_carbon}

Apart from general accelerations that can be applied to any task in principle, there have been extensive investigations in domain-specific accelerations. In this paper, we will focus on point cloud processing, video understanding and natural language processing because they are widely used in mobile applications, such as autonomous driving and mobile vision/NLP. On the one hand, they are much more computationally expensive than conventional 2D vision due to their large memory footprint. On the other hand, they also provide unique opportunities for acceleration by exploiting and removing the spatial and temporal redundancies: spatial redundancy (point clouds), temporal redundancy (videos), and token-level redundancy (natural languages).

Not all algorithmic improvements, however, can be translated into the measured acceleration on hardware. For instance, sparse and low-bit computations (which are introduced by fine-grained pruning and quantization) are not natively supported by the general-purpose inference library (\eg, cuDNN~\cite{chetlur2014cudnn}) as well as hardware (\eg, CPUs, GPUs). This gap has been gradually bridged by recent efforts on designing specialized software systems~\cite{chen2018tvm,jia2018beyond,jia2019taso,ding2020ios}
and hardware systems~\cite{han2016eie,albericio2016cnvlutin,parashar2017scnn,chen2017eyeriss,sharma2018bit,wang2020spatten,sparch}.
The specialized software systems explore intra and inter operation parallelism inside the neural network, and optimize the computation graph and memory scheduling, via heuristic rules and even learning-based methods. The specialized hardware systems directly support the sparsity of the pruned networks and mixed-precision quantization from the hardware architecture level.
The specialization of software and hardware systems opens up a new design space orthogonal to the algorithm space, which can be further exploited to unlock the unfulfilled potentials of specialization. Researchers thus have explored diverse co-design solutions, such as automatically assigning the computation resource on various platforms for NN models~\cite{mirhoseini2017device, mirhoseini2018hierarchical, jiang2020hardware, kao2020confuciux},  automatically sizing the hardware architecture~\cite{yang2020co, zoph2017neural}, and even jointly searching the neural network and accelerator design including connectivity between processing elements and loop scheduling~\cite{lin2021enhcs}.

There are many existing surveys related to model compression~\cite{cheng2017survey,deng2020model,choudhary2020comprehensive}, automated machine learning~\cite{elsken2019neural,wistuba2019survey,he2021automl}, efficient hardware architecture design~\cite{sze2017efficient}, and optimizations for specific tasks~\cite{tay2020efficient}. This paper aims to cover a wider spectrum of methods and applications for efficient deep learning: from manual to automated, from new primitives/operations design to design space exploration, from training to inference, from algorithm to hardware, and from general-purpose to application-specific optimizations. We believe that this survey paper will provide a more holistic view of this field.

The remainder of the paper will be structured as follows (\fig{fig:teaser}):
\begin{itemize}
    \item \sect{sect:compression} discusses various model compression methods, including pruning, low-rank factorization, quantization, knowledge distillation and compact model design.
    \item \sect{sect:automl} studies AutoML frameworks for model compression and neural architecture search.
    \item \sect{sect:training} describes efficient on-device training (general/transfer learning techniques).
    \item \sect{sect:domain} studies application-specific acceleration for point clouds, videos and languages.
    \item \sect{sect:system} introduces the efficient software/hardware design for deep learning.
\end{itemize}

%% file: figText/fig_teaser.tex
\begin{figure}[t]
\centering
\includegraphics[width=\linewidth]{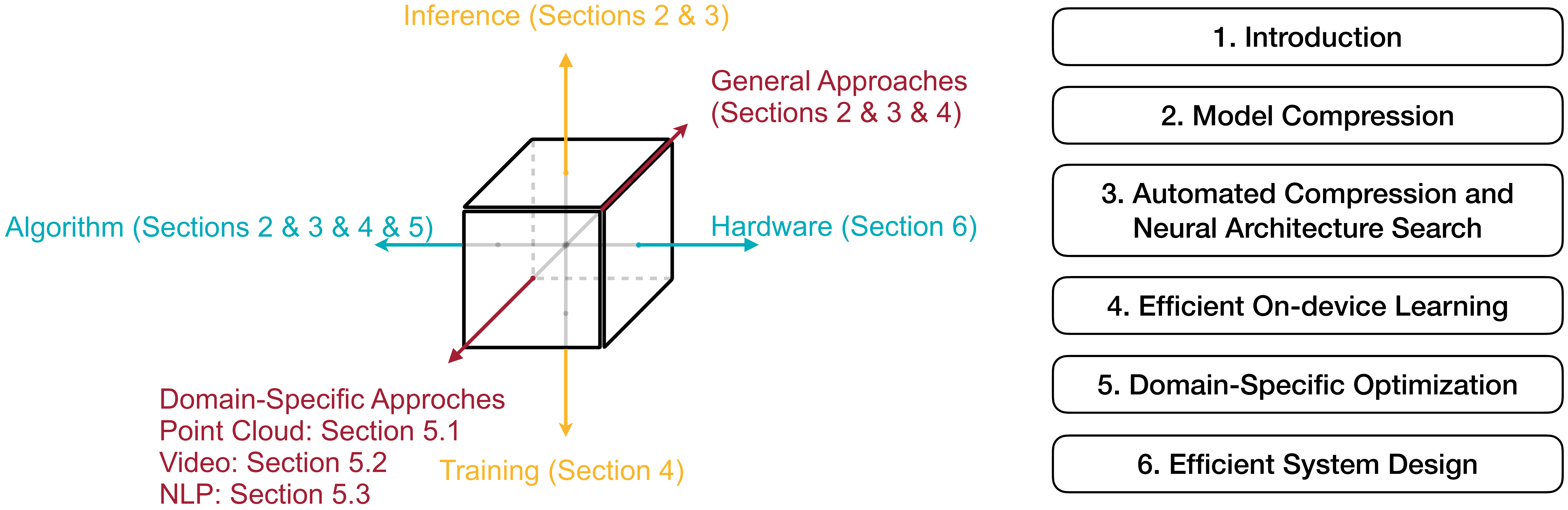}
\caption{Left: Spectrum of efficient deep learning solutions (from inference to training, from algorithm to software/hardware system, from general to domain-specific). Right: Overview of paper organization.}
\label{fig:teaser}
\end{figure}

%% file: figText/fig_carbon.tex
\begin{figure}[t]
\centering
\includegraphics[width=0.7\linewidth]{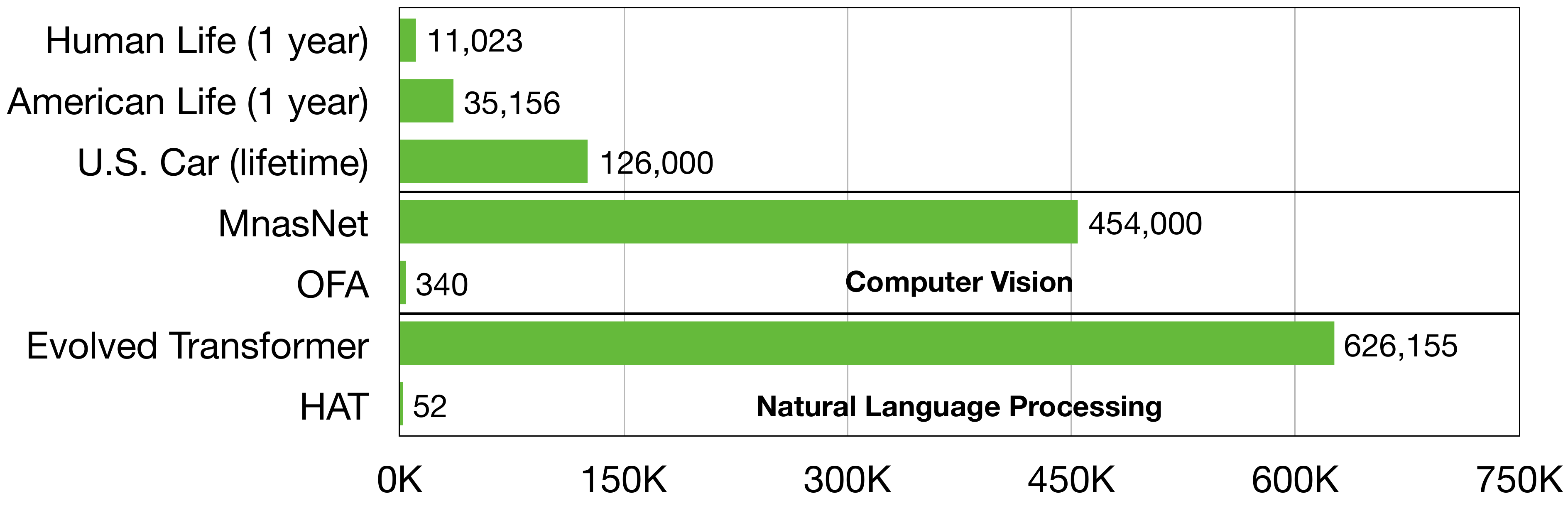}
\caption{Deep learning can introduce large carbon footprints: \eg, Evolved Transformer~\cite{so2019evolved} produces the lifetime carbon emissions of five cars. Thus, efficient deep learning is critical for green and sustainable AI.}
\label{fig:carbon}
\end{figure}

%% file: text/compression.tex
\section{Model Compression}
\label{sect:compression}

\subsection{Parameter Pruning}
\label{sec:pruning}

\begin{figure}
\centering
\begin{minipage}{.45\textwidth}
  \centering
  \includegraphics[width=\linewidth]{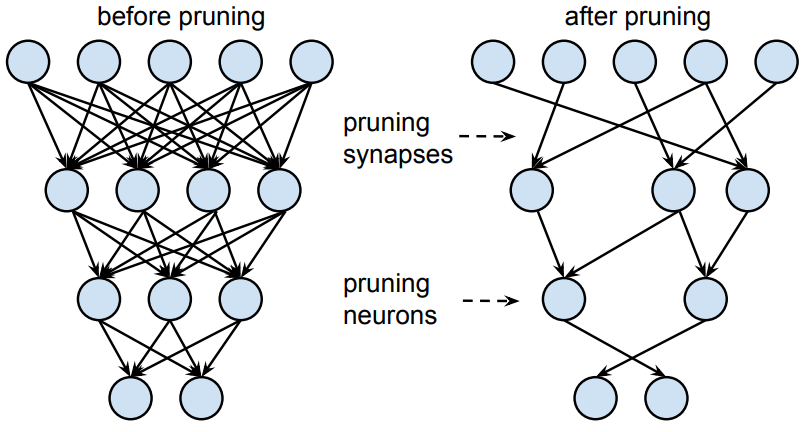}
  \captionof{figure}{Parameter pruning~\cite{han2015learning}.}
  \label{fig:pruning}
\end{minipage}%
\hfill
\begin{minipage}{.54\textwidth}
  \centering
  \includegraphics[width=\linewidth]{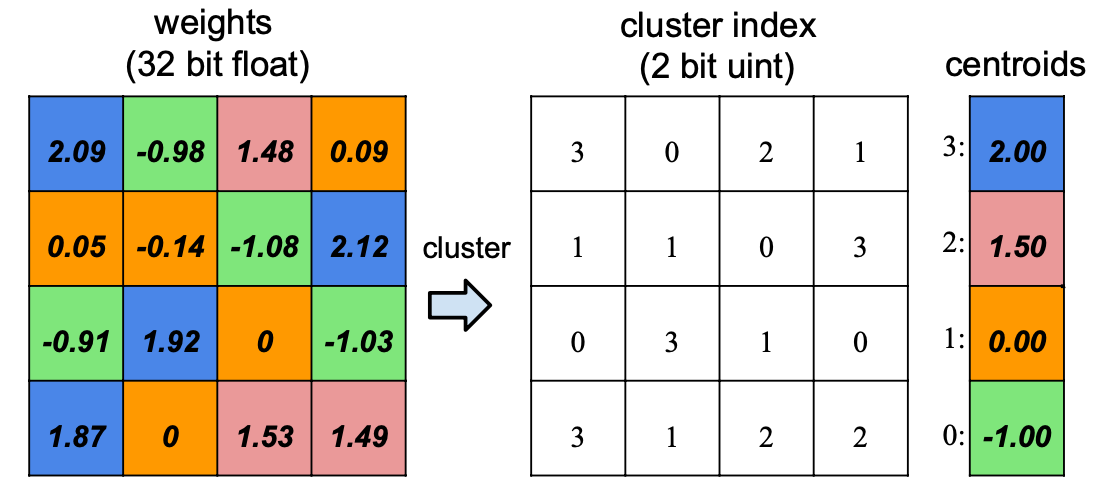}
  \captionof{figure}{$k$-means weight quantization~\cite{han2017deep}. }
  \label{fig:quantization}
\end{minipage}
\end{figure}

Deep neural networks are usually over-parameterized. 
Pruning removes the redundant elements in neural networks to reduce the model size and computation cost (\fig{fig:pruning}).

\myparagraph{Granularity.}
Pruning can be performed at different \emph{granularities}:

\begin{itemize}[leftmargin=*]
    \item \emph{Fine-grained pruning} removes individual elements from the weight tensor. Early approaches include Optimal Brain Damage~\cite{lecun1989optimal} and Optimal Brain Surgeon~\cite{hassibi1993second}, which reduce the number of connections based on the Hessian of the loss function. Han~\etal~\cite{han2015learning} propose a three-step method, train-prune-retrain, to prune the redundant connections in a deep neural network. It reduces the number of parameters of AlexNet by a factor of 9$\times$, and VGG-16 by 13$\times$, with no loss of accuracy. Srinivas~\etal~\cite{srinivas2015data} propose a data-free pruning method to remove the redundant neurons. In fine-grained pruning, the set of weights to be pruned can be chosen arbitrarily, it can achieve a very high compression ratio on CNN~\cite{han2015learning}, RNN~\cite{giles1994pruning}, LSTM~\cite{han2017ese} and Transformers~\cite{cheong2019transformers} without hurting accuracy.
    \item \emph{Pattern-based pruning} is a special kind of fine-grained pruning which has better hardware acceleration with compiler optimization~\cite{ma2020pconv, tan2020pcnn, niu2020patdnn}. It assigns a fixed set of masks to each 3$\times$3 kernel. The number of the masks is usually limited (4-6) to ensure hardware efficiency. Despite the intra-kernel fine-grained pruning pattern, pattern-based pruning can be accelerated with compiler optimization by reordering the computation loops, reducing the control-flow overhead.
    \item \emph{Coarse-grained pruning} or \emph{structured pruning} removes a regular tensor block for better hardware efficiency. Depending on the block size, entire vectors, kernels~\cite{niu2020patdnn}, or channels~\cite{he2017channel, wen2016learning, li2017pruning, molchanov2017pruning} are removed. Coarse-grained pruning like channel pruning can bring direct hardware acceleration on GPUs using standard deep learning libraries, but it usually comes at noticeable accuracy drop compared with fine-grained sparsity~\cite{li2017pruning}. Pruning using a smaller granularity usually brings smaller accuracy drop at the same compression rate.
\end{itemize}

\myparagraph{Hardware Acceleration.}
Regular pruning schemes are more hardware-friendly, making it easier for inference acceleration on existing hardware like GPUs, while more irregular pruning schemes better preserve the accuracy at the same compression rate. With specialized hardware accelerators~\cite{han2016eie, chen2017eyeriss, han2017ese, chen2019eyeriss,zhang2016cambricon,yu2017scalpel} and compiler-based optimization techniques~\cite{ma2020pconv, niu2020patdnn}, it is also possible to gain a considerable acceleration speed for more irregular pruning methods. 

\myparagraph{Importance Criteria.}
After choosing a pruning granularity, determining which weights to be pruned is also essential to the pruned models' performance. There have been several importance criteria heuristics to estimate the importance of each weight \emph{after} the model is trained; the less important weights are pruned according to the criteria. The most straight-forward heuristic is based on the magnitude: \ie, absolute weight values $|w|$~\cite{han2015learning, han2017deep}, where the weights of larger magnitude are considered as more important. It also extends to coarse-grained pruning like channel pruning, where the norm of tensor $\lVert\mathbf{W}\rVert_2$ is used as the criterion. Other criteria include second-order derivatives (the Hessian of the loss function)~\cite{lecun1989optimal, hassibi1993second}, loss-approximating Taylor expansion~\cite{molchanov2017pruning}, and output sensitivity~\cite{engelbrecht2001new}. Recently, Frankle and Carbin propose Lottery Ticket Hypothesis~\cite{frankle2018lottery} to find a sparse sub-network within the dense, randomly-initialized deep networks \emph{before} training, which can be trained to achieve the same accuracy. Experiments show that the method can find sparse sub-networks with less than 10-20\% of weights while reaching the same level of accuracy on MNIST~\cite{lecun2010mnist} and CIFAR~\cite{krizhevsky2009learning}. It is later scaled up to larger-scale setting (\eg, ResNet-50 and Inception-v3 on ImageNet), where the sparse sub-network can be found at the early phase of training~\cite{frankle2020linear} instead of initialization.

\myparagraph{Training Methods.}
Directly removing the weights in a deep neural network will significantly hurt the accuracy at a large compression ratio. Therefore, some training/fine-tuning is needed to recover the performance loss. Fine-tuning can be done after pruning to recover the performance drop~\cite{he2017channel}. It can be extended to iterative pruning~\cite{han2015learning, han2016deep}, where multiple iterations of pruning and fine-tuning are performed to further boost the accuracy. To avoid incorrect pruning of weights, dynamic pruning~\cite{guo2016dynamic} incorporates connection splicing into the whole process and make it as a continual network maintenance. Runtime pruning~\cite{lin2017runtime} chooses the pruning ratio according to each input sample, assigning a more aggressive pruning strategy for easier samples to achieve a better accuracy-computation trade-off. Another implementation trains compact DNNs using sparsity constraints. The sparsity constraints are usually implemented using $L_0$, $L_1$, or $L_2$-norm regularization applied to the weights, which are added to the training loss for joint optimization. Han~\etal~\cite{han2015learning} apply $L_1/L_2$ regularization to each individual weight during training.
Lebedev~\etal~\cite{lebedev2016fast} apply group sparsity constraints on convolutional filters to achieve structured sparsity.

\subsection{Low-Rank Factorization}

Low-rank factorization uses matrix/tensor decomposition to reduce the complexity of convolutional or fully-connected layers in deep neural networks. The idea of using low-rank filters to accelerate convolution has been long investigated in signal processing area.

The most widely used decomposition is Truncated Singular Value Decomposition (SVD)~\cite{golub1996matrix}, which is effective for accelerating fully-connected layers~\cite{xue2013restructuring, denton2014exploiting, girshick2015fast}. Given a fully-connected layer with weight $W\in \mathbb{R}^{m\times k}$, the SVD is defined as $W = USV^\text{T}$, where $U\in\mathbb{R}^{m\times m}, S\in\mathbb{R}^{m\times k}, V\in\mathbb{R}^{k\times k}$. $S$ is a diagonal matrix with the singular values on the diagonal. If the weight falls in a low-rank structure, it can be approximated by keeping only $t$ largest entries of $S$, where $t \ll \min(m, k)$. The computation $Wx$ can be reduced from $O(mk)$ to $O(mt + tk)$ for each sample. 

For 4D convolutional weights, Jaderberg \etal~\cite{jaderberg2014speeding} propose to factorize
$k\times k$ kernels into $1\times k$ and $k\times1$ kernels, which is also adopted in Inception-V3 design~\cite{szegedy2016rethinking}. Zhang \etal~\cite{zhang2016accelerating} propose to factorize a convolution weight of $n\times c\times k\times k$ into $n^\prime\times c\times k\times k$ and $n\times n^\prime\times1\times 1$, where $n^\prime \ll n$. 
Canonical Polyadic (CP) decomposition
can be used to decompose higher dimensional kernels like convolutional weights~\cite{lebedev2014speeding}. It computes a low-rank CP-decomposition of the 4D convolution kernel tensor into a sum of a small number of rank-one tensors. At inference time, the original convolution is replaced with a sequence of four convolutional layers with smaller kernels. Kim~\etal~\cite{kim2015compression} use Tucker Decomposition (the higher order extension of SVD) to factorize the convolutional kernels, getting higher compression ratio compared to using SVD. 

\subsection{Quantization}
\label{sec:quantization}

Network quantization compresses the network by reducing the bits per weight required to represent the deep network (\fig{fig:quantization}). The quantized network can have a faster inference speed with hardware support.

\myparagraph{Rounding Schemes.}
To quantize a full-precision weight (32-bit floating-point value) to lower precision, rounding is used to map the floating-point value into one of the quantization buckets. 

\begin{itemize}[leftmargin=*]

\item Early explorations~\cite{han2017deep, gong2014compressing, wu2016quantized} apply \emph{$k$-means clustering} to find the shared weights for each layer of a trained network:  \ie, all the weights that fall into the same cluster will share the same weight. Specifically, when partitioning $n$ original weights $W=\{w_1, w_2, ..., w_n\}$ into $k$ clusters $C=\{c_1, c_2, ..., c_k\}, n \gg k$, we minimize the within-cluster sum of squares (WCSS): 
\begin{equation}
    \argmin_C \sum_{i=1}^k \sum_{w\in c_i} |w-c_i|^2.
\end{equation}
This can be combined with pruning and Huffman coding to perform model compression~\cite{han2017deep}, which can compress the model size of VGG-16 by 49$\times$ with no loss of accuracy.

\item \emph{Linear/uniform} quantization~\cite{jacob2018quantization} directly rounds the floating-point value into the nearest quantized values after range truncation, and the gradient is propagated using STE approximation~\cite{bengio2013estimating}. Suppose the clipping range is $[a, b]$, and the number of quantization levels is $n$, the forward of quantizing floating-point value $x$ into quantized value $q$ is:
\begin{equation}
    q = \text{round}\left(\frac{\text{clamp}(x, a, b) - a}{\text{s}(a, b, n)}\right) \text{s}(a, b, n) + a,
\end{equation}
where $\text{clamp}(x, a, b) = \min(\max(x, a), b)$ and $\text{s}(a, b, n) = (b-a) / (n-1)$.
The back-propagation gradient is approximated by
$\partial \mathcal{L} / \partial q = \partial \mathcal{L} / \partial x$.
For relatively high-bit quantization (\eg, 8), $a$ and $b$ can be set as the minimum and maximum value of the weight tensor. It is also beneficial to choose the optimal $a, b$ values that reach the minimum Kullback-Leibler divergence between the floating-point weights and quantization weights. Apart from using the truncation values, some work~\cite{zhou2018dorefa} uses activation function such as tanh to map the range of the weights into $[-1, 1]$, making it easier for quantization.
\end{itemize}

\myparagraph{Bit-Precision.}
We can trade-off the model size and accuracy by using different bit-precisions. A lower bit-precision can lead to a smaller model size, but it may come at the cost of accuracy drop. Full-precision networks use {FP32} for both weights and activations. Half-precision networks use {FP16} to reduce the model size by half. {INT8} quantization for both weights and activations~\cite{jacob2018quantization} is widely used for integer-arithmetic-only inference, which can be accelerated on CPUs and GPUs.

Lower precision models include Ternary Weight Networks~\cite{li2016ternary}, where the weights are quantized to $\{-1, 0, +1\}$ or $\{-E, 0, +E\}$ (where $E$ is the mean absolute weight value). Trained Ternary Quantization~\cite{zhu2017trained} uses two learnable full-precision scaling coefficients $W_l^p$ and $W_l^n$ for each layer $l$ and quantizes the weights to $\{-W_l^n, 0, +W_l^P\}$. The extreme case for low-bit quantization is binary weight neural networks (\eg, BinaryConnect~\cite{courbariaux2015binaryconnect}, BinaryNet~\cite{courbariaux2016binarized}, XNOR~\cite{rastegari2016xnor}), where weights are represented with only 1 bit. The binary weights/activations are usually learned directly during network training. BinaryConnect~\cite{courbariaux2015binaryconnect} discusses both deterministic binarization:
\begin{equation}
    w_b = \text{sign}(w) =
    \begin{cases}
      +1 & \text{if } x\geq 0, \\
      -1 & \text{otherwise, }\\
    \end{cases}       
\end{equation}
and stochastic binarization:
\begin{equation}
    w_b = 
    \begin{cases}
      +1 & \text{with probability } p = \sigma(w), \\
      -1 & \text{with probability } 1-p.\\
    \end{cases}       
\end{equation}
where $\sigma$ is the ``hard sigmoid'' function:
\begin{equation}
    \sigma(x) = \max\left(0, \min\left(1, \frac{x+1}{2}\right)\right).
\end{equation}

\myparagraph{Quantization Schemes.}
For quantization of higher precisions (\eg, INT8), it is possible to perform \emph{post-training quantization}, where the weights and activations are quantized after the full-precision model training. The quantization range for activations is determined by computing the distribution on training set. Applying post-training INT8 quantization usually leads to minor or no loss of accuracy. Recent work~\cite{banner2019post} also studies the post-training quantization of INT4 models.

\emph{Quantization-aware training} can reduce the quantization accuracy loss by emulating inference-time quantization during training~\cite{jacob2018quantization}. The forward pass during training is consistent with testing time, which helps the on-device deployment. During training, the ``fake quantization operator'' is injected into the convolutional layers, and the batch normalization~\cite{ioffe2015batch} layers are folded.

Both post-training quantization and quantization-aware training require the access to the training data to get a good quantization performance, which is not always feasible on some privacy-sensitive applications. 
\emph{Data-free quantization} aims to reduce the bit-precisions with no access to the training data. Nagel~\etal~\cite{nagel2019data} propose to perform INT8 quantization in a data-free manner equalizing the weight ranges in the network.   ZeroQ~\cite{cai2020zeroq} optimizes for a Distilled Dataset to match the statistics of batch normalization across different layers of the network for data-free quantization.

\subsection{Knowledge Distillation}


Knowledge distillation (KD)~\cite{bucilua2006model, hinton2015distilling} can transfer the ``dark knowledge'' learned in a large model (denoted as the teacher) to a smaller model (denoted as the student) to improve the performance of the smaller one. The small model is either a compressed model or a shallower/narrower model. Bucilua~\etal~\cite{bucilua2006model} achieve the goal by training the student network to match output logits; Hinton~\etal~\cite{hinton2015distilling} introduce the idea of temperature in the softmax output and trained the student to mimic the softened distribution of the teacher model's softmax output. KD shows promising results in various image classification tasks despite the simple implementation.

Apart from the final output, intermediate activations also contain useful information. FitNet~\cite{romero2014fitnets} trains the student to mimic the full feature map of the teacher model through regression. Attention Transfer (AT)~\cite{zagoruyko2017paying} transfers the attention map of the activation from teacher to student, which is the summation of the feature map across channel dimension. Both methods require the intermediate activation to share the same spatial resolution, which limits the choice of the student model. 

KD-based method is also applicable to other applications beyond classification, including object detection~\cite{chen2017learning}, semantic segmentation~\cite{liu2019structured}, language modeling~\cite{sanh2019distilbert} and image synthesis~\cite{li2020gan}.

\subsection{Manual Neural Architecture Design} 
Besides compressing an existing deep neural network, another widely adopted approach to improving efficiency is to design new neural network architectures. A CNN model typically consists of convolution layers, pooling layers, and fully-connected layers, where most of the computation comes from convolution layers. For example, in ResNet-50 \cite{he2016deep}, more than 99\% multiply-accumulate operations (MACs) are from convolution layers. Therefore, designing efficient convolution layers is the core of building efficient CNN architectures. There are three widely used efficient convolution layers including 1$\times$1/pointwise convolution, group convolution, and depthwise convolution:
\begin{itemize}[leftmargin=*]

\item \textbf{1$\times$1 Convolution.} 1$\times$1 convolution (also called pointwise convolution) is a special kind of standard convolution layer, where the kernel size $K$ is 1. Replacing a $K \times K$ standard convolution layer with a 1$\times$1 convolution layer will reduce \#MACs and \#Params by $K^2$ times. In practice, as the 1$\times$1 convolution itself cannot aggregate spatial information, it is combined with other convolution layers to form CNN architectures. For example, 1$\times$1 convolution is usually used to reduce/increase the channel dimension of the feature map in CNN. 

\item \textbf{Group Convolution.} Different from 1$\times$1 convolution that reduces the cost by decreasing the kernel size dimension, group convolution reduces the cost by decreasing the channel dimension. Specifically, the input feature map is split into $G$ groups along the channel dimension. Each group is then fed to a standard $K \times K$ convolution of size $(O_c/G) \times (I_c/G) \times K \times K$. Finally, the outputs are concatenated along the channel dimension. Compared to a standard $K \times K$ convolution, \#MACs and \#Params are reduced by $G$ times in a group convolution.

\item \textbf{Depthwise Convolution.} The number of groups $G$ is an adjustable hyperparameter in group convolutions. A larger $G$ leads to lower computational cost and fewer parameters. An extreme case is that $G$ equals the number of input channels $I_c$. In that case, the group convolution layer is called a depthwise convolution. While the computational cost of a depthwise convolution is lower than a group/normal convolution, its modeling capacity is lower than the group/normal convolution. In practice, depthwise convolution is usually used for edge devices (\eg, mobile), while group/normal convolution is usually used for cloud devices (\eg, GPU).

\end{itemize}

\noindent Based on these efficient convolution layers, there are three representative manually design efficient CNN architectures, including SqueezeNet \cite{iandola2016squeezenet}, MobileNets~\cite{howard2017mobilenets,sandler2018mobilenetv2}, and ShuffleNets~\cite{ma2018shufflenet,zhang2018shufflenet}.

\begin{table}[!t]
\setlength{\tabcolsep}{7pt}
\small\centering
\begin{tabular}{lcccc}
\toprule
& \#Params (M) & \#MACs (M) & Top-1 Accuracy (\%) & Top-5 Accuracy (\%) \\
\midrule
AlexNet~\cite{krizhevsky2012imagenet} & 60 & 720 & 57.2 & 80.3 \\
GoogleNet~\cite{szegedy2015going} & 6.8 & 1550 & 69.8 & 89.5 \\
VGG-16~\cite{simonyan2015very} & 138 & 15300 & 71.5 & -- \\
ResNet-50~\cite{he2016deep} & 25.5 & 4100 & 76.1 & 92.9 \\
\midrule
SqueezeNet~\cite{iandola2016squeezenet} & 1.2 & 1700 & 57.4 & 80.5 \\
\midrule
MobileNetV1~\cite{howard2017mobilenets} & 4.2 & 569 & 70.6 & 89.5 \\
MobileNetV2~\cite{sandler2018mobilenetv2} & 3.4 & 300 & 72.0 & -- \\
MobileNetV2-1.4~\cite{sandler2018mobilenetv2} & 6.9 & 585 & 74.7 & -- \\
\midrule
ShuffleNetV1-1.5$\times$~\cite{zhang2018shufflenet} & 3.4 & 292 & 71.5 & -- \\
ShuffleNetV2-1.5$\times$~\cite{ma2018shufflenet} & 3.5 & 299 & 72.6 & -- \\
ShuffleNetV2-2$\times$~\cite{ma2018shufflenet} & 7.4 & 591 & 74.9 & -- \\
\bottomrule
\end{tabular}
\caption{Summarized results of manually-designed CNN architectures on the ImageNet dataset.}
\label{tab:manually_designed_cnn_results}
\end{table}

\begin{itemize}[leftmargin=*]

\item \textbf{SqueezeNet.} SqueezeNet~\cite{iandola2016squeezenet} targets extremely compact model sizes for mobile applications. It has only 1.2 million parameters but achieves an accuracy similar to AlexNet (Table~\ref{tab:manually_designed_cnn_results}). SqueezeNet has 26 convolution layers and no fully-connected layer. The last feature map goes through a global average pooling and forms a 1000-dimension vector to feed the softmax layer. SqueezeNet has eight \emph{Fire} modules. Each fire module contains a squeeze layer with 1$\times$1 convolution and a pair of 1$\times$1 and 3$\times$3 convolutions. SqueezeNet achieves a top-1 accuracy of 57.4\% and a top-5 accuracy of 80.5\% on ImageNet~\cite{deng2009imagenet}. 

\item \textbf{MobileNets.} MobileNetV1~\cite{howard2017mobilenets} is based on a building block called \emph{depthwise separable convolution}, which consists of a 3$\times$3 depthwise convolution layer and a 1$\times$1 convolution layer. The input image first goes through a 3$\times$3 standard convolution layer with stride 2, then 13 depthwise separable convolution blocks. Finally, the feature map goes through a global average pooling and forms a 1280-dimension vector fed to the final fully-connected layer with 1000 output units.
With 569M MACs and 4.2M parameters, MobileNetV1 achieves 70.6\% top-1 accuracy on ImageNet (Table~\ref{tab:manually_designed_cnn_results}). MobileNetV2~\cite{sandler2018mobilenetv2}, an improved version of MobileNetV1, also uses 3$\times$3 depthwise convolution and 1$\times$1 convolution to compose its building blocks. Unlike MobileNetV1, the building block in MobileNetV2 has three layers, including a 3$\times$3 depthwise convolution layer and two 1$\times$1 convolution layers. The intuition is that the capacity of depthwise convolution is much lower than the standard convolution, thus more channels are needed to improve its capacity. From the cost perspective, the \#MACs and \#Params of a depthwise convolution only grow linearly (rather than quadratically like standard convolution) as the number of channels increases. Thus, even having a large channel number, the cost of a depthwise convolution layer is still moderate. Therefore, in MobileNetV2, the input feature map first goes through a 1$\times$1 convolution to increase the channel dimension by a factor called \emph{expand ratio}. Then the expanded feature map is fed to a 3$\times$3 depthwise convolution, followed by another 1$\times$1 convolution to reduce the channel dimension back to the original value. This structure is called \emph{inverted bottleneck} and the block is called \emph{mobile inverted bottleneck block}. Besides the mobile inverted bottleneck block, MobileNetV2 has another two improvements over MobileNetV1. First, MobileNetV2 has skip connections for blocks in which the stride is 1. Second, the activation function of the last 1$\times$1 convolution in each block is removed. Combining these improvements, MobileNetV2 achieves 72.0\% top-1 accuracy on ImageNet with only 300M MACs and 3.4M parameters (Table~\ref{tab:manually_designed_cnn_results}). 

\item \textbf{ShuffleNets.} ShuffleNetV1 also utilizes 3$\times$3 depthwise convolution rather than standard convolution in its building blocks, similar to MobileNets. Besides, ShuffleNetV1 introduces two new operations, pointwise group convolution and channel shuffle. The pointwise group convolution's motivation is to reduce the computational cost of 1$\times$1 convolution layers. However, it has a side effect: a group cannot see information from other groups. This will significantly hurt accuracy. The channel shuffle operation is thus introduced to address this side effect by exchanging feature maps between different groups. After shuffling, each group will contain information from all groups. On ImageNet, ShuffleNetV1 achieves 71.5\% top-1 accuracy with 292M MACs (Table~\ref{tab:manually_designed_cnn_results}).
Following a similar idea, in ShuffleNetV2, the input feature map is divided into two groups at the beginning of each building block to reduce the computational cost. One group goes through the convolution branch that consists of a 3$\times$3 depthwise convolution layer and two 1$\times$1 convolution layers. The other group goes through a skip connection when the stride is 1 and goes through a 3$\times$3 depthwise separable convolution when the stride is 2. The outputs are concatenated along the channel dimension, followed by a channel shuffle operation to exchange information between groups.
With 299M MACs, ShuffleNetV2 achieves 72.6\% top-1 accuracy on ImageNet (Table~\ref{tab:manually_designed_cnn_results}). 
\end{itemize}

\subsection{Future Directions}

Most of the existing work studies model compression using hardware-unrelated metrics like MACs or model size, or using direct metrics like latency given a pre-defined hardware/software system. There could be a large potential on co-designing the model compression scheme and compiler optimization or hardware design. On the other hand, model compression usually starts from a pre-defined/pre-trained deep network, and compresses it for a more efficient deployment. The compression space is largely based on the pre-defined network architecture. Therefore, if the network is not originally designed for a specific edge hardware (\eg, deploy ResNets on mobile phones), compression may not be able to close the huge gap compared to other efficient network designs like MobileNets. In such cases, designing a new network architecture from scratch could bring larger efficiency improvement. We will discuss more about automatic network architecture design with Neural Architecture Search (NAS) in the next section.

%% file: text/automl.tex
\section{Automated Compression and Neural Architecture Search}
\label{sect:automl}

The success of the aforementioned model compression strategies and efficient neural network architectures relies on hand-crafted heuristics that require domain experts to explore the large design space, trading off among model size, latency, energy, and accuracy. This is time-consuming and sub-optimal. In this section, we describe automated methods for tackling this challenge.

\subsection{Automated Model Compression}
Model compression methods can improve the efficiency of the deployed models. However, the performance of model compression is largely affected by the hyperparameters. For example, different layers in deep networks have different capacities and sensitivities (\eg, the first layer in CNN is usually very sensitive to pruning). Therefore, we should apply different pruning ratios for different layers of the network to achieve the optimal performance. 
The design space is so large that human heuristic is usually sub-optimal, and manual model compression is time-consuming. To this end, automated model compression is proposed to find good compression policy without human effort.


\myparagraph{Automated Pruning.}
Conventional model pruning techniques rely on hand-crafted features and require domain experts to explore the large design space trading off among model size,
speed, and accuracy, which is usually sub-optimal and time-consuming.
AutoML for Model Compression (AMC)~\cite{he2018amc} leverages reinforcement learning to efficiently sample the design space and find the optimal pruning policy for a given network. 
The reward is calculated as a function of accuracy and FLOP.
AMC outperforms manually tuned compression baselines like Han~\etal~\cite{han2017thesis} in a fully automated manner. A recent work MetaPruning~\cite{liu2019metapruning} first trains a PruningNet, a kind of meta network, which is able to generate weight parameters for any pruned structure given the target network, and then uses it to search for the best pruning policy under different constraints. The meta network can be used to directly measure the compression accuracy without fine-tuning.

\myparagraph{Automated Quantization.}
Mixed-precision quantization also requires extensive effort deciding the optimal bit-width for each layer to achieve the best accuracy-performance trade-off. Hardware-Aware Automated Quantization
(HAQ)~\cite{wang2019haq} is proposed to automate the process. HAQ leverages the reinforcement learning to automatically determine the quantization policy. Compared with conventional methods, HAQ is fully automated and can specialize the quantization policy for different neural network architectures and hardware architectures. Done~\etal~\cite{dong2019hawq} propose a second-order quantization method Hessian AWare Quantization (HAWQ) for mixed precision quantization. HAWQ allows for the automatic selection of the relative quantization precision of each layer, based on the layer's Hessian spectrum. It also shows superior performance when applied to large language models~\cite{shen2020q}.

\subsection{Automated Neural Architecture Design}

Neural Architecture Search (NAS) refers to automatic methods for neural network architecture design. 
In the conventional NAS formulation~\cite{zoph2017neural}, designing neural network architectures is modeled as a sequence generation problem, where an auto-regressive RNN controller is introduced to generate neural network architectures. This RNN controller is trained by repeatedly sampling neural network architectures, evaluating the sampled neural network architectures, and updating the controller based on the feedback.

To find a good neural network architecture in the vast search space, this process typically requires to train and evaluate tens of thousands of neural networks (\eg, 12,800 in Zoph~\etal~\cite{zoph2018learning}) on the target task, leading to prohibitive computational cost ($10^4$ GPU hours). To address this challenge, many techniques are proposed that try to improve different components of NAS, including search space, search algorithm, and performance evaluation strategy. 

\begin{figure}[t]
    \centering
    \includegraphics[width=0.75\linewidth]{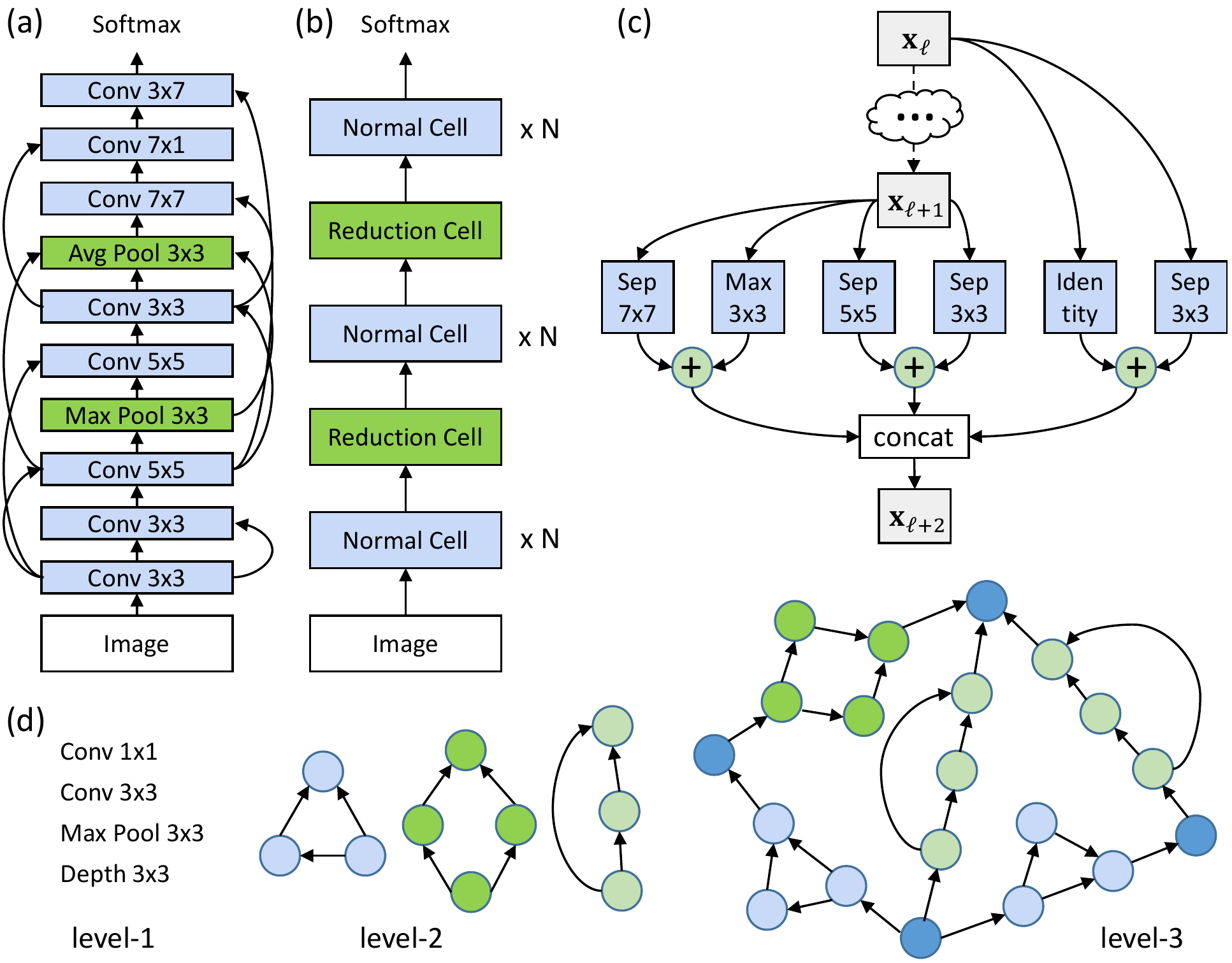}
    \caption{Design spaces for neural architecture search~\cite{deng2020model}: (a) network-level search space~\cite{zoph2017neural}; (b) cell-level search space~\cite{zoph2018learning}; (c) an example of learned cell structure~\cite{liu2018progressive}; (d) three-level hierarchical search space~\cite{liu2017hierarchical}.}
    \label{fig:nas}
\end{figure}

\myparagraph{Search Space.}
All NAS methods need a pre-defined search space that contains basic network elements and how they connect with each other. For example, the typical basic elements of CNN models consist of (1) convolutions~\cite{zoph2018learning,real2019regularized}: standard convolution (1$\times$1, 3$\times$3, 5$\times$5), asymmetric convolution (1$\times$3 and 3$\times$1, 1$\times$7 and 7$\times$1), depthwise-separable convolution (3$\times$3, 5$\times$5), dilated convolution (3$\times$3); (2) poolings: average pooling (3$\times$3), max pooling (3$\times$3); (3) activation functions~\cite{ramachandran2017searching}. Then these basic elements are stacked sequentially~\cite{baker2017designing} with identity connections~\cite{zoph2017neural}. The full network-level search space grows exponentially as the network deepens. When the depth is 20, this search space contains more than 10$^{36}$ different neural network architectures~\cite{zoph2018learning}.

Instead of directly searching on such an exponentially large space, restricting the search space is a very effective approach for search acceleration. Specifically, researchers propose to search for basic building cells (Figure \ref{fig:nas} (b)) that can be stacked to construct neural networks, rather than the entire neural network architecture~\cite{zoph2018learning,zhong2018practical}. As such, the architecture complexity is independent of the network depth, and the learned cells are transferable across different datasets. It enables NAS to search on a small proxy dataset (\eg, CIFAR-10), and then transfer to another large-scale dataset (\eg, ImageNet) by adapting the number of cells. This proxy-based approach greatly reduces the search cost. However, it usually provides worse performances than directly searching on the target large-scale dataset. Within the cell, the complexity is further reduced by supporting hierarchical topologies~\cite{liu2017hierarchical}, or increasing the number of elements (blocks) in a progressive manner~\cite{liu2018progressive}.

\begin{table}[!t]
\setlength{\tabcolsep}{3pt}
\small\centering
\begin{tabular}{lccccccc}
\toprule
& \multirow{2.5}{*}{Algorithm} & \multirow{2.5}{*}{GPU Days} & \multicolumn{2}{c}{CIFAR-10} & \multicolumn{2}{c}{ImageNet} \\
\cmidrule(lr){4-5}\cmidrule(lr){6-7}
& & & \#Params (M) & Accuracy (\%) & \#MACs (M) & Accuracy (\%) \\
\midrule
MetaQNN~\cite{baker2017designing} & Q-learning & 100 (--) & -- & 93.1 & -- & -- \\
NAS~\cite{zoph2017neural} & REINFORCE & 22400 (K40) & 37.4 & 96.4 & -- & -- \\
EAS~\cite{cai2018efficient} & REINFORCE	& 10 (GTX 1080) & 10.7 & 96.6 & -- & -- \\
BlockQNN~\cite{zhong2018practical} & Q-learning	& 96 (Titan X) & 39.8 & 96.5 & -- & 75.7 \\
NASNet~\cite{zoph2018learning} & PPO & 2000 (P100) & 3.3 & 96.6 & 564 & 74.0 \\
PNASNet~\cite{liu2018progressive} & SMBO & 250 (P100) & 3.2 & 96.6 & 588 & 74.2 \\
AmoebaNet~\cite{real2019regularized} & EA & 3150 (K40) & 34.9 & 97.9 & 570 & 75.7 \\
DARTS~\cite{liu2019darts} & Gradient & 4 (GTX 1080Ti) & 3.3 & 97.2 & 574 & 73.3 \\
\bottomrule
\end{tabular}
\caption{Results of different search algorithms for neural architecture search.}
\label{tab:nas_table}
\end{table}

\myparagraph{Search Algorithm.}
NAS methods usually have two stages at each search step: (1) the generator produces an architecture, and then (2) the evaluator trains the network and obtains the performance. As getting the performance of a sampled neural network architecture involves training a neural network, which is very expensive, search algorithms that affect the sample efficiency play an important role in improving the search speed of NAS. 
Most of the search algorithms used in NAS fall into 5 categories: 
random search, reinforcement learning (RL), evolutionary algorithms, Bayesian optimization, and gradient-based methods. Among them, RL, evolutionary algorithms, and gradient-based methods empirically provide the most competitive results (Table~\ref{tab:nas_table}).

RL-based methods model the architecture generation process as a Markov Decision Process, treat the accuracy of the sampled architecture as the reward and update the architecture generation model with RL algorithms, including Q-learning~\cite{baker2017designing,zhong2018practical}, REINFORCE~\cite{zoph2017neural} and PPO~\cite{zoph2018learning}. Instead of training an architecture generation model, evolutionary methods~\cite{real2019regularized,liu2017hierarchical} maintain a population of neural network architectures. This population is updated through mutation and recombination. While both RL-based methods and evolutionary methods optimize neural network architectures in the discrete space, DARTS~\cite{liu2019darts} proposes continuous relaxation of the architecture representation. The output $y$ is modeled as the weighted sum of candidate operations' outputs ($\{o_i(x)\}$): 
\begin{equation}
    y = \sum_i \alpha_i o_i(x), \qquad \alpha_i \geq 0, \sum_i \alpha_i = 1,
\end{equation}
where $\alpha_i$ is the architecture parameter representing the probability of choosing candidate operation $o_i$. Such continuous relaxation allows optimizing neural network architectures in the continuous space using gradient descent, which greatly improves the search efficiency. 
Besides the above techniques, the search efficiency can be improved by exploring the architecture space with network transformation operations, starting from an existing network, and reusing the weights~\cite{cai2018efficient,cai2018path,elsken2018efficient}.

\myparagraph{Performance Evaluation.}
To guide the search process, NAS methods need to get the performances (typically accuracy on the validation set) of sampled neural architectures. The trivial approach to get these performances is to train sampled neural network architectures on the training data and measure their accuracy on the validation set. However, it will result in excessive computational cost~\cite{zoph2017neural,zoph2018learning,real2019regularized}. This motivates many techniques that aim at speeding up the performance evaluation step. Alternatively, the evaluation step can also be accelerated using Hypernetwork~\cite{brock2018smash}, which can directly generate weights of a neural architecture without training it. As such, only a single Hypernetwork needs to be trained, which greatly saves the search cost. Similarly, one-shot NAS methods~\cite{pham2018efficient,liu2019darts,cai2019proxylessnas} focus on training a single super-net, from which small sub-networks directly inherit weights without training cost. 

\begin{figure*}[t]
    \centering
    \includegraphics[width=0.95\linewidth]{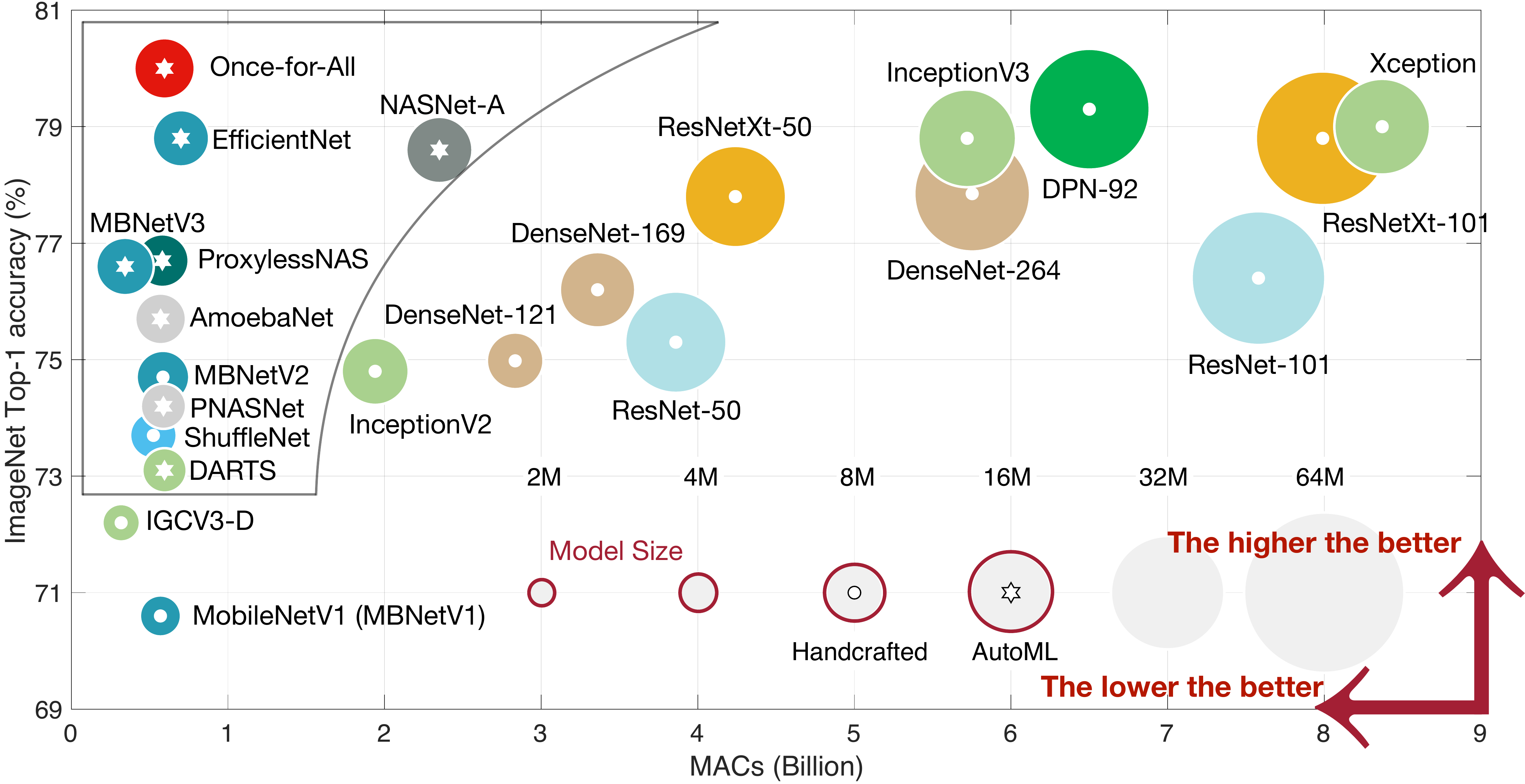}
    \caption{Summarized results of auto-designed and human-designed CNN models on the ImageNet dataset~\cite{cai2020once}.}
    \label{fig:auto_vs_manual}
\end{figure*}

\begin{wrapfigure}{r}{.45\linewidth}
    \centering
    \includegraphics[width=0.9\linewidth]{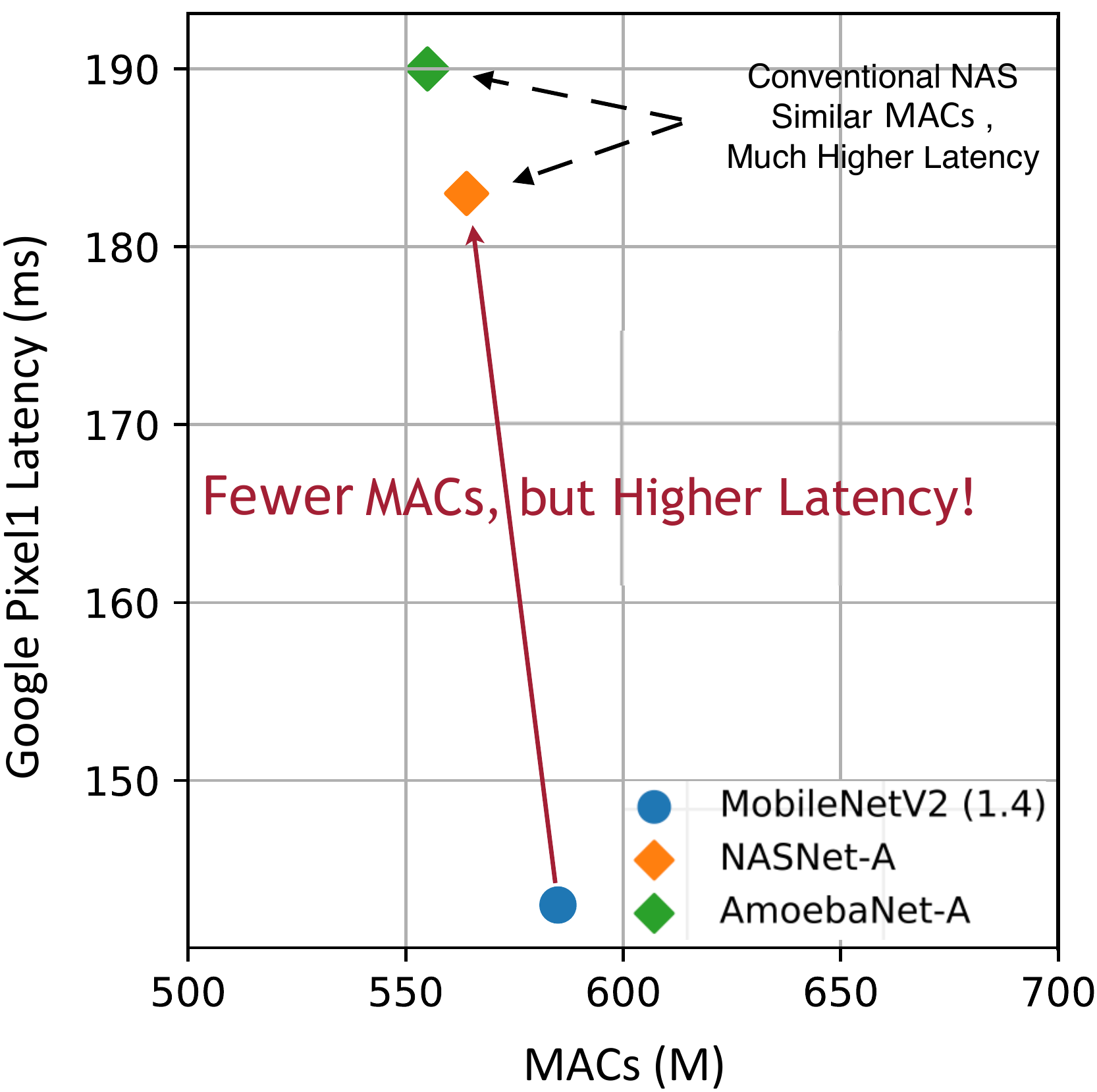}
    \caption{\#MACs does not reflect the real hardware efficiency. NASNet-A and AmoebaNet-A (auto-designed CNN models) have fewer MACs than MobileNetV2-1.4 (human-designed CNN model). However, they run slower than MobileNetV2-1.4 on the Google Pixel 1 phone.}
    \label{fig:macs_not_equal_latency}
\end{wrapfigure}

\myparagraph{Auto-Designed \vs Human-Designed.}
Figure~\ref{fig:auto_vs_manual} reports the summarized results of auto-designed and human-designed CNN models on ImageNet. NAS saves engineer labor costs and provides better CNN models over human-designed CNNs. Besides ImageNet classification, auto-designed CNN models have also outperformed manually designed CNN models on object detection~\cite{zoph2018learning,chen2019detnas,ghiasi2019fpn,tan2020efficientdet} and semantic segmentation~\cite{liu2019auto,chen2018searching}.

\myparagraph{Hardware-Aware Neural Architecture Search.}
While NAS has shown promising results, achieving significant MACs reduction without sacrificing accuracy, in real-world applications, we care about the real hardware efficiency (\eg, latency, energy) rather than \#MACs. Unfortunately, reduction in \#MACs does not directly translate to measured speedup. Figure~\ref{fig:macs_not_equal_latency} shows the comparison between auto-designed CNN models (NASNet-A and AmoebaNet-A) and human-designed CNN model (MobileNetV2-1.4). Although NASNet-A and AmoebaNet-A have fewer MACs, they actually run slower on hardware than MobileNetV2-1.4.

It is because \#MACs only reflects the computation complexity of convolutions. Other factors like data movement cost, parallelism, cost of element-wise operations that significantly affect real hardware efficiency are not taken into consideration. This problem motivates hardware-aware NAS techniques~\cite{tan2019mnasnet,cai2019proxylessnas,wu2019fbnet} that directly incorporate hardware feedback into the architecture search process.
For example, in MNAS~\cite{tan2019mnasnet}, each sampled neural network architecture is measured on the target hardware to collect its latency information, in addition to its accuracy information. A multi-objective reward is defined based on accuracy and latency: $\text{reward} = \text{accuracy} \times (\text{latency}/T)^\omega$, where $T$ is the target latency and $\omega$ is a hyperparameter. 

\myparagraph{Latency Prediction.}
Measuring the latency on-device is accurate but not ideal for scalable neural architecture search. There are two reasons: (i) \emph{Slow.} As suggested in TensorFlow-Lite, 
we need to average hundreds of runs to produce a precise measurement, approximately 20 seconds. This is far more slower than a single forward/backward execution. (ii) \emph{Expensive.} A lot of mobile devices and software engineering work are required to build an automatic pipeline to gather the latency from a mobile farm. Instead of direct measurement, an economical solution is to build a prediction model to estimate the latency~\cite{cai2019proxylessnas}. In practice, this is implemented by sampling neural network architectures from the candidate space and profiling their latency on the target hardware platform. The collected data is then used to build the latency prediction model. For hardware platforms that sequentially execute operations, like mobile device and FPGA, a simple latency lookup table that maps each operation to its estimated latency is sufficient to provide very accurate latency predictions~\cite{cai2019proxylessnas,cai2020once}. For hardware platforms where this sequential execution assumption does not hold, like GPU and CPU, we can train a neural network to predict the latency~\cite{wang2020hat}.

\myparagraph{Diverse Deployment Scenarios.}
Although specialized CNNs are superior over non-specialized ones, designing specialized CNNs for every scenario is still difficult, either with human-based methods or hardware-aware NAS. Since such methods need to \emph{repeat} the network design process and \emph{retrain} the designed network from scratch for \emph{each} case. Their total cost grows linearly as the number of deployment scenarios increases (Table~\ref{tab:comparison_with_nas}), which will result in excessive energy consumption and $CO_2$ emission~\cite{strubell2019energy}. It makes them unable to handle the vast amount of hardware devices (23.14 billion IoT devices till 2018) and highly dynamic deployment environments (different battery conditions, different latency requirements). To handle this challenge, one promising direction is to train a single neural network that supports diverse architecture configurations, amortizing the training cost. Wu~\etal~\cite{wu2018blockdrop}, Liu~\etal~\cite{liu2018dynamic} and Wang~\etal~\cite{wang2018skipnet} propose to learn a controller or gating modules to adaptively drop layers; Huang~\etal~\cite{huang2017multi} introduce early-exit branches in the computation graph;
Lin~\etal~\cite{lin2017runtime} adaptively prune channels based on the input feature map; Kuen~\etal~\cite{kuen2018stochastic} introduce stochastic downsampling point to reduce the feature map size adaptively; Slimmable Nets~\cite{yu2018slimmable,yu2019universally} propose to train a model to support multiple width multipliers (\eg, 4 different global width multipliers), building upon existing human-designed neural networks (\eg, MobileNetV2 0.35, 0.5, 0.75, 1.0). Recently, Cai~\etal~\cite{cai2020once} introduce techniques to train a more powerful once-for-all (OFA) network, which enables a much more diverse architecture space (depth, width, kernel size, and resolution) and a significantly larger number of architectural settings.

\begin{table*}[!t]
\setlength{\tabcolsep}{3pt}
\small\centering
\resizebox{1\linewidth}{!}{
\begin{tabular}{lcccccccc}
\toprule
& \multirow{2.5}{*}{Accuracy} & \multirow{2.5}{*}{\#MACs} & \multirow{2.5}{*}{\shortstack[c]{Mobile\\Latency}} & \multirow{2.5}{*}{\shortstack[c]{Searching Cost\\(GPU Hours)}} & \multirow{2.5}{*}{\shortstack[c]{Training Cost\\(GPU Hours)}} & \multicolumn{3}{c}{Total Cost $(N = 40)$} \\
\cmidrule(lr){7-9}
& & & & & & GPU Hours & $\text{CO}_2\text{e}$ (lbs) & AWS Cost \\
\midrule
MobileNetV2~\cite{sandler2018mobilenetv2} & 72.0\% & 300M & 66ms & 0 & 150$N$ & 6k & 1.7k & \$18.4k \\
MobileNetV2 \#1200~\cite{sandler2018mobilenetv2} & 73.5\% & 300M & 66ms & 0 & 1200$N$ & 48k & 13.6k & \$146.9k  \\
\midrule
NASNet-A~\cite{zoph2018learning} & 74.0\% & 564M & -- & 48,000$N$ & -- & 1,920k & 544.5k & \$5875.2k  \\
DARTS~\cite{liu2019darts} & 73.1\% & 595M & -- & 96$N$ & 250$N$ & 14k & 4.0k & \$42.8k \\
\midrule
MnasNet~\cite{tan2019mnasnet} & 74.0\% & 317M & 70ms & 40,000$N$ & -- & 1,600k & 453.8k & \$4896.0k \\
FBNet-C~\cite{wu2019fbnet} & 74.9\% & 375M & -- & 216$N$ & 360$N$ & 23k & 6.5k & \$70.4k \\
ProxylessNAS~\cite{cai2019proxylessnas} & 74.6\% & 320M & 71ms & 200$N$ & 300$N$ & 20k & 5.7k & \$61.2k \\
SinglePathNAS~\cite{guo2020single} & 74.7\% & 328M & -- & 288 + 24$N$ & 384$N$ & 17k & 4.8k & \$52.0k \\
AutoSlim~\cite{yu2019autoslim} & 74.2\% & 305M & 63ms & 180 & 300$N$ & 12k & 3.4k & \$36.7k \\
MobileNetV3-Large~\cite{howard2019searching} & 75.2\% & 219M & 58ms & -- & 180$N$ & 7.2k & 1.8k & \$22.2k \\
OFA~\cite{cai2020once} & 76.0\% & 230M & 58ms & 40 & 1200 & 1.2k  & 0.34k & \$3.7k \\
OFA \#75~\cite{cai2020once} & 76.9\% & 230M & 58ms & 40 & 1200 + 75$N$ & 4.2k & 1.2k & \$13.0k \\
\bottomrule
\end{tabular}
}
\caption{Summarized results of different neural architecture search frameworks~\cite{cai2020once}. In this table, the accuracy is evaluated on the ImageNet dataset, the mobile latency is measured with the Pixel 1 phone, the $\text{CO}_2$ emission (``$\text{CO}_2\text{e}$'') is calculated following Strubell~\etal~\cite{strubell2019energy}, and the AWS cost is estimated based on the price of on-demand P3.16xlarge instances.}
\label{tab:comparison_with_nas}
\end{table*}

\myparagraph{Design the Search Space.}
Search space design is crucial to the final NAS performance. Existing network design space is usually derived from manual network design (\eg, the search space used in ProxylessNAS~\cite{cai2019proxylessnas} is derived from MobileNetV2~\cite{sandler2018mobilenetv2}). Recent works investigate the design of search space itself. 
Radosavovic~\etal~\cite{radosavovic2020designing} propose to design a network design space that parametrize populations of networks, which consist of simple, regular networks called RegNet. The core insight of such parametrization is to represent the widths and depths by a quantized linear function. The optimized network design space contains good network architectures that can be found by random search. Recently, Lin~\etal~\cite{lin2020mcunet} propose TinyNAS, a two-stage neural architecture search method for memory-constrained deployment on microcontrollers (MCUs). Due to the lack of search space design for MCUs, TinyNAS first optimizes the search space itself to improve the performance of neural architecture search.

\myparagraph{AutoML for TinyML.}
TinyML is a new frontier for edge deep learning computing. AutoML-based methods have been applied to TinyML area. SpArSe~\cite{fedorov2019sparse} employs a Bayesian optimization framework that jointly selects model architecture and optimizations such as pruning to meet memory constraints.
MCUNet~\cite{lin2020mcunet} co-designs the efficient neural architecture and efficient compiler/runtime to enable ImageNet-scale applications on off-the-shelf microncontrollers. MicroNet~\cite{banbury2020micronets} observes that on average, model latency varies linearly with model operation (op) count for models in the search space. It then employs differentiable NAS to search for models with low memory usage and low op count. 
Rusci~\etal~\cite{rusci2020leveraging} use reinforcement learning
(RL) to find a good mixed-precision quantization policy in order to help fit an ImageNet model on MCUs.

\subsection{Joint Compression and Neural Architecture Search}

As designing efficient neural network architectures and model compression are orthogonal to each other, in practice, we can combine these two techniques to further boost efficiency. 

\begin{figure}[t]
    \centering
    \includegraphics[width=0.75\linewidth]{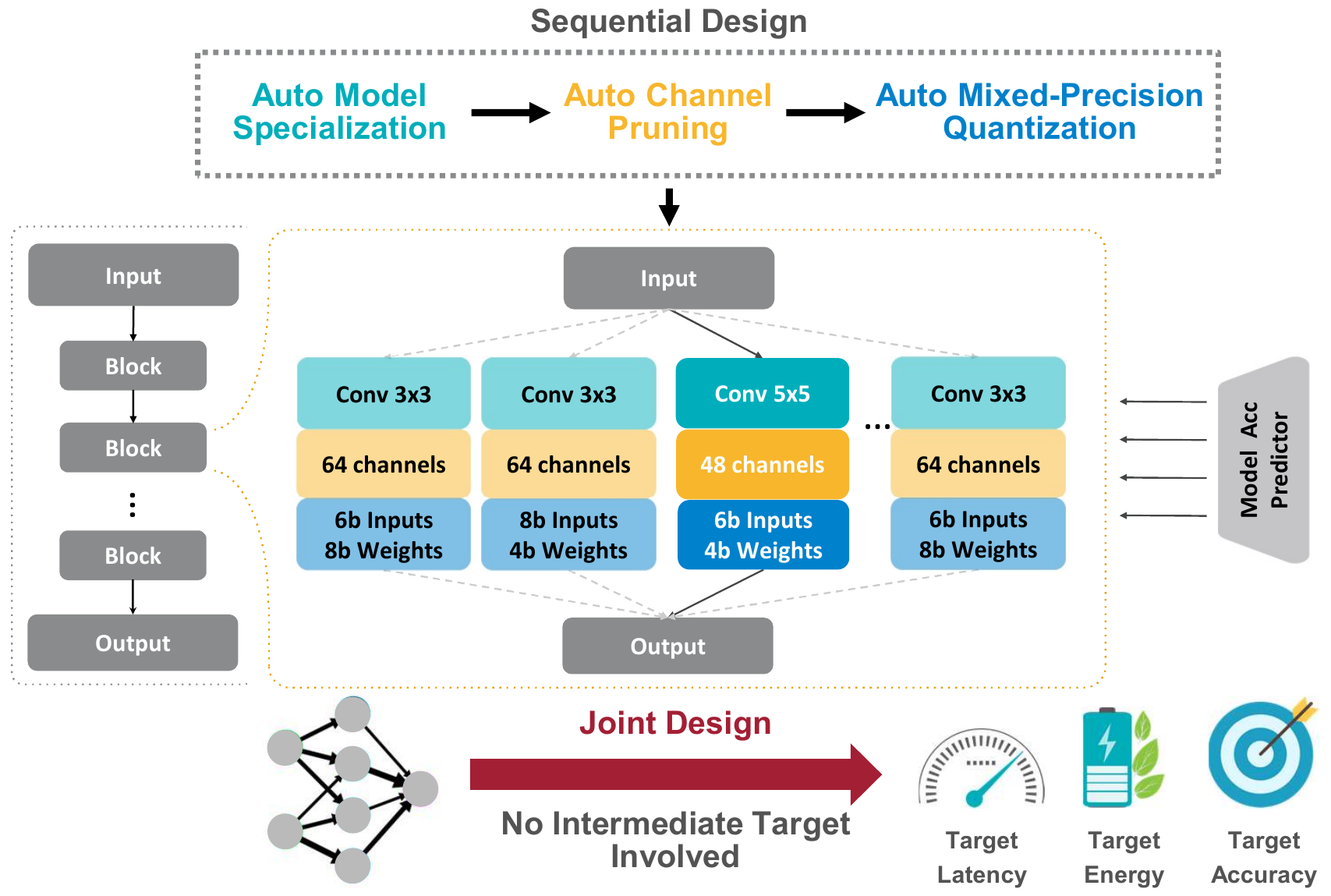}
    \caption{Sequential optimization framework (upper) and joint optimization framework (lower) for designing efficient neural architecture and searching best pruning/quantization policy~\cite{wang2020apq}.}
\label{fig:apq_joint_vs_sequential}
\end{figure}

\myparagraph{Sequential Optimization.}
A straightforward approach to doing this is applying these techniques separately (Figure~\ref{fig:apq_joint_vs_sequential} upper). For example, Cai~\etal~\cite{cai2019automl} employs a sequential AutoML pipeline to accelerate neural network inference on hardware, which starts with searching an efficient neural network architecture with hardware-aware NAS~\cite{cai2019proxylessnas}, then applies automated channel pruning~\cite{he2018amc} and mixed-precision quantization~\cite{wang2019haq} to compress the searched neural network. 
A critical drawback of this straightforward approach is that optimizing in separate stages will lead to sub-optimal results: \eg, the best network architecture for the full-precision model is not necessarily optimal after pruning and quantization. Besides, each step has its own optimization objective (\eg, accuracy, latency, energy) and thus requires considerable human efforts and computational cost to tune the intermediate targets. 

\myparagraph{Joint Optimization.}
Unlike sequential optimization approaches, a joint optimization approach tackles this problem in an end-to-end manner, which is preferred (Figure~\ref{fig:apq_joint_vs_sequential} lower). However, directly extending existing AutoML techniques to the joint model optimization setting can be problematic. Firstly, the joint search space is much larger (multiplicative) than the stage-wise search, making the search difficult.
Pruning and quantization usually require a time-consuming fine-tuning process to restore accuracy~\cite{wang2019haq,yang2018netadapt}, which dramatically increases the search cost.

APQ~\cite{wang2020apq} proposes to use the \emph{quantization-aware accuracy predictor} to accelerate this joint optimization. The predictor takes the model architecture and the quantization scheme as input, and can quickly predicts its accuracy. Instead of fine-tuning the pruned and quantized network to get the accuracy, it uses the estimated accuracy generated by the predictor, which can be obtained with negligible cost (since the predictor requires only a few FC layers). Training this quantization-aware accuracy predictor requires collecting a lot of (quantized model, quantized accuracy) data points, where getting each of the data points could be quite expensive: (1) we need to train the network to get the initial FP32 weights, (2) and further fine-tuning to get the quantized INT8 weights to evaluate the accuracy. Both stages are quite expensive, requiring hundreds of GPU hours. 

To reduce the cost of stage 1, APQ employs a pre-trained once-for-all network~\cite{cai2020once} that supports all sub-networks while achieving on-par or even higher accuracy compared to training from scratch. To reduce the cost of stage 2, APQ proposes a \emph{predictor-transfer technique}.
APQ first trains an FP32 model accuracy predictor using the cheap (FP32 model, FP32 accuracy) data points collected with the weight-sharing once-for-all network  (evaluation only, no training required). Then APQ transfers the predictor to the quantized model domain by fine-tuning it on a small number of expensive (quantized model, quantized accuracy) data points. The transfer technique dramatically improves the sample efficiency on the quantized network domain and reduces the overall cost to train the predictor. Under the same latency/energy constraint, APQ can attain better accuracy than the sequentially-optimized model (74.1\% \vs 71.8\%). This is reasonable since the per-stage optimization might fall into local optimal results while the joint design approach does not.

\subsection{Limitations and Future Directions}
Methodologically, current AutoML research mainly focuses over searching on restricted design spaces, such as searching several pre-defined architectural hyperparameters (\eg, depth, width, kernel size, resolution) based on existing human-designed neural network architectures. While this can simplify the task and reduce the search cost, it limits the optimization headroom, making it impossible to discover novel primitive operations or building blocks. Besides, different hardware platforms typically require different search spaces. For example, depthwise convolution provides a very good trade-off between latency and accuracy on mobile platforms, while it is less effective for GPUs since it cannot fully utilize the hardware parallelism. Thus, an important future research direction is to break this limitation and extend AutoML to more general and diverse design spaces. This also requires designing better AutoML algorithms to handle more complicated design spaces. Besides, another future research direction is to extend AutoML to more machine learning applications, which requires combining AutoML techniques with domain-specific insights. 

%% file: text/training.tex
\section{Efficient On-Device Learning}
\label{sect:training}

In real-world edge AI applications, intelligent edge devices keep collecting \emph{new} data through the sensor every day while being expected to provide high-quality and customized services without sacrificing privacy. 
These pose new challenges to efficient AI techniques that could not only run inference but also continually adapt the models to newly collected data (\ie, on-device learning). 

\begin{figure}
\centering
\begin{minipage}[t]{.48\textwidth}
  \centering
  \includegraphics[width=\linewidth]{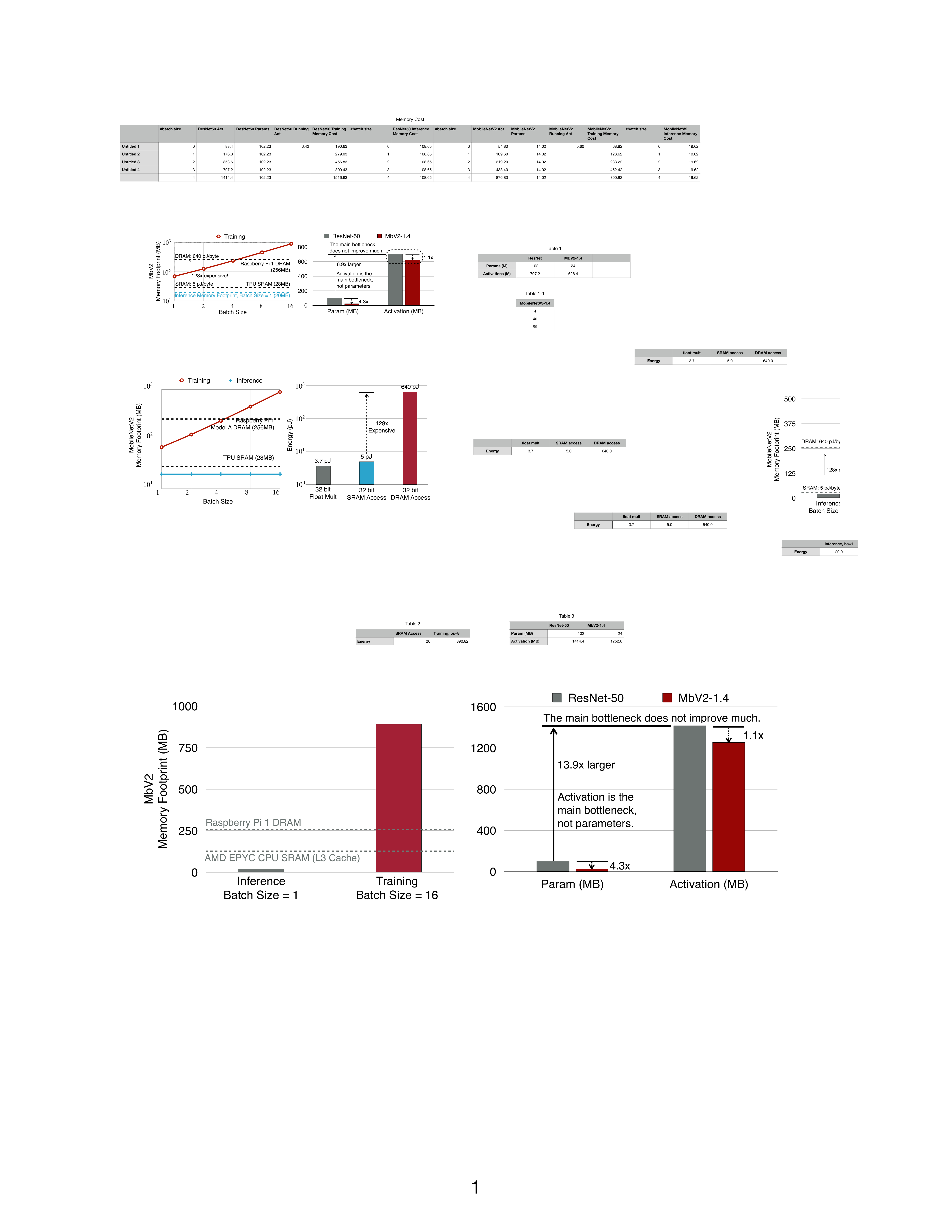}
  \captionof{figure}{The memory footprint required during training is much larger than inference.}
  \label{fig:memory_training_versus_inference}
\end{minipage}
\hfill
\begin{minipage}[t]{.48\textwidth}
  \centering
  \includegraphics[width=\linewidth]{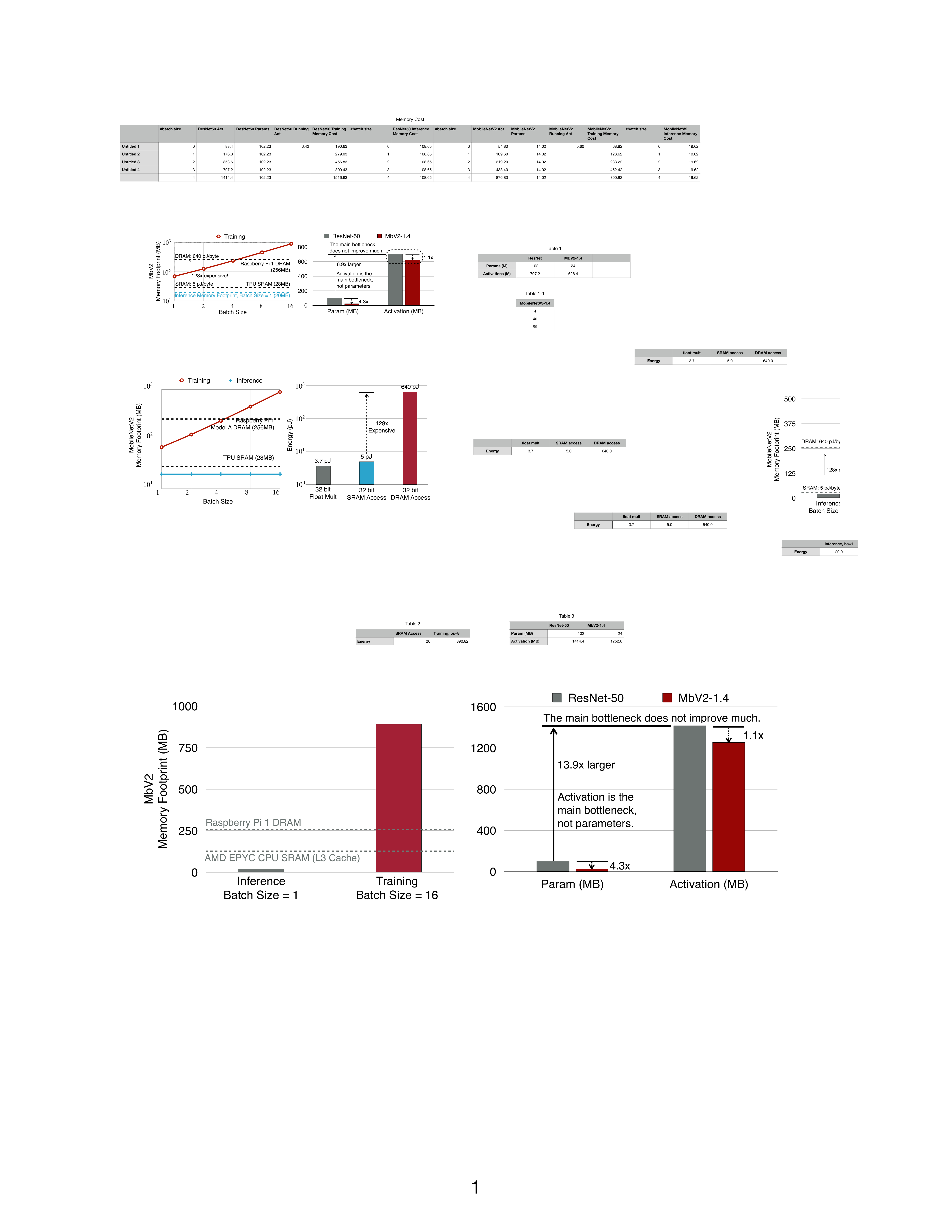}
  \captionof{figure}{Memory cost comparison between ResNet-50 and MobileNetV2-1.4 (under batch size 16). Recent advances in efficient model design only reduce the size of parameters, but the activation size, which is the bottleneck for training, does not improve much.}
  \label{fig:activation_is_the_main_bottleneck}
\end{minipage}
\end{figure}

Though on-device learning can enable many appealing applications, it is an extremely challenging problem. First, edge devices are \emph{memory-constrained}. For example, a Raspberry Pi 1 Model A only has 256MB of memory, which is sufficient for inference, but by far insufficient for training (Figure~\ref{fig:memory_training_versus_inference}), even with a lightweight neural network (MobileNetV2 \cite{sandler2018mobilenetv2}). Furthermore, the memory is shared by various on-device applications (\eg, other deep learning models) and the operating system. A single application may only be allocated a small fraction of the total memory, which makes this challenge more critical. Second, edge devices are \emph{energy-constrained}. DRAM access consumes two orders of magnitude more energy than on-chip SRAM access. The large memory footprint of activations cannot fit into the limited on-chip SRAM, thus it has to access DRAM. For instance, the training memory of MobileNetV2, under batch size 16, is close to 1GB, which is by far larger than the SRAM size of an AMD EPYC CPU (Figure~\ref{fig:memory_training_versus_inference}), not to mention lower-end edge platforms. If the training memory can fit on-chip SRAM, it will drastically improve the speed and energy efficiency.

In this section, we describe efficient training techniques towards the goal of efficient on-device learning. The key difference between inference and training is that training requires storing all intermediate activations for back-propagation while inference doesn't. Besides, the activation size grows linearly \wrt the training batch size. As a result, activation arises to be the major bottleneck for training. For example, under batch size 16, the activation size of ResNet50~\cite{he2016deep} is 13.9$\times$ larger than its parameter size (Figure~\ref{fig:activation_is_the_main_bottleneck}). Therefore, reducing the activation size is the core objective of most efficient training techniques.

\subsection{General Efficient Training Techniques}

\myparagraph{Gradient Checkpointing.}
One typical approach to reducing the training activation size is to discard a subset of intermediate activations~\cite{gruslys2016memory,chen2016training}. During the backward pass, the discarded intermediate activations will be re-computed to obtain the gradient. We can divide the neural network into $k$ segments, and the activation size of training this model is reduced from $O(n)$ to $O(n/k) + O(k)$. Setting $k=\sqrt{n}$, the activation size becomes $O(2\sqrt{n})$. 

\myparagraph{Activation Pruning.}
Besides dropping intermediate activations, another approach to reduce the activation size is activation pruning. Liu~\etal~\cite{liu2019dynamic} build a dynamic sparse computation graph to prune activations during training. Similar to weight pruning, non-critical neurons are removed to reduce the memory footprint and save computational cost. These non-critical neurons can be selected according to their output activations.  If the output activation has a small or negative value, it will be small/zero after ReLU. Thus, removing them does not significantly affect the final result. This process is input-dependent and does not remove any neurons permanently.

\myparagraph{Low-Bit Training.}
Apart from inference, \emph{training} with quantized weights, activaions, and gradients can reduce the cost of deep learning training. Training with a mixed 16-bit and 32-bit floating-point types in a model has been widely supported by deep learning frameworks such as TensorFlow and PyTorch. Hardware like NVIDIA Volta GPU architecture also paves the way for mixed-precision training. Additionally, custom data formats like BFloat16 and TensorFloat-32 allow for a larger dynamic range to preserve accuracy.
With techniques like loss scaling, such mixed-precision training can reduce the memory consumption and improve training speed with no loss of accuracy. DoReFa-Net uses 1-bit weights, 2-bit activations, and 6-bit gradients for faster training and inference, which can obtain comparable accuracy compared to FP32 for AlexNet~\cite{krizhevsky2012imagenet} on ImageNet~\cite{deng2009imagenet}. Lin~\etal~\cite{lin2016neural} stochastically binarize the weights to reduce the time of FP multiplication in training. 

\myparagraph{Gradient Compression.}
For large-scale distributed training, the quantization is no longer enough as the maximum saving is 32$\times$ (from FP32 to a single bit), while the gap between high-end networking and normal one is 100$\times$ (100Gbps infini-band \vs 1Gbps Ethernet).
To improve the scalability of multi-node training, a more effective method is needed to reduce bandwidth requirements. 
Thus, methods on reducing the transferred bits have been proposed, such as gradient quantization~\cite{onebit,terngrad} and gradient compression~\cite{lin2018deep,wangni2018gradient,sun2019optimizing}. By applying quantization or compression before exchanging gradients, the transferred bits can be reduced by a large margin (up to 600$\times$ as in DGC~\cite{lin2018deep}). The accuracy can be well preserved with warm-up training and error compensation. 

\subsection{Efficient Transfer Learning}

The aforementioned efficient training techniques mainly target the general case where neural networks are trained from scratch. This learning paradigm is suitable for cloud-based learning, where the number of data samples is sufficient. However, for on-device learning scenarios where the number of data samples is limited, it will be difficult to train deep neural networks from scratch. Alternatively, we can transfer a pre-trained neural network to the target on-device task (\ie, transfer learning), which is much more data-efficient. This also allows us to take advantage of existing powerful pre-trained neural networks \cite{devlin2018bert,brown2020language}, which take extensive human efforts to design and huge computational resources to train. 

Specifically, neural networks pre-trained on large-scale datasets (\eg, ImageNet \cite{deng2009imagenet}) are widely used as a fixed feature extractor for transfer learning, then only the last layer needs to be fine-tuned \cite{chatfield2014return,donahue2014decaf,gan2015devnet,sharif2014cnn}. 
This approach does not require to store the intermediate activations of the feature extractor, and thus is memory-efficient. However, the capacity of this approach is limited, resulting in poor accuracy, especially on datasets \cite{maji2013fine}  whose distribution is far from ImageNet (\eg, only 45.9\% Aircraft top1 accuracy achieved by Inception-V3 \cite{mudrakarta2019k}). Alternatively, fine-tuning the full network can achieve better accuracy \cite{kornblith2019better,cui2018large}. But it requires a vast memory footprint and hence is not friendly for training on edge devices. Recently, Mudrakarta~\etal~\cite{mudrakarta2019k} and  Frankle~\etal~\cite{frankle2020training} propose to only update parameters of the batch normalization (BN) \cite{ioffe2015batch} layers, which greatly reduces the number of trainable parameters. Unfortunately, parameter-efficiency does not translate to memory-efficiency (Figure~\ref{fig:activation_is_the_main_bottleneck} right). It still requires a large amount of memory (\eg, 326MB under batch size 8) to store the input activations of the BN layers.

Instead of focusing on reducing the number of trainable parameters, Cai~\etal~\cite{cai2020tinytl} propose \emph{tiny transfer learning} (TinyTL) that targets reducing the training memory footprint. The key insight of TinyTL is that the intermediate activations are only required to update weights, while updating biases does not need them. TinyTL proposes to freeze the weights of the pre-trained feature extractor while only update the biases. To compensate for the capacity loss due to freezing the weights, TinyTL introduces lite residual learning that exploits a new class of generalized memory-efficient bias modules to refine the intermediate feature maps. On Cars, TinyTL provides the same level of accuracy as fine-tuning the full network while reducing the memory cost by 4.6$\times$~\cite{cai2020tinytl}.

\subsection{Federated Learning}

The privacy of personal data is gaining growing attention recent years and it leads to increasing demand of training without breaking privacy.
Federated learning~\cite{mcmahan2016communication} is such a protocol that allows multi clients to jointly train a model without explicitly sharing their data. While it is common to have many edge devices in deployments, federated learning provides a way to utilize all of them and address the concerns of security as the local data never leaves the client. There have been applications such as keyboard content suggestions~\cite{haque2017towards} and medical treatments analysis~\cite{jochems2017developing}.

Different from clusters equipped with high-end network infrastructures, the edge devices are usually connected with less powerful network (\ie, Wi-Fi). In this case, the bandwidth is low and the latency is high and conventional methods scale poorly. 
To eliminate the bottleneck, federated average~\cite{mcmahan2016communication}, gradient compression~\cite{lin2018deep,caldas2018expanding},  and quantization~\cite{jacob2018quantization} greatly reduce the transferred bits to reduce the bandwidth requirements and delayed updated~\cite{zhu2019distributed, zhu2021dga} deals with the latency issue.

\subsection{Discussions and Future Directions}
TinyTL~\cite{cai2020tinytl} reduces the transfer learning memory from more than 250MB to only 16MB, making it promising for in-memory computing for training. For single-device cases, reducing the training memory footprint and the energy consumption is the key challenge for efficient on-device learning. For better efficiency, a promising direction is to put the whole training process into the cache (SRAM), which is much more energy-efficient and faster than DRAM training. To approach this goal, we need to combine advances from both the algorithm domain and the hardware/software system domain. For multi-device cases,
the key challenge is the connection quality. Previous distributed training is designed for high-end networking infrastructure like Infini-Band, but the networking of edge devices (\ie, Wi-Fi) can be unstable and slow. It is also important to protect the data ownership as direct message exchange may leak private user data. To achieve the target, we need to consider from both the system and privacy perspectives.

%% file: text/domain/main.tex
\section{Domain-Specific Optimization}
\label{sect:domain}

Apart from task-agnostic accelerations that can be applied to any domain, there have also been extensive investigations in optimizing for specific tasks. In this paper, we will focus on point cloud, video and natural language processing. On the one hand, they are more computationally expensive than conventional 2D vision due to their large memory bandwidth. On the other hand, they also offer unique opportunities for acceleration: spatial sparsity (point clouds), temporal redundancy (videos), and token redundancy (natural languages).

\input{text/domain/3d}
\input{text/domain/video}

\input{text/domain/nlp}

%% file: text/domain/3d.tex
\subsection{Efficient Point Cloud Processing}

Recently, emerging applications such as AR/VR and autonomous driving have been developing rapidly. In these applications, it is crucial to efficiently process 3D data, usually point clouds out of LiDAR sensors. However, such goal is particularly challenging in 3D since 3D deep learning models are usually an order of magnitude more expensive than image CNNs given similar input size. Different from image data which is represented as dense matrices or tensors, 3D point clouds are usually represented as a sparse set of points: $\bm{x} = \{\bm{x}_k\} = \{(\bm{p}_k, \bm{f}_k)\}$, where $\bm{p}_k$ is the 3D coordinate of the $k$\textsuperscript{th} point, and $\bm{f}_k$ is the feature corresponding to $\bm{p}_k$. Due to the sparse nature of 3D point clouds, they can not be effectively processed by conventional image CNNs, but by specialized DNNs composed of point cloud convolution operations. The major challenges for point cloud convolution are two-folded: large memory footprint introduced by the additional spatial dimension, and irregular memory access pattern introduced by sparse data format.

\myparagraph{Point Cloud Convolution.}
The general form of point cloud convolution can be written as:
\begin{equation}
\label{eqn:conv}
    \bm{y}_k = \sum_{\bm{x}_i \in \mathcal{N}(\bm{x}_k)} \mathcal{K}(\bm{x}_k, \bm{x}_i) \times \mathcal{F}(\bm{x}_i),
\end{equation}
During the convolution, we iterate the center $\bm{x}_k$ over the entire input. For each center, we first index its neighbor $\bm{x}_i$ in neighborhood $\mathcal{N}(\bm{x}_k)$, then convolve the neighboring features $\mathcal{F}(\bm{x}_i)$ with the kernel $\mathcal{K}(\bm{x}_k, \bm{x}_i)$, and finally produces the corresponding output $\bm{y}_k$. 

\input{figText/fig_3d_primitives}

\begin{itemize}[leftmargin=*]

\item \myparagraph{Voxel-Based Convolution.}
Early research on 3D deep learning relies on volumetric representation to process point cloud data~\cite{maturana2015voxnet,cicek20163d,zhou2018voxelnet,wu20153d,qi2016volumetric} (\fig{fig:3d_primitives:voxel}). The point cloud coordinates $\bm{p}_k$ are first quantized into integers, and the point cloud is converted to the dense tensor representation via voxelization. Maturana \etal~\cite{maturana2015voxnet} propose to generalize 2D CNNs to vanilla 3D CNNs to further extract features from the voxel grids. Qi~\etal~\cite{qi2016volumetric} propose subvolume supervision and anisotropic kernels for 3D CNNs, and systematically analyzed the relationship between 3D CNNs and multi-view CNNs. Chang~\etal~\cite{chang2015shapenet} further extend 3D CNNs to object segmentation, which is later improved by VoxSegNet~\cite{wang2019voxsegnet} with dilated convolutions and squeeze-and-excitation operations. Tchapmi~\etal~\cite{tchapmi2017segcloud} propose SEGCloud that uses trilinear interpolation to alleviate the information loss caused by voxelization. Voxel-based methods enjoy the regular memory access pattern thanks to the dense volumetric representation. However, the memory footprint of these methods grows cubically as the resolution grows. As a result, these methods cannot take in input with resolution higher than 64$\times$64$\times$64, which corresponds to 40\% information loss~\cite{liu2019point}.

\item \myparagraph{Point-Based Convolution.}
Another stream of research directly applies deep neural networks on point clouds without converting them to voxel grids~\cite{qi2017pointnet,qi2017pointnet++,qi2018frustum,wang2018sgpn,wang2018dynamic,lan2019modeling,shi2019pointrcnn,yang2019std,qi2019deep,xie2020pointcontrast,jiang2020pointgroup} (\fig{fig:3d_primitives:point}). PointNet~\cite{qi2017pointnet} takes advantage of the symmetric function to process the unordered point sets in 3D. Later research~\cite{qi2017pointnet++} proposed to stack PointNets hierarchically to model neighborhood information and increase model capacity. Instead of stacking PointNets as basic blocks, PointCNN~\cite{li2018pointcnn} and SpiderCNN~\cite{xu2018spidercnn} abstract away the symmetric function using dynamically generated convolution kernels or learned neighborhood permutation function and PointConv~\cite{wu2019pointconv} considers point cloud density while generating dynamic kernels. By applying different kernel to different neighborhood points, these methods have better model capacity comparing with PointNet++ and usually achieve better performance. PointConv achieves better performance on indoor scenes while PointCNN has superior accuracy on 3D objects. InterpCNN~\cite{mao2019interpolated} and KPConv~\cite{thomas2019kpconv} propose to generate convolution kernels via interpolation. This is more efficient than learning-based dynamic kernel generation since it does not require generating different IC$\times$OC kernel matrix for each point. DGCNN~\cite{wang2018dynamic} and Deep GCNs~\cite{li2019deepgcns} that model point clouds as graphs and applies graph convolution layers to extract hierarchical features. Graph-based methods are strong in modeling small objects but usually cannot scale up to large scenes with more than $10^5$ points since the adjacency matrix alone can take up 37.3 GB of GPU memory. Point-based methods have smaller memory footprint comparing with voxel-based methods. Nevertheless, the sparsity of point cloud also brings about large irregular memory access and dynamic kernel generation cost, which takes up to 50\% to 90\% of total runtime~\cite{liu2019point}. Therefore, most computations are wasted on dealing with the irregularity of point cloud representation.

\item \myparagraph{Efficient Voxel-/Point-Based Convolution.} As both voxel and point-based methods are inefficient, increased attention has been paid to the efficient design of point cloud convolution operations. To reduce the memory footprint and computation of vanilla voxel-based methods, OctNet~\cite{riegler2017octnet} proposes to place shallow octrees within regular volumetric grids and apply convolution on this hybrid grid-octree data structure. By limiting the convolution to the octants of 3D shape boundaries instead of the interior volume, O-CNN~\cite{wang2017ocnn} and AO-CNN~\cite{wang2018adaptive} achieve much better speed under smaller input resolution. Other than using octrees to reduce the memory cost of volumetric CNNs, Graham~\etal~\cite{graham2015sparse,graham20183d} propose SparseConvNet (\fig{fig:3d_efficient_primitives:sparseconv}) that skips the non-activated regions during computation and only stores activated points. As a result, SparseConvNet can achieve orders of magnitude lower computation and memory footprint for ultra sparse, large point clouds. Choy~\etal~\cite{choy20194d} further optimize SparseConvNet computation with customized matrix multiplication CUDA kernels and hashmap-based kernel map generation. The resulting MinkowskiNet is widely used in 3D instance segmentation~\cite{lahoud20193d,han2020occuseg} and 3D detection~\cite{shi2019parta2,shi2019pvrcnn,yan2018second}. It also supports high-dimensional geometric feature learning, which is applied in point cloud video understanding~\cite{choy20194d} and 3D registration~\cite{choy2019fully,choy2020deep}. Another stream of research focuses on accelerating point-based methods. KPConv~\cite{thomas2019kpconv} proposes to prebuild $k$-d trees for input point clouds and store them on the disk. As a result, at training time we only need to query the $k$-d trees on the fly instead of doing the entire distance computation and sorting as in vanilla implementations of point-based methods. This reduces the irregular data access cost during training. RandLA-Net~\cite{hu2019randla} further proposes to downsample the input aggressively via random sampling to reduce computation and memory cost of point-based methods.

\input{figText/tab_3d_shape}
\input{figText/tab_3d_indoor}
\input{figText/tab_3d_outdoor}

\item \myparagraph{Efficient Hybrid Convolution.} Recently, there are also hybrid approaches that take advantage of the virtue of both voxel-based and point-based methods. Liu~\etal~\cite{liu2019point,liu2020hardware} propose Point-Voxel Convolution (\fig{fig:3d_efficient_primitives:pvconv}) that does convolution on a voxel-based branch and keeps fine-grained information on a point-based branch. The voxel-based branch does not have to maintain high resolution thanks to the high resolution point-based branch. On the other hand, the point-based branch applies simple multi-layer perceptron to transform the high resolution features, which does not require inefficient irregular memory access operations and dynamic kernel generation. Consequently, these two branches enjoy the benefit of both regular memory access pattern and small memory footprint, which eventually leads to superior efficiency. Later research Grid-GCN~\cite{xu2020grid} extends similar idea to graph neural networks and also achieve significant speedup over existing GCN-based methods. To achieve better efficiency on larger input point clouds, Tang~\etal~\cite{tang2020searching} propose SPVCNN that upgrades the voxel-based branch in Point-Voxel Convolution to a sparse tensor branch, and applies sparse 3D convolutions to aggregate neighborhood features. 

\end{itemize}

\noindent
We summarize the results of different point cloud CNNs in \tab{tab:3d:shapenet:results}, \tab{tab:3d:indoor:results} and \tab{tab:3d:semantickitti:results}.

\begin{itemize}[leftmargin=*]

\item \myparagraph{3D Object Part Segmentation.}
We summarize the results of recent 3D deep learning methods on ShapeNet~\cite{chang2015shapenet} in \tab{tab:3d:shapenet:results}. The latency of KPConv~\cite{thomas2019kpconv} (which adopts heterogeneous batching) is estimated by projecting input size to 16384, and the results of MinkowskiNet~\cite{choy20194d} are cited from Xie~\etal~\cite{xie2020pointcontrast}. Most point-based methods~\cite{qi2017pointnet++,huang2018recurrent,wang2018dynamic,xu2018spidercnn,wu2019pointconv,li2018pointcnn} suffer from high irregular memory access and dynamic kernel computation cost and therefore runs slowly on GPUs. KPConv~\cite{thomas2019kpconv} and PVCNN~\cite{liu2019point} achieve state-of-the-art accuracy while maintaining good efficiency (up to $2.7\times$ speedup, $1.9\times$ parameters reduction over PointCNN).

\item \myparagraph{3D Indoor Scene Segmentation.}
We summarize the results on S3DIS~\cite{armeni20163d,armeni2017joint} and ScanNet~\cite{dai2017scannet} in \tab{tab:3d:indoor:results}. The latency and \#MACs of all methods except MinkowskiNet~\cite{choy20194d} are obtained on a batch of 32768 points. We measure the scene-averaged latency and \#MACs across the whole S3DIS dataset for MinkowskiNet since it does not consume sliding window-based input as other methods. We observe that MinkowskiNet achieves best accuracy and efficiency on indoor semantic segmentation tasks while KPConv~\cite{thomas2019kpconv} is the best among point-based methods. 

\item \myparagraph{3D Outdoor Scene Segmentation.}
We summarize results of different outdoor scene segmentation methods in \tab{tab:3d:semantickitti:results}, where we report per-scene latency and \#MACs statistics. Due to the large spatial size of outdoor LiDAR scans, point-based methods are usually very inefficient since they have to inference on large number of sliding windows. Methods based on sparse convolution~\cite{choy20194d,tang2020searching} have the best accuracy and efficiency among all methods. SPVNAS~\cite{tang2020searching} applies neural architecture search and achieves $3\times$ measured speedup and $8\times$ computation reduction over MinkowskiNet~\cite{choy20194d}.
\end{itemize}

\myparagraph{Neural Architecture Search for Point Clouds.}
Apart from point convolution operation design, neural architecture design also plays an important role in efficient point cloud processing. Early research~\cite{qi2017pointnet++,thomas2019kpconv,choy20194d} takes advantage of existing image CNN designs (\eg residual connections, U-Net structure) to manually assemble point cloud convolution layers. However, these manually designed networks are often inefficient and suboptimal. Consequently, recent research starts to focus on automated neural architecture design for efficient point cloud processing.

V-NAS~\cite{zhu2019vnas} is tailored for volumetric representation. It leverages differentiable neural architecture search to explore a hybrid design space composed of 2D, 3D and pseudo-3D convolution operations. However, V-NAS is targeted for medical image segmentation and cannot be directly applied to 3D point cloud processing. For point-based methods, SGAS~\cite{li2020sgas} explores a DARTS~\cite{liu2019darts}-like network topology with graph convolution layers as candidate operations in the search space. Built upon SGAS, LC-NAS~\cite{li2020lcnas} further incorporates the hardware feedback into the pipeline by training a differentiable latency regressor to predict the network latency on target platforms. The predicted latency acts as a regularization term and soft constraint during training. SGAS and LC-NAS are currently designed for 3D object classification or point cloud part segmentation and have not demonstrated their effectiveness in large-scale outdoor scene understanding.

Recently, Tang \etal~\cite{tang2020searching,liu2021pvnas} propose 3D-NAS framework as a general tool to automatically design point cloud processing networks under resource constraints. The 3D-NAS framework supports different candidate networks with fine grained channel numbers, elastic network depth and resolution, allowing the \#MACs of the subnets to span over a 16$\times$ range. To support such a large range of models, 3D-NAS follows a two-stage pipeline, which is similar to ProxylessNAS~\cite{cai2019proxylessnas} and OFA~\cite{cai2020once}. In the first stage, a super network that contains all the subnets in the design space is trained. Thanks to the weight sharing, heterogenous sampling and progressive depth shrinkage techniques, the subnets are able to achieve the same level of accuracy comparing with training from scratch. Subsequently, a biology-inspired evolutionary architecture search process is performed to derive the best candidate network under efficiency constraints (\eg, \#MACs, latency). Different from LC-NAS, 3D-NAS enforces hard efficiency constraints, where candidates exceeding the computation budget are directly discarded. 3D-NAS can be applied on both 3D object segmentation and large-scale LiDAR scene processing, and consistently achieves significantly better performance-latency tradeoff than the manually-designed ones.

%% file: figText/fig_3d_primitives.tex
\begin{figure}[t]
    \centering
    \begin{subfigure}{0.48\textwidth}
        \centering
        \includegraphics[width=\linewidth]{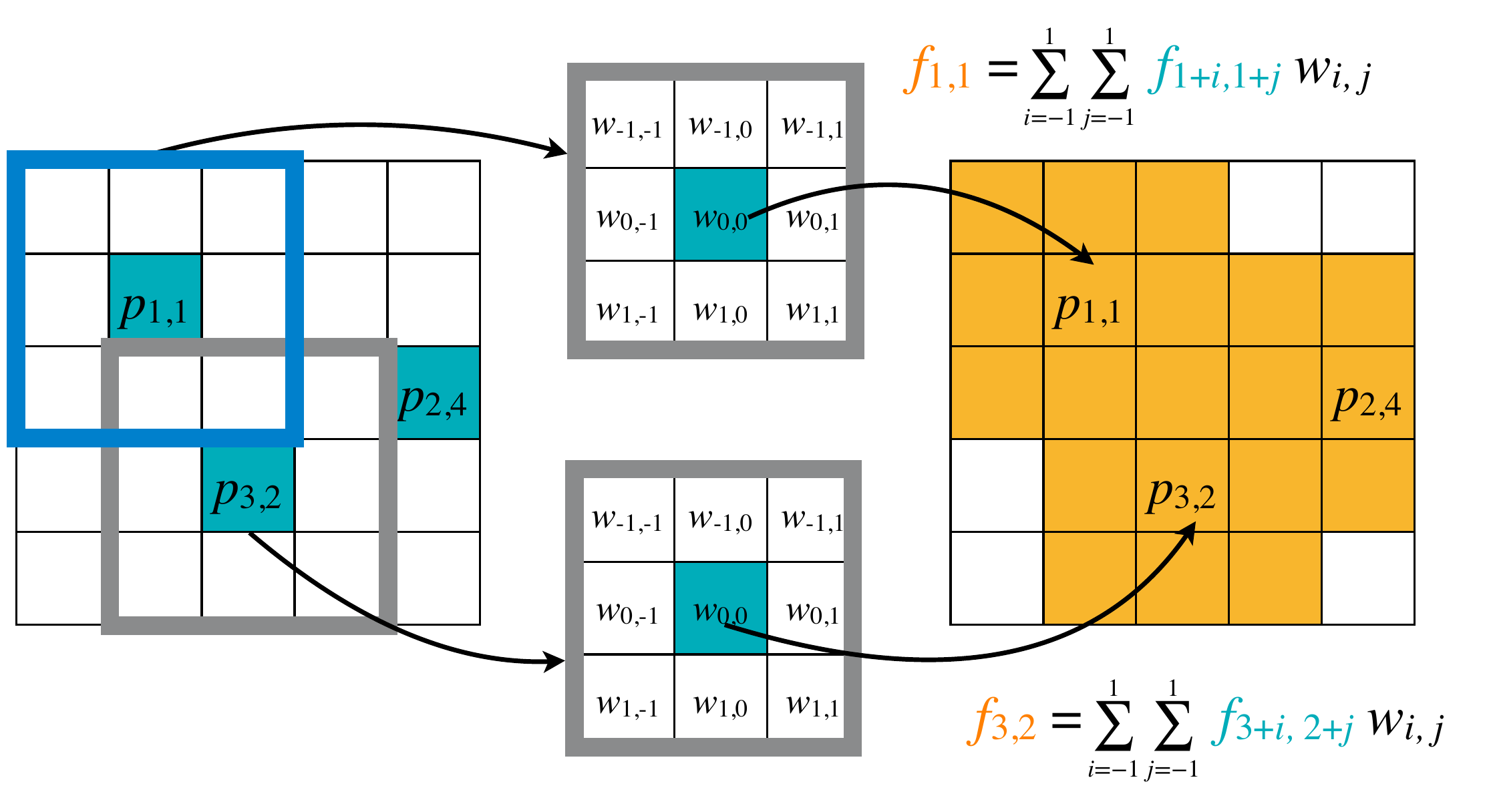}
        \caption{Voxel-based convolution~\cite{maturana2015voxnet}.}
        \label{fig:3d_primitives:voxel}
    \end{subfigure}
    \begin{subfigure}{0.48\textwidth}
        \centering
        \includegraphics[width=\linewidth]{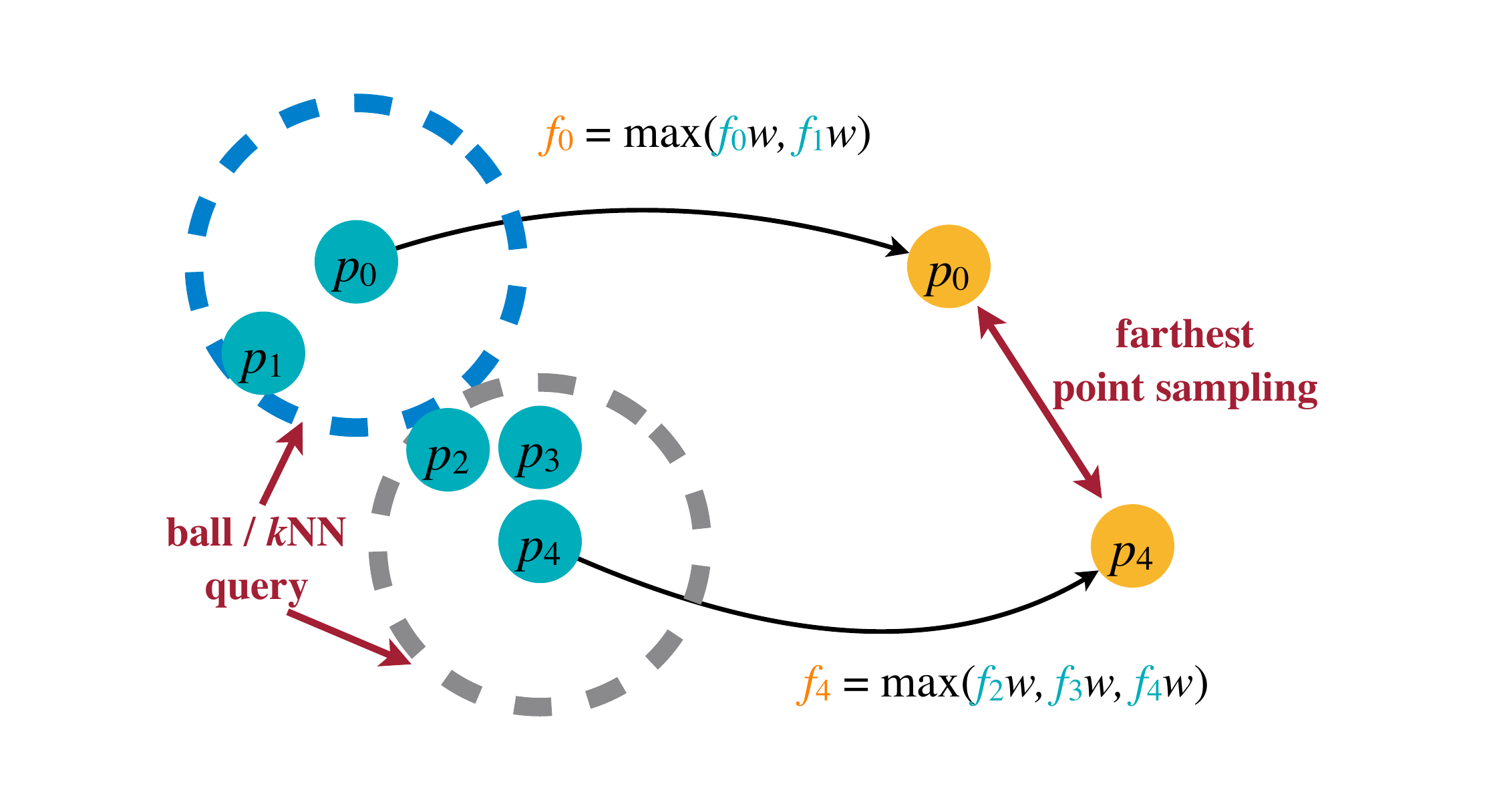}
        \caption{Point-based convolution~\cite{qi2017pointnet++,li2018pointcnn,thomas2019kpconv}.}
        \label{fig:3d_primitives:point}
    \end{subfigure}
    \begin{subfigure}{0.48\textwidth}
        \centering
        \includegraphics[width=\linewidth]{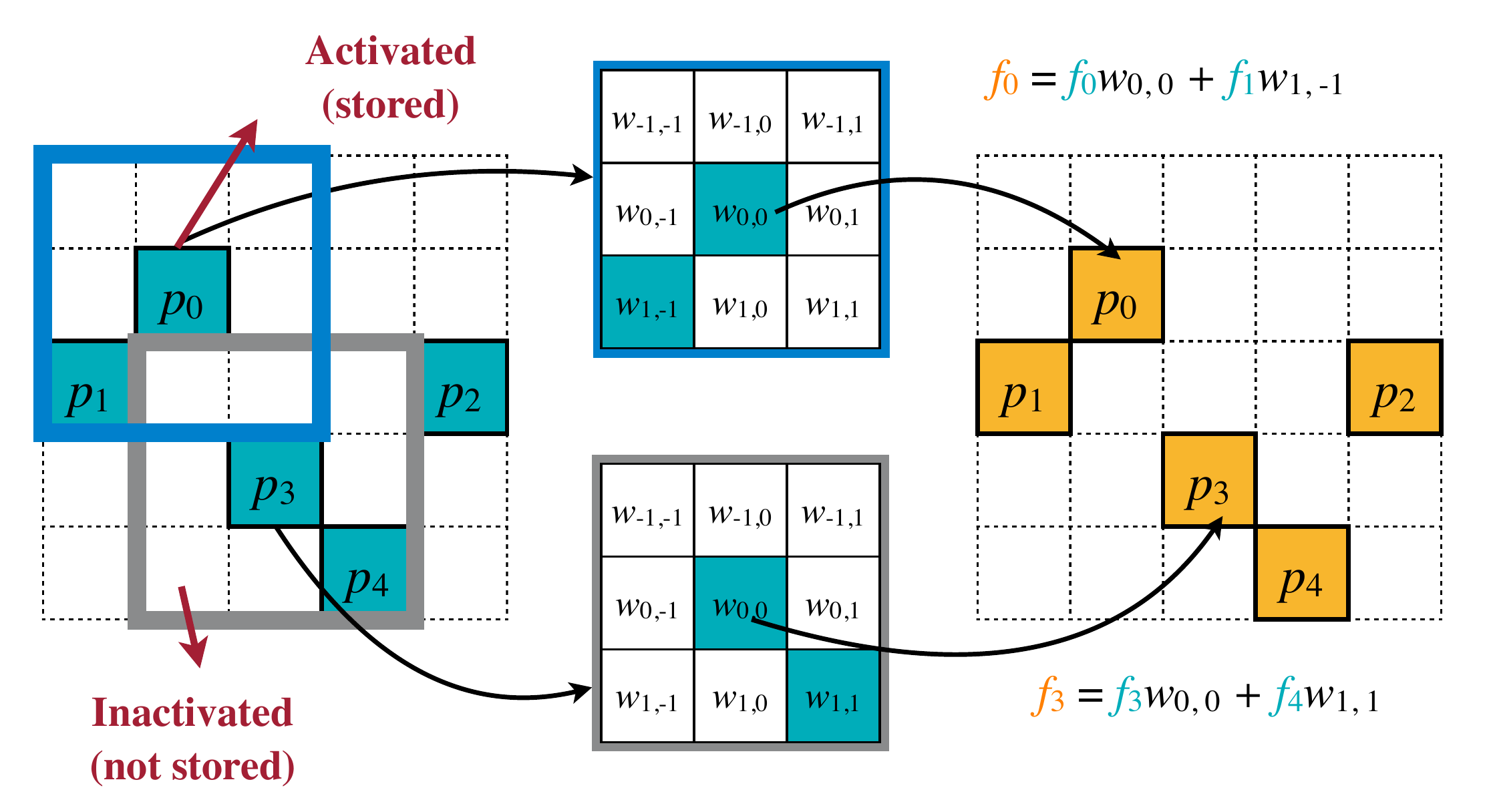}
        \caption{Sparse convolution~\cite{graham2015sparse,graham20183d,choy20194d}.}
        \label{fig:3d_efficient_primitives:sparseconv}
    \end{subfigure}
    \begin{subfigure}{0.48\textwidth}
        \centering
        \includegraphics[width=\linewidth]{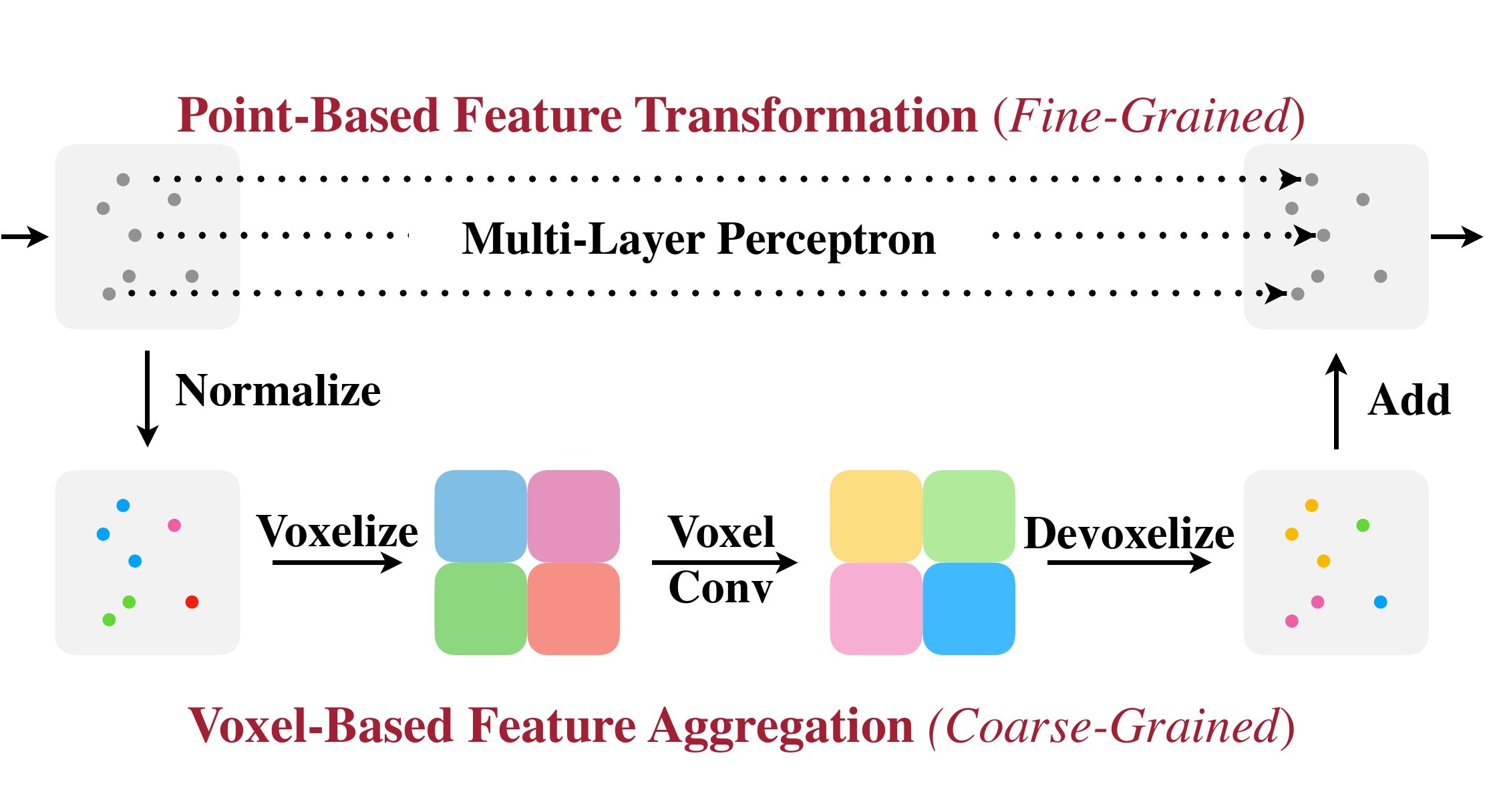}
        \caption{Point-voxel convolution~\cite{liu2019point,tang2020searching,shi2019pvrcnn}.}
        \label{fig:3d_efficient_primitives:pvconv}
    \end{subfigure}
    \caption{Overview of different 3D point cloud convolutions, where (a) and (b) are conventional approaches while (c) and (d) are emerging efficient approaches.}
    \label{fig:3d_primitives}
\end{figure}

%% file: figText/tab_3d_shape.tex
\begin{table}[t]
\setlength{\tabcolsep}{10pt}
\small\centering
\begin{tabular}{lcccc}
\toprule
& \#Params (M) & \#MACs (G) & GPU Latency (ms) & Mean IoU (\%) \\
\midrule
3D-UNet~\cite{cicek20163d}      & 8.1          & 2996.9 & 682.1        & 84.6 \\
\midrule

PointNet~\cite{qi2017pointnet}     & 2.5          & 10.3                        & 21.7         & 83.7 \\
PointNet++~\cite{qi2017pointnet++}   & 1.8          & 4.9                         & 77.9         & 85.1 \\
RS-Net~\cite{huang2018recurrent}       & 6.9          & 1.4                         & 74.6         & 84.9 \\
DGCNN~\cite{wang2018dynamic}        & 1.5          & 18.5                        & 87.8         & 85.1 \\
SpiderCNN~\cite{xu2018spidercnn}    & 2.6          & 10.6                        & 170.7        & 85.3 \\
PointConv~\cite{wu2019pointconv}    & 21.6         & 11.6                        & 163.7        & 85.7 \\
PointCNN~\cite{li2018pointcnn}     & 8.3          & 26.9                        & 135.8        & 86.2 \\
KPConv~\cite{thomas2019kpconv}       & 14.2         & --                           & 56.8         & 86.4 \\
\midrule
MinkowskiNet~\cite{choy20194d} & 21.7         & --                           & --            & 85.1 \\
PVCNN~\cite{liu2019point}        & 4.2          & 15.3                        & 50.7         & 86.2 \\
\bottomrule
\end{tabular}
\caption{Summarized results of 3D object segmentation on the ShapeNet dataset. In this table, the latency is measured on 8$\times$2048 input points with a single NVIDIA GTX1080Ti GPU, and the number of MACs is estimated based on 2048 input points.}
\label{tab:3d:shapenet:results}
\end{table}

%% file: figText/tab_3d_indoor.tex
\begin{table*}[!t]
\setlength{\tabcolsep}{6pt}
\small
\centering
\begin{tabular}{cccccccc}
\toprule
& \multirow{2.5}{*}{\#Params (M)} & \multirow{2.5}{*}{\#MACs (G)} & \multirow{2.5}{*}{Latency (ms)} & \multicolumn{2}{c}{S3DIS} & \multicolumn{2}{c}{ScanNet} \\
\cmidrule(lr){5-6}\cmidrule(lr){7-8}
&                               &                             &                               & mAcc        & mIoU        & mAcc         & mIoU         \\
\midrule
3D-UNet~\cite{cicek20163d}         & 14                            & 2796.8                      & 574.7                         & 86.1        & 54.9        & --            & --            \\
MinkowskiNet~\cite{choy20194d}     & 21.7                          & 28.2                        & 61.3                          & --           & 65.4        & --            & 73.6         \\
\midrule
PointNet~\cite{qi2017pointnet}     & 1.2                           & 28.8                        & 20.9                          & 82.5        & 43.0        & --            & --            \\
PointNet++~\cite{qi2017pointnet++} & 1.0                           & 9.6                         & 26.8                          & --           & --           & 84.5         & 33.9         \\
RS-Net~\cite{huang2018recurrent}   & 6.9                           & 17.6                        & 111.5                         & --           & 51.9        & 84.9         & 34.9         \\
DGCNN~\cite{wang2018dynamic}         & 1.0                           & 295.2                       & 178.1                         & 83.6        & 47.9        & --            & --            \\
PointConv~\cite{wu2019pointconv}   & 21.7                          & 112.0                       & 210.3                         & --           & --           & --            & 66.6         \\
PointCNN~\cite{li2018pointcnn}     & 11.5                          & 140.0                       & 282.3                         & 85.9        & 57.3        & 85.1         & 45.8         \\
KPConv~\cite{thomas2019kpconv}     & 14.1                          & 25.6                        & 71.3                          & --           & 67.1        & --            & 68.4         \\
\midrule
PVCNN~\cite{liu2019point}          & 2.6                           & 104.0                       & 47.3                          & 86.7        & 56.1        & --            & --            \\
PVCNN++~\cite{liu2019point}        & 13.7                          & 209.6                       & 69.5                          & 87.1        & 59.0        & --            & --            \\
\bottomrule
\end{tabular}
\caption{Summarized results of 3D indoor scene segmentation on S3DIS~\cite{armeni20163d,armeni2017joint} and ScanNet~\cite{dai2017scannet} datasets. In this table, the latency is measured with a single NVIDIA GTX1080Ti GPU. Both latency and \#MACs are estimated on the full scene for MinkowskiNet and on 8$\times$4096 points or equivalent size for all other methods.}
\label{tab:3d:indoor:results}
\end{table*}

%% file: figText/tab_3d_outdoor.tex
\begin{table}[t]
\setlength{\tabcolsep}{7pt}
\small\centering
\begin{tabular}{lcccc}
    \toprule
     & \#Params (M) & \#MACs (G) & GPU Latency (ms) & Mean IoU (\%) \\
    \midrule
    PointNet~\cite{qi2017pointnet} & 3.0$^{*}$ & -- & 500$^{*}$ & 14.6 \\
    SPGraph~\cite{landrieu2018large} & 0.3$^{*}$ & -- & 5200$^{*}$ & 17.4 \\
    PointNet++~\cite{qi2017pointnet++} & 6.0$^{*}$ & -- & 5900$^{*}$ & 20.1 \\
    PVCNN~\cite{liu2019point} & 2.5 & 42.4 & 146 & 39.0 \\
    PVCNN (sliding window)~\cite{liu2019point} & 2.5 & 42400 & 2500 & 56.4 \\
    TangentConv~\cite{tatarchenko2018tangent} & 0.4$^{*}$ & -- & 3000$^{*}$ & 40.9 \\
    RandLA-Net~\cite{hu2019randla} & 1.2 & 66.5 & 880 (\textcolor{red}{256}+\textcolor{blue}{624}) & 53.9 \\
    KPConv~\cite{thomas2019kpconv} & 18.3 & 207.3 & -- & 58.8 \\
    \midrule
    MinkowskiNet~\cite{choy20194d} & 21.7 & 114.0 & 294 & 63.1 \\
    SPVNAS~\cite{tang2020searching} & 2.6 & 15.0 & 110 & 63.7 \\
    SPVNAS~\cite{tang2020searching} & 12.5 & 73.8 & 259 & 66.4 \\
    \bottomrule
\end{tabular}
\caption{Summarized results of 3D outdoor scene segmentation on the SemanticKITTI~\cite{behley2019semantickitti} dataset. Here, \textcolor{red}{red numbers} correspond to the computation time, and \textcolor{blue}{blue numbers} correspond to the post-processing time. $^{*}$: results directly taken from Behley~\etal~\cite{behley2019semantickitti}.}
\label{tab:3d:semantickitti:results}
\end{table}

%% file: text/domain/video.tex
\subsection{Efficient Video Recognition}

Efficient video understanding is an important step towards real-world deployment, both on the cloud and on the edge. For example, there are over $10^5$ hours of videos uploaded to YouTube every day to be processed for recommendation and ads ranking; tera-bytes of sensitive videos in hospitals need to be processed locally on edge devices to protect privacy. All these industry applications require both accurate and efficient video understanding.

\myparagraph{2D \vs 3D CNN.}
Video recognition is different from image recognition, mainly due to the temporal information. 
Using the \textbf{2D CNN} on each of the video frames is a straightforward way to conduct video recognition~\cite{karpathy2014large, simonyan2014two, wang2016temporal, gan2015devnet, feichtenhofer2016spatiotemporal, feichtenhofer2016convolutional, bilen2016dynamic}. For example, Temporal Segment Networks (TSN)~\cite{wang2016temporal} extracted averaged features from strided sampled frames. Such methods are more efficient compared to 3D counterparts but cannot infer the temporal order or more complicated temporal relationships. 
\textbf{3D CNNs} can jointly learn spatio-temporal features~\cite{tran2015learning, carreira2017quo, szegedy2015going}. For example, Tran~\etal~\cite{tran2015learning} proposed a 3D CNN based on VGG models, named C3D, to learn spatio-temporal features from a frame sequence. 3D CNNs usually achieve superior action recognition performance given enough data. However, 3D CNNs are computationally heavy, making the deployment difficult. 

\myparagraph{RGB \vs Flow.}
Using optical flow as an extra input modality can improve the video recognition accuracy~\cite{zach2007duality}, which is also called ``two-stream'' method. For example, Simonyan~\etal~\cite{simonyan2014two} designed a two-stream CNN for RGB input (spatial stream) and optical flow~\cite{zach2007duality} input (temporal stream) respectively. Despite the higher accuracy, the optical flow is expensive to compute. Therefore, it is usually not used for efficient edge video processing. A recent work also tried to distill the optical flow stream information into the RGB stream during training~\cite{stroud2020d3d} so that it does not require the temporal stream during testing.


\myparagraph{Online \vs Offline.}
Most of the video understanding methods focus on offline video processing, where all the video frames are available and can be processed as a batch. On some deployment scenarios like smart camera, autonomous driving, \etc, it is required to perform online video understanding on a streaming video input to give low-latency feedback, usually on a per-input frame basis. TSM~\cite{lin2019tsm} can be used for online video recognition while still modeling temporal information. It does not incur duplicated computation unlike Zhu~\etal~\cite{zhu2017flow}.

\begin{figure}[t]
    \centering
    \includegraphics[width=\linewidth]{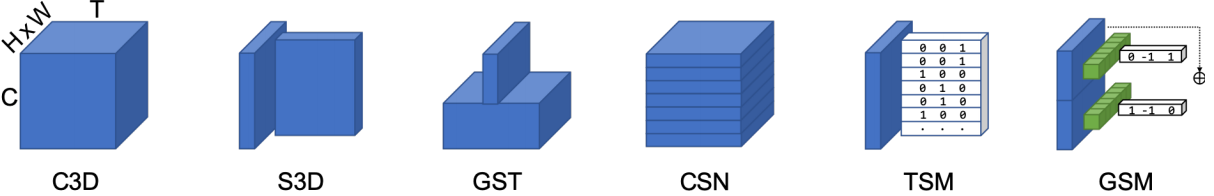}
    \caption{Spatio-temporal decomposition of 3D convolutions for efficient video recognition~\cite{sudhakaran2020gate}.}
    \label{fig:st_conv}
\end{figure}

\vspace{6pt}\noindent
Using 3D convolutions for spatio-temporal modeling can be redundant~\cite{lin2019tsm}. To leverage the temporal redundancy and improve the efficiency, researchers have introduced efficient primitives to reduce the cost of spatio-temporal modeling by \emph{decomposing} the spatial and temporal modeling. One method is to decompose the spatial and temporal dimension of the 3D convolutional kernels (see \fig{fig:st_conv}). P3D~\cite{qiu2017learning}, R(2+1)D~\cite{tran2018closer}, and S3D~\cite{xie2018rethinking} decompose the 3D kernels into a 2D spatial convolution and a 1D temporal convolution, which preserves the accuracy well. Another design is to separate the channel interaction and spatio-temporal interaction with group convolutions and depthwise convolutions~\cite{tran2019video}, or to decompose the feature channels into spatial and temporal groups in parallel with group convolution~\cite{luo2019grouped}. Temporal Shift Module (TSM)~\cite{lin2019tsm} demonstrates that the temporal modeling part can be performed with a hardware-efficient address shift without incurring extra computation.  Gate-Shift Module (GST)~\cite{sudhakaran2020gate} replaces the hard-wired channel split in TSM with a learnable spatial gating block to further improve the accuracy. Depthwise Temporal Aggregation Module (TAM)~\cite{fan2019more} enables the exchange of temporal information between frames by weighted channel-wise aggregation. The efficient primitives with spatial-temporal decomposition greatly improves the accuracy \vs speed/computation trade-off. TSM models greatly outperform 3D CNNs (I3D) and an efficient mixed 2D/3D design (ECO) at the same computation per video.
The efficiency of the model design also improves the training scalability significantly. Lin~\etal~\cite{lin2019training} scale up the training to 1,536 GPUs, finishing the large-scale training on Kinetics~\cite{carreira2017quo} in 1 hour.

\myparagraph{Neural Architecture Search for Videos.}
Neural architecture search has been employed to improve the efficiency of video understanding networks.
EvaNet~\cite{piergiovanni2019evolving} is the first attempt to perform neural architecture search for video architectures. AssembleNet~\cite{ryoo2020assemblenet} uses NAS to design new methods for fusing different sub-networks with different input modalities (RGB and optical flow) and temporal resolutions. 
Tiny Video Networks~\cite{piergiovanni2019tiny} automatically design highly efficient models for video understanding. X3D~\cite{feichtenhofer2020x3d} progressively expands a tiny 2D image classification architecture along multiple network axes, in space,
time, width and depth, to get a family of efficient video networks. Recently, Wang~\etal~\cite{wang2020pv} propose Practical Video Neural Architecture Search (PV-NAS) to efficiently search across a tremendously large scale of architectures in a novel spatial-temporal network search space using the gradient based search methods.

%% file: text/domain/nlp.tex
\subsection{Efficient Natural Language Processing}
\label{sec:nlp}


Natural Language Processing (NLP) is the key technique for numerous real-world applications, including machine translation, document summarization, and chatbots. Tradition NLP models are based on Recurrently Neural Network (RNN) and Convolutional Neural Network (CNN)~\cite{kim2014convolutional, NIPS2015_5782, hu2014convolutional, mikolov2010recurrent, liu2014recursive, sutskever2014sequence}. Lately, the NLP area is witnessing much faster advancements by virtue of the invention of the \emph{attention mechanism}~\cite{bahdanau2015neural, vaswani2017attention}. Attention-based NN models such as Transformer~\cite{vaswani2017attention}, BERT~\cite{devlin2018bert}, GPT-2~\cite{radford2019language}, GPT-3~\cite{brown2020language}, and Switch Transformer~\cite{fedus2021switch} provide significant performance improvements over models based on CNN and RNN. BERT~\cite{devlin2018bert} even outperforms human performance on the challenging question answering~\cite{rajpurkar2016squad} and sentence classification~\cite{wang2018glue} tasks.
To pursue higher performance, the sizes of recent attention-based models are increasing exponentially. The exploding model size and computation complexity bring severe efficiency issues, making it extremely challenging to deploy NLP models on resource-limited edge devices. For instance, in order to translate a sentence with only 30 tokens, a Transformer-Big model needs to execute 13G MACs and takes 20 seconds on a Raspberry Pi. Such long latency will hurt the user experience and make real-time NLP applications impossible on mobile devices. Therefore, efficient NLP techniques are of pressing demand.

\input{figText/fig_nlp_trend}

NLP tasks can be categorized into two types: discriminative and generative. For discriminative ones, the models need to summarize the input information and make predictions. Discriminative tasks include token-level classification, sentence-level classification, and regression. Meanwhile, models for generative tasks need to summarize the input information firstly and then generate new tokens. Examples of generation tasks include Language Modeling (LM)~\cite{radford2019language}, machine translation~\cite{vaswani2017attention} and text summarization~\cite{wu2019pay}.
Figure~\ref{fig:nlp_background} illustrates BERT (for discriminative tasks) and GPT-2 and Transformer (for generative tasks). BERT only contains the summarization stage, while GPT-2 and Transformer models first perform summarization and then generation stage.

In the summarization stage (Figure~\ref{fig:nlp_background} left), the input tokens are firstly embedded into vectors and processed by blocks. Inside each block, \texttt{block\_in} is firstly multiplied with three matrices to get Query (Q), Key (K), and Value (V). Then Q, K, and V are processed by attention to obtain the intermediate features \texttt{attention\_out}. A residual layer adds the \texttt{attention\_out} with \texttt{block\_in} and performs layer normalization. Furthermore, a Feed-Forward Network (FFN) layer containing two Fully-Connected (FC) layers is applied. Finally, another residual operation is conducted and outputs \texttt{block\_out}.
The same block is repeated multiple times, such as 12 times for BERT-Base. The last block is followed by one classification layer in BERT to get the \emph{final result}. In contrast, GPT-2 applies an LM head to generate one new token and then enter the generation stage.

\input{figText/fig_nlp_background}

The generation stage (Figure~\ref{fig:nlp_background} right) has two main differences from the summarization stage: (i) Each iteration only processes \emph{one single token} instead of the whole sentence. (ii) Ks and Vs from the summarization stage are concatenated with current K and V, and sent to attention \emph{in batch}, while the query is still \emph{one single vector}. After the last block, another new token will be generated. The generation stage ends when the `end of sentence' token is generated, or the sentence length reaches a pre-defined limit. 

\input{algorithm/algo_attention}

The attention mechanism is shown in Algorithm~\ref{algo:attention}. In the summarization stage, K, Q, and V are matrices with the same dimension, while in the generation stage, Q is one single vector, and K, V are matrices. 
Attention has multiple \textit{heads}, each processing a chunk of K, Q, and V. Different heads capture various dependencies between tokens, some for long-term, and some for short-term. Inside each head, $\mathrm{Q} \times \mathrm{K}^\text{T} / \mathrm{sqrt}(D)$ gives attention scores, where $D$ is the dimension of K, Q and V in one head. The attention scores indicate whether too tokens are related. Intuitively, each token is looking for the most important tokens to it. After that, a row-wise softmax computes the attention probabilities. The exponential of softmax further enlarges the attention scores for highly-related token pairs. The feature of the head is then computed with prob $\times$ V. This step lets each token fetch information from their cared tokens. Finally, multiple heads are concatenated together as the attention output. If there is more than one head, an additional FC layer is applied to the attention output $O$.

\myparagraph{Efficient NLP Primitive.}
Although the attention mechanism is the current workhorse in mainstream NLP models, research progress has been made to improve attention or replace it with new primitive operations. 
DeFormer~\cite{cao2020deformer}, EdgeBert~\cite{tambe2020edgebert}, Block-wise self-attention~\cite{qiu2019blockwise}, and Parmar \etal~\cite{parmar2018image} restrict the size of attention receptive field to reduce cost. Moreover, \cite{bapna-etal-2018-training} explores to use learned linear combinations of encoder outputs as the decoder inputs. \cite{wang-etal-2019-learning-deep} also proposes to train deep Transformers by propagating multiple layers together in the encoder. The Hardware-Aware Transformer (HAT)~\cite{wang2020hat} proposes arbitrary encoder-decoder attention to break the information bottleneck between the encoder and decoder by allowing decoder layers to get information from arbitrary and multiple encoder layers. Heterogeneous layer is also proposed in HAT to make different encoder and decoder layers to have various architecture, such as different numbers of heads, embedding dimensions. 

Some researchers propose to replace the attention mechanism with other operations. Wu \etal~\cite{wu2019pay} introduce a convolution-based module to replace the attention. The benefit of using convolution is its better local feature extraction ability. Also, the computation scales linearly instead of quadratic growth in attention. They propose lightweight depth-wise separable convolution to reduce the model size and develop dynamic convolutions built on depth-wise convolution to make the kernel adaptive to the current input context. Wu \etal~\cite{wu2020lite} further this direction by proposing a hybrid Long-Short Range Attention (LSRA) in which one branch is convolution, and another is attention. By specializing convolution and attention branch for short-range and long-range feature extractions, it achieves better accuracy-efficiency trade-offs. Memory-compressed attention~\cite{liu2018generating} employs convolution layers to shrink the sentence length. Reformer~\cite{kitaev2019reformer} replaces dot-product attention with locality-sensitive hashing to reduce computation complexity, and replace standard residuals with reversible residual layers which reduces peak memory consumption because the residual information dimension is reduced compared to normal residual layers. Routing Transformer~\cite{roy2020efficient} leverages k-means clustering of the tokens. Set Transformer~\cite{lee2019set} is proposed to model interactions among elements in the input set with an auxiliary memory. Average Attention Network (ANN)~\cite{zhang2018accelerating} replaces the original dynamically computed attention weights with fixed average weights. 
SRU~\cite{lei2017simple} proposes recurrent units to replace attention.
Linformer~\cite{wang2020linformer} approximates the attention with low-rank matrix thus reducing attention complexity from $O(n^2)$ to $O(n)$. Reservoir Transformer~\cite{shen2021reservoir} interleaves non-linear reservoir layers with regular attention layers. DeeBert~\cite{xin2020deebert} explores early exit of BERT layers to reduce computation cost.  Another research stream focuses on applying non- or partially-autoregressive models to cut down the iteration number for decoding~\cite{gu2019levenshtein, akoury-etal-2019-syntactically, wei-etal-2019-imitation,  gu2018nonautoregressive} by generating more than one token in one model forward pass.

\myparagraph{Automated Compression and Neural Architecture Search for NLP.}
Besides new primitive operations, research efforts have been made towards automated model compression and neural architecture search for NLP models. The numerous FC layers in NLP models are promising for weight pruning and quantization. GOBO~\cite{zadeh2020gobo} proposes to compress BERT model down to 3 bits, thus significantly reducing DRAM access. Q-BERT~\cite{shen2020q} proposes a group-wise quantization with a Hessian-based mix-precision strategy. I-BERT~\cite{kim2021bert} proposes integer-only BERT. TernaryBERT~\cite{zhang2020ternarybert} and BinaryBERT~\cite{bai2020binarybert} further quantize the model down to Ternary (\{-1, 0, +1\}) and binary schemes. Tambe \etal~\cite{tambe2020algorithm} propose an adaptive floating-point data format in consideration of the large dynamic range of NLP models' weights. Weight pruning is also widely used in NLP model size reduction~\cite{gordon2020compressing, yan2020micronet, li2020train}. The major differences between them are the number of bits and the granularity of quantization. According to Bai~\etal~\cite{bai2020binarybert}, on different tasks, the BERT accuracy goes down from 3\% to 7\% when quantized to 4 bits, and another 4\% to 13\% when quantized to 2 bits.

Besides, NLP models have large opportunities for activation pruning and quantization because of the redundancy of human languages. Multiple on-the-fly pruning methods are proposed to reduce the redundancy by token/head pruning~\cite{wang2020spatten, goyal2020power, michel2019sixteen, voita2019analyzing, kim2020length, wang2020efficient}. Specifically, SpAtten~\cite{wang2020spatten} proposes \emph{cascade token/head pruning} to reduce both DRAM access and computation. It prunes the tokens according to cumulative token importance scores obtained by accumulating attention probabilities (indicators for token influence) across layers. The heads are pruned based on cumulative head importance scores, which are computed by accumulating each head's magnitude across layers. Longformer~\cite{beltagy2020longformer} and Sparse Transformer~\cite{child2019generating} also leverage activation pruning by sparsifying the attention connections -- only attending to one token from every several tokens. For quantization, SpAtten~\cite{wang2020spatten} proposes \emph{progressive quantization} for attention inputs. It quantizes more aggressively for attention with dominated attention probabilities and more conservatively for others. Concretely, it first fetches MSBs of attention inputs to compute the attention probabilities. If the maximum probability is smaller than a threshold, indicating the distribution is flat, it will fetch LSBs on-chip and recompute attention probabilities.

\input{table/tab_hat}

Besides pruning and quantization, research efforts have also been invested on the automatic search for efficient NLP model architecture. Considering different properties of various hardware platforms~\cite{cong2018understanding}, Hardware-Aware Transformer (HAT)~\cite{wang2020hat} is proposed. It finds that FLOPs cannot reflect the measured latency, and different hardware prefers different Transformer architecture. Therefore, a specialized NLP model for each hardware is necessary. Specifically, it first constructs a large design space with arbitrary encoder-decoder attention and heterogeneous layers. Then HAT trains a SuperTransformer that covers all candidates in the design space, and efficiently produces many SubTransformers with weight sharing.
Finally, it performs an evolutionary search with a hardware latency constraint to find a specialized SubTransformer dedicated to run fast on the target hardware. Table~\ref{tab:hat} shows the performance comparison between HAT and state-of-the-art models.
Similarly, AutoADR~\cite{chen2020autoadr} also searches for a model under certain hardware constraints (memory, latency). AdaBert~\cite{chen2020adabert} leverages differentiable supernet-based NAS and knowledge distillation to automatically compress BERT into task-adaptive small models for specific tasks. AutoEmb~\cite{zhao2020autoemb} searches for the embedding size according to the input popularity. AttentionNAS~\cite{zhu2020a3d} designs a large design space containing various operations to search for an efficient attention cell.

\subsection{Discussions and Future Directions}
In this section, we introduced efficient approaches for emerging applications such as 3D point cloud processing, video understanding and natural language processing. For efficient 3D deep learning, we believe there will be new research on hardware-aware automated point cloud network design for different tasks, such as 3D object detection and panoptic segmentation. System-algorithm codesign will be another important topic: current state-of-the-art implementations of 3D deep learning modules only consider optimizing single layer or single operation, which is usually model-agnostic and has large room for improvement. For efficient video understanding, new primitive design and automated architecture search will still play an important role in future research. Since video annotation is usually more expensive than image labeling, data efficient video understanding will also receive attention in the future. For efficient natural language processing, one future research direction can be One-For-All Transformer, in which we can train one super transformer and deploy it to various edge devices without large training cost. Since most of the current state-of-the-art models require pre-training on a large corpus, another future research direction can be increasing the efficiency of the pre-training stage such as incorporating sparsity and quantization. That may help reduce the prohibitive bar for model pre-training and cultivate more efficient backbone model architectures. Since the auto-regressive GPT-2 model spends most of the runtime in the generation stage, combining existing partial or non-autoregressive model architectures with attention operation has the potential to significantly reduce the latency. Finally, algorithm-hardware co-design across the stack is another promising methodology to address the challenge of efficient NLP.

%% file: figText/fig_nlp_trend.tex
    
    

%% file: figText/fig_nlp_background.tex
\begin{figure}[t]
    \centering
    \includegraphics[width=0.6\columnwidth]{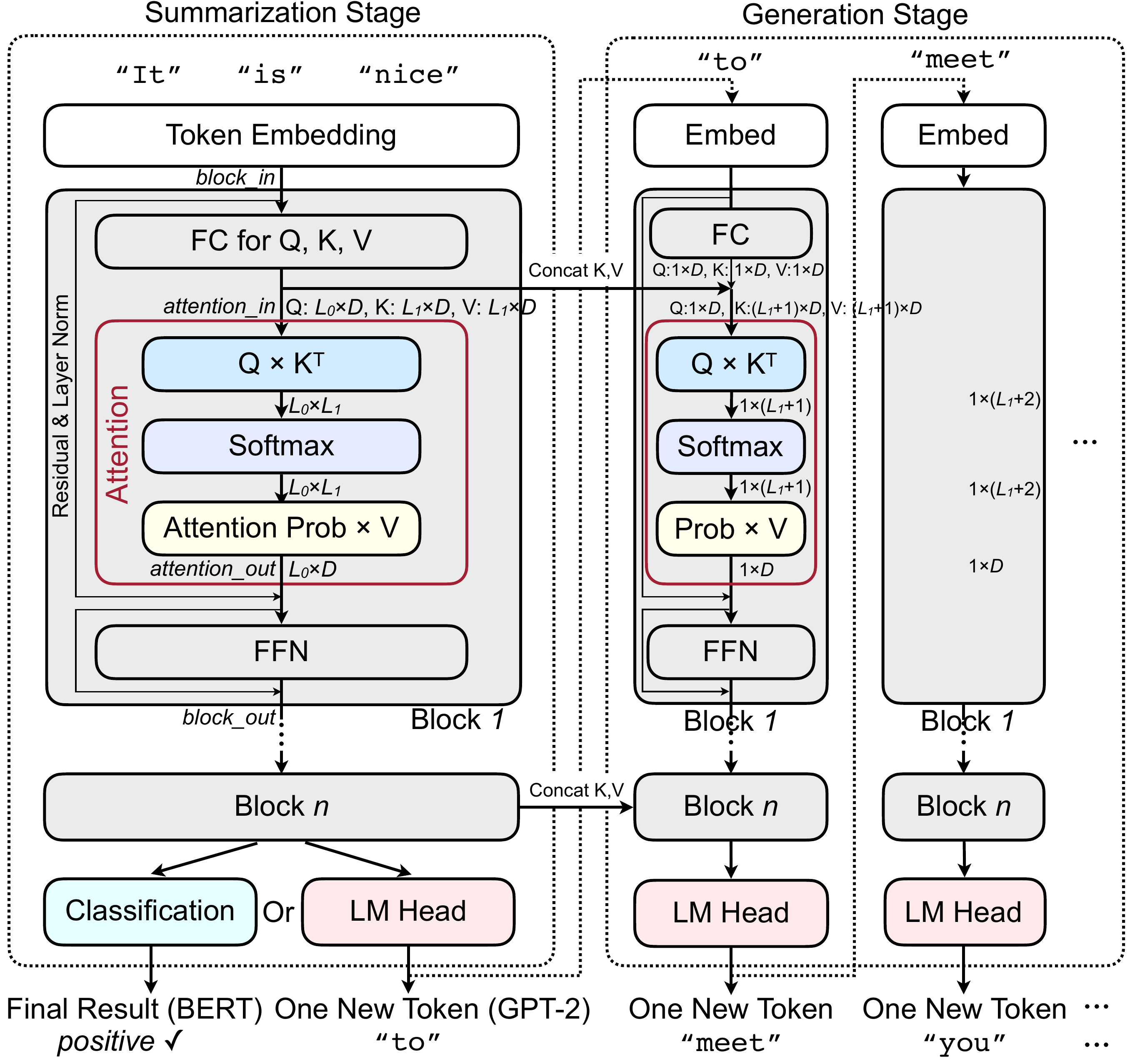}
    
    \caption{Attention-based NLP model architectures~\cite{wang2020spatten}. BERT only contains the summarization stage, while Transformer and GPT-2 contain both summarization and generation stages.}
    
    \label{fig:nlp_background}
\end{figure}

%% file: algorithm/algo_attention.tex
\begin{algorithm}[!t]
\setlength{\textfloatsep}{0pt}
\SetAlCapNameFnt{\footnotesize}
\SetAlCapFnt{\footnotesize}
\SetKwInOut{Input}{Input}
\footnotesize{
\SetAlgoLined
\textbf{Inputs:} 
query $Q_\text{in} \in \mathbb{R}^{L_0 \times D_\text{in}}$,
key $K_\text{in} \in \mathbb{R}^{L_1 \times D_\text{in}}$, 
value $V_\text{in} \in \mathbb{R}^{L_1 \times D_\text{in}}$, number of heads $h$ \\
~\\
Split $Q_\text{in}, K_\text{in}, V_\text{in}$ to $h$ chunks: $Q \in \mathbb{R}^{ h \times L_0 \times D}, K \in \mathbb{R}^{ h \times L_1 \times D}, V \in \mathbb{R}^{ h \times L_1 \times D}, \text{where}~D = D_\text{in}/h$ \\
\For{$i = 1 \mathbf{ \ to \ } h$}{

\colorbox{qxkcolor}{\vbox{ 
$\text{score} \in \mathbb{R}^{  L_0 \times L_1}$ \\ 
$\text{score} = Q_i \cdot K_i^\text{T} / \text{sqrt}(D)$ \\
}}

\colorbox{softmaxcolor}{\vbox{ 
$\text{prob} \in \mathbb{R}^{  L_0 \times L_1}$ \\
\For{$k = 1 \mathbf{ \ to \ } L_0$}{
$\text{prob}_k =  \text{softmax}(\text{score}_k)$
}
}}

\colorbox{probxvcolor}{\vbox{ 
$O_i = \text{prob} \cdot V_i$ \\
}}
}

Concatenate $\{O_i \mid 1 \leq i \leq h\}$ as output $O \in \mathbb{R}^{ h \times L_0 \times D}$
}
\caption{Attention\label{algo:attention}}
\end{algorithm}

%% file: table/tab_hat.tex
\begin{table*}[t]
\setlength{\tabcolsep}{3pt}
\small\centering
\resizebox{1\linewidth}{!}{
\begin{tabular}{clccccccccc}
\toprule
& & Latency (s) & \#Params (M) & FLOPs (G) & BLEU & GPU Hours & CO\textsubscript{2}e (lbs) & Cloud Comp. Cost \\
\midrule
\multirow{2}{*}{\shortstack[c]{IWSLT'14\\De-En}} & Transformer~\cite{vaswani2017attention} & 3.3 & 32 & 1.5  & 34.5 & 2 & 5 & \$12 - \$40  \\
& HAT~\cite{wang2020hardware} & 2.1 & 23 & 1.1 & 34.5 & 4 & 9 & \$24 - \$80  \\
\midrule
\multirow{4}{*}{\shortstack[c]{WMT'14\\En-Fr}} & Transformer~\cite{vaswani2017attention} & 23.2 & 176 & 10.6 & 41.2 & 240 & 68 & \$178 - \$595 \\
& ET~\cite{so2019evolved} & 20.9 & 175 & 10.8 & 41.3 & 2,192,000 & 626,000 & \$1.6M - \$5.5M  \\
& HAT~\cite{wang2020hardware} & 7.8 & 48 & 3.4 & 41.4 & 216 & 61 & \$159 - \$534  \\
& HAT~\cite{wang2020hardware} & 9.1 & 57 & 3.9 & 41.8 & 224 & 64 & \$166 - \$555  \\
\midrule
\multirow{4}{*}{\shortstack[c]{WMT'14\\En-De}} & Transformer~\cite{vaswani2017attention} & 20.5 & 176 & 10.6 & 28.4 & 184 & 52 & \$136 - \$456  \\
& ET~\cite{so2019evolved} & 7.6 & 47 & 2.9 & 28.2 & 2,192,000 & 626,000 & \$1.6M - \$5.5M  \\
& HAT~\cite{wang2020hardware} & 6.0 & 44 & 2.7 & 28.2 & 184 & 52 & \$136 - \$456  \\
& HAT~\cite{wang2020hardware} & 6.9 & 48 & 3.0 & 28.4 & 200 & 57 & \$147 - \$495 \\
\bottomrule
\end{tabular}
}
\caption{Summarized results of Transformer~\cite{vaswani2017attention}, Evolved Transformer~\cite{so2019evolved} and HAT~\cite{wang2020hardware}. In this table, the latency is measured on the Raspberry Pi ARM CPU, the training cost is estimated with a single NVIDIA V100 GPU, and the CO\textsubscript{2} emission and the cloud computing cost are computed following Strubell~\etal~\cite{strubell2019energy}.}
\label{tab:hat}
\end{table*}

%% file: text/system.tex
\section{Efficient System Design}
\label{sect:system}
Approaches introduced in the previous sections are top-down solutions for efficient deep learning processing, where performance gains come from the algorithmic optimization. However, these \textit{theoretical} benefits (\eg, FLOPs and model size) cannot be easily converted into the \textit{real} improvements in measured speedup and energy efficiency. Therefore, specialized software/hardware systems are required to bridge the gap. On the other hand, these specialized software/hardware systems open up a new design space orthogonal to the algorithm space, bringing the opportunities for holistic optimization by jointly optimizing the algorithm and software/hardware systems.
\subsection{Software System}

Deep learning software systems include general training \& inference libraries (PyTorch~\cite{paszke2019pytorch}, TensorFlow~\cite{abadi2016tensorflow} and MXNet~\cite{chen2015mxnet}), hardware-acceleration libraries (cuDNN~\cite{chetlur2014cudnn} and MKLDNN), and model serving libraries (TVM~\cite{chen2018tvm} and MNN~\cite{alibaba2020mnn}).

Researchers have been developing specialized deep learning systems, some with \emph{learning-based} methods for optimization. TVM~\cite{chen2018tvm} is a compiler that for deep learning workloads across diverse hardware backends. It  automates optimization of low-level programs to hardware characteristics with a learning-based method. 
FlexFlow~\cite{jia2018beyond} introduces functional-preserving graph transformations to optimize DNN executions. TASO~\cite{jia2019taso} further introduces an automated generation of substitution rules. MetaFlow and TASO can take the whole graph into consideration and search for high optimized substitution strategies and achieve up to 3x speedup over existing DNN frameworks. 
For high-end server accelerators,  IOS~\cite{ding2020ios} explores the inter operator parallelism beyond intra operator ones, as a single operator can no longer fully utilize the accelerator because of hardware advances. For low-end micro-controllers, TinyEngine~\cite{lin2020mcunet} generates model-specific memory scheduling for deep learning inference on memory-constrained microcontrollers.

Recently, there are also emerging domain-specific software systems. For example, in point cloud processing, Kaolin~\cite{jatavallabhula2019kaolin} collects implementations of state-of-the-art point cloud operators. PyTorch3D~\cite{ravi2020accelerating} features both point-based 3D module acceleration and efficient differentiable rendering. PyTorch3D accelerates implementation for $D$-dimensional $k$-NN with specialized tuning for different $(D, k)$ pairs. Besides, graph convolution operations are improved in memory efficiency in PyTorch3D through fusing \texttt{gather+scatter\_add} within a single CUDA kernel. SpConv~\cite{yan2018second}, MinkowskiEngine~\cite{choy20194d} and TorchSparse~\cite{tang2022torchsparse} are specialized for sparse tensor operations (especially sparse convolution) on 3D point clouds. These libraries differ in implementations of kernel map construction (\ie convolution rule generation) and convolution. SpConv implements kernel map construction with volumetric binary tensor lookup, and MinkowskiEngine features CPU parallel hashtable based solution. TorchSparse further optimizes the efficiency of kernel map construction with a GPU hashmap and parallel zero eliminator. For convolution implementation, MinkowskiEngine features direct computation while SpConv prefers the \texttt{Gather-MatMul-Scatter} dataflow. TorchSparse improves the memory efficiency of \texttt{Gather-MatMul-Scatter} dataflow while also provides support for direct computation of convolution. 

\subsection{Hardware System}

For most CPUs and GPUs, data for ALU can only be fetched from the memory hierarchy and cannot communicate directly with each other; thus the common approach to accelerate neural networks is vectorization (SIMD) and multi-threading (SIMT). Nonetheless, domain-specific accelerators are able to obtain additional performance and efficiency gain via four main techniques~\cite{dally2020domain}: data specialization, parallelism, local and optimized memory, and reduced overhead. Therefore, designing specialized hardware system is a popular bottom-up way for efficient deep learning processing.

\myparagraph{General Deep Learning Accelerators.}
General deep learning accelerators are mostly spatial architectures using dataflow processing~\cite{sze2017efficient} where ALUs form a processing chain so that they can pass data from one to another directly. Since the area, power as well as performance of most accelerators are dominated by memory, data handling characteristics, \ie, dataflow, can be used to classify the deep learning accelerators. Weight stationary dataflow is widely used~\cite{sankaradas2009massively, sriram2010towards, chakradhar2010dynamically, park20154, cavigelli2015origami} where the weights are store in the local memory and reused as much as possible. On the contrary, output stationary dataflow~\cite{gupta2015deep, du2015shidiannao, peemen2013memory} maximizes the reuse of output partial sums which are accumulated locally. Instead of maximizing the reuse of single type of data, row stationary dataflow~\cite{chen2017eyeriss, chen2019eyeriss, gao2017tetris, gupta2015deep} assigns 1-D row convolution to each PE in order to maximize the reuse of all data types (inputs, weights and outputs). Besides, no-local-reuse dataflow~\cite{zhang2015optimizing, chen2014diannao, chen2014dadiannao} increases the global buffer capacity and minimizes the off-chip memory bandwidth by eliminating the local memory. Lu~\etal~\cite{lu2020hardware} also present an NLP accelerator with support of efficient matrix partition and on-chip PE reuse. Many researchers also leverage specialized operators or devices for NLP acceleration. ATT~\cite{guo2020att} proposes a ReRAM-based architecture to support attention operations and adds non-uniform redundancy to improve accuracy. Park~\etal~\cite{park2020memory} propose an FPGA-based accelerator for the Question Answering task. RAMANN~\cite{ahmadzadeh2021a2p} designs a specialized 9T SRAM cell for efficient dot production in memory-augmented networks. 

\myparagraph{Hardware Support For Sparsity.}
Several works~\cite{park2016faster, wang2021sparsednn} propose to leverage existing general-purpose CPUs/GPUs to support sparsity in pruned NN models. Others propose domain-specific hardware accelerators which bring much high efficiency at the cost of longer design cycle and larger design automation burden~\cite{mao2019park, wang2020gcn, wang2018learning}. Compressing the sparse data is a straightforward way to save energy by reducing the data transfer cost. An example is Eyeriss~\cite{chen2017eyeriss}, which uses a run length encoding scheme to compress activations being transferred to/from DRAM and gates the multiplier for zero activations.

Skipping the multiplication whose operand is zero is another natural way to saving time. EIE~\cite{han2016eie} accelerates the sparse matrix-vector multiplication specifically for the fully-connected layers. It stores the weights in a CSC format along with the start location of each column. When the input activation is not zero, the compressed weight column is read and the corresponding output is updated as shown in \fig{fig:hardware_spmm}. For sparse convolution acceleration, Cnvlutin~\cite{albericio2016cnvlutin} similarly selects weights for multiplication based on only non-zero input activations while Cambricon-X~\cite{zhang2016cambricon} contrarily selects the input activations based on the non-zero weights. SCNN~\cite{parashar2017scnn} further supports convolution in a compressed format for both input activations and weights. It uses an input stationary dataflow to deliver the compressed weights and activations to a multiplier array, and performs Cartesian product in parallel to compute the partial sums followed by a scatter network to accumulate these
scattered partial sums as shown in \fig{fig:hardware_spmm}.
\input{figText/hardware/fig_spmm}

Generalized Sparse Matrix-Matrix Multiplication (SpGEMM) accelerators can also be used for sparse deep learning inference~\cite{pal2018outerspace, sparch, qin2020sigma}. SpArch~\cite{sparch} is a specialized accelerator for sparse-sparse matrix multiplication. It jointly optimizes the input and output matrix data reuse. SpArch first designs a highly parallelized \textit{streaming-based merger} to pipeline the multiplication and merge stage of partial matrices so that partial matrices are merged on chip immediately after produced, and then uses a \textit{condensed matrix representation} to reduce the number of partial matrices by three orders of magnitude.
SpArch further applies a \textit{Huffman tree scheduler} to improve the scalability of the merger for larger sparse matrices and uses a \textit{row prefetcher} with near-optimal buffer replacement policy to deal with the increased input matrix read. A number of designs are also presented to accelerate sparse linear algebra on FPGA platforms. Zhuo~\etal~\cite{zhuo2005sparse} introduce a tree of binary operators to increase the energy efficiency of SpMV on FPGAs. Jamro~\etal~\cite{jamro2015algorithms} prove that separating indices comparison and computing operations could increase the throughput. The index comparison and computing in the SpAtten system design are also separated. Zou~\etal~\cite{zou2013high} propose a new sparse matrix storage method called ``BVCSR'' to compress the indices of non-zero elements, thus increasing the valid bandwidth of FPGA. Elkurdi~\etal~\cite{elkurdi2008fpga} and Grigoracs~\etal~\cite{grigoracs2016optimising} propose an architecture for large-scale SpMV in the FEM problem. Elkurdi~\etal~\cite{elkurdi2008fpga} co-design an FPGA SpMV architecture with a matrix stripping and partitioning algorithms that enable the architecture to process arbitrarily large matrices without changing the PE quantities.

Apart from sparsity in weights and activations, some accelerators also explore the sparsity of query, key and value vectors in the attention layer. Pruning attention is fundamentally different from weight pruning because, as introduced in Section~\ref{sec:nlp}, there are no weights in the attention part.
$A^3$~\cite{ham20203} first sorts each dimension of the key vectors among all keys. Then it uses a pre-specified number of largest/smallest elements in the keys to conduct multiplications with a query and get \emph{partial} attention scores. The corresponding key will be pruned if a score is smaller than a threshold. MNNFast~\cite{jang2019mnnfast} removes V vectors whose attention probabilities are smaller than a threshold. SpAtten~\cite{wang2020spatten} proposes an accelerator architecture to support attention computation with specialized high-parallelism top-k engine for token/head selections, specialized memory hierarchy, and fully-pipelined datapath to translate theoretical savings to real speedup and energy reduction.

\input{table/tab_nlp_acc}

\myparagraph{Hardware Support For Quantization.}
Quantized models can reduce the model size and storage for deployment, but it requires hardware support on low-precision arithmetic for inference acceleration.
INT8 quantization is supported on mobile ARM CPUs (\eg, Qualcomm Hexagon, ARM Neon), x86 CPUs,  NVIDIA GPUs with TensorRT, and Xilinx FPGAs with DNNDK. NVIDIA's Turing architecture is able to further support INT4 inference, which brings an additional 59\% speedup compared to INT8. Since lower bit widths are less supported on existing hardware, specialized hardware implementations for binary and ternary neural networks have also studied~\cite{andri2016yodann, umuroglu2017finn, ando2017brein}. To achieve better accuracy \vs cost trade-off, 
hardware support for \emph{mixed-precision} quantization has been extensively explored. NVIDIA's Turing Tensor Core supports 1-bit, 4-bit, 8-bit and 16-bit arithmetic operations; Imagination launched a flexible neural network IP that supports per-layer bit-width adjustment for both weights and activations. Stripes~\cite{judd2016stripes} and UNPU~\cite{lee2018unpu} use bit-serial computation to support one mixed-precision operand (either inputs or weights). BISMO~\cite{umuroglu2018bismo} and Loom~\cite{sharify2018loom} further adopt temporal fusion to provide full bit flexibility.  Deeprecon~\cite{rzayev2017deeprecon} and BitFusion~\cite{sharma2018bit} dynamically compose and decompose 2-bit multipliers in space to construct 2/4/8-bit multiply-add units. SpAtten~\cite{wang2020spatten} develops hardware support for the progressive quantization technique to accelerate the memory-bounded NLP models (\eg, GPT-2).

\myparagraph{ML-Based Design and Optimization of Hardware.}
The surge in the demand for the computational resources for training and inferring the NN models leads to the wide usage of heterogeneous environment with a combination of many CPUs, GPUs and even FPGAs. The operations in NN are explicitly and manually placed onto the particular computing devices for model parallelism and data parallelism. Such approach does not scale well or produce the optimal results as neural networks become more and more complicated. Recent works exploit deep reinforcement learning to automatically optimize the hardware resource assignment for training and inference of neural networks given the growing diversity of hardware devices. 

A simplest setting is device placement optimization problem, which tries to map given neural networks to given hardware devices. Mirhoseini~\etal~\cite{mirhoseini2017device} use a sequence-to-sequence model to process the sequence of operations and output the placement for each operation. The execution time of each proposal is applied as a reward signal to train the proposal network. Mirhoseini~\etal~\cite{mirhoseini2018hierarchical} then propose an end-to-end solution with a two-level hierarchical model for proposal, the first model groups the operations in the compute graph and the second model places these groups onto devices, getting rid of manually group operations as a pre-processing step.

Jiang~\etal~\cite{jiang2020hardware} combine the FPGA device placement with neural architecture design by conducting device placement optimization and NAS iteratively: the best pipelined FPGA configuration is identified for the proposed neural architecture candidates, and then the superior network candidates are trained for controller update using policy gradient reinforcement learning. Kao~\etal~\cite{kao2020confuciux} further try to allocate the compute resource (such as \#PEs, buffer sizes) even inside single accelerator accordingly: apart from using reinforcement learning to propose a rough assignment, genetic algorithm is applied afterwards to finetune the proposal. 

These works only focus on using machine learning algorithms to allocate the hardware resource given the fixed hardware design, and the freedom in the hardware design, such as accelerator architecture and design parameters, is neglected.
\input{figText/hardware/tab_correlation}
Meanwhile, existing accelerators mostly target common neural architectures and do not reap the power of NAS. The design space of hardware and neural architectures deeply entangle with each other, as illustrated in Table~\ref{tab:hardware_correlation} where the correlations are complicated and vary from hardware to hardware. For instance, tiled convolution kernels should fit into the on-chip weight buffers in both accelerators, and the product of input channels and weight bitwidth should be multiples of the compute array \#rows in BitFusion. It is important to co-design the neural and hardware architectures by considering all the correlations and make them fit. Perfectly matched neural architecture and hardware architecture improve the utilization of the compute array and on-chip memory, maximizing efficiency and performance. 

A straightforward approach is to integrate NAS with designing hardware accelerator. NASAIC~\cite{yang2020co} narrows down the hardware design space to the selection of limited accelerator templates, and exploits the meta-controller similar to vanilla NAS~\cite{zoph2017neural} to predict the parameters of hardware resource allocation for different hardware template selections, achieving 17.77\%, 2.49$\times$ and 2.32$\times$ reduction in latency, energy and area with less than 1.6\% accuracy loss on CIFAR-scale datasets. 
NHAS~\cite{linneural} further expands the neural architecture design space to allow mixed-precision quantization, and expands the hardware architecture design space with hardware architectural sizing parameters including compute array size, input/weight/output buffer sizes and global buffer sizes. NHAS exploits the evolution algorithms to improve the sample efficiency during search, achieving 1.92$\times$, 1.79$\times$ speedup and 1.63$\times$, 1.49$\times$ energy savings without hurting the accuracy of ResNet-18 and ResNet-50, respectively.

These works mainly focus on sizing the architectural hyperparameters while neglect searching the PE connectivities and compiler mappings.
Neural Accelerator Architecture Search (NAAS)~\cite{lin2021enhcs} pushes beyond architectural sizing and searches PE connectivities and compiler mapping at the same time. NAAS models the PE connectivity as the choices of parallel dimensions in the computation loop nests, and proposes ``importance-based'' encoding method to encode \textit{non-numerical} parameters, such as indexing of parallelism in hardware optimization and ordering of for-loops in mapping optimization, into \textit{numerical} parameters. Furthermore, NAAS integrates the Once-For-All NAS algorithm in the optimization loop. NAAS achieves 4.88$\times$ energy-delay-product improvement in total as well as 2.7\% top-1 accuracy improvement on ImageNet dataset than Eyeriss running ResNet50, with a new hardware architecture design.

\subsubsection{Discussions and Future Directions.}

Designing deep learning hardware systems require expertise of both algorithm and hardware. Thus exploiting machine learning to design and optimize deep learning hardware is a heated topic. However, current machine-learning-based hardware architecture design frameworks focus on sizing the architectural hyperparameters while neglect either the lower level design such as PE connectivities or the mapping strategies compiled by software system. Therefore, an important future research direction is to free more design freedom in the hardware architecture design and jointly optimize hardware and software systems. Current works also heavily depend on the results of simulation instead of results after a complete hardware design cycle including synthesis and layout due to time limitation. Thus, another future research direction is to apply machine learning on each possible step in the hardware design cycle to automate the whole procedure and provide a better hardware solution than that guided by heuristic rules.

%% file: figText/hardware/fig_spmm.tex
\begin{figure}[t]
    \centering
    \includegraphics[width=0.8\linewidth]{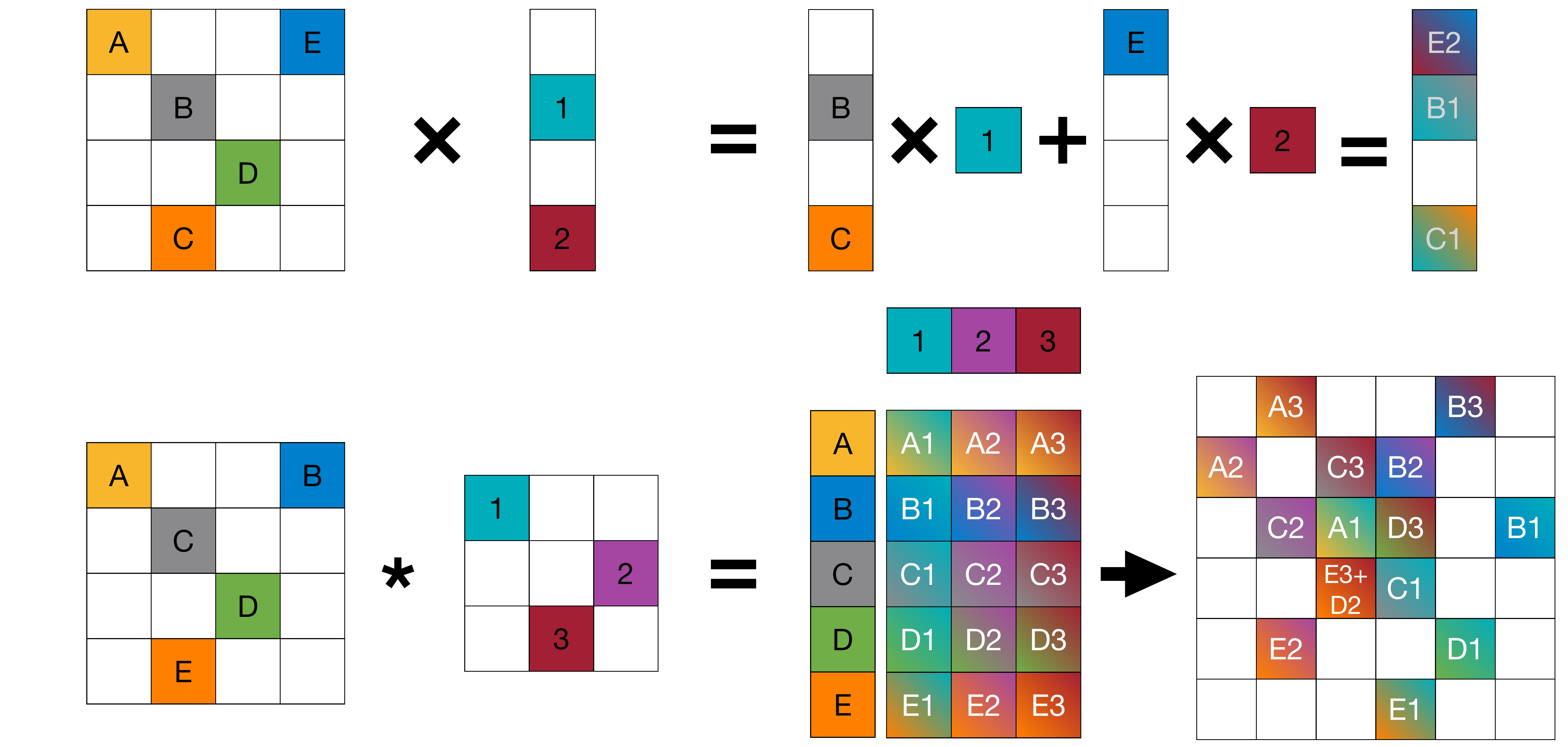}
    \caption{Upper: Sparse matrix-vector multiplication using compressed sparse column (CSC) format in EIE~\cite{han2016eie}. Lower: Sparse convolution using cartesian product in SCNN~\cite{parashar2017scnn}.}
    \label{fig:hardware_spmm}
\end{figure}

%% file: table/tab_nlp_acc.tex


%% file: figText/hardware/tab_correlation.tex
\begin{table}[t]
\small\centering
\resizebox{1\linewidth}{!}{
\begin{tabular}{c|c|c|c|c|c|c}
\toprule
\multirow{2.5}{*}{\shortstack[c]{Hardware Architecture\\Design Space}} & \multicolumn{6}{c}{\centering Neural Architecture Design Space} \\
\cmidrule{2-7}
& Input Channels & Output Channels & Kernel Size & Feature Map Size & Input Bits & Weight Bits \\
\midrule
Array \#rows & B &   & E &   & B & B \\
\midrule
Array \#cols  &   & B &   & E &   &   \\
\midrule
IBUF Size    & B/E &   &   & B/E & B/E &  \\
\midrule
WBUF Size    & B/E & B/E & B/E &   & B & B/E \\
\midrule
OBUF Size    &   & B/E &   & B/E &   &  \\
\bottomrule
\end{tabular}
}
\caption{Correlation between neural and hardware architecture design space~\cite{linneural}. This relationship differs from hardware to hardware (B: BitFusion~\cite{sharma2018bit} and E: Eyeriss~\cite{chen2017eyeriss}).}
\label{tab:hardware_correlation}%
\end{table}%

%% file: text/conclusion.tex
\section{Conclusion}

Over the past few years, deep neural networks have achieved unprecedented success in the field of artificial intelligence; however, their superior performance comes at the cost of high computational complexity. This limits their applications on many edge devices, where the hardware resources are tightly constrained by the form factor, battery and heat dissipation.

In this paper, we offer a systematic overview of efficient deep learning to enable both researchers and practitioners to quickly get started in this field. We first introduce various model compression approaches that have become the industry standards, such as pruning, factorization, quantization and efficient model design. To reduce the design cost of these handcrafted solutions, we then describe many recent efforts on neural architecture search, automated pruning and quantization, which can outperform the manual design with minimal human efforts. Apart from inference, we also cover the efficient on-device training to enable user customization based on the local data. We showcase several task-specific accelerations for point cloud, video and natural language processing by leveraging their spatial sparsity and temporal/token redundancy. Finally, we introduce the efficient software/hardware system to support all these algorithmic improvements.

Efficient deep learning is the key enabler for many real-world AI applications. As for the future direction, it is critical to explore the extreme of efficient deep learning, ranging from cloud AI to mobile and tiny AI. To achieve this goal, we will have to accelerate deep learning from all possible perspectives, from training to inference, and from software to hardware. Furthermore, it is promising to study the co-design of algorithm, software and hardware system, and exploit the unique properties of different domains for optimization.

%% file: main.bbl

\begin{thebibliography}{368}


\ifx \showCODEN    \undefined \def \showCODEN     #1{\unskip}     \fi
\ifx \showDOI      \undefined \def \showDOI       #1{#1}\fi
\ifx \showISBNx    \undefined \def \showISBNx     #1{\unskip}     \fi
\ifx \showISBNxiii \undefined \def \showISBNxiii  #1{\unskip}     \fi
\ifx \showISSN     \undefined \def \showISSN      #1{\unskip}     \fi
\ifx \showLCCN     \undefined \def \showLCCN      #1{\unskip}     \fi
\ifx \shownote     \undefined \def \shownote      #1{#1}          \fi
\ifx \showarticletitle \undefined \def \showarticletitle #1{#1}   \fi
\ifx \showURL      \undefined \def \showURL       {\relax}        \fi
\providecommand\bibfield[2]{#2}
\providecommand\bibinfo[2]{#2}
\providecommand\natexlab[1]{#1}
\providecommand\showeprint[2][]{arXiv:#2}

\bibitem[\protect\citeauthoryear{Abadi, Barham, Chen, Chen, Davis, Dean, Devin,
  Ghemawat, Irving, Isard, Isard, Kudlur, Levenberg, Monga, Moore, Murray,
  Steiner, Tucker, Vasudevan, Warden, Wicke, Yu, and Zheng}{Abadi
  et~al\mbox{.}}{2016}]%
        {abadi2016tensorflow}
\bibfield{author}{\bibinfo{person}{Mart{\'\i}n Abadi}, \bibinfo{person}{Paul
  Barham}, \bibinfo{person}{Jianmin Chen}, \bibinfo{person}{Zhifeng Chen},
  \bibinfo{person}{Andy Davis}, \bibinfo{person}{Jeffrey Dean},
  \bibinfo{person}{Matthieu Devin}, \bibinfo{person}{Sanjay Ghemawat},
  \bibinfo{person}{Geoffrey Irving}, \bibinfo{person}{Michael Isard},
  \bibinfo{person}{Michael Isard}, \bibinfo{person}{Manjunath Kudlur},
  \bibinfo{person}{Josh Levenberg}, \bibinfo{person}{Rajat Monga},
  \bibinfo{person}{Sherry Moore}, \bibinfo{person}{Derek~G Murray},
  \bibinfo{person}{Benoit Steiner}, \bibinfo{person}{Paul Tucker},
  \bibinfo{person}{Vijay Vasudevan}, \bibinfo{person}{Pete Warden},
  \bibinfo{person}{Martin Wicke}, \bibinfo{person}{Yuan Yu}, {and}
  \bibinfo{person}{Xiaoqiang Zheng}.} \bibinfo{year}{2016}\natexlab{}.
\newblock \showarticletitle{{TensorFlow: A System for Large-Scale Machine
  Learning}}. In \bibinfo{booktitle}{\emph{USENIX Symposium on Operating
  Systems Design and Implementation}}.
\newblock


\bibitem[\protect\citeauthoryear{Ahmadzadeh, Kamal, Afzali-Kusha, and
  Pedram}{Ahmadzadeh et~al\mbox{.}}{2021}]%
        {ahmadzadeh2021a2p}
\bibfield{author}{\bibinfo{person}{Mohsen Ahmadzadeh}, \bibinfo{person}{Mehdi
  Kamal}, \bibinfo{person}{Ali Afzali-Kusha}, {and} \bibinfo{person}{Massoud
  Pedram}.} \bibinfo{year}{2021}\natexlab{}.
\newblock \showarticletitle{{A2P-MANN: Adaptive Attention Inference Hops Pruned
  Memory-Augmented Neural Networks}}.
\newblock \bibinfo{journal}{\emph{arXiv preprint arXiv:2101.09693}}
  (\bibinfo{year}{2021}).
\newblock


\bibitem[\protect\citeauthoryear{Akoury, Krishna, and Iyyer}{Akoury
  et~al\mbox{.}}{2019}]%
        {akoury-etal-2019-syntactically}
\bibfield{author}{\bibinfo{person}{Nader Akoury}, \bibinfo{person}{Kalpesh
  Krishna}, {and} \bibinfo{person}{Mohit Iyyer}.}
  \bibinfo{year}{2019}\natexlab{}.
\newblock \showarticletitle{{Syntactically Supervised Transformers for Faster
  Neural Machine Translation}}. In \bibinfo{booktitle}{\emph{Conference of the
  Association for Computational Linguistics}}.
\newblock


\bibitem[\protect\citeauthoryear{Albericio, Judd, Hetherington, Aamodt, Jerger,
  and Moshovos}{Albericio et~al\mbox{.}}{2016}]%
        {albericio2016cnvlutin}
\bibfield{author}{\bibinfo{person}{Jorge Albericio}, \bibinfo{person}{Patrick
  Judd}, \bibinfo{person}{Tayler~H Hetherington}, \bibinfo{person}{Tor~M
  Aamodt}, \bibinfo{person}{Natalie D~Enright Jerger}, {and}
  \bibinfo{person}{Andreas Moshovos}.} \bibinfo{year}{2016}\natexlab{}.
\newblock \showarticletitle{{Cnvlutin: Ineffectual-Neuron-Free Deep Neural
  Network Computing}}. In \bibinfo{booktitle}{\emph{International Symposium on
  Computer Architecture}}.
\newblock


\bibitem[\protect\citeauthoryear{Amini, Rosman, Karaman, and Rus}{Amini
  et~al\mbox{.}}{2019}]%
        {amini2019variational}
\bibfield{author}{\bibinfo{person}{Alexander Amini}, \bibinfo{person}{Guy
  Rosman}, \bibinfo{person}{Sertac Karaman}, {and} \bibinfo{person}{Daniela
  Rus}.} \bibinfo{year}{2019}\natexlab{}.
\newblock \showarticletitle{{Variational End-to-End Navigation and
  Localization}}. In \bibinfo{booktitle}{\emph{IEEE International Conference on
  Robotics and Automation}}.
\newblock


\bibitem[\protect\citeauthoryear{Ando, Ueyoshi, Orimo, Yonekawa, Sato,
  Nakahara, Ikebe, Asai, Takamaeda-Yamazaki, Kuroda, and Motomur}{Ando
  et~al\mbox{.}}{2017}]%
        {ando2017brein}
\bibfield{author}{\bibinfo{person}{Kota Ando}, \bibinfo{person}{Kodai Ueyoshi},
  \bibinfo{person}{Kentaro Orimo}, \bibinfo{person}{Haruyoshi Yonekawa},
  \bibinfo{person}{Shimpei Sato}, \bibinfo{person}{Hiroki Nakahara},
  \bibinfo{person}{Masayuki Ikebe}, \bibinfo{person}{Tetsuya Asai},
  \bibinfo{person}{Shinya Takamaeda-Yamazaki}, \bibinfo{person}{Tadahiro
  Kuroda}, {and} \bibinfo{person}{Masato Motomur}.}
  \bibinfo{year}{2017}\natexlab{}.
\newblock \showarticletitle{{BRein Memory: A 13-Layer 4.2 K Neuron/0.8 M
  Synapse Binary/Ternary Reconfigurable In-Memory Deep Neural Network
  Accelerator in 65 nm CMOS}}. In \bibinfo{booktitle}{\emph{Symposium on VLSI
  Circuits}}.
\newblock


\bibitem[\protect\citeauthoryear{Andri, Cavigelli, Rossi, and Benini}{Andri
  et~al\mbox{.}}{2016}]%
        {andri2016yodann}
\bibfield{author}{\bibinfo{person}{Renzo Andri}, \bibinfo{person}{Lukas
  Cavigelli}, \bibinfo{person}{Davide Rossi}, {and} \bibinfo{person}{Luca
  Benini}.} \bibinfo{year}{2016}\natexlab{}.
\newblock \showarticletitle{{YodaNN: An Ultra-Low Power Convolutional Neural
  Network Accelerator Based on Binary Weights}}. In
  \bibinfo{booktitle}{\emph{IEEE Computer Society Annual Symposium on VLSI}}.
\newblock


\bibitem[\protect\citeauthoryear{Armeni, Sax, Zamir, and Savarese}{Armeni
  et~al\mbox{.}}{2017}]%
        {armeni2017joint}
\bibfield{author}{\bibinfo{person}{Iro Armeni}, \bibinfo{person}{Alexandar
  Sax}, \bibinfo{person}{Amir~R Zamir}, {and} \bibinfo{person}{Silvio
  Savarese}.} \bibinfo{year}{2017}\natexlab{}.
\newblock \showarticletitle{{Joint 2D-3D-Semantic Data for Indoor Scene
  Understanding}}.
\newblock \bibinfo{journal}{\emph{arXiv preprint arXiv:1702.01105}}
  (\bibinfo{year}{2017}).
\newblock


\bibitem[\protect\citeauthoryear{Armeni, Sener, Zamir, Jiang, Brilakis,
  Fischer, and Savarese}{Armeni et~al\mbox{.}}{2016}]%
        {armeni20163d}
\bibfield{author}{\bibinfo{person}{Iro Armeni}, \bibinfo{person}{Ozan Sener},
  \bibinfo{person}{Amir~R Zamir}, \bibinfo{person}{Helen Jiang},
  \bibinfo{person}{Ioannis Brilakis}, \bibinfo{person}{Martin Fischer}, {and}
  \bibinfo{person}{Silvio Savarese}.} \bibinfo{year}{2016}\natexlab{}.
\newblock \showarticletitle{{3D Semantic Parsing of Large-Scale Indoor
  Spaces}}. In \bibinfo{booktitle}{\emph{IEEE Conference on Computer Vision and
  Pattern Recognition}}.
\newblock


\bibitem[\protect\citeauthoryear{Bahdanau, Cho, and Bengio}{Bahdanau
  et~al\mbox{.}}{2015}]%
        {bahdanau2015neural}
\bibfield{author}{\bibinfo{person}{Dzmitry Bahdanau},
  \bibinfo{person}{Kyunghyun Cho}, {and} \bibinfo{person}{Yoshua Bengio}.}
  \bibinfo{year}{2015}\natexlab{}.
\newblock \showarticletitle{{Neural Machine Translation by Jointly Learning to
  Align and Translate}}. In \bibinfo{booktitle}{\emph{International Conference
  on Learning Representations}}.
\newblock


\bibitem[\protect\citeauthoryear{Bai, Zhang, Hou, Shang, Jin, Jiang, Liu, Lyu,
  and King}{Bai et~al\mbox{.}}{2020}]%
        {bai2020binarybert}
\bibfield{author}{\bibinfo{person}{Haoli Bai}, \bibinfo{person}{Wei Zhang},
  \bibinfo{person}{Lu Hou}, \bibinfo{person}{Lifeng Shang},
  \bibinfo{person}{Jing Jin}, \bibinfo{person}{Xin Jiang}, \bibinfo{person}{Qun
  Liu}, \bibinfo{person}{Michael Lyu}, {and} \bibinfo{person}{Irwin King}.}
  \bibinfo{year}{2020}\natexlab{}.
\newblock \showarticletitle{{BinaryBERT: Pushing the Limit of BERT
  Quantization}}. In \bibinfo{booktitle}{\emph{Conference of the Association
  for Computational Linguistics}}.
\newblock


\bibitem[\protect\citeauthoryear{Baker, Gupta, Naik, and Raskar}{Baker
  et~al\mbox{.}}{2017}]%
        {baker2017designing}
\bibfield{author}{\bibinfo{person}{Bowen Baker}, \bibinfo{person}{Otkrist
  Gupta}, \bibinfo{person}{Nikhil Naik}, {and} \bibinfo{person}{Ramesh
  Raskar}.} \bibinfo{year}{2017}\natexlab{}.
\newblock \showarticletitle{{Designing Neural Network Architectures using
  Reinforcement Learning}}. In \bibinfo{booktitle}{\emph{International
  Conference on Learning Representations}}.
\newblock


\bibitem[\protect\citeauthoryear{Banbury, Zhou, Fedorov, Navarro, Thakkar,
  Gope, Reddi, Mattina, and Whatmough}{Banbury et~al\mbox{.}}{2020}]%
        {banbury2020micronets}
\bibfield{author}{\bibinfo{person}{Colby Banbury}, \bibinfo{person}{Chuteng
  Zhou}, \bibinfo{person}{Igor Fedorov}, \bibinfo{person}{Ramon~Matas Navarro},
  \bibinfo{person}{Urmish Thakkar}, \bibinfo{person}{Dibakar Gope},
  \bibinfo{person}{Vijay~Janapa Reddi}, \bibinfo{person}{Matthew Mattina},
  {and} \bibinfo{person}{Paul~N Whatmough}.} \bibinfo{year}{2020}\natexlab{}.
\newblock \showarticletitle{{MicroNets: Neural Network Architectures for
  Deploying TinyML Applications on Commodity Microcontrollers}}. In
  \bibinfo{booktitle}{\emph{Conference on Machine Learning and Systems}}.
\newblock


\bibitem[\protect\citeauthoryear{Banner, Nahshan, and Soudry}{Banner
  et~al\mbox{.}}{2019}]%
        {banner2019post}
\bibfield{author}{\bibinfo{person}{Ron Banner}, \bibinfo{person}{Yury Nahshan},
  {and} \bibinfo{person}{Daniel Soudry}.} \bibinfo{year}{2019}\natexlab{}.
\newblock \showarticletitle{{Post Training 4-Bit Quantization of Convolutional
  Networks for Rapid-Deployment}}. In \bibinfo{booktitle}{\emph{Conference on
  Neural Information Processing Systems}}.
\newblock


\bibitem[\protect\citeauthoryear{Bapna, Chen, Firat, Cao, and Wu}{Bapna
  et~al\mbox{.}}{2018}]%
        {bapna-etal-2018-training}
\bibfield{author}{\bibinfo{person}{Ankur Bapna}, \bibinfo{person}{Mia Chen},
  \bibinfo{person}{Orhan Firat}, \bibinfo{person}{Yuan Cao}, {and}
  \bibinfo{person}{Yonghui Wu}.} \bibinfo{year}{2018}\natexlab{}.
\newblock \showarticletitle{{Training Deeper Neural Machine Translation Models
  with Transparent Attention}}. In \bibinfo{booktitle}{\emph{Conference on
  Empirical Methods in Natural Language Processing}}.
\newblock


\bibitem[\protect\citeauthoryear{Behley, Garbade, Milioto, Quenzel, Behnke,
  Stachniss, and Gall}{Behley et~al\mbox{.}}{2019}]%
        {behley2019semantickitti}
\bibfield{author}{\bibinfo{person}{Jens Behley}, \bibinfo{person}{Martin
  Garbade}, \bibinfo{person}{Andres Milioto}, \bibinfo{person}{Jan Quenzel},
  \bibinfo{person}{Sven Behnke}, \bibinfo{person}{Cyrill Stachniss}, {and}
  \bibinfo{person}{Juergen Gall}.} \bibinfo{year}{2019}\natexlab{}.
\newblock \showarticletitle{{SemanticKITTI: A Dataset for Semantic Scene
  Understanding of LiDAR Sequences}}. In
  \bibinfo{booktitle}{\emph{International Conference on Computer Vision}}.
\newblock


\bibitem[\protect\citeauthoryear{Beltagy, Peters, and Cohan}{Beltagy
  et~al\mbox{.}}{2020}]%
        {beltagy2020longformer}
\bibfield{author}{\bibinfo{person}{Iz Beltagy}, \bibinfo{person}{Matthew~E
  Peters}, {and} \bibinfo{person}{Arman Cohan}.}
  \bibinfo{year}{2020}\natexlab{}.
\newblock \showarticletitle{{Longformer: The Long-Document Transformer}}.
\newblock \bibinfo{journal}{\emph{arXiv preprint arXiv:2004.05150}}
  (\bibinfo{year}{2020}).
\newblock


\bibitem[\protect\citeauthoryear{Bengio, L{\'e}onard, and Courville}{Bengio
  et~al\mbox{.}}{2013}]%
        {bengio2013estimating}
\bibfield{author}{\bibinfo{person}{Yoshua Bengio}, \bibinfo{person}{Nicholas
  L{\'e}onard}, {and} \bibinfo{person}{Aaron Courville}.}
  \bibinfo{year}{2013}\natexlab{}.
\newblock \showarticletitle{{Estimating or Propagating Gradients Through
  Stochastic Neurons for Conditional Computation}}.
\newblock \bibinfo{journal}{\emph{arXiv preprint arXiv:1308.3432}}
  (\bibinfo{year}{2013}).
\newblock


\bibitem[\protect\citeauthoryear{Bilen, Fernando, Gavves, Vedaldi, and
  Gould}{Bilen et~al\mbox{.}}{2016}]%
        {bilen2016dynamic}
\bibfield{author}{\bibinfo{person}{Hakan Bilen}, \bibinfo{person}{Basura
  Fernando}, \bibinfo{person}{Efstratios Gavves}, \bibinfo{person}{Andrea
  Vedaldi}, {and} \bibinfo{person}{Stephen Gould}.}
  \bibinfo{year}{2016}\natexlab{}.
\newblock \showarticletitle{{Dynamic Image Networks for Action Recognition}}.
  In \bibinfo{booktitle}{\emph{IEEE Conference on Computer Vision and Pattern
  Recognition}}.
\newblock


\bibitem[\protect\citeauthoryear{Bojarski, Del~Testa, Dworakowski, Firner,
  Flepp, Goyal, Jackel, Monfort, Muller, Zhang, Zhang, Zhao, and
  Zieba}{Bojarski et~al\mbox{.}}{2016}]%
        {bojarski2016end}
\bibfield{author}{\bibinfo{person}{Mariusz Bojarski}, \bibinfo{person}{Davide
  Del~Testa}, \bibinfo{person}{Daniel Dworakowski}, \bibinfo{person}{Bernhard
  Firner}, \bibinfo{person}{Beat Flepp}, \bibinfo{person}{Prasoon Goyal},
  \bibinfo{person}{Lawrence~D Jackel}, \bibinfo{person}{Mathew Monfort},
  \bibinfo{person}{Urs Muller}, \bibinfo{person}{Jiakai Zhang},
  \bibinfo{person}{Xin Zhang}, \bibinfo{person}{Jake Zhao}, {and}
  \bibinfo{person}{Karol Zieba}.} \bibinfo{year}{2016}\natexlab{}.
\newblock \showarticletitle{{End to End Learning for Self-Driving Cars}}.
\newblock \bibinfo{journal}{\emph{arXiv preprint arXiv:1604.07316}}
  (\bibinfo{year}{2016}).
\newblock


\bibitem[\protect\citeauthoryear{Brock, Lim, Ritchie, and Weston}{Brock
  et~al\mbox{.}}{2018}]%
        {brock2018smash}
\bibfield{author}{\bibinfo{person}{Andrew Brock}, \bibinfo{person}{Theodore
  Lim}, \bibinfo{person}{James~M Ritchie}, {and} \bibinfo{person}{Nick
  Weston}.} \bibinfo{year}{2018}\natexlab{}.
\newblock \showarticletitle{{SMASH: One-Shot Model Architecture Search through
  HyperNetworks}}. In \bibinfo{booktitle}{\emph{International Conference on
  Learning Representations}}.
\newblock


\bibitem[\protect\citeauthoryear{Brown, Mann, Ryder, Subbiah, Kaplan, Dhariwal,
  Neelakantan, Shyam, Sastry, Askell, Agarwal, Herbert-Voss, Krueger, Henighan,
  Child, Ramesh, Ziegler, Wu, Winter, Hesse, Chen, Sigler, Litwin, Gray, Chess,
  Clark, Berner, McCandlish, Radford, Sutskever, and Amodei}{Brown
  et~al\mbox{.}}{2020}]%
        {brown2020language}
\bibfield{author}{\bibinfo{person}{Tom~B Brown}, \bibinfo{person}{Benjamin
  Mann}, \bibinfo{person}{Nick Ryder}, \bibinfo{person}{Melanie Subbiah},
  \bibinfo{person}{Jared Kaplan}, \bibinfo{person}{Prafulla Dhariwal},
  \bibinfo{person}{Arvind Neelakantan}, \bibinfo{person}{Pranav Shyam},
  \bibinfo{person}{Girish Sastry}, \bibinfo{person}{Amanda Askell},
  \bibinfo{person}{Sandhini Agarwal}, \bibinfo{person}{Ariel Herbert-Voss},
  \bibinfo{person}{Gretchen Krueger}, \bibinfo{person}{Tom Henighan},
  \bibinfo{person}{Rewon Child}, \bibinfo{person}{Aditya Ramesh},
  \bibinfo{person}{Daniel Ziegler}, \bibinfo{person}{Jeffrey Wu},
  \bibinfo{person}{Clemens Winter}, \bibinfo{person}{Chris Hesse},
  \bibinfo{person}{Mark Chen}, \bibinfo{person}{Eric Sigler},
  \bibinfo{person}{Mateusz Litwin}, \bibinfo{person}{Scott Gray},
  \bibinfo{person}{Benjamin Chess}, \bibinfo{person}{Jack Clark},
  \bibinfo{person}{Christopher Berner}, \bibinfo{person}{Sam McCandlish},
  \bibinfo{person}{Alec Radford}, \bibinfo{person}{Ilya Sutskever}, {and}
  \bibinfo{person}{Dario Amodei}.} \bibinfo{year}{2020}\natexlab{}.
\newblock \showarticletitle{{Language Models are Few-Shot Learners}}. In
  \bibinfo{booktitle}{\emph{Conference on Neural Information Processing
  Systems}}.
\newblock


\bibitem[\protect\citeauthoryear{Bucilu{\v{a}}, Caruana, and
  Niculescu-Mizil}{Bucilu{\v{a}} et~al\mbox{.}}{2006}]%
        {bucilua2006model}
\bibfield{author}{\bibinfo{person}{Cristian Bucilu{\v{a}}},
  \bibinfo{person}{Rich Caruana}, {and} \bibinfo{person}{Alexandru
  Niculescu-Mizil}.} \bibinfo{year}{2006}\natexlab{}.
\newblock \showarticletitle{{Model Compression}}. In
  \bibinfo{booktitle}{\emph{International Conference on Knowledge Discovery and
  Data Mining}}.
\newblock


\bibitem[\protect\citeauthoryear{Cai, Chen, Zhang, Yu, and Wang}{Cai
  et~al\mbox{.}}{2018a}]%
        {cai2018efficient}
\bibfield{author}{\bibinfo{person}{Han Cai}, \bibinfo{person}{Tianyao Chen},
  \bibinfo{person}{Weinan Zhang}, \bibinfo{person}{Yong Yu}, {and}
  \bibinfo{person}{Jun Wang}.} \bibinfo{year}{2018}\natexlab{a}.
\newblock \showarticletitle{{Efficient Architecture Search by Network
  Transformation}}. In \bibinfo{booktitle}{\emph{AAAI Conference on Artificial
  Intelligence}}.
\newblock


\bibitem[\protect\citeauthoryear{Cai, Gan, Wang, Zhang, and Han}{Cai
  et~al\mbox{.}}{2020a}]%
        {cai2020once}
\bibfield{author}{\bibinfo{person}{Han Cai}, \bibinfo{person}{Chuang Gan},
  \bibinfo{person}{Tianzhe Wang}, \bibinfo{person}{Zhekai Zhang}, {and}
  \bibinfo{person}{Song Han}.} \bibinfo{year}{2020}\natexlab{a}.
\newblock \showarticletitle{{Once for All: Train One Network and Specialize it
  for Efficient Deployment}}. In \bibinfo{booktitle}{\emph{International
  Conference on Learning Representations}}.
\newblock


\bibitem[\protect\citeauthoryear{Cai, Gan, Zhu, and Han}{Cai
  et~al\mbox{.}}{2020b}]%
        {cai2020tinytl}
\bibfield{author}{\bibinfo{person}{Han Cai}, \bibinfo{person}{Chuang Gan},
  \bibinfo{person}{Ligeng Zhu}, {and} \bibinfo{person}{Song Han}.}
  \bibinfo{year}{2020}\natexlab{b}.
\newblock \showarticletitle{{TinyTL: Reduce Memory, Not Parameters for
  Efficient On-Device Learning}}. In \bibinfo{booktitle}{\emph{Conference on
  Neural Information Processing Systems}}.
\newblock


\bibitem[\protect\citeauthoryear{Cai, Lin, Lin, Liu, Wang, Wang, Zhu, and
  Han}{Cai et~al\mbox{.}}{2019a}]%
        {cai2019automl}
\bibfield{author}{\bibinfo{person}{Han Cai}, \bibinfo{person}{Ji Lin},
  \bibinfo{person}{Yujun Lin}, \bibinfo{person}{Zhijian Liu},
  \bibinfo{person}{Kuan Wang}, \bibinfo{person}{Tianzhe Wang},
  \bibinfo{person}{Ligeng Zhu}, {and} \bibinfo{person}{Song Han}.}
  \bibinfo{year}{2019}\natexlab{a}.
\newblock \showarticletitle{{AutoML for Architecting Efficient and Specialized
  Neural Networks}}.
\newblock \bibinfo{journal}{\emph{IEEE Micro}} \bibinfo{volume}{40},
  \bibinfo{number}{1} (\bibinfo{year}{2019}), \bibinfo{pages}{75--82}.
\newblock


\bibitem[\protect\citeauthoryear{Cai, Yang, Zhang, Han, and Yu}{Cai
  et~al\mbox{.}}{2018b}]%
        {cai2018path}
\bibfield{author}{\bibinfo{person}{Han Cai}, \bibinfo{person}{Jiacheng Yang},
  \bibinfo{person}{Weinan Zhang}, \bibinfo{person}{Song Han}, {and}
  \bibinfo{person}{Yong Yu}.} \bibinfo{year}{2018}\natexlab{b}.
\newblock \showarticletitle{{Path-Level Network Transformation for Efficient
  Architecture Search}}. In \bibinfo{booktitle}{\emph{International Conference
  on Machine Learning}}.
\newblock


\bibitem[\protect\citeauthoryear{Cai, Zhu, and Han}{Cai et~al\mbox{.}}{2019b}]%
        {cai2019proxylessnas}
\bibfield{author}{\bibinfo{person}{Han Cai}, \bibinfo{person}{Ligeng Zhu},
  {and} \bibinfo{person}{Song Han}.} \bibinfo{year}{2019}\natexlab{b}.
\newblock \showarticletitle{{ProxylessNAS: Direct Neural Architecture Search on
  Target Task and Hardware}}. In \bibinfo{booktitle}{\emph{International
  Conference on Learning Representations}}.
\newblock


\bibitem[\protect\citeauthoryear{Cai, Yao, Dong, Gholami, Mahoney, and
  Keutzer}{Cai et~al\mbox{.}}{2020c}]%
        {cai2020zeroq}
\bibfield{author}{\bibinfo{person}{Yaohui Cai}, \bibinfo{person}{Zhewei Yao},
  \bibinfo{person}{Zhen Dong}, \bibinfo{person}{Amir Gholami},
  \bibinfo{person}{Michael~W Mahoney}, {and} \bibinfo{person}{Kurt Keutzer}.}
  \bibinfo{year}{2020}\natexlab{c}.
\newblock \showarticletitle{{ZeroQ: A Novel Zero Shot Quantization Framework}}.
  In \bibinfo{booktitle}{\emph{IEEE Conference on Computer Vision and Pattern
  Recognition}}.
\newblock


\bibitem[\protect\citeauthoryear{Caldas, Kone{\v{c}}ny, McMahan, and
  Talwalkar}{Caldas et~al\mbox{.}}{2018}]%
        {caldas2018expanding}
\bibfield{author}{\bibinfo{person}{Sebastian Caldas}, \bibinfo{person}{Jakub
  Kone{\v{c}}ny}, \bibinfo{person}{H~Brendan McMahan}, {and}
  \bibinfo{person}{Ameet Talwalkar}.} \bibinfo{year}{2018}\natexlab{}.
\newblock \showarticletitle{{Expanding the Reach of Federated Learning by
  Reducing Client Resource Requirements}}.
\newblock \bibinfo{journal}{\emph{arXiv preprint arXiv:1812.07210}}
  (\bibinfo{year}{2018}).
\newblock


\bibitem[\protect\citeauthoryear{Cao, Trivedi, Balasubramanian, and
  Balasubramanian}{Cao et~al\mbox{.}}{2020}]%
        {cao2020deformer}
\bibfield{author}{\bibinfo{person}{Qingqing Cao}, \bibinfo{person}{Harsh
  Trivedi}, \bibinfo{person}{Aruna Balasubramanian}, {and}
  \bibinfo{person}{Niranjan Balasubramanian}.} \bibinfo{year}{2020}\natexlab{}.
\newblock \showarticletitle{{DeFormer: Decomposing Pre-trained Transformers for
  Faster Question Answering}}. In \bibinfo{booktitle}{\emph{Conference of the
  Association for Computational Linguistics}}.
\newblock


\bibitem[\protect\citeauthoryear{Carreira and Zisserman}{Carreira and
  Zisserman}{2017}]%
        {carreira2017quo}
\bibfield{author}{\bibinfo{person}{Joao Carreira} {and} \bibinfo{person}{Andrew
  Zisserman}.} \bibinfo{year}{2017}\natexlab{}.
\newblock \showarticletitle{{Quo Vadis, Action Recognition? A New Model and the
  Kinetics Dataset}}. In \bibinfo{booktitle}{\emph{IEEE Conference on Computer
  Vision and Pattern Recognition}}.
\newblock


\bibitem[\protect\citeauthoryear{Cavigelli, Gschwend, Mayer, Willi, Muheim, and
  Benini}{Cavigelli et~al\mbox{.}}{2015}]%
        {cavigelli2015origami}
\bibfield{author}{\bibinfo{person}{Lukas Cavigelli}, \bibinfo{person}{David
  Gschwend}, \bibinfo{person}{Christoph Mayer}, \bibinfo{person}{Samuel Willi},
  \bibinfo{person}{Beat Muheim}, {and} \bibinfo{person}{Luca Benini}.}
  \bibinfo{year}{2015}\natexlab{}.
\newblock \showarticletitle{{Origami: A Convolutional Network Accelerator}}. In
  \bibinfo{booktitle}{\emph{Great Lakes Symposium on VLSI}}.
\newblock


\bibitem[\protect\citeauthoryear{Chakradhar, Sankaradas, Jakkula, and
  Cadambi}{Chakradhar et~al\mbox{.}}{2010}]%
        {chakradhar2010dynamically}
\bibfield{author}{\bibinfo{person}{Srimat Chakradhar}, \bibinfo{person}{Murugan
  Sankaradas}, \bibinfo{person}{Venkata Jakkula}, {and}
  \bibinfo{person}{Srihari Cadambi}.} \bibinfo{year}{2010}\natexlab{}.
\newblock \showarticletitle{{A Dynamically Configurable Coprocessor for
  Convolutional Neural Networks}}. In \bibinfo{booktitle}{\emph{International
  Symposium on Computer Architecture}}.
\newblock


\bibitem[\protect\citeauthoryear{Chang, Funkhouser, Guibas, Hanrahan, Huang,
  Li, Savarese, Savva, Song, Su, Xiao, Yi, and Yu}{Chang et~al\mbox{.}}{2015}]%
        {chang2015shapenet}
\bibfield{author}{\bibinfo{person}{Angel~X Chang}, \bibinfo{person}{Thomas
  Funkhouser}, \bibinfo{person}{Leonidas Guibas}, \bibinfo{person}{Pat
  Hanrahan}, \bibinfo{person}{Qixing Huang}, \bibinfo{person}{Zimo Li},
  \bibinfo{person}{Silvio Savarese}, \bibinfo{person}{Manolis Savva},
  \bibinfo{person}{Shuran Song}, \bibinfo{person}{Hao Su},
  \bibinfo{person}{Jianxiong Xiao}, \bibinfo{person}{Li Yi}, {and}
  \bibinfo{person}{Fisher Yu}.} \bibinfo{year}{2015}\natexlab{}.
\newblock \showarticletitle{{ShapeNet: An Information-Rich 3D Model
  Repository}}.
\newblock \bibinfo{journal}{\emph{arXiv preprint arXiv:1512.03012}}
  (\bibinfo{year}{2015}).
\newblock


\bibitem[\protect\citeauthoryear{Chatfield, Simonyan, Vedaldi, and
  Zisserman}{Chatfield et~al\mbox{.}}{2014}]%
        {chatfield2014return}
\bibfield{author}{\bibinfo{person}{Ken Chatfield}, \bibinfo{person}{Karen
  Simonyan}, \bibinfo{person}{Andrea Vedaldi}, {and} \bibinfo{person}{Andrew
  Zisserman}.} \bibinfo{year}{2014}\natexlab{}.
\newblock \showarticletitle{{Return of the Devil in the Details: Delving Deep
  into Convolutional Nets}}. In \bibinfo{booktitle}{\emph{British Machine
  Vision Conference}}.
\newblock


\bibitem[\protect\citeauthoryear{Chen, Li, Qiu, Wang, Li, Ding, Deng, Huang,
  Lin, and Zhou}{Chen et~al\mbox{.}}{2020a}]%
        {chen2020adabert}
\bibfield{author}{\bibinfo{person}{Daoyuan Chen}, \bibinfo{person}{Yaliang Li},
  \bibinfo{person}{Minghui Qiu}, \bibinfo{person}{Zhen Wang},
  \bibinfo{person}{Bofang Li}, \bibinfo{person}{Bolin Ding},
  \bibinfo{person}{Hongbo Deng}, \bibinfo{person}{Jun Huang},
  \bibinfo{person}{Wei Lin}, {and} \bibinfo{person}{Jingren Zhou}.}
  \bibinfo{year}{2020}\natexlab{a}.
\newblock \showarticletitle{{AdaBERT: Task-Adaptive BERT Compression with
  Differentiable Neural Architecture Search}}. In
  \bibinfo{booktitle}{\emph{International Joint Conference on Artificial
  Intelligence}}.
\newblock


\bibitem[\protect\citeauthoryear{Chen, Choi, Yu, Han, and Chandraker}{Chen
  et~al\mbox{.}}{2017a}]%
        {chen2017learning}
\bibfield{author}{\bibinfo{person}{Guobin Chen}, \bibinfo{person}{Wongun Choi},
  \bibinfo{person}{Xiang Yu}, \bibinfo{person}{Tony Han}, {and}
  \bibinfo{person}{Manmohan Chandraker}.} \bibinfo{year}{2017}\natexlab{a}.
\newblock \showarticletitle{{Learning Efficient Object Detection Models with
  Knowledge Distillation}}. In \bibinfo{booktitle}{\emph{Conference on Neural
  Information Processing Systems}}.
\newblock


\bibitem[\protect\citeauthoryear{Chen, Collins, Zhu, Papandreou, Zoph, Schroff,
  Adam, and Shlens}{Chen et~al\mbox{.}}{2018a}]%
        {chen2018searching}
\bibfield{author}{\bibinfo{person}{Liang-Chieh Chen}, \bibinfo{person}{Maxwell
  Collins}, \bibinfo{person}{Yukun Zhu}, \bibinfo{person}{George Papandreou},
  \bibinfo{person}{Barret Zoph}, \bibinfo{person}{Florian Schroff},
  \bibinfo{person}{Hartwig Adam}, {and} \bibinfo{person}{Jon Shlens}.}
  \bibinfo{year}{2018}\natexlab{a}.
\newblock \showarticletitle{{Searching for Efficient Multi-Scale Architectures
  for Dense Image Prediction}}. In \bibinfo{booktitle}{\emph{Conference on
  Neural Information Processing Systems}}.
\newblock


\bibitem[\protect\citeauthoryear{Chen, Du, Sun, Wang, Wu, Chen, and Temam}{Chen
  et~al\mbox{.}}{2014a}]%
        {chen2014diannao}
\bibfield{author}{\bibinfo{person}{Tianshi Chen}, \bibinfo{person}{Zidong Du},
  \bibinfo{person}{Ninghui Sun}, \bibinfo{person}{Jia Wang},
  \bibinfo{person}{Chengyong Wu}, \bibinfo{person}{Yunji Chen}, {and}
  \bibinfo{person}{Olivier Temam}.} \bibinfo{year}{2014}\natexlab{a}.
\newblock \showarticletitle{{DianNao: A Small-Footprint High-Throughput
  Accelerator for Ubiquitous Machine-Learning}}. In
  \bibinfo{booktitle}{\emph{ACM International Conference on Architectural
  Support for Programming Languages and Operating Systems}}.
\newblock


\bibitem[\protect\citeauthoryear{Chen, Li, Li, Lin, Wang, Wang, Xiao, Xu,
  Zhang, and Zhang}{Chen et~al\mbox{.}}{2015}]%
        {chen2015mxnet}
\bibfield{author}{\bibinfo{person}{Tianqi Chen}, \bibinfo{person}{Mu Li},
  \bibinfo{person}{Yutian Li}, \bibinfo{person}{Min Lin},
  \bibinfo{person}{Naiyan Wang}, \bibinfo{person}{Minjie Wang},
  \bibinfo{person}{Tianjun Xiao}, \bibinfo{person}{Bing Xu},
  \bibinfo{person}{Chiyuan Zhang}, {and} \bibinfo{person}{Zheng Zhang}.}
  \bibinfo{year}{2015}\natexlab{}.
\newblock \showarticletitle{{MXNet: A Flexible and Efficient Machine Learning
  Library for Heterogeneous Distributed Systems}}.
\newblock \bibinfo{journal}{\emph{arXiv preprint arXiv:1512.01274}}
  (\bibinfo{year}{2015}).
\newblock


\bibitem[\protect\citeauthoryear{Chen, Moreau, Jiang, Zheng, Yan, Shen, Cowan,
  Wang, Hu, Ceze, Guestrin, and Krishnamurthy}{Chen et~al\mbox{.}}{2018b}]%
        {chen2018tvm}
\bibfield{author}{\bibinfo{person}{Tianqi Chen}, \bibinfo{person}{Thierry
  Moreau}, \bibinfo{person}{Ziheng Jiang}, \bibinfo{person}{Lianmin Zheng},
  \bibinfo{person}{Eddie Yan}, \bibinfo{person}{Haichen Shen},
  \bibinfo{person}{Meghan Cowan}, \bibinfo{person}{Leyuan Wang},
  \bibinfo{person}{Yuwei Hu}, \bibinfo{person}{Luis Ceze},
  \bibinfo{person}{Carlos Guestrin}, {and} \bibinfo{person}{Arvind
  Krishnamurthy}.} \bibinfo{year}{2018}\natexlab{b}.
\newblock \showarticletitle{{TVM: An Automated End-to-End Optimizing Compiler
  for Deep Learning}}. In \bibinfo{booktitle}{\emph{USENIX Symposium on
  Operating Systems Design and Implementation}}.
\newblock


\bibitem[\protect\citeauthoryear{Chen, Xu, Zhang, and Guestrin}{Chen
  et~al\mbox{.}}{2016}]%
        {chen2016training}
\bibfield{author}{\bibinfo{person}{Tianqi Chen}, \bibinfo{person}{Bing Xu},
  \bibinfo{person}{Chiyuan Zhang}, {and} \bibinfo{person}{Carlos Guestrin}.}
  \bibinfo{year}{2016}\natexlab{}.
\newblock \showarticletitle{{Training Deep Nets with Sublinear Memory Cost}}.
\newblock \bibinfo{journal}{\emph{arXiv preprint arXiv:1604.06174}}
  (\bibinfo{year}{2016}).
\newblock


\bibitem[\protect\citeauthoryear{Chen, Luo, Liu, Zhang, He, Wang, Li, Chen, Xu,
  Sun, and Temam}{Chen et~al\mbox{.}}{2014b}]%
        {chen2014dadiannao}
\bibfield{author}{\bibinfo{person}{Yunji Chen}, \bibinfo{person}{Tao Luo},
  \bibinfo{person}{Shaoli Liu}, \bibinfo{person}{Shijin Zhang},
  \bibinfo{person}{Liqiang He}, \bibinfo{person}{Jia Wang},
  \bibinfo{person}{Ling Li}, \bibinfo{person}{Tianshi Chen},
  \bibinfo{person}{Zhiwei Xu}, \bibinfo{person}{Ninghui Sun}, {and}
  \bibinfo{person}{Olivier Temam}.} \bibinfo{year}{2014}\natexlab{b}.
\newblock \showarticletitle{{DaDianNao: A Machine-Learning Supercomputer}}. In
  \bibinfo{booktitle}{\emph{International Symposium on Microarchitecture}}.
\newblock


\bibitem[\protect\citeauthoryear{Chen, Yang, Zhang, Meng, Xiao, and Sun}{Chen
  et~al\mbox{.}}{2019b}]%
        {chen2019detnas}
\bibfield{author}{\bibinfo{person}{Yukang Chen}, \bibinfo{person}{Tong Yang},
  \bibinfo{person}{Xiangyu Zhang}, \bibinfo{person}{Gaofeng Meng},
  \bibinfo{person}{Xinyu Xiao}, {and} \bibinfo{person}{Jian Sun}.}
  \bibinfo{year}{2019}\natexlab{b}.
\newblock \showarticletitle{{DetNAS: Backbone Search for Object Detection}}. In
  \bibinfo{booktitle}{\emph{Conference on Neural Information Processing
  Systems}}.
\newblock


\bibitem[\protect\citeauthoryear{Chen, Yang, Sun, Wang, Xu, Shen, Zhou, Tong,
  Bai, and Zhang}{Chen et~al\mbox{.}}{2020b}]%
        {chen2020autoadr}
\bibfield{author}{\bibinfo{person}{Yiren Chen}, \bibinfo{person}{Yaming Yang},
  \bibinfo{person}{Hong Sun}, \bibinfo{person}{Yujing Wang},
  \bibinfo{person}{Yu Xu}, \bibinfo{person}{Wei Shen}, \bibinfo{person}{Rong
  Zhou}, \bibinfo{person}{Yunhai Tong}, \bibinfo{person}{Jing Bai}, {and}
  \bibinfo{person}{Ruofei Zhang}.} \bibinfo{year}{2020}\natexlab{b}.
\newblock \showarticletitle{{AutoADR: Automatic Model Design for Ad
  Relevance}}. In \bibinfo{booktitle}{\emph{International Conference on
  Information \& Knowledge Management}}.
\newblock


\bibitem[\protect\citeauthoryear{Chen, Krishna, Emer, and Sze}{Chen
  et~al\mbox{.}}{2017b}]%
        {chen2017eyeriss}
\bibfield{author}{\bibinfo{person}{Yu-Hsin Chen}, \bibinfo{person}{Tushar
  Krishna}, \bibinfo{person}{Joel~S Emer}, {and} \bibinfo{person}{Vivienne
  Sze}.} \bibinfo{year}{2017}\natexlab{b}.
\newblock \showarticletitle{{Eyeriss: An Energy-Efficient Reconfigurable
  Accelerator for Deep Convolutional Neural Networks}}.
\newblock \bibinfo{journal}{\emph{International Journal of Space-Based and
  Situated Computing}} \bibinfo{volume}{52}, \bibinfo{number}{1}
  (\bibinfo{year}{2017}), \bibinfo{pages}{127--138}.
\newblock


\bibitem[\protect\citeauthoryear{Chen, Yang, Emer, and Sze}{Chen
  et~al\mbox{.}}{2019a}]%
        {chen2019eyeriss}
\bibfield{author}{\bibinfo{person}{Yu-Hsin Chen}, \bibinfo{person}{Tien-Ju
  Yang}, \bibinfo{person}{Joel Emer}, {and} \bibinfo{person}{Vivienne Sze}.}
  \bibinfo{year}{2019}\natexlab{a}.
\newblock \showarticletitle{{Eyeriss v2: A Flexible Accelerator for Emerging
  Deep Neural Networks on Mobile Devices}}.
\newblock \bibinfo{journal}{\emph{IEEE Journal on Emerging and Selected Topics
  in Circuits and Systems}} \bibinfo{volume}{9}, \bibinfo{number}{2}
  (\bibinfo{year}{2019}), \bibinfo{pages}{292--308}.
\newblock


\bibitem[\protect\citeauthoryear{Cheng, Wang, Zhou, and Zhang}{Cheng
  et~al\mbox{.}}{2017}]%
        {cheng2017survey}
\bibfield{author}{\bibinfo{person}{Yu Cheng}, \bibinfo{person}{Duo Wang},
  \bibinfo{person}{Pan Zhou}, {and} \bibinfo{person}{Tao Zhang}.}
  \bibinfo{year}{2017}\natexlab{}.
\newblock \showarticletitle{{Model Compression and Acceleration for Deep Neural
  Networks: The Principles, Progress, and Challenges}}.
\newblock \bibinfo{journal}{\emph{IEEE Signal Processing Magazine}}
  \bibinfo{volume}{35}, \bibinfo{number}{1} (\bibinfo{year}{2017}),
  \bibinfo{pages}{126--136}.
\newblock


\bibitem[\protect\citeauthoryear{Cheong and Daniel}{Cheong and Daniel}{2019}]%
        {cheong2019transformers}
\bibfield{author}{\bibinfo{person}{Robin Cheong} {and} \bibinfo{person}{Robel
  Daniel}.} \bibinfo{year}{2019}\natexlab{}.
\newblock \bibinfo{booktitle}{\emph{{transformers.zip: Compressing Transformers
  with Pruning and Quantization}}}.
\newblock \bibinfo{type}{{T}echnical {R}eport}. \bibinfo{institution}{Stanford
  University, Stanford, California}.
\newblock


\bibitem[\protect\citeauthoryear{Chetlur, Woolley, Vandermersch, Cohen, Tran,
  Catanzaro, and Shelhamer}{Chetlur et~al\mbox{.}}{2014}]%
        {chetlur2014cudnn}
\bibfield{author}{\bibinfo{person}{Sharan Chetlur}, \bibinfo{person}{Cliff
  Woolley}, \bibinfo{person}{Philippe Vandermersch}, \bibinfo{person}{Jonathan
  Cohen}, \bibinfo{person}{John Tran}, \bibinfo{person}{Bryan Catanzaro}, {and}
  \bibinfo{person}{Evan Shelhamer}.} \bibinfo{year}{2014}\natexlab{}.
\newblock \showarticletitle{{cuDNN: Efficient Primitives for Deep Learning}}.
\newblock \bibinfo{journal}{\emph{arXiv preprint arXiv:1410.0759}}
  (\bibinfo{year}{2014}).
\newblock


\bibitem[\protect\citeauthoryear{Child, Gray, Radford, and Sutskever}{Child
  et~al\mbox{.}}{2019}]%
        {child2019generating}
\bibfield{author}{\bibinfo{person}{Rewon Child}, \bibinfo{person}{Scott Gray},
  \bibinfo{person}{Alec Radford}, {and} \bibinfo{person}{Ilya Sutskever}.}
  \bibinfo{year}{2019}\natexlab{}.
\newblock \showarticletitle{{Generating Long Sequences with Sparse
  Transformers}}.
\newblock \bibinfo{journal}{\emph{arXiv preprint arXiv:1904.10509}}
  (\bibinfo{year}{2019}).
\newblock


\bibitem[\protect\citeauthoryear{Choudhary, Mishra, Goswami, and
  Sarangapani}{Choudhary et~al\mbox{.}}{2020}]%
        {choudhary2020comprehensive}
\bibfield{author}{\bibinfo{person}{Tejalal Choudhary}, \bibinfo{person}{Vipul
  Mishra}, \bibinfo{person}{Anurag Goswami}, {and} \bibinfo{person}{Jagannathan
  Sarangapani}.} \bibinfo{year}{2020}\natexlab{}.
\newblock \showarticletitle{{A Comprehensive Survey on Model Compression and
  Acceleration}}.
\newblock \bibinfo{journal}{\emph{Artificial Intelligence Review}}
  (\bibinfo{year}{2020}).
\newblock


\bibitem[\protect\citeauthoryear{Choy, Dong, and Koltun}{Choy
  et~al\mbox{.}}{2020}]%
        {choy2020deep}
\bibfield{author}{\bibinfo{person}{Christopher Choy}, \bibinfo{person}{Wei
  Dong}, {and} \bibinfo{person}{Vladlen Koltun}.}
  \bibinfo{year}{2020}\natexlab{}.
\newblock \showarticletitle{{Deep Global Registration}}. In
  \bibinfo{booktitle}{\emph{IEEE Conference on Computer Vision and Pattern
  Recognition}}.
\newblock


\bibitem[\protect\citeauthoryear{Choy, Gwak, and Savarese}{Choy
  et~al\mbox{.}}{2019a}]%
        {choy20194d}
\bibfield{author}{\bibinfo{person}{Christopher Choy}, \bibinfo{person}{JunYoung
  Gwak}, {and} \bibinfo{person}{Silvio Savarese}.}
  \bibinfo{year}{2019}\natexlab{a}.
\newblock \showarticletitle{{4D Spatio-Temporal ConvNets: Minkowski
  Convolutional Neural Networks}}. In \bibinfo{booktitle}{\emph{IEEE Conference
  on Computer Vision and Pattern Recognition}}.
\newblock


\bibitem[\protect\citeauthoryear{Choy, Park, and Koltun}{Choy
  et~al\mbox{.}}{2019b}]%
        {choy2019fully}
\bibfield{author}{\bibinfo{person}{Christopher Choy}, \bibinfo{person}{Jaesik
  Park}, {and} \bibinfo{person}{Vladlen Koltun}.}
  \bibinfo{year}{2019}\natexlab{b}.
\newblock \showarticletitle{{Fully Convolutional Geometric Features}}. In
  \bibinfo{booktitle}{\emph{International Conference on Computer Vision}}.
\newblock


\bibitem[\protect\citeauthoryear{Cicek, Abdulkadir, Lienkamp, Brox, and
  Ronneberger}{Cicek et~al\mbox{.}}{2016}]%
        {cicek20163d}
\bibfield{author}{\bibinfo{person}{Ozgun Cicek}, \bibinfo{person}{Ahmed
  Abdulkadir}, \bibinfo{person}{Soeren~S Lienkamp}, \bibinfo{person}{Thomas
  Brox}, {and} \bibinfo{person}{Olaf Ronneberger}.}
  \bibinfo{year}{2016}\natexlab{}.
\newblock \showarticletitle{{3D U-Net: Learning Dense Volumetric Segmentation
  from Sparse Annotation}}. In \bibinfo{booktitle}{\emph{International
  Conference on Medical Image Computing and Computer Assisted Intervention}}.
\newblock


\bibitem[\protect\citeauthoryear{Codevilla, Miiller, L{\'o}pez, Koltun, and
  Dosovitskiy}{Codevilla et~al\mbox{.}}{2018}]%
        {codevilla2018end}
\bibfield{author}{\bibinfo{person}{Felipe Codevilla}, \bibinfo{person}{Matthias
  Miiller}, \bibinfo{person}{Antonio L{\'o}pez}, \bibinfo{person}{Vladlen
  Koltun}, {and} \bibinfo{person}{Alexey Dosovitskiy}.}
  \bibinfo{year}{2018}\natexlab{}.
\newblock \showarticletitle{{End-to-End Driving Via Conditional Imitation
  Learning}}. In \bibinfo{booktitle}{\emph{IEEE International Conference on
  Robotics and Automation}}.
\newblock


\bibitem[\protect\citeauthoryear{Cong, Fang, Lo, Wang, Xu, and Zhang}{Cong
  et~al\mbox{.}}{2018}]%
        {cong2018understanding}
\bibfield{author}{\bibinfo{person}{Jason Cong}, \bibinfo{person}{Zhenman Fang},
  \bibinfo{person}{Michael Lo}, \bibinfo{person}{Hanrui Wang},
  \bibinfo{person}{Jingxian Xu}, {and} \bibinfo{person}{Shaochong Zhang}.}
  \bibinfo{year}{2018}\natexlab{}.
\newblock \showarticletitle{{Understanding performance differences of FPGAs and
  GPUs}}. In \bibinfo{booktitle}{\emph{IEEE Symposium on Field-Programmable
  Custom Computing Machines}}.
\newblock


\bibitem[\protect\citeauthoryear{Courbariaux, Bengio, and David}{Courbariaux
  et~al\mbox{.}}{2015}]%
        {courbariaux2015binaryconnect}
\bibfield{author}{\bibinfo{person}{Matthieu Courbariaux},
  \bibinfo{person}{Yoshua Bengio}, {and} \bibinfo{person}{Jean-Pierre David}.}
  \bibinfo{year}{2015}\natexlab{}.
\newblock \showarticletitle{{BinaryConnect: Training Deep Neural Networks with
  Binary Weights during Propagations}}. In \bibinfo{booktitle}{\emph{Conference
  on Neural Information Processing Systems}}.
\newblock


\bibitem[\protect\citeauthoryear{Courbariaux, Hubara, Soudry, El-Yaniv, and
  Bengio}{Courbariaux et~al\mbox{.}}{2016}]%
        {courbariaux2016binarized}
\bibfield{author}{\bibinfo{person}{Matthieu Courbariaux}, \bibinfo{person}{Itay
  Hubara}, \bibinfo{person}{Daniel Soudry}, \bibinfo{person}{Ran El-Yaniv},
  {and} \bibinfo{person}{Yoshua Bengio}.} \bibinfo{year}{2016}\natexlab{}.
\newblock \showarticletitle{{Binarized Neural Networks: Training Deep Neural
  Networks with Weights and Activations Constrained to +1 or -1}}.
\newblock \bibinfo{journal}{\emph{arXiv preprint arXiv:1602.02830}}
  (\bibinfo{year}{2016}).
\newblock


\bibitem[\protect\citeauthoryear{Cui, Song, Sun, Howard, and Belongie}{Cui
  et~al\mbox{.}}{2018}]%
        {cui2018large}
\bibfield{author}{\bibinfo{person}{Yin Cui}, \bibinfo{person}{Yang Song},
  \bibinfo{person}{Chen Sun}, \bibinfo{person}{Andrew Howard}, {and}
  \bibinfo{person}{Serge Belongie}.} \bibinfo{year}{2018}\natexlab{}.
\newblock \showarticletitle{{Large Scale Fine-Grained Categorization and
  Domain-Specific Transfer Learning}}. In \bibinfo{booktitle}{\emph{IEEE
  Conference on Computer Vision and Pattern Recognition}}.
\newblock


\bibitem[\protect\citeauthoryear{Dai, Chang, Savva, Halber, Funkhouser, and
  Nie\ss{}ner}{Dai et~al\mbox{.}}{2017}]%
        {dai2017scannet}
\bibfield{author}{\bibinfo{person}{Angela Dai}, \bibinfo{person}{Angel~X
  Chang}, \bibinfo{person}{Manolis Savva}, \bibinfo{person}{Maciej Halber},
  \bibinfo{person}{Thomas Funkhouser}, {and} \bibinfo{person}{Matthias
  Nie\ss{}ner}.} \bibinfo{year}{2017}\natexlab{}.
\newblock \showarticletitle{{ScanNet: Richly-Annotated 3D Reconstructions of
  Indoor Scenes}}. In \bibinfo{booktitle}{\emph{IEEE Conference on Computer
  Vision and Pattern Recognition}}.
\newblock


\bibitem[\protect\citeauthoryear{Dai, Yang, Ye, Cheng, Luo, Song, Chen, and
  Zhao}{Dai et~al\mbox{.}}{2020}]%
        {dai2020sparsetrain}
\bibfield{author}{\bibinfo{person}{Pengcheng Dai}, \bibinfo{person}{Jianlei
  Yang}, \bibinfo{person}{Xucheng Ye}, \bibinfo{person}{Xingzhou Cheng},
  \bibinfo{person}{Junyu Luo}, \bibinfo{person}{Linghao Song},
  \bibinfo{person}{Yiran Chen}, {and} \bibinfo{person}{Weisheng Zhao}.}
  \bibinfo{year}{2020}\natexlab{}.
\newblock \showarticletitle{{SparseTrain: Exploiting Dataflow Sparsity for
  Efficient Convolutional Neural Networks Training}}. In
  \bibinfo{booktitle}{\emph{Design Automation Conference}}.
\newblock


\bibitem[\protect\citeauthoryear{Dally, Turakhia, and Han}{Dally
  et~al\mbox{.}}{2020}]%
        {dally2020domain}
\bibfield{author}{\bibinfo{person}{William~J Dally}, \bibinfo{person}{Yatish
  Turakhia}, {and} \bibinfo{person}{Song Han}.}
  \bibinfo{year}{2020}\natexlab{}.
\newblock \showarticletitle{{Domain-Specific Hardware Accelerators}}.
\newblock \bibinfo{journal}{\emph{Communications of the ACM}}
  \bibinfo{volume}{63}, \bibinfo{number}{7} (\bibinfo{year}{2020}).
\newblock


\bibitem[\protect\citeauthoryear{Deng, Dong, Socher, Li, Li, and Fei-Fei}{Deng
  et~al\mbox{.}}{2009}]%
        {deng2009imagenet}
\bibfield{author}{\bibinfo{person}{Jia Deng}, \bibinfo{person}{Wei Dong},
  \bibinfo{person}{Richard Socher}, \bibinfo{person}{Li-Jia Li},
  \bibinfo{person}{Kai Li}, {and} \bibinfo{person}{Li Fei-Fei}.}
  \bibinfo{year}{2009}\natexlab{}.
\newblock \showarticletitle{{ImageNet: A Large-Scale Hierarchical Image
  Database}}. In \bibinfo{booktitle}{\emph{IEEE Conference on Computer Vision
  and Pattern Recognition}}.
\newblock


\bibitem[\protect\citeauthoryear{Deng, Li, Han, Shi, and Xie}{Deng
  et~al\mbox{.}}{2020}]%
        {deng2020model}
\bibfield{author}{\bibinfo{person}{Lei Deng}, \bibinfo{person}{Guoqi Li},
  \bibinfo{person}{Song Han}, \bibinfo{person}{Luping Shi}, {and}
  \bibinfo{person}{Yuan Xie}.} \bibinfo{year}{2020}\natexlab{}.
\newblock \showarticletitle{{Model Compression and Hardware Acceleration for
  Neural Networks: A Comprehensive Survey}}.
\newblock \bibinfo{journal}{\emph{Proceedings of the IEEE}}
  (\bibinfo{year}{2020}).
\newblock


\bibitem[\protect\citeauthoryear{Deng, Li, Huang, Yao, Yu, Seide, Seltzer,
  Zweig, He, Williams, Gong, and Acero}{Deng et~al\mbox{.}}{2013}]%
        {deng2013recent}
\bibfield{author}{\bibinfo{person}{Li Deng}, \bibinfo{person}{Jinyu Li},
  \bibinfo{person}{Jui-Ting Huang}, \bibinfo{person}{Kaisheng Yao},
  \bibinfo{person}{Dong Yu}, \bibinfo{person}{Frank Seide},
  \bibinfo{person}{Michael Seltzer}, \bibinfo{person}{Geoff Zweig},
  \bibinfo{person}{Xiaodong He}, \bibinfo{person}{Jason Williams},
  \bibinfo{person}{Yifan Gong}, {and} \bibinfo{person}{Alex Acero}.}
  \bibinfo{year}{2013}\natexlab{}.
\newblock \showarticletitle{{Recent Advances in Deep Learning for Speech
  Research at Microsoft}}. In \bibinfo{booktitle}{\emph{IEEE International
  Conference on Acoustics, Speech and Signal Processing}}.
\newblock


\bibitem[\protect\citeauthoryear{Denton, Zaremba, Bruna, LeCun, and
  Fergus}{Denton et~al\mbox{.}}{2014}]%
        {denton2014exploiting}
\bibfield{author}{\bibinfo{person}{Emily~L Denton}, \bibinfo{person}{Wojciech
  Zaremba}, \bibinfo{person}{Joan Bruna}, \bibinfo{person}{Yann LeCun}, {and}
  \bibinfo{person}{Rob Fergus}.} \bibinfo{year}{2014}\natexlab{}.
\newblock \showarticletitle{{Exploiting Linear Structure Within Convolutional
  Networks for Efficient Evaluation}}. In \bibinfo{booktitle}{\emph{Conference
  on Neural Information Processing Systems}}.
\newblock


\bibitem[\protect\citeauthoryear{Devlin, Chang, Lee, and Toutanova}{Devlin
  et~al\mbox{.}}{2018}]%
        {devlin2018bert}
\bibfield{author}{\bibinfo{person}{Jacob Devlin}, \bibinfo{person}{Ming-Wei
  Chang}, \bibinfo{person}{Kenton Lee}, {and} \bibinfo{person}{Kristina
  Toutanova}.} \bibinfo{year}{2018}\natexlab{}.
\newblock \showarticletitle{{BERT: Pre-Training of Deep Bidirectional
  Transformers for Language Understanding}}. In
  \bibinfo{booktitle}{\emph{Conference of the North American Chapter of the
  Association for Computational Linguistics}}.
\newblock


\bibitem[\protect\citeauthoryear{Ding, Zhu, Jia, Pekhimenko, and Han}{Ding
  et~al\mbox{.}}{2020}]%
        {ding2020ios}
\bibfield{author}{\bibinfo{person}{Yaoyao Ding}, \bibinfo{person}{Ligeng Zhu},
  \bibinfo{person}{Zhihao Jia}, \bibinfo{person}{Gennady Pekhimenko}, {and}
  \bibinfo{person}{Song Han}.} \bibinfo{year}{2020}\natexlab{}.
\newblock \showarticletitle{{IOS: Inter-Operator Scheduler for CNN
  Acceleration}}.
\newblock \bibinfo{journal}{\emph{arXiv preprint arXiv:2011.01302}}
  (\bibinfo{year}{2020}).
\newblock


\bibitem[\protect\citeauthoryear{Donahue, Jia, Vinyals, Hoffman, Zhang, Tzeng,
  and Darrell}{Donahue et~al\mbox{.}}{2014}]%
        {donahue2014decaf}
\bibfield{author}{\bibinfo{person}{Jeff Donahue}, \bibinfo{person}{Yangqing
  Jia}, \bibinfo{person}{Oriol Vinyals}, \bibinfo{person}{Judy Hoffman},
  \bibinfo{person}{Ning Zhang}, \bibinfo{person}{Eric Tzeng}, {and}
  \bibinfo{person}{Trevor Darrell}.} \bibinfo{year}{2014}\natexlab{}.
\newblock \showarticletitle{{DeCAF: A Deep Convolutional Activation Feature for
  Generic Visual Recognition}}. In \bibinfo{booktitle}{\emph{International
  Conference on Machine Learning}}.
\newblock


\bibitem[\protect\citeauthoryear{Dong, Yao, Gholami, Mahoney, and Keutzer}{Dong
  et~al\mbox{.}}{2019}]%
        {dong2019hawq}
\bibfield{author}{\bibinfo{person}{Zhen Dong}, \bibinfo{person}{Zhewei Yao},
  \bibinfo{person}{Amir Gholami}, \bibinfo{person}{Michael~W Mahoney}, {and}
  \bibinfo{person}{Kurt Keutzer}.} \bibinfo{year}{2019}\natexlab{}.
\newblock \showarticletitle{{HAWQ: Hessian AWare Quantization of Neural
  Networks with Mixed-Precision}}. In \bibinfo{booktitle}{\emph{International
  Conference on Computer Vision}}.
\newblock


\bibitem[\protect\citeauthoryear{Du, Fasthuber, Chen, Ienne, Li, Luo, Feng,
  Chen, and Temam}{Du et~al\mbox{.}}{2015}]%
        {du2015shidiannao}
\bibfield{author}{\bibinfo{person}{Zidong Du}, \bibinfo{person}{Robert
  Fasthuber}, \bibinfo{person}{Tianshi Chen}, \bibinfo{person}{Paolo Ienne},
  \bibinfo{person}{Ling Li}, \bibinfo{person}{Tao Luo},
  \bibinfo{person}{Xiaobing Feng}, \bibinfo{person}{Yunji Chen}, {and}
  \bibinfo{person}{Olivier Temam}.} \bibinfo{year}{2015}\natexlab{}.
\newblock \showarticletitle{{ShiDianNao: Shifting Vision Processing Closer to
  the Sensor}}. In \bibinfo{booktitle}{\emph{International Symposium on
  Computer Architecture}}.
\newblock


\bibitem[\protect\citeauthoryear{Elkurdi, Fern{\'a}ndez, Souleimanov,
  Giannacopoulos, and Gross}{Elkurdi et~al\mbox{.}}{2008}]%
        {elkurdi2008fpga}
\bibfield{author}{\bibinfo{person}{Yousef Elkurdi}, \bibinfo{person}{David
  Fern{\'a}ndez}, \bibinfo{person}{Evgueni Souleimanov},
  \bibinfo{person}{Dennis Giannacopoulos}, {and} \bibinfo{person}{Warren~J
  Gross}.} \bibinfo{year}{2008}\natexlab{}.
\newblock \showarticletitle{{FPGA Architecture and Implementation of Sparse
  Matrix-Vector Multiplication for the Finite Element Method}}.
\newblock \bibinfo{journal}{\emph{Computer Physics Communications}}
  \bibinfo{volume}{178}, \bibinfo{number}{8} (\bibinfo{year}{2008}),
  \bibinfo{pages}{558--570}.
\newblock


\bibitem[\protect\citeauthoryear{Elsken, Metzen, and Hutter}{Elsken
  et~al\mbox{.}}{2018}]%
        {elsken2018efficient}
\bibfield{author}{\bibinfo{person}{Thomas Elsken}, \bibinfo{person}{Jan~Hendrik
  Metzen}, {and} \bibinfo{person}{Frank Hutter}.}
  \bibinfo{year}{2018}\natexlab{}.
\newblock \showarticletitle{{Efficient Multi-Objective Neural Architecture
  Search via Lamarckian Evolution}}. In \bibinfo{booktitle}{\emph{International
  Conference on Learning Representations}}.
\newblock


\bibitem[\protect\citeauthoryear{Elsken, Metzen, and Hutter}{Elsken
  et~al\mbox{.}}{2019}]%
        {elsken2019neural}
\bibfield{author}{\bibinfo{person}{Thomas Elsken}, \bibinfo{person}{Jan~Hendrik
  Metzen}, {and} \bibinfo{person}{Frank Hutter}.}
  \bibinfo{year}{2019}\natexlab{}.
\newblock \showarticletitle{{Neural Architecture Search: A Survey}}.
\newblock \bibinfo{journal}{\emph{Journal of Machine Learning Research}}
  \bibinfo{volume}{20}, \bibinfo{number}{55} (\bibinfo{year}{2019}),
  \bibinfo{pages}{1--21}.
\newblock


\bibitem[\protect\citeauthoryear{Engelbrecht}{Engelbrecht}{2001}]%
        {engelbrecht2001new}
\bibfield{author}{\bibinfo{person}{Andries~Petrus Engelbrecht}.}
  \bibinfo{year}{2001}\natexlab{}.
\newblock \showarticletitle{{A New Pruning Heuristic Based on Variance Analysis
  of Sensitivity Information}}.
\newblock \bibinfo{journal}{\emph{IEEE Transactions on Neural Networks}}
  \bibinfo{volume}{12}, \bibinfo{number}{6} (\bibinfo{year}{2001}),
  \bibinfo{pages}{1386--1399}.
\newblock


\bibitem[\protect\citeauthoryear{Fan, Chen, Kuehne, Pistoia, and Cox}{Fan
  et~al\mbox{.}}{2019}]%
        {fan2019more}
\bibfield{author}{\bibinfo{person}{Quanfu Fan}, \bibinfo{person}{Chun-Fu Chen},
  \bibinfo{person}{Hilde Kuehne}, \bibinfo{person}{Marco Pistoia}, {and}
  \bibinfo{person}{David Cox}.} \bibinfo{year}{2019}\natexlab{}.
\newblock \showarticletitle{{More Is Less: Learning Efficient Video
  Representations by Big-Little Network and Depthwise Temporal Aggregation}}.
  In \bibinfo{booktitle}{\emph{Conference on Neural Information Processing
  Systems}}.
\newblock


\bibitem[\protect\citeauthoryear{Fedorov, Adams, Mattina, and
  Whatmough}{Fedorov et~al\mbox{.}}{2019}]%
        {fedorov2019sparse}
\bibfield{author}{\bibinfo{person}{Igor Fedorov}, \bibinfo{person}{Ryan~P
  Adams}, \bibinfo{person}{Matthew Mattina}, {and} \bibinfo{person}{Paul~N
  Whatmough}.} \bibinfo{year}{2019}\natexlab{}.
\newblock \showarticletitle{{SpArSe: Sparse Architecture Search for CNNs on
  Resource-Constrained Microcontrollers}}. In
  \bibinfo{booktitle}{\emph{Conference on Neural Information Processing
  Systems}}.
\newblock


\bibitem[\protect\citeauthoryear{Fedus, Zoph, and Shazeer}{Fedus
  et~al\mbox{.}}{2021}]%
        {fedus2021switch}
\bibfield{author}{\bibinfo{person}{William Fedus}, \bibinfo{person}{Barret
  Zoph}, {and} \bibinfo{person}{Noam Shazeer}.}
  \bibinfo{year}{2021}\natexlab{}.
\newblock \showarticletitle{{Switch Transformers: Scaling to Trillion Parameter
  Models with Simple and Efficient Sparsity}}.
\newblock \bibinfo{journal}{\emph{arXiv preprint arXiv:2101.03961}}
  (\bibinfo{year}{2021}).
\newblock


\bibitem[\protect\citeauthoryear{Feichtenhofer}{Feichtenhofer}{2020}]%
        {feichtenhofer2020x3d}
\bibfield{author}{\bibinfo{person}{Christoph Feichtenhofer}.}
  \bibinfo{year}{2020}\natexlab{}.
\newblock \showarticletitle{{X3D: Expanding Architectures for Efficient Video
  Recognition}}. In \bibinfo{booktitle}{\emph{IEEE Conference on Computer
  Vision and Pattern Recognition}}.
\newblock


\bibitem[\protect\citeauthoryear{Feichtenhofer, Pinz, and Wildes}{Feichtenhofer
  et~al\mbox{.}}{2016a}]%
        {feichtenhofer2016spatiotemporal}
\bibfield{author}{\bibinfo{person}{Christoph Feichtenhofer},
  \bibinfo{person}{Axel Pinz}, {and} \bibinfo{person}{Richard Wildes}.}
  \bibinfo{year}{2016}\natexlab{a}.
\newblock \showarticletitle{{Spatiotemporal Residual Networks for Video Action
  Recognition}}. In \bibinfo{booktitle}{\emph{Conference on Neural Information
  Processing Systems}}.
\newblock


\bibitem[\protect\citeauthoryear{Feichtenhofer, Pinz, and
  Zisserman}{Feichtenhofer et~al\mbox{.}}{2016b}]%
        {feichtenhofer2016convolutional}
\bibfield{author}{\bibinfo{person}{Christoph Feichtenhofer},
  \bibinfo{person}{Axel Pinz}, {and} \bibinfo{person}{Andrew Zisserman}.}
  \bibinfo{year}{2016}\natexlab{b}.
\newblock \showarticletitle{{Convolutional Two-Stream Network Fusion for Video
  Action Recognition}}. In \bibinfo{booktitle}{\emph{IEEE Conference on
  Computer Vision and Pattern Recognition}}.
\newblock


\bibitem[\protect\citeauthoryear{Frankle and Carbin}{Frankle and
  Carbin}{2018}]%
        {frankle2018lottery}
\bibfield{author}{\bibinfo{person}{Jonathan Frankle} {and}
  \bibinfo{person}{Michael Carbin}.} \bibinfo{year}{2018}\natexlab{}.
\newblock \showarticletitle{{The Lottery Ticket Hypothesis: Finding Sparse,
  Trainable Neural Networks}}. In \bibinfo{booktitle}{\emph{International
  Conference on Learning Representations}}.
\newblock


\bibitem[\protect\citeauthoryear{Frankle, Dziugaite, Roy, and Carbin}{Frankle
  et~al\mbox{.}}{2020a}]%
        {frankle2020linear}
\bibfield{author}{\bibinfo{person}{Jonathan Frankle},
  \bibinfo{person}{Gintare~Karolina Dziugaite}, \bibinfo{person}{Daniel Roy},
  {and} \bibinfo{person}{Michael Carbin}.} \bibinfo{year}{2020}\natexlab{a}.
\newblock \showarticletitle{{Linear Mode Connectivity and the Lottery Ticket
  Hypothesis}}. In \bibinfo{booktitle}{\emph{International Conference on
  Machine Learning}}.
\newblock


\bibitem[\protect\citeauthoryear{Frankle, Schwab, and Morcos}{Frankle
  et~al\mbox{.}}{2020b}]%
        {frankle2020training}
\bibfield{author}{\bibinfo{person}{Jonathan Frankle}, \bibinfo{person}{David~J
  Schwab}, {and} \bibinfo{person}{Ari~S Morcos}.}
  \bibinfo{year}{2020}\natexlab{b}.
\newblock \showarticletitle{{Training BatchNorm and Only BatchNorm: On the
  Expressive Power of Random Features in CNNs}}.
\newblock \bibinfo{journal}{\emph{arXiv preprint arXiv:2003.00152}}
  (\bibinfo{year}{2020}).
\newblock


\bibitem[\protect\citeauthoryear{Gan, Wang, Yang, Yeung, and Hauptmann}{Gan
  et~al\mbox{.}}{2015}]%
        {gan2015devnet}
\bibfield{author}{\bibinfo{person}{Chuang Gan}, \bibinfo{person}{Naiyan Wang},
  \bibinfo{person}{Yi Yang}, \bibinfo{person}{Dit-Yan Yeung}, {and}
  \bibinfo{person}{Alex~G Hauptmann}.} \bibinfo{year}{2015}\natexlab{}.
\newblock \showarticletitle{{DevNet: A Deep Event Network for Multimedia Event
  Detection and Evidence Recounting}}. In \bibinfo{booktitle}{\emph{IEEE
  Conference on Computer Vision and Pattern Recognition}}.
\newblock


\bibitem[\protect\citeauthoryear{Gao, Pu, Yang, Horowitz, and Kozyrakis}{Gao
  et~al\mbox{.}}{2017}]%
        {gao2017tetris}
\bibfield{author}{\bibinfo{person}{Mingyu Gao}, \bibinfo{person}{Jing Pu},
  \bibinfo{person}{Xuan Yang}, \bibinfo{person}{Mark Horowitz}, {and}
  \bibinfo{person}{Christos Kozyrakis}.} \bibinfo{year}{2017}\natexlab{}.
\newblock \showarticletitle{{TETRIS: Scalable and Efficient Neural Network
  Acceleration with 3D Memory}}. In \bibinfo{booktitle}{\emph{International
  Conference on Architectural Support for Programming Languages and Operating
  Systems}}.
\newblock


\bibitem[\protect\citeauthoryear{Ghiasi, Lin, and Le}{Ghiasi
  et~al\mbox{.}}{2019}]%
        {ghiasi2019fpn}
\bibfield{author}{\bibinfo{person}{Golnaz Ghiasi}, \bibinfo{person}{Tsung-Yi
  Lin}, {and} \bibinfo{person}{Quoc~V Le}.} \bibinfo{year}{2019}\natexlab{}.
\newblock \showarticletitle{{NAS-FPN: Learning Scalable Feature Pyramid
  Architecture for Object Detection}}. In \bibinfo{booktitle}{\emph{IEEE
  Conference on Computer Vision and Pattern Recognition}}.
\newblock


\bibitem[\protect\citeauthoryear{Giles and Omlin}{Giles and Omlin}{1994}]%
        {giles1994pruning}
\bibfield{author}{\bibinfo{person}{C~Lee Giles} {and}
  \bibinfo{person}{Christian~W Omlin}.} \bibinfo{year}{1994}\natexlab{}.
\newblock \showarticletitle{{Pruning Recurrent Neural Networks for Improved
  Generalization Performance}}.
\newblock \bibinfo{journal}{\emph{IEEE Transactions on Neural Networks}}
  \bibinfo{volume}{5}, \bibinfo{number}{5} (\bibinfo{year}{1994}),
  \bibinfo{pages}{848--851}.
\newblock


\bibitem[\protect\citeauthoryear{Girshick}{Girshick}{2015}]%
        {girshick2015fast}
\bibfield{author}{\bibinfo{person}{Ross Girshick}.}
  \bibinfo{year}{2015}\natexlab{}.
\newblock \showarticletitle{{Fast R-CNN}}. In
  \bibinfo{booktitle}{\emph{International Conference on Computer Vision}}.
\newblock


\bibitem[\protect\citeauthoryear{Golub and Van~Loan}{Golub and
  Van~Loan}{1996}]%
        {golub1996matrix}
\bibfield{author}{\bibinfo{person}{Gene~H Golub} {and}
  \bibinfo{person}{Charles~F Van~Loan}.} \bibinfo{year}{1996}\natexlab{}.
\newblock \showarticletitle{{Matrix Computations}}.
\newblock  (\bibinfo{year}{1996}).
\newblock


\bibitem[\protect\citeauthoryear{Gong, Liu, Yang, and Bourdev}{Gong
  et~al\mbox{.}}{2014}]%
        {gong2014compressing}
\bibfield{author}{\bibinfo{person}{Yunchao Gong}, \bibinfo{person}{Liu Liu},
  \bibinfo{person}{Ming Yang}, {and} \bibinfo{person}{Lubomir Bourdev}.}
  \bibinfo{year}{2014}\natexlab{}.
\newblock \showarticletitle{{Compressing Deep Convolutional Networks using
  Vector Quantization}}.
\newblock \bibinfo{journal}{\emph{arXiv preprint arXiv:1412.6115}}
  (\bibinfo{year}{2014}).
\newblock


\bibitem[\protect\citeauthoryear{Gordon, Duh, and Andrews}{Gordon
  et~al\mbox{.}}{2020}]%
        {gordon2020compressing}
\bibfield{author}{\bibinfo{person}{Mitchell~A Gordon}, \bibinfo{person}{Kevin
  Duh}, {and} \bibinfo{person}{Nicholas Andrews}.}
  \bibinfo{year}{2020}\natexlab{}.
\newblock \showarticletitle{{Compressing BERT: Studying the Effects of Weight
  Pruning on Transfer Learning}}.
\newblock \bibinfo{journal}{\emph{arXiv preprint arXiv:2002.08307}}
  (\bibinfo{year}{2020}).
\newblock


\bibitem[\protect\citeauthoryear{Goyal}{Goyal}{2020}]%
        {goyal2020power}
\bibfield{author}{\bibinfo{person}{Saurabh~others Goyal}.}
  \bibinfo{year}{2020}\natexlab{}.
\newblock \showarticletitle{{PoWER-BERT: Accelerating BERT inference for
  Classification Tasks}}. In \bibinfo{booktitle}{\emph{International Conference
  on Machine Learning}}.
\newblock


\bibitem[\protect\citeauthoryear{Graham}{Graham}{2015}]%
        {graham2015sparse}
\bibfield{author}{\bibinfo{person}{Benjamin Graham}.}
  \bibinfo{year}{2015}\natexlab{}.
\newblock \showarticletitle{{Sparse 3D Convolutional Neural Networks}}. In
  \bibinfo{booktitle}{\emph{British Machine Vision Conference}}.
\newblock


\bibitem[\protect\citeauthoryear{Graham, Engelcke, and van~der Maaten}{Graham
  et~al\mbox{.}}{2018}]%
        {graham20183d}
\bibfield{author}{\bibinfo{person}{Benjamin Graham}, \bibinfo{person}{Martin
  Engelcke}, {and} \bibinfo{person}{Laurens van~der Maaten}.}
  \bibinfo{year}{2018}\natexlab{}.
\newblock \showarticletitle{{3D Semantic Segmentation With Submanifold Sparse
  Convolutional Networks}}. In \bibinfo{booktitle}{\emph{IEEE Conference on
  Computer Vision and Pattern Recognition}}.
\newblock


\bibitem[\protect\citeauthoryear{Grigora{\c{s}}, Burovskiy, Luk, and
  Sherwin}{Grigora{\c{s}} et~al\mbox{.}}{2016}]%
        {grigoracs2016optimising}
\bibfield{author}{\bibinfo{person}{Paul Grigora{\c{s}}}, \bibinfo{person}{Pavel
  Burovskiy}, \bibinfo{person}{Wayne Luk}, {and} \bibinfo{person}{Spencer
  Sherwin}.} \bibinfo{year}{2016}\natexlab{}.
\newblock \showarticletitle{{Optimising Sparse Matrix Vector Multiplication for
  Large Scale FEM Problems on FPGA}}. In
  \bibinfo{booktitle}{\emph{International Conference on Field Programmable
  Logic and Applications}}.
\newblock


\bibitem[\protect\citeauthoryear{Gruslys, Munos, Danihelka, Lanctot, and
  Graves}{Gruslys et~al\mbox{.}}{2016}]%
        {gruslys2016memory}
\bibfield{author}{\bibinfo{person}{Audrunas Gruslys}, \bibinfo{person}{R{\'e}mi
  Munos}, \bibinfo{person}{Ivo Danihelka}, \bibinfo{person}{Marc Lanctot},
  {and} \bibinfo{person}{Alex Graves}.} \bibinfo{year}{2016}\natexlab{}.
\newblock \showarticletitle{{Memory-Efficient Backpropagation Through Time}}.
  In \bibinfo{booktitle}{\emph{Conference on Neural Information Processing
  Systems}}.
\newblock


\bibitem[\protect\citeauthoryear{Gu, Bradbury, Xiong, Li, and Socher}{Gu
  et~al\mbox{.}}{2018}]%
        {gu2018nonautoregressive}
\bibfield{author}{\bibinfo{person}{Jiatao Gu}, \bibinfo{person}{James
  Bradbury}, \bibinfo{person}{Caiming Xiong}, \bibinfo{person}{Victor~OK Li},
  {and} \bibinfo{person}{Richard Socher}.} \bibinfo{year}{2018}\natexlab{}.
\newblock \showarticletitle{{Non-Autoregressive Neural Machine Translation}}.
  In \bibinfo{booktitle}{\emph{International Conference on Learning
  Representations}}.
\newblock


\bibitem[\protect\citeauthoryear{Gu, Wang, and Zhao}{Gu et~al\mbox{.}}{2019}]%
        {gu2019levenshtein}
\bibfield{author}{\bibinfo{person}{Jiatao Gu}, \bibinfo{person}{Changhan Wang},
  {and} \bibinfo{person}{Jake Zhao}.} \bibinfo{year}{2019}\natexlab{}.
\newblock \showarticletitle{{Levenshtein Transformer}}. In
  \bibinfo{booktitle}{\emph{Conference on Neural Information Processing
  Systems}}.
\newblock


\bibitem[\protect\citeauthoryear{Guo, Peng, Zhang, Chen, and LeCompte}{Guo
  et~al\mbox{.}}{2020a}]%
        {guo2020att}
\bibfield{author}{\bibinfo{person}{Haoqiang Guo}, \bibinfo{person}{Lu Peng},
  \bibinfo{person}{Jian Zhang}, \bibinfo{person}{Qing Chen}, {and}
  \bibinfo{person}{Travis~D LeCompte}.} \bibinfo{year}{2020}\natexlab{a}.
\newblock \showarticletitle{{ATT: A Fault-Tolerant ReRAM Accelerator for
  Attention-Based Neural Networks}}. In \bibinfo{booktitle}{\emph{International
  Conference on Computer Design}}.
\newblock


\bibitem[\protect\citeauthoryear{Guo, Yao, and Chen}{Guo et~al\mbox{.}}{2016}]%
        {guo2016dynamic}
\bibfield{author}{\bibinfo{person}{Yiwen Guo}, \bibinfo{person}{Anbang Yao},
  {and} \bibinfo{person}{Yurong Chen}.} \bibinfo{year}{2016}\natexlab{}.
\newblock \showarticletitle{{Dynamic Network Surgery for Efficient DNNs}}. In
  \bibinfo{booktitle}{\emph{Conference on Neural Information Processing
  Systems}}.
\newblock


\bibitem[\protect\citeauthoryear{Guo, Zhang, Mu, Heng, Liu, Wei, and Sun}{Guo
  et~al\mbox{.}}{2020b}]%
        {guo2020single}
\bibfield{author}{\bibinfo{person}{Zichao Guo}, \bibinfo{person}{Xiangyu
  Zhang}, \bibinfo{person}{Haoyuan Mu}, \bibinfo{person}{Wen Heng},
  \bibinfo{person}{Zechun Liu}, \bibinfo{person}{Yichen Wei}, {and}
  \bibinfo{person}{Jian Sun}.} \bibinfo{year}{2020}\natexlab{b}.
\newblock \showarticletitle{{Single Path One-Shot Neural Architecture Search
  with Uniform Sampling}}. In \bibinfo{booktitle}{\emph{European Conference on
  Computer Vision}}.
\newblock


\bibitem[\protect\citeauthoryear{Gupta, Agrawal, Gopalakrishnan, and
  Narayanan}{Gupta et~al\mbox{.}}{2015}]%
        {gupta2015deep}
\bibfield{author}{\bibinfo{person}{Suyog Gupta}, \bibinfo{person}{Ankur
  Agrawal}, \bibinfo{person}{Kailash Gopalakrishnan}, {and}
  \bibinfo{person}{Pritish Narayanan}.} \bibinfo{year}{2015}\natexlab{}.
\newblock \showarticletitle{{Deep Learning with Limited Numerical Precision}}.
  In \bibinfo{booktitle}{\emph{International Conference on Machine Learning}}.
\newblock


\bibitem[\protect\citeauthoryear{Ham, Jung, Kim, Oh, Park, Song, Park, Lee,
  Park, Lee, and Jeong}{Ham et~al\mbox{.}}{2020}]%
        {ham20203}
\bibfield{author}{\bibinfo{person}{Tae~Jun Ham}, \bibinfo{person}{Sung~Jun
  Jung}, \bibinfo{person}{Seonghak Kim}, \bibinfo{person}{Young~H Oh},
  \bibinfo{person}{Yeonhong Park}, \bibinfo{person}{Yoonho Song},
  \bibinfo{person}{Jung-Hun Park}, \bibinfo{person}{Sanghee Lee},
  \bibinfo{person}{Kyoung Park}, \bibinfo{person}{Jae~W Lee}, {and}
  \bibinfo{person}{Deog-Kyoon Jeong}.} \bibinfo{year}{2020}\natexlab{}.
\newblock \showarticletitle{{A3: Accelerating Attention Mechanisms in Neural
  Networks with Approximation}}. In \bibinfo{booktitle}{\emph{IEEE
  International Symposium on High-Performance Computer Architecture}}.
\newblock


\bibitem[\protect\citeauthoryear{Han, Kim, and Kim}{Han et~al\mbox{.}}{2017b}]%
        {han2017deep}
\bibfield{author}{\bibinfo{person}{Dongyoon Han}, \bibinfo{person}{Jiwhan Kim},
  {and} \bibinfo{person}{Junmo Kim}.} \bibinfo{year}{2017}\natexlab{b}.
\newblock \showarticletitle{{Deep Pyramidal Residual Networks}}. In
  \bibinfo{booktitle}{\emph{IEEE Conference on Computer Vision and Pattern
  Recognition}}.
\newblock


\bibitem[\protect\citeauthoryear{Han, Zheng, Xu, and Fang}{Han
  et~al\mbox{.}}{2020}]%
        {han2020occuseg}
\bibfield{author}{\bibinfo{person}{Lei Han}, \bibinfo{person}{Tian Zheng},
  \bibinfo{person}{Lan Xu}, {and} \bibinfo{person}{Lu Fang}.}
  \bibinfo{year}{2020}\natexlab{}.
\newblock \showarticletitle{{OccuSeg: Occupancy-Aware 3D Instance
  Segmentation}}. In \bibinfo{booktitle}{\emph{IEEE Conference on Computer
  Vision and Pattern Recognition}}.
\newblock


\bibitem[\protect\citeauthoryear{Han}{Han}{2017}]%
        {han2017thesis}
\bibfield{author}{\bibinfo{person}{Song Han}.} \bibinfo{year}{2017}\natexlab{}.
\newblock \emph{\bibinfo{title}{{Efficient Methods and Hardware for Deep
  Learning}}}.
\newblock \bibinfo{thesistype}{Ph.D. Dissertation}. \bibinfo{school}{Stanford
  University}.
\newblock


\bibitem[\protect\citeauthoryear{Han, Kang, Mao, Hu, Li, Li, Xie, Luo, Yao,
  Wang, Yang, and Dally}{Han et~al\mbox{.}}{2017a}]%
        {han2017ese}
\bibfield{author}{\bibinfo{person}{Song Han}, \bibinfo{person}{Junlong Kang},
  \bibinfo{person}{Huizi Mao}, \bibinfo{person}{Yiming Hu},
  \bibinfo{person}{Xin Li}, \bibinfo{person}{Yubin Li},
  \bibinfo{person}{Dongliang Xie}, \bibinfo{person}{Hong Luo},
  \bibinfo{person}{Song Yao}, \bibinfo{person}{Yu Wang},
  \bibinfo{person}{Huazhong Yang}, {and} \bibinfo{person}{William~J Dally}.}
  \bibinfo{year}{2017}\natexlab{a}.
\newblock \showarticletitle{{ESE: Efficient Speech Recognition Engine with
  Sparse LSTM on FPGA}}. In \bibinfo{booktitle}{\emph{ACM/SIGDA International
  Symposium on Field-Programmable Gate Arrays}}.
\newblock


\bibitem[\protect\citeauthoryear{Han, Liu, Mao, Pu, Pedram, Horowitz, and
  Dally}{Han et~al\mbox{.}}{2016a}]%
        {han2016eie}
\bibfield{author}{\bibinfo{person}{Song Han}, \bibinfo{person}{Xingyu Liu},
  \bibinfo{person}{Huizi Mao}, \bibinfo{person}{Jing Pu},
  \bibinfo{person}{Ardavan Pedram}, \bibinfo{person}{Mark~A Horowitz}, {and}
  \bibinfo{person}{William~J Dally}.} \bibinfo{year}{2016}\natexlab{a}.
\newblock \showarticletitle{{EIE: Efficient Inference Engine on Compressed Deep
  Neural Network}}. In \bibinfo{booktitle}{\emph{International Symposium on
  Computer Architecture}}.
\newblock


\bibitem[\protect\citeauthoryear{Han, Mao, and Dally}{Han
  et~al\mbox{.}}{2016b}]%
        {han2016deep}
\bibfield{author}{\bibinfo{person}{Song Han}, \bibinfo{person}{Huizi Mao},
  {and} \bibinfo{person}{William~J Dally}.} \bibinfo{year}{2016}\natexlab{b}.
\newblock \showarticletitle{{Deep Compression: Compressing Deep Neural Networks
  with Pruning, Trained Quantization and Huffman Coding}}. In
  \bibinfo{booktitle}{\emph{International Conference on Learning
  Representations}}.
\newblock


\bibitem[\protect\citeauthoryear{Han, Pool, Tran, and Dally}{Han
  et~al\mbox{.}}{2015}]%
        {han2015learning}
\bibfield{author}{\bibinfo{person}{Song Han}, \bibinfo{person}{Jeff Pool},
  \bibinfo{person}{John Tran}, {and} \bibinfo{person}{William~J Dally}.}
  \bibinfo{year}{2015}\natexlab{}.
\newblock \showarticletitle{{Learning Both Weights and Connections for
  Efficient Neural Networks}}. In \bibinfo{booktitle}{\emph{Conference on
  Neural Information Processing Systems}}.
\newblock


\bibitem[\protect\citeauthoryear{Haque, Guo, Alahi, Yeung, Luo, Rege, Jopling,
  Downing, Beninati, Singh, Platchek, Milstein, and Fei-Fei}{Haque
  et~al\mbox{.}}{2017}]%
        {haque2017towards}
\bibfield{author}{\bibinfo{person}{Albert Haque}, \bibinfo{person}{Michelle
  Guo}, \bibinfo{person}{Alexandre Alahi}, \bibinfo{person}{Serena Yeung},
  \bibinfo{person}{Zelun Luo}, \bibinfo{person}{Alisha Rege},
  \bibinfo{person}{Jeffrey Jopling}, \bibinfo{person}{Lance Downing},
  \bibinfo{person}{William Beninati}, \bibinfo{person}{Amit Singh},
  \bibinfo{person}{Terry Platchek}, \bibinfo{person}{Arnold Milstein}, {and}
  \bibinfo{person}{Li Fei-Fei}.} \bibinfo{year}{2017}\natexlab{}.
\newblock \showarticletitle{{Towards Vision-Based Smart Hospitals: A System for
  Tracking and Monitoring Hand Hygiene Compliance}}. In
  \bibinfo{booktitle}{\emph{Machine Learning for Healthcare Conference}}.
\newblock


\bibitem[\protect\citeauthoryear{Hassibi and Stork}{Hassibi and Stork}{1993}]%
        {hassibi1993second}
\bibfield{author}{\bibinfo{person}{Babak Hassibi} {and}
  \bibinfo{person}{David~G Stork}.} \bibinfo{year}{1993}\natexlab{}.
\newblock \showarticletitle{{Second Order Derivatives for Network Pruning:
  Optimal Brain Surgeon}}. In \bibinfo{booktitle}{\emph{Conference on Neural
  Information Processing Systems}}.
\newblock


\bibitem[\protect\citeauthoryear{He, Zhang, Ren, and Sun}{He
  et~al\mbox{.}}{2016}]%
        {he2016deep}
\bibfield{author}{\bibinfo{person}{Kaiming He}, \bibinfo{person}{Xiangyu
  Zhang}, \bibinfo{person}{Shaoqing Ren}, {and} \bibinfo{person}{Jian Sun}.}
  \bibinfo{year}{2016}\natexlab{}.
\newblock \showarticletitle{{Deep Residual Learning for Image Recognition}}. In
  \bibinfo{booktitle}{\emph{IEEE Conference on Computer Vision and Pattern
  Recognition}}.
\newblock


\bibitem[\protect\citeauthoryear{He, Zhao, and Chu}{He et~al\mbox{.}}{2021}]%
        {he2021automl}
\bibfield{author}{\bibinfo{person}{Xin He}, \bibinfo{person}{Kaiyong Zhao},
  {and} \bibinfo{person}{Xiaowen Chu}.} \bibinfo{year}{2021}\natexlab{}.
\newblock \showarticletitle{{AutoML: A Survey of the State-of-the-Art}}.
\newblock \bibinfo{journal}{\emph{Knowledge-Based Systems}}
  \bibinfo{volume}{212} (\bibinfo{year}{2021}), \bibinfo{pages}{106622}.
\newblock


\bibitem[\protect\citeauthoryear{He, Lin, Liu, Wang, Li, and Han}{He
  et~al\mbox{.}}{2018}]%
        {he2018amc}
\bibfield{author}{\bibinfo{person}{Yihui He}, \bibinfo{person}{Ji Lin},
  \bibinfo{person}{Zhijian Liu}, \bibinfo{person}{Hanrui Wang},
  \bibinfo{person}{Li-Jia Li}, {and} \bibinfo{person}{Song Han}.}
  \bibinfo{year}{2018}\natexlab{}.
\newblock \showarticletitle{{AMC: AutoML for Model Compression and Acceleration
  on Mobile Devices}}. In \bibinfo{booktitle}{\emph{European Conference on
  Computer Vision}}.
\newblock


\bibitem[\protect\citeauthoryear{He, Zhang, and Sun}{He et~al\mbox{.}}{2017}]%
        {he2017channel}
\bibfield{author}{\bibinfo{person}{Yihui He}, \bibinfo{person}{Xiangyu Zhang},
  {and} \bibinfo{person}{Jian Sun}.} \bibinfo{year}{2017}\natexlab{}.
\newblock \showarticletitle{{Channel Pruning for Accelerating Very Deep Neural
  Networks}}. In \bibinfo{booktitle}{\emph{International Conference on Computer
  Vision}}.
\newblock


\bibitem[\protect\citeauthoryear{Hinton, Deng, Yu, Dahl, Mohamed, Jaitly,
  Senior, Vanhoucke, Nguyen, Sainath, and Kingsbury}{Hinton
  et~al\mbox{.}}{2012}]%
        {hinton2012deep}
\bibfield{author}{\bibinfo{person}{Geoffrey Hinton}, \bibinfo{person}{Li Deng},
  \bibinfo{person}{Dong Yu}, \bibinfo{person}{George~E Dahl},
  \bibinfo{person}{Abdel-rahman Mohamed}, \bibinfo{person}{Navdeep Jaitly},
  \bibinfo{person}{Andrew Senior}, \bibinfo{person}{Vincent Vanhoucke},
  \bibinfo{person}{Patrick Nguyen}, \bibinfo{person}{Tara~N Sainath}, {and}
  \bibinfo{person}{Brian Kingsbury}.} \bibinfo{year}{2012}\natexlab{}.
\newblock \showarticletitle{{Deep Neural Networks for Acoustic Modeling in
  Speech Recognition: The Shared Views of Four Research Groups}}.
\newblock \bibinfo{journal}{\emph{IEEE Signal Processing Magazine}}
  \bibinfo{volume}{29}, \bibinfo{number}{6} (\bibinfo{year}{2012}),
  \bibinfo{pages}{82--97}.
\newblock


\bibitem[\protect\citeauthoryear{Hinton, Vinyals, and Dean}{Hinton
  et~al\mbox{.}}{2015}]%
        {hinton2015distilling}
\bibfield{author}{\bibinfo{person}{Geoffrey Hinton}, \bibinfo{person}{Oriol
  Vinyals}, {and} \bibinfo{person}{Jeff Dean}.}
  \bibinfo{year}{2015}\natexlab{}.
\newblock \showarticletitle{{Distilling the Knowledge in a Neural Network}}.
\newblock \bibinfo{journal}{\emph{arXiv preprint arXiv:1503.02531}}
  (\bibinfo{year}{2015}).
\newblock


\bibitem[\protect\citeauthoryear{Howard, Sandler, Chu, Chen, Chen, Tan, Wang,
  Zhu, Pang, Vasudevan, Le, and Adam}{Howard et~al\mbox{.}}{2019}]%
        {howard2019searching}
\bibfield{author}{\bibinfo{person}{Andrew Howard}, \bibinfo{person}{Mark
  Sandler}, \bibinfo{person}{Grace Chu}, \bibinfo{person}{Liang-Chieh Chen},
  \bibinfo{person}{Bo Chen}, \bibinfo{person}{Mingxing Tan},
  \bibinfo{person}{Weijun Wang}, \bibinfo{person}{Yukun Zhu},
  \bibinfo{person}{Ruoming Pang}, \bibinfo{person}{Vijay Vasudevan},
  \bibinfo{person}{Quoc~V Le}, {and} \bibinfo{person}{Hartwig Adam}.}
  \bibinfo{year}{2019}\natexlab{}.
\newblock \showarticletitle{{Searching for MobileNetV3}}. In
  \bibinfo{booktitle}{\emph{International Conference on Computer Vision}}.
\newblock


\bibitem[\protect\citeauthoryear{Howard, Zhu, Chen, Kalenichenko, Wang, Weyand,
  Andreetto, and Adam}{Howard et~al\mbox{.}}{2017}]%
        {howard2017mobilenets}
\bibfield{author}{\bibinfo{person}{Andrew~G Howard}, \bibinfo{person}{Menglong
  Zhu}, \bibinfo{person}{Bo Chen}, \bibinfo{person}{Dimitry Kalenichenko},
  \bibinfo{person}{Weijun Wang}, \bibinfo{person}{Tobias Weyand},
  \bibinfo{person}{Marco Andreetto}, {and} \bibinfo{person}{Hartwig Adam}.}
  \bibinfo{year}{2017}\natexlab{}.
\newblock \showarticletitle{{MobileNets: Efficient Convolutional Neural
  Networks for Mobile Vision Applications}}.
\newblock \bibinfo{journal}{\emph{arXiv preprint arXiv:1704.04861}}
  (\bibinfo{year}{2017}).
\newblock


\bibitem[\protect\citeauthoryear{Hu, Lu, Li, and Chen}{Hu
  et~al\mbox{.}}{2014}]%
        {hu2014convolutional}
\bibfield{author}{\bibinfo{person}{Baotian Hu}, \bibinfo{person}{Zhengdong Lu},
  \bibinfo{person}{Hang Li}, {and} \bibinfo{person}{Qingcai Chen}.}
  \bibinfo{year}{2014}\natexlab{}.
\newblock \showarticletitle{{Convolutional Neural Network Architectures for
  Matching Natural Language Sentences}}. In
  \bibinfo{booktitle}{\emph{Conference on Neural Information Processing
  Systems}}.
\newblock


\bibitem[\protect\citeauthoryear{Hu, Yang, Xie, Rosa, Guo, Wang, Trigoni, and
  Markham}{Hu et~al\mbox{.}}{2020}]%
        {hu2019randla}
\bibfield{author}{\bibinfo{person}{Qingyong Hu}, \bibinfo{person}{Bo Yang},
  \bibinfo{person}{Linhai Xie}, \bibinfo{person}{Stefano Rosa},
  \bibinfo{person}{Yulan Guo}, \bibinfo{person}{Zhihua Wang},
  \bibinfo{person}{Niki Trigoni}, {and} \bibinfo{person}{Andrew Markham}.}
  \bibinfo{year}{2020}\natexlab{}.
\newblock \showarticletitle{{RandLA-Net: Efficient Semantic Segmentation of
  Large-Scale Point Clouds}}. In \bibinfo{booktitle}{\emph{IEEE Conference on
  Computer Vision and Pattern Recognition}}.
\newblock


\bibitem[\protect\citeauthoryear{Huang, Chen, Li, Wu, van~der Maaten, and
  Weinberger}{Huang et~al\mbox{.}}{2018a}]%
        {huang2017multi}
\bibfield{author}{\bibinfo{person}{Gao Huang}, \bibinfo{person}{Danlu Chen},
  \bibinfo{person}{Tianhong Li}, \bibinfo{person}{Felix Wu},
  \bibinfo{person}{Laurens van~der Maaten}, {and} \bibinfo{person}{Kilian~Q
  Weinberger}.} \bibinfo{year}{2018}\natexlab{a}.
\newblock \showarticletitle{{Multi-Scale Dense Networks for Resource Efficient
  Image Classification}}. In \bibinfo{booktitle}{\emph{International Conference
  on Learning Representations}}.
\newblock


\bibitem[\protect\citeauthoryear{Huang, Wang, and Neumann}{Huang
  et~al\mbox{.}}{2018b}]%
        {huang2018recurrent}
\bibfield{author}{\bibinfo{person}{Qiangui Huang}, \bibinfo{person}{Weiyue
  Wang}, {and} \bibinfo{person}{Ulrich Neumann}.}
  \bibinfo{year}{2018}\natexlab{b}.
\newblock \showarticletitle{{Recurrent Slice Networks for 3D Segmentation on
  Point Clouds}}. In \bibinfo{booktitle}{\emph{IEEE Conference on Computer
  Vision and Pattern Recognition}}.
\newblock


\bibitem[\protect\citeauthoryear{Iandola, Han, Moskewicz, Ashraf, Dally, and
  Keutzer}{Iandola et~al\mbox{.}}{2016}]%
        {iandola2016squeezenet}
\bibfield{author}{\bibinfo{person}{Forrest~N Iandola}, \bibinfo{person}{Song
  Han}, \bibinfo{person}{Matthew~W Moskewicz}, \bibinfo{person}{Khalid Ashraf},
  \bibinfo{person}{William~J Dally}, {and} \bibinfo{person}{Kurt Keutzer}.}
  \bibinfo{year}{2016}\natexlab{}.
\newblock \showarticletitle{{SqueezeNet: AlexNet-Level Accuracy with 50$\times$
  Fewer Parameters and $<$0.5MB Model Size}}.
\newblock \bibinfo{journal}{\emph{arXiv preprint arXiv:1602.07360}}
  (\bibinfo{year}{2016}).
\newblock


\bibitem[\protect\citeauthoryear{Ignatov, Timofte, Chou, Wang, Wu, Hartley, and
  Van~Gool}{Ignatov et~al\mbox{.}}{2018}]%
        {ignatov2018ai}
\bibfield{author}{\bibinfo{person}{Andrey Ignatov}, \bibinfo{person}{Radu
  Timofte}, \bibinfo{person}{William Chou}, \bibinfo{person}{Ke Wang},
  \bibinfo{person}{Max Wu}, \bibinfo{person}{Tim Hartley}, {and}
  \bibinfo{person}{Luc Van~Gool}.} \bibinfo{year}{2018}\natexlab{}.
\newblock \showarticletitle{{AI Benchmark: Running Deep Neural Networks on
  Android Smartphones}}.
\newblock \bibinfo{journal}{\emph{arXiv preprint arXiv:1810.01109}}
  (\bibinfo{year}{2018}).
\newblock


\bibitem[\protect\citeauthoryear{Ioffe and Szegedy}{Ioffe and Szegedy}{2015}]%
        {ioffe2015batch}
\bibfield{author}{\bibinfo{person}{Sergey Ioffe} {and}
  \bibinfo{person}{Christian Szegedy}.} \bibinfo{year}{2015}\natexlab{}.
\newblock \showarticletitle{{Batch Normalization: Accelerating Deep Network
  Training by Reducing Internal Covariate Shift}}. In
  \bibinfo{booktitle}{\emph{International Conference on Machine Learning}}.
\newblock


\bibitem[\protect\citeauthoryear{Jacob, Kligys, Chen, Zhu, Tang, Howard, Adam,
  and Kalenichenko}{Jacob et~al\mbox{.}}{2018}]%
        {jacob2018quantization}
\bibfield{author}{\bibinfo{person}{Benoit Jacob}, \bibinfo{person}{Skirmantas
  Kligys}, \bibinfo{person}{Bo Chen}, \bibinfo{person}{Menglong Zhu},
  \bibinfo{person}{Matthew Tang}, \bibinfo{person}{Andrew Howard},
  \bibinfo{person}{Hartwig Adam}, {and} \bibinfo{person}{Dmitry Kalenichenko}.}
  \bibinfo{year}{2018}\natexlab{}.
\newblock \showarticletitle{{Quantization and Training of Neural Networks for
  Efficient Integer-Arithmetic-Only Inference}}. In
  \bibinfo{booktitle}{\emph{IEEE Conference on Computer Vision and Pattern
  Recognition}}.
\newblock


\bibitem[\protect\citeauthoryear{Jaderberg, Vedaldi, and Zisserman}{Jaderberg
  et~al\mbox{.}}{2014}]%
        {jaderberg2014speeding}
\bibfield{author}{\bibinfo{person}{Max Jaderberg}, \bibinfo{person}{Andrea
  Vedaldi}, {and} \bibinfo{person}{Andrew Zisserman}.}
  \bibinfo{year}{2014}\natexlab{}.
\newblock \showarticletitle{{Speeding Up Convolutional Neural Networks with Low
  Rank Expansions}}. In \bibinfo{booktitle}{\emph{British Machine Vision
  Conference}}.
\newblock


\bibitem[\protect\citeauthoryear{Jamro, Pabi{\'s}, Russek, and Wiatr}{Jamro
  et~al\mbox{.}}{2015}]%
        {jamro2015algorithms}
\bibfield{author}{\bibinfo{person}{Ernest Jamro}, \bibinfo{person}{Tomasz
  Pabi{\'s}}, \bibinfo{person}{Pawe{\l} Russek}, {and}
  \bibinfo{person}{Kazimierz Wiatr}.} \bibinfo{year}{2015}\natexlab{}.
\newblock \showarticletitle{{The Algorithms for FPGA Implementation of Sparse
  Matrices Multiplication}}.
\newblock \bibinfo{journal}{\emph{Computing and Informatics}}
  \bibinfo{volume}{33}, \bibinfo{number}{3} (\bibinfo{year}{2015}),
  \bibinfo{pages}{667--684}.
\newblock


\bibitem[\protect\citeauthoryear{Jang, Kim, Jo, Lee, and Kim}{Jang
  et~al\mbox{.}}{2019}]%
        {jang2019mnnfast}
\bibfield{author}{\bibinfo{person}{Hanhwi Jang}, \bibinfo{person}{Joonsung
  Kim}, \bibinfo{person}{Jae-Eon Jo}, \bibinfo{person}{Jaewon Lee}, {and}
  \bibinfo{person}{Jangwoo Kim}.} \bibinfo{year}{2019}\natexlab{}.
\newblock \showarticletitle{{MnnFast: A Fast and Scalable System Architecture
  for Memory-Augmented Neural Networks}}. In
  \bibinfo{booktitle}{\emph{International Symposium on Computer Architecture}}.
\newblock


\bibitem[\protect\citeauthoryear{Jia, Padon, Thomas, Warszawski, Zaharia, and
  Aiken}{Jia et~al\mbox{.}}{2019}]%
        {jia2019taso}
\bibfield{author}{\bibinfo{person}{Zhihao Jia}, \bibinfo{person}{Oded Padon},
  \bibinfo{person}{James Thomas}, \bibinfo{person}{Todd Warszawski},
  \bibinfo{person}{Matei Zaharia}, {and} \bibinfo{person}{Alex Aiken}.}
  \bibinfo{year}{2019}\natexlab{}.
\newblock \showarticletitle{{TASO: Optimizing Deep Learning Computation with
  Automatic Generation of Graph Substitutions}}. In
  \bibinfo{booktitle}{\emph{ACM Symposium on Operating Systems Principles}}.
\newblock


\bibitem[\protect\citeauthoryear{Jia, Zaharia, and Aiken}{Jia
  et~al\mbox{.}}{2018}]%
        {jia2018beyond}
\bibfield{author}{\bibinfo{person}{Zhihao Jia}, \bibinfo{person}{Matei
  Zaharia}, {and} \bibinfo{person}{Alex Aiken}.}
  \bibinfo{year}{2018}\natexlab{}.
\newblock \showarticletitle{{Beyond Data and Model Parallelism for Deep Neural
  Networks}}. In \bibinfo{booktitle}{\emph{International Conference on Machine
  Learning}}.
\newblock


\bibitem[\protect\citeauthoryear{Jiang, Zhao, Shi, Liu, Fu, and Jia}{Jiang
  et~al\mbox{.}}{2020c}]%
        {jiang2020pointgroup}
\bibfield{author}{\bibinfo{person}{Li Jiang}, \bibinfo{person}{Hengshuang
  Zhao}, \bibinfo{person}{Shaoshuai Shi}, \bibinfo{person}{Shu Liu},
  \bibinfo{person}{Chi-Wing Fu}, {and} \bibinfo{person}{Jiaya Jia}.}
  \bibinfo{year}{2020}\natexlab{c}.
\newblock \showarticletitle{{PointGroup: Dual-Set Point Grouping for 3D
  Instance Segmentation}}. In \bibinfo{booktitle}{\emph{IEEE Conference on
  Computer Vision and Pattern Recognition}}.
\newblock


\bibitem[\protect\citeauthoryear{Jiang, Yang, Sha, Zhuge, Gu, Dasgupta, Shi,
  and Hu}{Jiang et~al\mbox{.}}{2020b}]%
        {jiang2020hardware}
\bibfield{author}{\bibinfo{person}{Weiwen Jiang}, \bibinfo{person}{Lei Yang},
  \bibinfo{person}{Edwin Hsing-Mean Sha}, \bibinfo{person}{Qingfeng Zhuge},
  \bibinfo{person}{Shouzhen Gu}, \bibinfo{person}{Sakyasingha Dasgupta},
  \bibinfo{person}{Yiyu Shi}, {and} \bibinfo{person}{Jingtong Hu}.}
  \bibinfo{year}{2020}\natexlab{b}.
\newblock \showarticletitle{{Hardware/Software Co-Exploration of Neural
  Architectures}}.
\newblock \bibinfo{journal}{\emph{IEEE Transactions on Computer-Aided Design of
  Integrated Circuits and Systems}} \bibinfo{volume}{39}, \bibinfo{number}{12}
  (\bibinfo{year}{2020}), \bibinfo{pages}{4805--4815}.
\newblock


\bibitem[\protect\citeauthoryear{Jiang, Wang, Chen, Wu, Wang, Zou, Yang, Cui,
  Cai, Yu, Lv, and Wu}{Jiang et~al\mbox{.}}{2020a}]%
        {alibaba2020mnn}
\bibfield{author}{\bibinfo{person}{Xiaotang Jiang}, \bibinfo{person}{Huan
  Wang}, \bibinfo{person}{Yiliu Chen}, \bibinfo{person}{Ziqi Wu},
  \bibinfo{person}{Lichuan Wang}, \bibinfo{person}{Bin Zou},
  \bibinfo{person}{Yafeng Yang}, \bibinfo{person}{Zongyang Cui},
  \bibinfo{person}{Yu Cai}, \bibinfo{person}{Tianhang Yu},
  \bibinfo{person}{Chengfei Lv}, {and} \bibinfo{person}{Zhihua Wu}.}
  \bibinfo{year}{2020}\natexlab{a}.
\newblock \showarticletitle{{MNN: A Universal and Efficient Inference Engine}}.
  In \bibinfo{booktitle}{\emph{Conference on Machine Learning and Systems}}.
\newblock


\bibitem[\protect\citeauthoryear{Jochems, Deist, El~Naqa, Kessler, Mayo,
  Reeves, Jolly, Matuszak, Ten~Haken, van Soest, Oberije, Faivre-Finn, Price,
  de~Ruysscher, Lambin, and Dekker}{Jochems et~al\mbox{.}}{2017}]%
        {jochems2017developing}
\bibfield{author}{\bibinfo{person}{Arthur Jochems}, \bibinfo{person}{Timo~M
  Deist}, \bibinfo{person}{Issam El~Naqa}, \bibinfo{person}{Marc Kessler},
  \bibinfo{person}{Chuck Mayo}, \bibinfo{person}{Jackson Reeves},
  \bibinfo{person}{Shruti Jolly}, \bibinfo{person}{Martha Matuszak},
  \bibinfo{person}{Randall Ten~Haken}, \bibinfo{person}{Johan van Soest},
  \bibinfo{person}{Cary Oberije}, \bibinfo{person}{Corinne Faivre-Finn},
  \bibinfo{person}{Gareth Price}, \bibinfo{person}{Dirk de Ruysscher},
  \bibinfo{person}{Philippe Lambin}, {and} \bibinfo{person}{Andre Dekker}.}
  \bibinfo{year}{2017}\natexlab{}.
\newblock \showarticletitle{{Developing and Validating a Survival Prediction
  Model for NSCLC Patients Through Distributed Learning Across 3 Countries}}.
\newblock \bibinfo{journal}{\emph{International Journal of Radiation Oncology,
  Biology, Physics}} \bibinfo{volume}{99}, \bibinfo{number}{2}
  (\bibinfo{year}{2017}), \bibinfo{pages}{344--352}.
\newblock


\bibitem[\protect\citeauthoryear{Judd, Albericio, Hetherington, Aamodt, and
  Moshovos}{Judd et~al\mbox{.}}{2016}]%
        {judd2016stripes}
\bibfield{author}{\bibinfo{person}{Patrick Judd}, \bibinfo{person}{Jorge
  Albericio}, \bibinfo{person}{Tayler Hetherington}, \bibinfo{person}{Tor~M
  Aamodt}, {and} \bibinfo{person}{Andreas Moshovos}.}
  \bibinfo{year}{2016}\natexlab{}.
\newblock \showarticletitle{{Stripes: Bit-Serial Deep Neural Network
  Computing}}. In \bibinfo{booktitle}{\emph{IEEE/ACM International Symposium on
  Microarchitecture}}.
\newblock


\bibitem[\protect\citeauthoryear{Kao, Jeong, and Krishna}{Kao
  et~al\mbox{.}}{2020}]%
        {kao2020confuciux}
\bibfield{author}{\bibinfo{person}{Sheng-Chun Kao}, \bibinfo{person}{Geonhwa
  Jeong}, {and} \bibinfo{person}{Tushar Krishna}.}
  \bibinfo{year}{2020}\natexlab{}.
\newblock \showarticletitle{{ConfuciuX: Autonomous Hardware Resource Assignment
  for DNN Accelerators using Reinforcement Learning}}. In
  \bibinfo{booktitle}{\emph{IEEE/ACM International Symposium on
  Microarchitecture}}.
\newblock


\bibitem[\protect\citeauthoryear{Karpathy, Toderici, Shetty, Leung, Sukthankar,
  and Fei-Fei}{Karpathy et~al\mbox{.}}{2014}]%
        {karpathy2014large}
\bibfield{author}{\bibinfo{person}{Andrej Karpathy}, \bibinfo{person}{George
  Toderici}, \bibinfo{person}{Sanketh Shetty}, \bibinfo{person}{Thomas Leung},
  \bibinfo{person}{Rahul Sukthankar}, {and} \bibinfo{person}{Li Fei-Fei}.}
  \bibinfo{year}{2014}\natexlab{}.
\newblock \showarticletitle{{Large-Scale Video Classification with
  Convolutional Neural Networks}}. In \bibinfo{booktitle}{\emph{IEEE Conference
  on Computer Vision and Pattern Recognition}}.
\newblock


\bibitem[\protect\citeauthoryear{Kim and Cho}{Kim and Cho}{2020}]%
        {kim2020length}
\bibfield{author}{\bibinfo{person}{Gyuwan Kim} {and} \bibinfo{person}{Kyunghyun
  Cho}.} \bibinfo{year}{2020}\natexlab{}.
\newblock \showarticletitle{{Length-Adaptive Transformer: Train Once with
  Length Drop, Use Anytime with Search}}.
\newblock \bibinfo{journal}{\emph{arXiv preprint arXiv:2010.07003}}
  (\bibinfo{year}{2020}).
\newblock


\bibitem[\protect\citeauthoryear{Kim, Gholami, Yao, Mahoney, and Keutzer}{Kim
  et~al\mbox{.}}{2021}]%
        {kim2021bert}
\bibfield{author}{\bibinfo{person}{Sehoon Kim}, \bibinfo{person}{Amir Gholami},
  \bibinfo{person}{Zhewei Yao}, \bibinfo{person}{Michael~W Mahoney}, {and}
  \bibinfo{person}{Kurt Keutzer}.} \bibinfo{year}{2021}\natexlab{}.
\newblock \showarticletitle{{I-BERT: Integer-only BERT Quantization}}.
\newblock \bibinfo{journal}{\emph{arXiv preprint arXiv:2101.01321}}
  (\bibinfo{year}{2021}).
\newblock


\bibitem[\protect\citeauthoryear{Kim}{Kim}{2014}]%
        {kim2014convolutional}
\bibfield{author}{\bibinfo{person}{Yoon Kim}.} \bibinfo{year}{2014}\natexlab{}.
\newblock \showarticletitle{{Convolutional Neural Networks for Sentence
  Classification}}. In \bibinfo{booktitle}{\emph{Conference on Empirical
  Methods in Natural Language Processing}}.
\newblock


\bibitem[\protect\citeauthoryear{Kim, Park, Yoo, Choi, Yang, and Shin}{Kim
  et~al\mbox{.}}{2015}]%
        {kim2015compression}
\bibfield{author}{\bibinfo{person}{Yong-Deok Kim}, \bibinfo{person}{Eunhyeok
  Park}, \bibinfo{person}{Sungjoo Yoo}, \bibinfo{person}{Taelim Choi},
  \bibinfo{person}{Lu Yang}, {and} \bibinfo{person}{Dongjun Shin}.}
  \bibinfo{year}{2015}\natexlab{}.
\newblock \showarticletitle{{Compression of Deep Convolutional Neural Networks
  for Fast and Low Power Mobile Applications}}.
\newblock \bibinfo{journal}{\emph{arXiv preprint arXiv:1511.06530}}
  (\bibinfo{year}{2015}).
\newblock


\bibitem[\protect\citeauthoryear{Kitaev, Kaiser, and Levskaya}{Kitaev
  et~al\mbox{.}}{2019}]%
        {kitaev2019reformer}
\bibfield{author}{\bibinfo{person}{Nikita Kitaev}, \bibinfo{person}{Lukasz
  Kaiser}, {and} \bibinfo{person}{Anselm Levskaya}.}
  \bibinfo{year}{2019}\natexlab{}.
\newblock \showarticletitle{{Reformer: The Efficient Transformer}}. In
  \bibinfo{booktitle}{\emph{International Conference on Learning
  Representations}}.
\newblock


\bibitem[\protect\citeauthoryear{Kone{\v{c}}n{\`y}, McMahan, Yu, Richt{\'a}rik,
  Suresh, and Bacon}{Kone{\v{c}}n{\`y} et~al\mbox{.}}{2016}]%
        {konevcny2016federated}
\bibfield{author}{\bibinfo{person}{Jakub Kone{\v{c}}n{\`y}},
  \bibinfo{person}{H~Brendan McMahan}, \bibinfo{person}{Felix~X Yu},
  \bibinfo{person}{Peter Richt{\'a}rik}, \bibinfo{person}{Ananda~Theertha
  Suresh}, {and} \bibinfo{person}{Dave Bacon}.}
  \bibinfo{year}{2016}\natexlab{}.
\newblock \showarticletitle{{Federated Learning: Strategies for Improving
  Communication Efficiency}}.
\newblock \bibinfo{journal}{\emph{arXiv preprint arXiv:1610.05492}}
  (\bibinfo{year}{2016}).
\newblock


\bibitem[\protect\citeauthoryear{Kornblith, Shlens, and Le}{Kornblith
  et~al\mbox{.}}{2019}]%
        {kornblith2019better}
\bibfield{author}{\bibinfo{person}{Simon Kornblith}, \bibinfo{person}{Jonathon
  Shlens}, {and} \bibinfo{person}{Quoc~V Le}.} \bibinfo{year}{2019}\natexlab{}.
\newblock \showarticletitle{{Do Better ImageNet Models Transfer Better?}}. In
  \bibinfo{booktitle}{\emph{IEEE Conference on Computer Vision and Pattern
  Recognition}}.
\newblock


\bibitem[\protect\citeauthoryear{Krishna Murthy~Jatavallabhula}{Krishna
  Murthy~Jatavallabhula}{2019}]%
        {jatavallabhula2019kaolin}
\bibfield{author}{\bibinfo{person}{Jean-Francois Lafleche Clement Fuji Tsang
  Artem Rozantsev Wenzheng Chen Tommy Xiang Rev Lebaredian Sanja~Fidler Krishna
  Murthy~Jatavallabhula, Edward~Smith}.} \bibinfo{year}{2019}\natexlab{}.
\newblock \showarticletitle{{Kaolin: A Pytorch Library for Accelerating 3D Deep
  Learning Research}}.
\newblock \bibinfo{journal}{\emph{arXiv preprint arXiv:1911.05063}}
  (\bibinfo{year}{2019}).
\newblock


\bibitem[\protect\citeauthoryear{Krizhevsky and Hinton}{Krizhevsky and
  Hinton}{2009}]%
        {krizhevsky2009learning}
\bibfield{author}{\bibinfo{person}{Alex Krizhevsky} {and}
  \bibinfo{person}{Geoffrey Hinton}.} \bibinfo{year}{2009}\natexlab{}.
\newblock \bibinfo{booktitle}{\emph{{Learning Multiple Layers of Features from
  Tiny Images}}}.
\newblock \bibinfo{type}{{T}echnical {R}eport}.
  \bibinfo{institution}{University of Toronto}.
\newblock


\bibitem[\protect\citeauthoryear{Krizhevsky, Sutskever, and Hinton}{Krizhevsky
  et~al\mbox{.}}{2012}]%
        {krizhevsky2012imagenet}
\bibfield{author}{\bibinfo{person}{Alex Krizhevsky}, \bibinfo{person}{Ilya
  Sutskever}, {and} \bibinfo{person}{Geoffrey~E Hinton}.}
  \bibinfo{year}{2012}\natexlab{}.
\newblock \showarticletitle{{ImageNet Classification with Deep Convolutional
  Neural Networks}}. In \bibinfo{booktitle}{\emph{Conference on Neural
  Information Processing Systems}}.
\newblock


\bibitem[\protect\citeauthoryear{Kuen, Kong, Lin, Wang, Yin, See, and Tan}{Kuen
  et~al\mbox{.}}{2018}]%
        {kuen2018stochastic}
\bibfield{author}{\bibinfo{person}{Jason Kuen}, \bibinfo{person}{Xiangfei
  Kong}, \bibinfo{person}{Zhe Lin}, \bibinfo{person}{Gang Wang},
  \bibinfo{person}{Jianxiong Yin}, \bibinfo{person}{Simon See}, {and}
  \bibinfo{person}{Yap-Peng Tan}.} \bibinfo{year}{2018}\natexlab{}.
\newblock \showarticletitle{{Stochastic Downsampling for Cost-Adjustable
  Inference and Improved Regularization in Convolutional Networks}}. In
  \bibinfo{booktitle}{\emph{IEEE Conference on Computer Vision and Pattern
  Recognition}}.
\newblock


\bibitem[\protect\citeauthoryear{Lahoud, Ghanem, Pollefeys, and Oswald}{Lahoud
  et~al\mbox{.}}{2019}]%
        {lahoud20193d}
\bibfield{author}{\bibinfo{person}{Jean Lahoud}, \bibinfo{person}{Bernard
  Ghanem}, \bibinfo{person}{Marc Pollefeys}, {and} \bibinfo{person}{Martin~R
  Oswald}.} \bibinfo{year}{2019}\natexlab{}.
\newblock \showarticletitle{{3D Instance Segmentation via Multi-Task Metric
  Learning}}. In \bibinfo{booktitle}{\emph{International Conference on Computer
  Vision}}.
\newblock


\bibitem[\protect\citeauthoryear{Lan, Yu, Yu, and Davis}{Lan
  et~al\mbox{.}}{2019}]%
        {lan2019modeling}
\bibfield{author}{\bibinfo{person}{Shiyi Lan}, \bibinfo{person}{Ruichi Yu},
  \bibinfo{person}{Gang Yu}, {and} \bibinfo{person}{Larry~S Davis}.}
  \bibinfo{year}{2019}\natexlab{}.
\newblock \showarticletitle{{Modeling Local Geometric Structure of 3D Point
  Clouds using Geo-CNN}}. In \bibinfo{booktitle}{\emph{IEEE Conference on
  Computer Vision and Pattern Recognition}}.
\newblock


\bibitem[\protect\citeauthoryear{Landrieu and Simonovsky}{Landrieu and
  Simonovsky}{2018}]%
        {landrieu2018large}
\bibfield{author}{\bibinfo{person}{Loic Landrieu} {and} \bibinfo{person}{Martin
  Simonovsky}.} \bibinfo{year}{2018}\natexlab{}.
\newblock \showarticletitle{{Large-Scale Point Cloud Semantic Segmentation With
  Superpoint Graphs}}. In \bibinfo{booktitle}{\emph{IEEE Conference on Computer
  Vision and Pattern Recognition}}.
\newblock


\bibitem[\protect\citeauthoryear{Lebedev, Ganin, Rakhuba, Oseledets, and
  Lempitsky}{Lebedev et~al\mbox{.}}{2014}]%
        {lebedev2014speeding}
\bibfield{author}{\bibinfo{person}{Vadim Lebedev}, \bibinfo{person}{Yaroslav
  Ganin}, \bibinfo{person}{Maksim Rakhuba}, \bibinfo{person}{Ivan Oseledets},
  {and} \bibinfo{person}{Victor Lempitsky}.} \bibinfo{year}{2014}\natexlab{}.
\newblock \showarticletitle{{Speeding-Up Convolutional Neural Networks Using
  Fine-tuned CP-Decomposition}}.
\newblock \bibinfo{journal}{\emph{arXiv preprint arXiv:1412.6553}}
  (\bibinfo{year}{2014}).
\newblock


\bibitem[\protect\citeauthoryear{Lebedev and Lempitsky}{Lebedev and
  Lempitsky}{2016}]%
        {lebedev2016fast}
\bibfield{author}{\bibinfo{person}{Vadim Lebedev} {and} \bibinfo{person}{Victor
  Lempitsky}.} \bibinfo{year}{2016}\natexlab{}.
\newblock \showarticletitle{{Fast ConvNets Using Group-Wise Brain Damage}}. In
  \bibinfo{booktitle}{\emph{IEEE Conference on Computer Vision and Pattern
  Recognition}}.
\newblock


\bibitem[\protect\citeauthoryear{LeCun, Cortes, and Burges}{LeCun
  et~al\mbox{.}}{2010}]%
        {lecun2010mnist}
\bibfield{author}{\bibinfo{person}{Yann LeCun}, \bibinfo{person}{Corinna
  Cortes}, {and} \bibinfo{person}{Christopher~JC Burges}.}
  \bibinfo{year}{2010}\natexlab{}.
\newblock \showarticletitle{{MNIST Handwritten Digit Database}}.
\newblock \bibinfo{journal}{\emph{AT\&T Labs [Online]. Available:
  http://yann.lecun.com/exdb/mnist}} (\bibinfo{year}{2010}).
\newblock


\bibitem[\protect\citeauthoryear{LeCun, Denker, Solla, Howard, and
  Jackel}{LeCun et~al\mbox{.}}{1989}]%
        {lecun1989optimal}
\bibfield{author}{\bibinfo{person}{Yann LeCun}, \bibinfo{person}{John~S
  Denker}, \bibinfo{person}{Sara~A Solla}, \bibinfo{person}{Richard~E Howard},
  {and} \bibinfo{person}{Lawrence~D Jackel}.} \bibinfo{year}{1989}\natexlab{}.
\newblock \showarticletitle{{Optimal Brain Damage}}. In
  \bibinfo{booktitle}{\emph{Conference on Neural Information Processing
  Systems}}.
\newblock


\bibitem[\protect\citeauthoryear{Lee, Kim, Kang, Shin, Kim, and Yoo}{Lee
  et~al\mbox{.}}{2018}]%
        {lee2018unpu}
\bibfield{author}{\bibinfo{person}{Jinmook Lee}, \bibinfo{person}{Changhyeon
  Kim}, \bibinfo{person}{Sanghoon Kang}, \bibinfo{person}{Dongjoo Shin},
  \bibinfo{person}{Sangyeob Kim}, {and} \bibinfo{person}{Hoi-Jun Yoo}.}
  \bibinfo{year}{2018}\natexlab{}.
\newblock \showarticletitle{{UNPU: A 50.6 TOPS/W Unified Deep Neural Network
  Accelerator with 1b-to-16b Fully-Variable Weight Bit-Precision}}. In
  \bibinfo{booktitle}{\emph{International Solid-State Circuits Conference}}.
\newblock


\bibitem[\protect\citeauthoryear{Lee, Lee, Kim, Kosiorek, Choi, and Teh}{Lee
  et~al\mbox{.}}{2019}]%
        {lee2019set}
\bibfield{author}{\bibinfo{person}{Juho Lee}, \bibinfo{person}{Yoonho Lee},
  \bibinfo{person}{Jungtaek Kim}, \bibinfo{person}{Adam Kosiorek},
  \bibinfo{person}{Seungjin Choi}, {and} \bibinfo{person}{Yee~Whye Teh}.}
  \bibinfo{year}{2019}\natexlab{}.
\newblock \showarticletitle{{Set Transformer: A Framework for Attention-Based
  Permutation-Invariant Neural Networks}}. In
  \bibinfo{booktitle}{\emph{International Conference on Machine Learning}}.
\newblock


\bibitem[\protect\citeauthoryear{Lei, Zhang, Wang, Dai, and Artzi}{Lei
  et~al\mbox{.}}{2017}]%
        {lei2017simple}
\bibfield{author}{\bibinfo{person}{Tao Lei}, \bibinfo{person}{Yu Zhang},
  \bibinfo{person}{Sida~I Wang}, \bibinfo{person}{Hui Dai}, {and}
  \bibinfo{person}{Yoav Artzi}.} \bibinfo{year}{2017}\natexlab{}.
\newblock \showarticletitle{{Simple Recurrent Units for Highly Parallelizable
  Recurrence}}. In \bibinfo{booktitle}{\emph{Conference on Empirical Methods in
  Natural Language Processing}}.
\newblock


\bibitem[\protect\citeauthoryear{Li, Zhang, and Liu}{Li et~al\mbox{.}}{2016}]%
        {li2016ternary}
\bibfield{author}{\bibinfo{person}{Fengfu Li}, \bibinfo{person}{Bo Zhang},
  {and} \bibinfo{person}{Bin Liu}.} \bibinfo{year}{2016}\natexlab{}.
\newblock \showarticletitle{{Ternary Weight Networks}}.
\newblock \bibinfo{journal}{\emph{arXiv preprint arXiv:1605.04711}}
  (\bibinfo{year}{2016}).
\newblock


\bibitem[\protect\citeauthoryear{Li, M\"{u}ller, Qian, Delgadillo, Abualshour,
  Thabet, and Ghanem}{Li et~al\mbox{.}}{2021}]%
        {li2019deepgcns}
\bibfield{author}{\bibinfo{person}{Guohao Li}, \bibinfo{person}{Matthias
  M\"{u}ller}, \bibinfo{person}{Guocheng Qian}, \bibinfo{person}{Itzel~C.
  Delgadillo}, \bibinfo{person}{Abdulellah Abualshour}, \bibinfo{person}{Ali
  Thabet}, {and} \bibinfo{person}{Bernard Ghanem}.}
  \bibinfo{year}{2021}\natexlab{}.
\newblock \showarticletitle{{DeepGCNs: Making GCNs Go as Deep as CNNs}}.
\newblock \bibinfo{journal}{\emph{IEEE Transactions on Pattern Analysis and
  Machine Intelligence}} (\bibinfo{year}{2021}).
\newblock


\bibitem[\protect\citeauthoryear{Li, Qian, Delgadillo, Muller, Thabet, and
  Ghanem}{Li et~al\mbox{.}}{2020b}]%
        {li2020sgas}
\bibfield{author}{\bibinfo{person}{Guohao Li}, \bibinfo{person}{Guocheng Qian},
  \bibinfo{person}{Itzel~C Delgadillo}, \bibinfo{person}{Matthias Muller},
  \bibinfo{person}{Ali Thabet}, {and} \bibinfo{person}{Bernard Ghanem}.}
  \bibinfo{year}{2020}\natexlab{b}.
\newblock \showarticletitle{{SGAS: Sequential Greedy Architecture Search}}. In
  \bibinfo{booktitle}{\emph{IEEE Conference on Computer Vision and Pattern
  Recognition}}.
\newblock


\bibitem[\protect\citeauthoryear{Li, Xu, Giancola, Thabet, and Ghanem}{Li
  et~al\mbox{.}}{2020d}]%
        {li2020lcnas}
\bibfield{author}{\bibinfo{person}{Guohao Li}, \bibinfo{person}{Mengmeng Xu},
  \bibinfo{person}{Silvio Giancola}, \bibinfo{person}{Ali Thabet}, {and}
  \bibinfo{person}{Bernard Ghanem}.} \bibinfo{year}{2020}\natexlab{d}.
\newblock \showarticletitle{{LC-NAS: Latency Constrained Neural Architecture
  Search for Point Cloud Networks}}.
\newblock \bibinfo{journal}{\emph{arXiv preprint arXiv:2008.10309}}
  (\bibinfo{year}{2020}).
\newblock


\bibitem[\protect\citeauthoryear{Li, Kadav, Durdanovic, Samet, and Graf}{Li
  et~al\mbox{.}}{2017}]%
        {li2017pruning}
\bibfield{author}{\bibinfo{person}{Hao Li}, \bibinfo{person}{Asim Kadav},
  \bibinfo{person}{Igor Durdanovic}, \bibinfo{person}{Hanan Samet}, {and}
  \bibinfo{person}{Hans~Peter Graf}.} \bibinfo{year}{2017}\natexlab{}.
\newblock \showarticletitle{{Pruning Filters for Efficient ConvNets}}. In
  \bibinfo{booktitle}{\emph{International Conference on Learning
  Representations}}.
\newblock


\bibitem[\protect\citeauthoryear{Li, Lin, Ding, Liu, Zhu, and Han}{Li
  et~al\mbox{.}}{2020a}]%
        {li2020gan}
\bibfield{author}{\bibinfo{person}{Muyang Li}, \bibinfo{person}{Ji Lin},
  \bibinfo{person}{Yaoyao Ding}, \bibinfo{person}{Zhijian Liu},
  \bibinfo{person}{Jun-Yan Zhu}, {and} \bibinfo{person}{Song Han}.}
  \bibinfo{year}{2020}\natexlab{a}.
\newblock \showarticletitle{{GAN Compression: Efficient Architectures for
  Interactive Conditional GANs}}. In \bibinfo{booktitle}{\emph{IEEE Conference
  on Computer Vision and Pattern Recognition}}.
\newblock


\bibitem[\protect\citeauthoryear{Li, Bu, Sun, Wu, Di, and Chen}{Li
  et~al\mbox{.}}{2018}]%
        {li2018pointcnn}
\bibfield{author}{\bibinfo{person}{Yangyan Li}, \bibinfo{person}{Rui Bu},
  \bibinfo{person}{Mingchao Sun}, \bibinfo{person}{Wei Wu},
  \bibinfo{person}{Xinhan Di}, {and} \bibinfo{person}{Baoquan Chen}.}
  \bibinfo{year}{2018}\natexlab{}.
\newblock \showarticletitle{{PointCNN: Convolution on $\mathcal{X}$-Transformed
  Points}}. In \bibinfo{booktitle}{\emph{Conference on Neural Information
  Processing Systems}}.
\newblock


\bibitem[\protect\citeauthoryear{Li, Wallace, Shen, Lin, Keutzer, Klein, and
  Gonzalez}{Li et~al\mbox{.}}{2020c}]%
        {li2020train}
\bibfield{author}{\bibinfo{person}{Zhuohan Li}, \bibinfo{person}{Eric Wallace},
  \bibinfo{person}{Sheng Shen}, \bibinfo{person}{Kevin Lin},
  \bibinfo{person}{Kurt Keutzer}, \bibinfo{person}{Dan Klein}, {and}
  \bibinfo{person}{Joey Gonzalez}.} \bibinfo{year}{2020}\natexlab{c}.
\newblock \showarticletitle{{Train Big, Then Compress: Rethinking Model Size
  for Efficient Training and Inference of Transformers}}. In
  \bibinfo{booktitle}{\emph{International Conference on Machine Learning}}.
\newblock


\bibitem[\protect\citeauthoryear{Lin, Chen, Lin, Cohn, Gan, and Han}{Lin
  et~al\mbox{.}}{2020a}]%
        {lin2020mcunet}
\bibfield{author}{\bibinfo{person}{Ji Lin}, \bibinfo{person}{Wei-Ming Chen},
  \bibinfo{person}{Yujun Lin}, \bibinfo{person}{John Cohn},
  \bibinfo{person}{Chuang Gan}, {and} \bibinfo{person}{Song Han}.}
  \bibinfo{year}{2020}\natexlab{a}.
\newblock \showarticletitle{{MCUNet: Tiny Deep Learning on IoT Devices}}.
\newblock \bibinfo{journal}{\emph{arXiv preprint arXiv:2007.10319}}
  (\bibinfo{year}{2020}).
\newblock


\bibitem[\protect\citeauthoryear{Lin, Gan, and Han}{Lin et~al\mbox{.}}{2019a}]%
        {lin2019training}
\bibfield{author}{\bibinfo{person}{Ji Lin}, \bibinfo{person}{Chuang Gan}, {and}
  \bibinfo{person}{Song Han}.} \bibinfo{year}{2019}\natexlab{a}.
\newblock \showarticletitle{{Training Kinetics in 15 Minutes: Large-Scale
  Distributed Training on Videos}}.
\newblock \bibinfo{journal}{\emph{arXiv preprint arXiv:1910.00932}}
  (\bibinfo{year}{2019}).
\newblock


\bibitem[\protect\citeauthoryear{Lin, Gan, and Han}{Lin et~al\mbox{.}}{2019b}]%
        {lin2019tsm}
\bibfield{author}{\bibinfo{person}{Ji Lin}, \bibinfo{person}{Chuang Gan}, {and}
  \bibinfo{person}{Song Han}.} \bibinfo{year}{2019}\natexlab{b}.
\newblock \showarticletitle{{TSM: Temporal Shift Module for Efficient Video
  Understanding}}. In \bibinfo{booktitle}{\emph{International Conference on
  Computer Vision}}.
\newblock


\bibitem[\protect\citeauthoryear{Lin, Rao, and Lu}{Lin et~al\mbox{.}}{2017}]%
        {lin2017runtime}
\bibfield{author}{\bibinfo{person}{Ji Lin}, \bibinfo{person}{Yongming Rao},
  {and} \bibinfo{person}{Jiwen Lu}.} \bibinfo{year}{2017}\natexlab{}.
\newblock \showarticletitle{{Runtime Neural Pruning}}. In
  \bibinfo{booktitle}{\emph{Conference on Neural Information Processing
  Systems}}.
\newblock


\bibitem[\protect\citeauthoryear{Lin, Hafdi, Wang, Liu, and Han}{Lin
  et~al\mbox{.}}{2020b}]%
        {linneural}
\bibfield{author}{\bibinfo{person}{Yujun Lin}, \bibinfo{person}{Driss Hafdi},
  \bibinfo{person}{Kuan Wang}, \bibinfo{person}{Zhijian Liu}, {and}
  \bibinfo{person}{Song Han}.} \bibinfo{year}{2020}\natexlab{b}.
\newblock \showarticletitle{{Neural-Hardware Architecture Search}}. In
  \bibinfo{booktitle}{\emph{Workshop on ML for Systems at NeurIPS}}.
\newblock


\bibitem[\protect\citeauthoryear{Lin, Han, Mao, Wang, and Dally}{Lin
  et~al\mbox{.}}{2018}]%
        {lin2018deep}
\bibfield{author}{\bibinfo{person}{Yujun Lin}, \bibinfo{person}{Song Han},
  \bibinfo{person}{Huizi Mao}, \bibinfo{person}{Yu Wang}, {and}
  \bibinfo{person}{William~J Dally}.} \bibinfo{year}{2018}\natexlab{}.
\newblock \showarticletitle{{Deep Gradient Compression: Reducing the
  Communication Bandwidth for Distributed Training}}. In
  \bibinfo{booktitle}{\emph{International Conference on Learning
  Representations}}.
\newblock


\bibitem[\protect\citeauthoryear{Lin, Yang, and Han}{Lin et~al\mbox{.}}{2021}]%
        {lin2021enhcs}
\bibfield{author}{\bibinfo{person}{Yujun Lin}, \bibinfo{person}{Mengtian Yang},
  {and} \bibinfo{person}{Song Han}.} \bibinfo{year}{2021}\natexlab{}.
\newblock \showarticletitle{{NAAS: Neural Accelerator Architecture Search}}. In
  \bibinfo{booktitle}{\emph{Design Automation Conference}}.
\newblock


\bibitem[\protect\citeauthoryear{Lin, Courbariaux, Memisevic, and Bengio}{Lin
  et~al\mbox{.}}{2016}]%
        {lin2016neural}
\bibfield{author}{\bibinfo{person}{Zhouhan Lin}, \bibinfo{person}{Matthieu
  Courbariaux}, \bibinfo{person}{Roland Memisevic}, {and}
  \bibinfo{person}{Yoshua Bengio}.} \bibinfo{year}{2016}\natexlab{}.
\newblock \showarticletitle{{Neural Networks with Few Multiplications}}. In
  \bibinfo{booktitle}{\emph{International Conference on Learning
  Representations}}.
\newblock


\bibitem[\protect\citeauthoryear{Liu, Guo, Chou, Mehra, Yeung, Downing,
  Salipur, Jopling, Campbell, Deru, Beninati, Milstein, and Fei-Fei}{Liu
  et~al\mbox{.}}{2018a}]%
        {liu20183d}
\bibfield{author}{\bibinfo{person}{Bingbin Liu}, \bibinfo{person}{Michelle
  Guo}, \bibinfo{person}{Edward Chou}, \bibinfo{person}{Rishab Mehra},
  \bibinfo{person}{Serena Yeung}, \bibinfo{person}{N~Lance Downing},
  \bibinfo{person}{Francesca Salipur}, \bibinfo{person}{Jeffrey Jopling},
  \bibinfo{person}{Brandi Campbell}, \bibinfo{person}{Kayla Deru},
  \bibinfo{person}{William Beninati}, \bibinfo{person}{Arnold Milstein}, {and}
  \bibinfo{person}{Li Fei-Fei}.} \bibinfo{year}{2018}\natexlab{a}.
\newblock \showarticletitle{{3D Point Cloud-Based Visual Prediction of ICU
  Mobility Care Activities}}. In \bibinfo{booktitle}{\emph{Machine Learning for
  Healthcare Conference}}.
\newblock


\bibitem[\protect\citeauthoryear{Liu, Chen, Schroff, Adam, Hua, Yuille, and
  Fei-Fei}{Liu et~al\mbox{.}}{2019b}]%
        {liu2019auto}
\bibfield{author}{\bibinfo{person}{Chenxi Liu}, \bibinfo{person}{Liang-Chieh
  Chen}, \bibinfo{person}{Florian Schroff}, \bibinfo{person}{Hartwig Adam},
  \bibinfo{person}{Wei Hua}, \bibinfo{person}{Alan~L Yuille}, {and}
  \bibinfo{person}{Li Fei-Fei}.} \bibinfo{year}{2019}\natexlab{b}.
\newblock \showarticletitle{{Auto-DeepLab: Hierarchical Neural Architecture
  Search for Semantic Image Segmentation}}. In \bibinfo{booktitle}{\emph{IEEE
  Conference on Computer Vision and Pattern Recognition}}.
\newblock


\bibitem[\protect\citeauthoryear{Liu, Zoph, Neumann, Shlens, Hua, Li, Fei-Fei,
  Yuille, Huang, and Murphy}{Liu et~al\mbox{.}}{2018d}]%
        {liu2018progressive}
\bibfield{author}{\bibinfo{person}{Chenxi Liu}, \bibinfo{person}{Barret Zoph},
  \bibinfo{person}{Maxim Neumann}, \bibinfo{person}{Jonathon Shlens},
  \bibinfo{person}{Wei Hua}, \bibinfo{person}{Li-Jia Li}, \bibinfo{person}{Li
  Fei-Fei}, \bibinfo{person}{Alan Yuille}, \bibinfo{person}{Jonathan Huang},
  {and} \bibinfo{person}{Kevin Murphy}.} \bibinfo{year}{2018}\natexlab{d}.
\newblock \showarticletitle{{Progressive Neural Architecture Search}}. In
  \bibinfo{booktitle}{\emph{European Conference on Computer Vision}}.
\newblock


\bibitem[\protect\citeauthoryear{Liu, Simonyan, Vinyals, Fernando, and
  Kavukcuoglu}{Liu et~al\mbox{.}}{2018c}]%
        {liu2017hierarchical}
\bibfield{author}{\bibinfo{person}{Hanxiao Liu}, \bibinfo{person}{Karen
  Simonyan}, \bibinfo{person}{Oriol Vinyals}, \bibinfo{person}{Chrisantha
  Fernando}, {and} \bibinfo{person}{Koray Kavukcuoglu}.}
  \bibinfo{year}{2018}\natexlab{c}.
\newblock \showarticletitle{{Hierarchical Representations for Efficient
  Architecture Search}}. In \bibinfo{booktitle}{\emph{International Conference
  on Learning Representations}}.
\newblock


\bibitem[\protect\citeauthoryear{Liu, Simonyan, and Yang}{Liu
  et~al\mbox{.}}{2019e}]%
        {liu2019darts}
\bibfield{author}{\bibinfo{person}{Haoxiao Liu}, \bibinfo{person}{Karen
  Simonyan}, {and} \bibinfo{person}{Yiming Yang}.}
  \bibinfo{year}{2019}\natexlab{e}.
\newblock \showarticletitle{{DARTS: Differentiable Architecture Search}}. In
  \bibinfo{booktitle}{\emph{International Conference on Learning
  Representations}}.
\newblock


\bibitem[\protect\citeauthoryear{Liu and Deng}{Liu and Deng}{2018}]%
        {liu2018dynamic}
\bibfield{author}{\bibinfo{person}{Lanlan Liu} {and} \bibinfo{person}{Jia
  Deng}.} \bibinfo{year}{2018}\natexlab{}.
\newblock \showarticletitle{{Dynamic Deep Neural Networks: Optimizing
  Accuracy-Efficiency Trade-Offs by Selective Execution}}. In
  \bibinfo{booktitle}{\emph{AAAI Conference on Artificial Intelligence}}.
\newblock


\bibitem[\protect\citeauthoryear{Liu, Deng, Hu, Zhu, Li, Ding, and Xie}{Liu
  et~al\mbox{.}}{2019c}]%
        {liu2019dynamic}
\bibfield{author}{\bibinfo{person}{Liu Liu}, \bibinfo{person}{Lei Deng},
  \bibinfo{person}{Xing Hu}, \bibinfo{person}{Maohua Zhu},
  \bibinfo{person}{Guoqi Li}, \bibinfo{person}{Yufei Ding}, {and}
  \bibinfo{person}{Yuan Xie}.} \bibinfo{year}{2019}\natexlab{c}.
\newblock \showarticletitle{{Dynamic Sparse Graph for Efficient Deep
  Learning}}. In \bibinfo{booktitle}{\emph{International Conference on Learning
  Representations}}.
\newblock


\bibitem[\protect\citeauthoryear{Liu, Saleh, Pot, Goodrich, Sepassi, Kaiser,
  and Shazeer}{Liu et~al\mbox{.}}{2018b}]%
        {liu2018generating}
\bibfield{author}{\bibinfo{person}{Peter~J Liu}, \bibinfo{person}{Mohammad
  Saleh}, \bibinfo{person}{Etienne Pot}, \bibinfo{person}{Ben Goodrich},
  \bibinfo{person}{Ryan Sepassi}, \bibinfo{person}{Lukasz Kaiser}, {and}
  \bibinfo{person}{Noam Shazeer}.} \bibinfo{year}{2018}\natexlab{b}.
\newblock \showarticletitle{{Generating Wikipedia by Summarizing Long
  Sequences}}. In \bibinfo{booktitle}{\emph{International Conference on
  Learning Representations}}.
\newblock


\bibitem[\protect\citeauthoryear{Liu, Liu, Yang, Li, and Zhou}{Liu
  et~al\mbox{.}}{2015}]%
        {liu2014recursive}
\bibfield{author}{\bibinfo{person}{Shujie Liu}, \bibinfo{person}{Shujie Liu},
  \bibinfo{person}{Nan Yang}, \bibinfo{person}{Mu Li}, {and}
  \bibinfo{person}{Ming Zhou}.} \bibinfo{year}{2015}\natexlab{}.
\newblock \showarticletitle{{A Recursive Recurrent Neural Network for
  Statistical Machine Translation}}. In \bibinfo{booktitle}{\emph{Conference of
  the Association for Computational Linguistics}}.
\newblock


\bibitem[\protect\citeauthoryear{Liu, Chen, Liu, Qin, Luo, and Wang}{Liu
  et~al\mbox{.}}{2019a}]%
        {liu2019structured}
\bibfield{author}{\bibinfo{person}{Yifan Liu}, \bibinfo{person}{Ke Chen},
  \bibinfo{person}{Chris Liu}, \bibinfo{person}{Zengchang Qin},
  \bibinfo{person}{Zhenbo Luo}, {and} \bibinfo{person}{Jingdong Wang}.}
  \bibinfo{year}{2019}\natexlab{a}.
\newblock \showarticletitle{{Structured Knowledge Distillation for Semantic
  Segmentation}}. In \bibinfo{booktitle}{\emph{IEEE Conference on Computer
  Vision and Pattern Recognition}}.
\newblock


\bibitem[\protect\citeauthoryear{Liu}{Liu}{2020}]%
        {liu2020hardware}
\bibfield{author}{\bibinfo{person}{Zhijian Liu}.}
  \bibinfo{year}{2020}\natexlab{}.
\newblock \emph{\bibinfo{title}{{Hardware-Efficient Deep Learning for 3D Point
  Cloud}}}.
\newblock \bibinfo{thesistype}{Master's\ Thesis}.
  \bibinfo{school}{Massachusetts Institute of Technology}.
\newblock


\bibitem[\protect\citeauthoryear{Liu, Amini, Zhu, Karaman, Han, and Rus}{Liu
  et~al\mbox{.}}{2021a}]%
        {liu2021efficient}
\bibfield{author}{\bibinfo{person}{Zhijian Liu}, \bibinfo{person}{Alexander
  Amini}, \bibinfo{person}{Sibo Zhu}, \bibinfo{person}{Sertac Karaman},
  \bibinfo{person}{Song Han}, {and} \bibinfo{person}{Daniela Rus}.}
  \bibinfo{year}{2021}\natexlab{a}.
\newblock \showarticletitle{{Efficient and Robust LiDAR-Based End-to-End
  Navigation}}. In \bibinfo{booktitle}{\emph{IEEE International Conference on
  Robotics and Automation}}.
\newblock


\bibitem[\protect\citeauthoryear{Liu, Li, Shen, Huang, Yan, and Zhang}{Liu
  et~al\mbox{.}}{2017}]%
        {liu2017learning}
\bibfield{author}{\bibinfo{person}{Zhuang Liu}, \bibinfo{person}{Jianguo Li},
  \bibinfo{person}{Zhiqiang Shen}, \bibinfo{person}{Gao Huang},
  \bibinfo{person}{Shoumeng Yan}, {and} \bibinfo{person}{Changshui Zhang}.}
  \bibinfo{year}{2017}\natexlab{}.
\newblock \showarticletitle{{Learning Efficient Convolutional Networks Through
  Network Slimming}}. In \bibinfo{booktitle}{\emph{International Conference on
  Computer Vision}}.
\newblock


\bibitem[\protect\citeauthoryear{Liu, Mu, Zhang, Guo, Yang, Cheng, and Sun}{Liu
  et~al\mbox{.}}{2019d}]%
        {liu2019metapruning}
\bibfield{author}{\bibinfo{person}{Zechun Liu}, \bibinfo{person}{Haoyuan Mu},
  \bibinfo{person}{Xiangyu Zhang}, \bibinfo{person}{Zichao Guo},
  \bibinfo{person}{Xin Yang}, \bibinfo{person}{Kwang-Ting Cheng}, {and}
  \bibinfo{person}{Jian Sun}.} \bibinfo{year}{2019}\natexlab{d}.
\newblock \showarticletitle{{MetaPruning: Meta Learning for Automatic Neural
  Network Channel Pruning}}. In \bibinfo{booktitle}{\emph{International
  Conference on Computer Vision}}.
\newblock


\bibitem[\protect\citeauthoryear{Liu, Tang, Lin, and Han}{Liu
  et~al\mbox{.}}{2019f}]%
        {liu2019point}
\bibfield{author}{\bibinfo{person}{Zhijian Liu}, \bibinfo{person}{Haotian
  Tang}, \bibinfo{person}{Yujun Lin}, {and} \bibinfo{person}{Song Han}.}
  \bibinfo{year}{2019}\natexlab{f}.
\newblock \showarticletitle{{Point-Voxel CNN for Efficient 3D Deep Learning}}.
  In \bibinfo{booktitle}{\emph{Conference on Neural Information Processing
  Systems}}.
\newblock


\bibitem[\protect\citeauthoryear{Liu, Tang, Zhao, Shao, and Han}{Liu
  et~al\mbox{.}}{2021b}]%
        {liu2021pvnas}
\bibfield{author}{\bibinfo{person}{Zhijian Liu}, \bibinfo{person}{Haotian
  Tang}, \bibinfo{person}{Shengyu Zhao}, \bibinfo{person}{Kevin Shao}, {and}
  \bibinfo{person}{Song Han}.} \bibinfo{year}{2021}\natexlab{b}.
\newblock \showarticletitle{{PVNAS: 3D Neural Architecture Search with
  Point-Voxel Convolution}}.
\newblock \bibinfo{journal}{\emph{IEEE Transactions on Pattern Analysis and
  Machine Intelligence}} (\bibinfo{year}{2021}).
\newblock


\bibitem[\protect\citeauthoryear{Liu, Wu, Gan, Zhu, and Han}{Liu
  et~al\mbox{.}}{2020}]%
        {liu2020datamix}
\bibfield{author}{\bibinfo{person}{Zhijian Liu}, \bibinfo{person}{Zhanghao Wu},
  \bibinfo{person}{Chuang Gan}, \bibinfo{person}{Ligeng Zhu}, {and}
  \bibinfo{person}{Song Han}.} \bibinfo{year}{2020}\natexlab{}.
\newblock \showarticletitle{{DataMix: Efficient Privacy-Preserving Edge-Cloud
  Inference}}. In \bibinfo{booktitle}{\emph{European Conference on Computer
  Vision}}.
\newblock


\bibitem[\protect\citeauthoryear{Lu, Wang, Liang, Lin, and Wang}{Lu
  et~al\mbox{.}}{2020}]%
        {lu2020hardware}
\bibfield{author}{\bibinfo{person}{Siyuan Lu}, \bibinfo{person}{Meiqi Wang},
  \bibinfo{person}{Shuang Liang}, \bibinfo{person}{Jun Lin}, {and}
  \bibinfo{person}{Zhongfeng Wang}.} \bibinfo{year}{2020}\natexlab{}.
\newblock \showarticletitle{{Hardware Accelerator for Multi-Head Attention and
  Position-Wise Feed-Forward in the Transformer}}. In
  \bibinfo{booktitle}{\emph{IEEE International System-on-Chip Conference}}.
\newblock


\bibitem[\protect\citeauthoryear{Luo and Yuille}{Luo and Yuille}{2019}]%
        {luo2019grouped}
\bibfield{author}{\bibinfo{person}{Chenxu Luo} {and} \bibinfo{person}{Alan~L
  Yuille}.} \bibinfo{year}{2019}\natexlab{}.
\newblock \showarticletitle{{Grouped Spatial-Temporal Aggregation for Efficient
  Action Recognition}}. In \bibinfo{booktitle}{\emph{International Conference
  on Computer Vision}}.
\newblock


\bibitem[\protect\citeauthoryear{Ma, Zhang, Zheng, and Sun}{Ma
  et~al\mbox{.}}{2018}]%
        {ma2018shufflenet}
\bibfield{author}{\bibinfo{person}{Ningning Ma}, \bibinfo{person}{Xiangyu
  Zhang}, \bibinfo{person}{Hai-Tao Zheng}, {and} \bibinfo{person}{Jian Sun}.}
  \bibinfo{year}{2018}\natexlab{}.
\newblock \showarticletitle{{ShuffleNet V2: Practical Guidelines for Efficient
  CNN Architecture Design}}. In \bibinfo{booktitle}{\emph{European Conference
  on Computer Vision}}.
\newblock


\bibitem[\protect\citeauthoryear{Ma, Guo, Niu, Lin, Tang, Ma, Ren, and Wang}{Ma
  et~al\mbox{.}}{2020}]%
        {ma2020pconv}
\bibfield{author}{\bibinfo{person}{Xiaolong Ma}, \bibinfo{person}{Fu-Ming Guo},
  \bibinfo{person}{Wei Niu}, \bibinfo{person}{Xue Lin}, \bibinfo{person}{Jian
  Tang}, \bibinfo{person}{Kaisheng Ma}, \bibinfo{person}{Bin Ren}, {and}
  \bibinfo{person}{Yanzhi Wang}.} \bibinfo{year}{2020}\natexlab{}.
\newblock \showarticletitle{{PCONV: The Missing but Desirable Sparsity in DNN
  Weight Pruning for Real-Time Execution on Mobile Devices}}. In
  \bibinfo{booktitle}{\emph{AAAI Conference on Artificial Intelligence}}.
\newblock


\bibitem[\protect\citeauthoryear{Maji, Rahtu, Kannala, Blaschko, and
  Vedaldi}{Maji et~al\mbox{.}}{2013}]%
        {maji2013fine}
\bibfield{author}{\bibinfo{person}{Subhransu Maji}, \bibinfo{person}{Esa
  Rahtu}, \bibinfo{person}{Juho Kannala}, \bibinfo{person}{Matthew Blaschko},
  {and} \bibinfo{person}{Andrea Vedaldi}.} \bibinfo{year}{2013}\natexlab{}.
\newblock \showarticletitle{{Fine-Grained Visual Classification of Aircraft}}.
\newblock \bibinfo{journal}{\emph{arXiv preprint arXiv:1306.5151}}
  (\bibinfo{year}{2013}).
\newblock


\bibitem[\protect\citeauthoryear{Mao, Negi, Narayan, Wang, Yang, Wang, Marcus,
  Addanki, Khani, He, Nathan, Cangialosi, Venkatakrishnan, Weng, Han, Kraska,
  and Alizadeh}{Mao et~al\mbox{.}}{2019a}]%
        {mao2019park}
\bibfield{author}{\bibinfo{person}{Hongzi Mao}, \bibinfo{person}{Parimarjan
  Negi}, \bibinfo{person}{Akshay Narayan}, \bibinfo{person}{Hanrui Wang},
  \bibinfo{person}{Jiacheng Yang}, \bibinfo{person}{Haonan Wang},
  \bibinfo{person}{Ryan Marcus}, \bibinfo{person}{Ravichandra Addanki},
  \bibinfo{person}{Mehrdad Khani}, \bibinfo{person}{Songtao He},
  \bibinfo{person}{Vikram Nathan}, \bibinfo{person}{Frank Cangialosi},
  \bibinfo{person}{Shaileshh Venkatakrishnan}, \bibinfo{person}{Wei-Hung Weng},
  \bibinfo{person}{Song Han}, \bibinfo{person}{Tim Kraska}, {and}
  \bibinfo{person}{Mohammad Alizadeh}.} \bibinfo{year}{2019}\natexlab{a}.
\newblock \showarticletitle{{Park: An Open Platform for Learning-Augmented
  Computer Systems}}. In \bibinfo{booktitle}{\emph{Conference on Neural
  Information Processing Systems}}.
\newblock


\bibitem[\protect\citeauthoryear{Mao, Wang, and Li}{Mao et~al\mbox{.}}{2019b}]%
        {mao2019interpolated}
\bibfield{author}{\bibinfo{person}{Jiageng Mao}, \bibinfo{person}{Xiaogang
  Wang}, {and} \bibinfo{person}{Hongsheng Li}.}
  \bibinfo{year}{2019}\natexlab{b}.
\newblock \showarticletitle{{Interpolated Convolutional Networks for 3D Point
  Cloud Understanding}}. In \bibinfo{booktitle}{\emph{International Conference
  on Computer Vision}}.
\newblock


\bibitem[\protect\citeauthoryear{Maturana and Scherer}{Maturana and
  Scherer}{2015}]%
        {maturana2015voxnet}
\bibfield{author}{\bibinfo{person}{Daniel Maturana} {and}
  \bibinfo{person}{Sebastian Scherer}.} \bibinfo{year}{2015}\natexlab{}.
\newblock \showarticletitle{{VoxNet: A 3D Convolutional Neural Network for
  Real-Time Object Recognition}}. In \bibinfo{booktitle}{\emph{IEEE/RSJ
  International Conference on Intelligent Robots and Systems}}.
\newblock


\bibitem[\protect\citeauthoryear{McMahan, Moore, Ramage, Hampson, and
  y~Arcas}{McMahan et~al\mbox{.}}{2016}]%
        {mcmahan2016communication}
\bibfield{author}{\bibinfo{person}{Brendan McMahan}, \bibinfo{person}{Eider
  Moore}, \bibinfo{person}{Daniel Ramage}, \bibinfo{person}{Seth Hampson},
  {and} \bibinfo{person}{Blaise~Aguera y Arcas}.}
  \bibinfo{year}{2016}\natexlab{}.
\newblock \showarticletitle{{Communication-Efficient Learning of Deep Networks
  from Decentralized Data}}. In \bibinfo{booktitle}{\emph{International
  Conference on Artificial Intelligence and Statistics}}.
\newblock


\bibitem[\protect\citeauthoryear{Michel, Levy, and Neubig}{Michel
  et~al\mbox{.}}{2019}]%
        {michel2019sixteen}
\bibfield{author}{\bibinfo{person}{Paul Michel}, \bibinfo{person}{Omer Levy},
  {and} \bibinfo{person}{Graham Neubig}.} \bibinfo{year}{2019}\natexlab{}.
\newblock \showarticletitle{{Are Sixteen Heads Really Better Than One?}}. In
  \bibinfo{booktitle}{\emph{Conference on Neural Information Processing
  Systems}}.
\newblock


\bibitem[\protect\citeauthoryear{Mikolov, Karafi{\'a}t, Burget,
  {\v{C}}ernock{\`y}, and Khudanpur}{Mikolov et~al\mbox{.}}{2010}]%
        {mikolov2010recurrent}
\bibfield{author}{\bibinfo{person}{Tom{\'a}{\v{s}} Mikolov},
  \bibinfo{person}{Martin Karafi{\'a}t}, \bibinfo{person}{Luk{\'a}{\v{s}}
  Burget}, \bibinfo{person}{Jan {\v{C}}ernock{\`y}}, {and}
  \bibinfo{person}{Sanjeev Khudanpur}.} \bibinfo{year}{2010}\natexlab{}.
\newblock \showarticletitle{{Recurrent Neural Network Based Language Model}}.
  In \bibinfo{booktitle}{\emph{Conference of the International Speech
  Communication Association}}.
\newblock


\bibitem[\protect\citeauthoryear{Mirhoseini, Goldie, Pham, Steiner, Le, and
  Dean}{Mirhoseini et~al\mbox{.}}{2018}]%
        {mirhoseini2018hierarchical}
\bibfield{author}{\bibinfo{person}{Azalia Mirhoseini}, \bibinfo{person}{Anna
  Goldie}, \bibinfo{person}{Hieu Pham}, \bibinfo{person}{Benoit Steiner},
  \bibinfo{person}{Quoc~V Le}, {and} \bibinfo{person}{Jeff Dean}.}
  \bibinfo{year}{2018}\natexlab{}.
\newblock \showarticletitle{{A Hierarchical Model for Device Placement}}. In
  \bibinfo{booktitle}{\emph{International Conference on Learning
  Representations}}.
\newblock


\bibitem[\protect\citeauthoryear{Mirhoseini, Pham, Le, Steiner, Larsen, Zhou,
  Kumar, Norouzi, Bengio, and Dean}{Mirhoseini et~al\mbox{.}}{2017}]%
        {mirhoseini2017device}
\bibfield{author}{\bibinfo{person}{Azalia Mirhoseini}, \bibinfo{person}{Hieu
  Pham}, \bibinfo{person}{Quoc~V Le}, \bibinfo{person}{Benoit Steiner},
  \bibinfo{person}{Rasmus Larsen}, \bibinfo{person}{Yuefeng Zhou},
  \bibinfo{person}{Naveen Kumar}, \bibinfo{person}{Mohammad Norouzi},
  \bibinfo{person}{Samy Bengio}, {and} \bibinfo{person}{Jeff Dean}.}
  \bibinfo{year}{2017}\natexlab{}.
\newblock \showarticletitle{{Device Placement Optimization with Reinforcement
  Learning}}. In \bibinfo{booktitle}{\emph{International Conference on Machine
  Learning}}.
\newblock


\bibitem[\protect\citeauthoryear{Molchanov, Tyree, Karras, Aila, and
  Kautz}{Molchanov et~al\mbox{.}}{2017}]%
        {molchanov2017pruning}
\bibfield{author}{\bibinfo{person}{Pavlo Molchanov}, \bibinfo{person}{Stephen
  Tyree}, \bibinfo{person}{Tero Karras}, \bibinfo{person}{Timo Aila}, {and}
  \bibinfo{person}{Jan Kautz}.} \bibinfo{year}{2017}\natexlab{}.
\newblock \showarticletitle{{Pruning Convolutional Neural Networks for Resource
  Efficient Transfer Learning}}. In \bibinfo{booktitle}{\emph{International
  Conference on Learning Representations}}.
\newblock


\bibitem[\protect\citeauthoryear{Mudrakarta, Sandler, Zhmoginov, and
  Howard}{Mudrakarta et~al\mbox{.}}{2019}]%
        {mudrakarta2019k}
\bibfield{author}{\bibinfo{person}{Pramod~Kaushik Mudrakarta},
  \bibinfo{person}{Mark Sandler}, \bibinfo{person}{Andrey Zhmoginov}, {and}
  \bibinfo{person}{Andrew Howard}.} \bibinfo{year}{2019}\natexlab{}.
\newblock \showarticletitle{{K for the Price of 1: Parameter-Efficient
  Multi-Task and Transfer Learning}}. In
  \bibinfo{booktitle}{\emph{International Conference on Learning
  Representations}}.
\newblock


\bibitem[\protect\citeauthoryear{Nagel, Baalen, Blankevoort, and Welling}{Nagel
  et~al\mbox{.}}{2019}]%
        {nagel2019data}
\bibfield{author}{\bibinfo{person}{Markus Nagel}, \bibinfo{person}{Mart~van
  Baalen}, \bibinfo{person}{Tijmen Blankevoort}, {and} \bibinfo{person}{Max
  Welling}.} \bibinfo{year}{2019}\natexlab{}.
\newblock \showarticletitle{{Data-Free Quantization Through Weight Equalization
  and Bias Correction}}. In \bibinfo{booktitle}{\emph{International Conference
  on Computer Vision}}.
\newblock


\bibitem[\protect\citeauthoryear{Niu, Ma, Lin, Wang, Qian, Lin, Wang, and
  Ren}{Niu et~al\mbox{.}}{2020}]%
        {niu2020patdnn}
\bibfield{author}{\bibinfo{person}{Wei Niu}, \bibinfo{person}{Xiaolong Ma},
  \bibinfo{person}{Sheng Lin}, \bibinfo{person}{Shihao Wang},
  \bibinfo{person}{Xuehai Qian}, \bibinfo{person}{Xue Lin},
  \bibinfo{person}{Yanzhi Wang}, {and} \bibinfo{person}{Bin Ren}.}
  \bibinfo{year}{2020}\natexlab{}.
\newblock \showarticletitle{{PatDNN: Achieving Real-Time DNN Execution on
  Mobile Devices with Pattern-Based Weight Pruning}}. In
  \bibinfo{booktitle}{\emph{International Conference on Architectural Support
  for Programming Languages and Operating Systems}}.
\newblock


\bibitem[\protect\citeauthoryear{Ota, Dao, Mezaris, and Natale}{Ota
  et~al\mbox{.}}{2017}]%
        {ota2017deep}
\bibfield{author}{\bibinfo{person}{Kaoru Ota}, \bibinfo{person}{Minh~Son Dao},
  \bibinfo{person}{Vasileios Mezaris}, {and} \bibinfo{person}{Francesco G.
  B.~De Natale}.} \bibinfo{year}{2017}\natexlab{}.
\newblock \showarticletitle{{Deep Learning for Mobile Multimedia: A Survey}}.
\newblock \bibinfo{journal}{\emph{ACM Transactions on Multimedia Computing,
  Communications, and Applications}} \bibinfo{volume}{13}, \bibinfo{number}{35}
  (\bibinfo{year}{2017}), \bibinfo{pages}{1--22}.
\newblock


\bibitem[\protect\citeauthoryear{Pal, Beaumont, Park, Amarnath, Feng,
  Chakrabarti, Kim, Blaauw, Mudge, and Dreslinski}{Pal et~al\mbox{.}}{2018}]%
        {pal2018outerspace}
\bibfield{author}{\bibinfo{person}{Subhankar Pal}, \bibinfo{person}{Jonathan
  Beaumont}, \bibinfo{person}{Dong-Hyeon Park}, \bibinfo{person}{Aporva
  Amarnath}, \bibinfo{person}{Siying Feng}, \bibinfo{person}{Chaitali
  Chakrabarti}, \bibinfo{person}{Hun-Seok Kim}, \bibinfo{person}{David Blaauw},
  \bibinfo{person}{Trevor Mudge}, {and} \bibinfo{person}{Ronald Dreslinski}.}
  \bibinfo{year}{2018}\natexlab{}.
\newblock \showarticletitle{{OuterSPACE: An Outer Product Based Sparse Matrix
  Multiplication Accelerator}}. In \bibinfo{booktitle}{\emph{IEEE International
  Symposium on High-Performance Computer Architecture}}.
\newblock


\bibitem[\protect\citeauthoryear{Parashar, Rhu, Mukkara, Puglielli, Venkatesan,
  Khailany, Emer, Keckler, and Dally}{Parashar et~al\mbox{.}}{2017}]%
        {parashar2017scnn}
\bibfield{author}{\bibinfo{person}{Angshuman Parashar}, \bibinfo{person}{Minsoo
  Rhu}, \bibinfo{person}{Anurag Mukkara}, \bibinfo{person}{Antonio Puglielli},
  \bibinfo{person}{Rangharajan Venkatesan}, \bibinfo{person}{Brucek Khailany},
  \bibinfo{person}{Joel Emer}, \bibinfo{person}{Stephen~W Keckler}, {and}
  \bibinfo{person}{William~J Dally}.} \bibinfo{year}{2017}\natexlab{}.
\newblock \showarticletitle{{SCNN: An Accelerator for Compressed-Sparse
  Convolutional Neural Networks}}. In \bibinfo{booktitle}{\emph{International
  Symposium on Computer Architecture}}.
\newblock


\bibitem[\protect\citeauthoryear{Park, Li, Wen, Tang, Li, Chen, and Dubey}{Park
  et~al\mbox{.}}{2016}]%
        {park2016faster}
\bibfield{author}{\bibinfo{person}{Jongsoo Park}, \bibinfo{person}{Sheng Li},
  \bibinfo{person}{Wei Wen}, \bibinfo{person}{Ping Tak~Peter Tang},
  \bibinfo{person}{Hai Li}, \bibinfo{person}{Yiran Chen}, {and}
  \bibinfo{person}{Pradeep Dubey}.} \bibinfo{year}{2016}\natexlab{}.
\newblock \showarticletitle{{Faster CNNs with Direct Sparse Convolutions and
  Guided Pruning}}.
\newblock \bibinfo{journal}{\emph{arXiv preprint arXiv:1608.01409}}
  (\bibinfo{year}{2016}).
\newblock


\bibitem[\protect\citeauthoryear{Park, Bong, Shin, Lee, Choi, and Yoo}{Park
  et~al\mbox{.}}{2015}]%
        {park20154}
\bibfield{author}{\bibinfo{person}{Seongwook Park},
  \bibinfo{person}{Kyeongryeol Bong}, \bibinfo{person}{Dongjoo Shin},
  \bibinfo{person}{Jinmook Lee}, \bibinfo{person}{Sungpill Choi}, {and}
  \bibinfo{person}{Hoi-Jun Yoo}.} \bibinfo{year}{2015}\natexlab{}.
\newblock \showarticletitle{{A 1.93TOPS/W Scalable Deep Learning/Inference
  Processor with Tetra-Parallel MIMD Architecture for Big-Data Applications}}.
  In \bibinfo{booktitle}{\emph{IEEE International Solid-State Circuits
  Conference}}.
\newblock


\bibitem[\protect\citeauthoryear{Park, Jang, Kim, Na, and Yoon}{Park
  et~al\mbox{.}}{2020}]%
        {park2020memory}
\bibfield{author}{\bibinfo{person}{Seongsik Park}, \bibinfo{person}{Jaehee
  Jang}, \bibinfo{person}{Seijoon Kim}, \bibinfo{person}{Byunggook Na}, {and}
  \bibinfo{person}{Sungroh Yoon}.} \bibinfo{year}{2020}\natexlab{}.
\newblock \showarticletitle{{Memory-Augmented Neural Networks on FPGA for
  Real-Time and Energy-Efficient Question Answering}}.
\newblock \bibinfo{journal}{\emph{IEEE Transactions on Very Large Scale
  Integration (VLSI) Systems}} (\bibinfo{year}{2020}).
\newblock


\bibitem[\protect\citeauthoryear{Parmar, Vaswani, Uszkoreit, Kaiser, Shazeer,
  Ku, and Tran}{Parmar et~al\mbox{.}}{2018}]%
        {parmar2018image}
\bibfield{author}{\bibinfo{person}{Niki Parmar}, \bibinfo{person}{Ashish
  Vaswani}, \bibinfo{person}{Jakob Uszkoreit}, \bibinfo{person}{Lukasz Kaiser},
  \bibinfo{person}{Noam Shazeer}, \bibinfo{person}{Alexander Ku}, {and}
  \bibinfo{person}{Dustin Tran}.} \bibinfo{year}{2018}\natexlab{}.
\newblock \showarticletitle{{Image Transformer}}. In
  \bibinfo{booktitle}{\emph{International Conference on Machine Learning}}.
\newblock


\bibitem[\protect\citeauthoryear{Paszke, Gross, Massa, Lerer, Bradbury, Chanan,
  Killeen, Lin, Gimelshein, Antiga, Desmaison, Kopf, Yang, DeVito, Raison,
  Tejani, Chilamkurthy, Steiner, Fang, Bai, and Chintala}{Paszke
  et~al\mbox{.}}{2019}]%
        {paszke2019pytorch}
\bibfield{author}{\bibinfo{person}{Adam Paszke}, \bibinfo{person}{Sam Gross},
  \bibinfo{person}{Francisco Massa}, \bibinfo{person}{Adam Lerer},
  \bibinfo{person}{James Bradbury}, \bibinfo{person}{Gregory Chanan},
  \bibinfo{person}{Trevor Killeen}, \bibinfo{person}{Zeming Lin},
  \bibinfo{person}{Natalia Gimelshein}, \bibinfo{person}{Luca Antiga},
  \bibinfo{person}{Alban Desmaison}, \bibinfo{person}{Andreas Kopf},
  \bibinfo{person}{Edward Yang}, \bibinfo{person}{Zachary DeVito},
  \bibinfo{person}{Martin Raison}, \bibinfo{person}{Alykhan Tejani},
  \bibinfo{person}{Sasank Chilamkurthy}, \bibinfo{person}{Benoit Steiner},
  \bibinfo{person}{Lu Fang}, \bibinfo{person}{Junjie Bai}, {and}
  \bibinfo{person}{Soumith Chintala}.} \bibinfo{year}{2019}\natexlab{}.
\newblock \showarticletitle{{PyTorch: An Imperative Style, High-Performance
  Deep Learning Library}}. In \bibinfo{booktitle}{\emph{Conference on Neural
  Information Processing Systems}}.
\newblock


\bibitem[\protect\citeauthoryear{Peemen, Setio, Mesman, and Corporaal}{Peemen
  et~al\mbox{.}}{2013}]%
        {peemen2013memory}
\bibfield{author}{\bibinfo{person}{Maurice Peemen}, \bibinfo{person}{Arnaud~AA
  Setio}, \bibinfo{person}{Bart Mesman}, {and} \bibinfo{person}{Henk
  Corporaal}.} \bibinfo{year}{2013}\natexlab{}.
\newblock \showarticletitle{{Memory-Centric Accelerator Design for
  Convolutional Neural Networks}}. In \bibinfo{booktitle}{\emph{IEEE
  International Conference on Computer Design}}.
\newblock


\bibitem[\protect\citeauthoryear{Pham, Guan, Zoph, Le, and Dean}{Pham
  et~al\mbox{.}}{2018}]%
        {pham2018efficient}
\bibfield{author}{\bibinfo{person}{Hieu Pham}, \bibinfo{person}{Melody~Y Guan},
  \bibinfo{person}{Barret Zoph}, \bibinfo{person}{Quoc~V Le}, {and}
  \bibinfo{person}{Jeff Dean}.} \bibinfo{year}{2018}\natexlab{}.
\newblock \showarticletitle{{Efficient Neural Architecture Search via Parameter
  Sharing}}. In \bibinfo{booktitle}{\emph{International Conference on Machine
  Learning}}.
\newblock


\bibitem[\protect\citeauthoryear{Piergiovanni, Angelova, and Ryoo}{Piergiovanni
  et~al\mbox{.}}{2019a}]%
        {piergiovanni2019tiny}
\bibfield{author}{\bibinfo{person}{AJ Piergiovanni}, \bibinfo{person}{Anelia
  Angelova}, {and} \bibinfo{person}{Michael~S Ryoo}.}
  \bibinfo{year}{2019}\natexlab{a}.
\newblock \showarticletitle{{Tiny Video Networks}}.
\newblock \bibinfo{journal}{\emph{arXiv preprint arXiv:1910.06961}}
  (\bibinfo{year}{2019}).
\newblock


\bibitem[\protect\citeauthoryear{Piergiovanni, Angelova, Toshev, and
  Ryoo}{Piergiovanni et~al\mbox{.}}{2019b}]%
        {piergiovanni2019evolving}
\bibfield{author}{\bibinfo{person}{AJ Piergiovanni}, \bibinfo{person}{Anelia
  Angelova}, \bibinfo{person}{Alexander Toshev}, {and}
  \bibinfo{person}{Michael~S Ryoo}.} \bibinfo{year}{2019}\natexlab{b}.
\newblock \showarticletitle{{Evolving Space-Time Neural Architectures for
  Videos}}. In \bibinfo{booktitle}{\emph{International Conference on Computer
  Vision}}.
\newblock


\bibitem[\protect\citeauthoryear{Qi, Litany, He, and Guibas}{Qi
  et~al\mbox{.}}{2019}]%
        {qi2019deep}
\bibfield{author}{\bibinfo{person}{Charles~R Qi}, \bibinfo{person}{Or Litany},
  \bibinfo{person}{Kaiming He}, {and} \bibinfo{person}{Leonidas~J Guibas}.}
  \bibinfo{year}{2019}\natexlab{}.
\newblock \showarticletitle{{Deep Hough Voting for 3D Object Detection in Point
  Clouds}}. In \bibinfo{booktitle}{\emph{International Conference on Computer
  Vision}}.
\newblock


\bibitem[\protect\citeauthoryear{Qi, Liu, Wu, Su, and Guibas}{Qi
  et~al\mbox{.}}{2018}]%
        {qi2018frustum}
\bibfield{author}{\bibinfo{person}{Charles~R Qi}, \bibinfo{person}{Wei Liu},
  \bibinfo{person}{Chenxia Wu}, \bibinfo{person}{Hao Su}, {and}
  \bibinfo{person}{Leonidas~J Guibas}.} \bibinfo{year}{2018}\natexlab{}.
\newblock \showarticletitle{{Frustum PointNets for 3D Object Detection from
  RGB-D Data}}. In \bibinfo{booktitle}{\emph{IEEE Conference on Computer Vision
  and Pattern Recognition}}.
\newblock


\bibitem[\protect\citeauthoryear{Qi, Su, Mo, and Guibas}{Qi
  et~al\mbox{.}}{2017a}]%
        {qi2017pointnet}
\bibfield{author}{\bibinfo{person}{Charles~R Qi}, \bibinfo{person}{Hao Su},
  \bibinfo{person}{Kaichun Mo}, {and} \bibinfo{person}{Leonidas~J Guibas}.}
  \bibinfo{year}{2017}\natexlab{a}.
\newblock \showarticletitle{{PointNet: Deep Learning on Point Sets for 3D
  Classification and Segmentation}}. In \bibinfo{booktitle}{\emph{IEEE
  Conference on Computer Vision and Pattern Recognition}}.
\newblock


\bibitem[\protect\citeauthoryear{Qi, Su, Niessner, Dai, Yan, and Guibas}{Qi
  et~al\mbox{.}}{2016}]%
        {qi2016volumetric}
\bibfield{author}{\bibinfo{person}{Charles~R Qi}, \bibinfo{person}{Hao Su},
  \bibinfo{person}{Matthias Niessner}, \bibinfo{person}{Angela Dai},
  \bibinfo{person}{Mengyuan Yan}, {and} \bibinfo{person}{Leonidas~J Guibas}.}
  \bibinfo{year}{2016}\natexlab{}.
\newblock \showarticletitle{{Volumetric and Multi-View CNNs for Object
  Classification on 3D Data}}. In \bibinfo{booktitle}{\emph{IEEE Conference on
  Computer Vision and Pattern Recognition}}.
\newblock


\bibitem[\protect\citeauthoryear{Qi, Yi, Su, and Guibas}{Qi
  et~al\mbox{.}}{2017b}]%
        {qi2017pointnet++}
\bibfield{author}{\bibinfo{person}{Charles~R Qi}, \bibinfo{person}{Li Yi},
  \bibinfo{person}{Hao Su}, {and} \bibinfo{person}{Leonidas~J Guibas}.}
  \bibinfo{year}{2017}\natexlab{b}.
\newblock \showarticletitle{{PointNet++: Deep Hierarchical Feature Learning on
  Point Sets in a Metric Space}}. In \bibinfo{booktitle}{\emph{Conference on
  Neural Information Processing Systems}}.
\newblock


\bibitem[\protect\citeauthoryear{Qin, Samajdar, Kwon, Nadella, Srinivasan, Das,
  Kaul, and Krishna}{Qin et~al\mbox{.}}{2020}]%
        {qin2020sigma}
\bibfield{author}{\bibinfo{person}{Eric Qin}, \bibinfo{person}{Ananda
  Samajdar}, \bibinfo{person}{Hyoukjun Kwon}, \bibinfo{person}{Vineet Nadella},
  \bibinfo{person}{Sudarshan Srinivasan}, \bibinfo{person}{Dipankar Das},
  \bibinfo{person}{Bharat Kaul}, {and} \bibinfo{person}{Tushar Krishna}.}
  \bibinfo{year}{2020}\natexlab{}.
\newblock \showarticletitle{{SIGMA: A Sparse and Irregular GEMM Accelerator
  with Flexible Interconnects for DNN Training}}. In
  \bibinfo{booktitle}{\emph{IEEE International Symposium on High-Performance
  Computer Architecture}}.
\newblock


\bibitem[\protect\citeauthoryear{Qiu, Ma, Levy, Yih, Wang, and Tang}{Qiu
  et~al\mbox{.}}{2019}]%
        {qiu2019blockwise}
\bibfield{author}{\bibinfo{person}{Jiezhong Qiu}, \bibinfo{person}{Hao Ma},
  \bibinfo{person}{Omer Levy}, \bibinfo{person}{Scott Wen-tau Yih},
  \bibinfo{person}{Sinong Wang}, {and} \bibinfo{person}{Jie Tang}.}
  \bibinfo{year}{2019}\natexlab{}.
\newblock \showarticletitle{{Blockwise Self-Attention for Long Document
  Understanding}}.
\newblock \bibinfo{journal}{\emph{arXiv preprint arXiv:1911.02972}}
  (\bibinfo{year}{2019}).
\newblock


\bibitem[\protect\citeauthoryear{Qiu, Yao, and Mei}{Qiu et~al\mbox{.}}{2017}]%
        {qiu2017learning}
\bibfield{author}{\bibinfo{person}{Zhaofan Qiu}, \bibinfo{person}{Ting Yao},
  {and} \bibinfo{person}{Tao Mei}.} \bibinfo{year}{2017}\natexlab{}.
\newblock \showarticletitle{{Learning Spatio-Temporal Representation with
  Pseudo-3D Residual Networks}}. In \bibinfo{booktitle}{\emph{International
  Conference on Computer Vision}}.
\newblock


\bibitem[\protect\citeauthoryear{Radford, Wu, Child, Luan, Amodei, and
  Sutskever}{Radford et~al\mbox{.}}{2019}]%
        {radford2019language}
\bibfield{author}{\bibinfo{person}{Alec Radford}, \bibinfo{person}{Jeff Wu},
  \bibinfo{person}{Rewon Child}, \bibinfo{person}{David Luan},
  \bibinfo{person}{Dario Amodei}, {and} \bibinfo{person}{Ilya Sutskever}.}
  \bibinfo{year}{2019}\natexlab{}.
\newblock \showarticletitle{{Language Models are Unsupervised Multitask
  Learners}}.
\newblock  (\bibinfo{year}{2019}).
\newblock


\bibitem[\protect\citeauthoryear{Radosavovic, Kosaraju, Girshick, He, and
  Doll{\'a}r}{Radosavovic et~al\mbox{.}}{2020}]%
        {radosavovic2020designing}
\bibfield{author}{\bibinfo{person}{Ilija Radosavovic},
  \bibinfo{person}{Raj~Prateek Kosaraju}, \bibinfo{person}{Ross Girshick},
  \bibinfo{person}{Kaiming He}, {and} \bibinfo{person}{Piotr Doll{\'a}r}.}
  \bibinfo{year}{2020}\natexlab{}.
\newblock \showarticletitle{{Designing Network Design Spaces}}. In
  \bibinfo{booktitle}{\emph{IEEE Conference on Computer Vision and Pattern
  Recognition}}.
\newblock


\bibitem[\protect\citeauthoryear{Rajpurkar, Zhang, Lopyrev, and
  Liang}{Rajpurkar et~al\mbox{.}}{2016}]%
        {rajpurkar2016squad}
\bibfield{author}{\bibinfo{person}{Pranav Rajpurkar}, \bibinfo{person}{Jian
  Zhang}, \bibinfo{person}{Konstantin Lopyrev}, {and} \bibinfo{person}{Percy
  Liang}.} \bibinfo{year}{2016}\natexlab{}.
\newblock \showarticletitle{{SQuAD: 100,000+ Questions for Machine
  Comprehension of Text}}. In \bibinfo{booktitle}{\emph{Conference on Empirical
  Methods in Natural Language Processing}}.
\newblock


\bibitem[\protect\citeauthoryear{Ramachandran, Zoph, and Le}{Ramachandran
  et~al\mbox{.}}{2017}]%
        {ramachandran2017searching}
\bibfield{author}{\bibinfo{person}{Prajit Ramachandran},
  \bibinfo{person}{Barret Zoph}, {and} \bibinfo{person}{Quoc~V Le}.}
  \bibinfo{year}{2017}\natexlab{}.
\newblock \showarticletitle{{Searching for Activation Functions}}.
\newblock \bibinfo{journal}{\emph{arXiv preprint arXiv:1710.05941}}
  (\bibinfo{year}{2017}).
\newblock


\bibitem[\protect\citeauthoryear{Rastegari, Ordonez, Redmon, and
  Farhadi}{Rastegari et~al\mbox{.}}{2016}]%
        {rastegari2016xnor}
\bibfield{author}{\bibinfo{person}{Mohammad Rastegari},
  \bibinfo{person}{Vicente Ordonez}, \bibinfo{person}{Joseph Redmon}, {and}
  \bibinfo{person}{Ali Farhadi}.} \bibinfo{year}{2016}\natexlab{}.
\newblock \showarticletitle{{XNOR-Net: ImageNet Classification Using Binary
  Convolutional Neural Networks}}. In \bibinfo{booktitle}{\emph{European
  Conference on Computer Vision}}.
\newblock


\bibitem[\protect\citeauthoryear{Ravi, Reizenstein, Novotny, Gordon, Lo,
  Johnson, and Gkioxari}{Ravi et~al\mbox{.}}{2020}]%
        {ravi2020accelerating}
\bibfield{author}{\bibinfo{person}{Nikhila Ravi}, \bibinfo{person}{Jeremy
  Reizenstein}, \bibinfo{person}{David Novotny}, \bibinfo{person}{Taylor
  Gordon}, \bibinfo{person}{Wan-Yen Lo}, \bibinfo{person}{Justin Johnson},
  {and} \bibinfo{person}{Georgia Gkioxari}.} \bibinfo{year}{2020}\natexlab{}.
\newblock \showarticletitle{{Accelerating 3D Deep Learning with PyTorch3D}}.
\newblock \bibinfo{journal}{\emph{arXiv preprint arXiv:2007.08501}}
  (\bibinfo{year}{2020}).
\newblock


\bibitem[\protect\citeauthoryear{Real, Aggarwal, Huang, and Le}{Real
  et~al\mbox{.}}{2019}]%
        {real2019regularized}
\bibfield{author}{\bibinfo{person}{Esteban Real}, \bibinfo{person}{Alok
  Aggarwal}, \bibinfo{person}{Yanping Huang}, {and} \bibinfo{person}{Quoc~V
  Le}.} \bibinfo{year}{2019}\natexlab{}.
\newblock \showarticletitle{{Regularized Evolution for Image Classifier
  Architecture Search}}. In \bibinfo{booktitle}{\emph{AAAI Conference on
  Artificial Intelligence}}.
\newblock


\bibitem[\protect\citeauthoryear{Riegler, Ulusoy, and Geiger}{Riegler
  et~al\mbox{.}}{2017}]%
        {riegler2017octnet}
\bibfield{author}{\bibinfo{person}{Gernot Riegler}, \bibinfo{person}{Ali~Osman
  Ulusoy}, {and} \bibinfo{person}{Andreas Geiger}.}
  \bibinfo{year}{2017}\natexlab{}.
\newblock \showarticletitle{{OctNet: Learning Deep 3D Representations at High
  Resolutions}}. In \bibinfo{booktitle}{\emph{IEEE Conference on Computer
  Vision and Pattern Recognition}}.
\newblock


\bibitem[\protect\citeauthoryear{Romero, Ballas, Kahou, Chassang, Gatta, and
  Bengio}{Romero et~al\mbox{.}}{2014}]%
        {romero2014fitnets}
\bibfield{author}{\bibinfo{person}{Adriana Romero}, \bibinfo{person}{Nicolas
  Ballas}, \bibinfo{person}{Samira~Ebrahimi Kahou}, \bibinfo{person}{Antoine
  Chassang}, \bibinfo{person}{Carlo Gatta}, {and} \bibinfo{person}{Yoshua
  Bengio}.} \bibinfo{year}{2014}\natexlab{}.
\newblock \showarticletitle{{FitNets: Hints for Thin Deep Nets}}.
\newblock \bibinfo{journal}{\emph{arXiv preprint arXiv:1412.6550}}
  (\bibinfo{year}{2014}).
\newblock


\bibitem[\protect\citeauthoryear{Roy, Saffar, Vaswani, and Grangier}{Roy
  et~al\mbox{.}}{2020}]%
        {roy2020efficient}
\bibfield{author}{\bibinfo{person}{Aurko Roy}, \bibinfo{person}{Mohammad
  Saffar}, \bibinfo{person}{Ashish Vaswani}, {and} \bibinfo{person}{David
  Grangier}.} \bibinfo{year}{2020}\natexlab{}.
\newblock \showarticletitle{{Efficient Content-Based Sparse Attention with
  Routing Transformers}}.
\newblock \bibinfo{journal}{\emph{Transactions of the Association for
  Computational Linguistics}} \bibinfo{volume}{9}, \bibinfo{number}{3}
  (\bibinfo{year}{2020}), \bibinfo{pages}{53--68}.
\newblock


\bibitem[\protect\citeauthoryear{Rusci, Fariselli, Capotondi, and Benini}{Rusci
  et~al\mbox{.}}{2020}]%
        {rusci2020leveraging}
\bibfield{author}{\bibinfo{person}{Manuele Rusci}, \bibinfo{person}{Marco
  Fariselli}, \bibinfo{person}{Alessandro Capotondi}, {and}
  \bibinfo{person}{Luca Benini}.} \bibinfo{year}{2020}\natexlab{}.
\newblock \showarticletitle{{Leveraging Automated Mixed-Low-Precision
  Quantization for Tiny Edge Microcontrollers}}.
\newblock \bibinfo{journal}{\emph{arXiv preprint arXiv:2008.05124}}
  (\bibinfo{year}{2020}).
\newblock


\bibitem[\protect\citeauthoryear{Ryoo, Piergiovanni, Tan, and Angelova}{Ryoo
  et~al\mbox{.}}{2020}]%
        {ryoo2020assemblenet}
\bibfield{author}{\bibinfo{person}{Michael~S Ryoo}, \bibinfo{person}{AJ
  Piergiovanni}, \bibinfo{person}{Mingxing Tan}, {and} \bibinfo{person}{Anelia
  Angelova}.} \bibinfo{year}{2020}\natexlab{}.
\newblock \showarticletitle{{AssembleNet: Searching for Multi-Stream Neural
  Connectivity in Video Architectures}}. In
  \bibinfo{booktitle}{\emph{International Conference on Learning
  Representations}}.
\newblock


\bibitem[\protect\citeauthoryear{Rzayev, Moradi, Albonesi, and Manchar}{Rzayev
  et~al\mbox{.}}{2017}]%
        {rzayev2017deeprecon}
\bibfield{author}{\bibinfo{person}{Tayyar Rzayev}, \bibinfo{person}{Saber
  Moradi}, \bibinfo{person}{David~H Albonesi}, {and} \bibinfo{person}{Rajit
  Manchar}.} \bibinfo{year}{2017}\natexlab{}.
\newblock \showarticletitle{{DeepRecon: Dynamically Reconfigurable Architecture
  for Accelerating Deep Neural Networks}}. In
  \bibinfo{booktitle}{\emph{International Joint Conference on Neural
  Networks}}.
\newblock


\bibitem[\protect\citeauthoryear{Sandler, Howard, Zhu, Zhmoginov, and
  Chen}{Sandler et~al\mbox{.}}{2018}]%
        {sandler2018mobilenetv2}
\bibfield{author}{\bibinfo{person}{Mark Sandler}, \bibinfo{person}{Andrew
  Howard}, \bibinfo{person}{Menglong Zhu}, \bibinfo{person}{Andrey Zhmoginov},
  {and} \bibinfo{person}{Liang-Chieh Chen}.} \bibinfo{year}{2018}\natexlab{}.
\newblock \showarticletitle{{MobileNetV2: Inverted Residuals and Linear
  Bottlenecks}}. In \bibinfo{booktitle}{\emph{IEEE Conference on Computer
  Vision and Pattern Recognition}}.
\newblock


\bibitem[\protect\citeauthoryear{Sanh, Debut, Chaumond, and Wolf}{Sanh
  et~al\mbox{.}}{2019}]%
        {sanh2019distilbert}
\bibfield{author}{\bibinfo{person}{Victor Sanh}, \bibinfo{person}{Lysandre
  Debut}, \bibinfo{person}{Julien Chaumond}, {and} \bibinfo{person}{Thomas
  Wolf}.} \bibinfo{year}{2019}\natexlab{}.
\newblock \showarticletitle{{DistilBERT, A Distilled Version of BERT: Smaller,
  Faster, Cheaper and Lighter}}. In \bibinfo{booktitle}{\emph{Conference on
  Neural Information Processing Systems}}.
\newblock


\bibitem[\protect\citeauthoryear{Sankaradas, Jakkula, Cadambi, Chakradhar,
  Durdanovic, Cosatto, and Graf}{Sankaradas et~al\mbox{.}}{2009}]%
        {sankaradas2009massively}
\bibfield{author}{\bibinfo{person}{Murugan Sankaradas},
  \bibinfo{person}{Venkata Jakkula}, \bibinfo{person}{Srihari Cadambi},
  \bibinfo{person}{Srimat Chakradhar}, \bibinfo{person}{Igor Durdanovic},
  \bibinfo{person}{Eric Cosatto}, {and} \bibinfo{person}{Hans~Peter Graf}.}
  \bibinfo{year}{2009}\natexlab{}.
\newblock \showarticletitle{{A Massively Parallel Coprocessor for Convolutional
  Neural Networks}}. In \bibinfo{booktitle}{\emph{International Conference on
  Application-Specific Systems, Architectures and Processors}}.
\newblock


\bibitem[\protect\citeauthoryear{Seide, Fu, Droppo, Li, and Yu}{Seide
  et~al\mbox{.}}{2014}]%
        {onebit}
\bibfield{author}{\bibinfo{person}{Frank Seide}, \bibinfo{person}{Hao Fu},
  \bibinfo{person}{Jasha Droppo}, \bibinfo{person}{Gang Li}, {and}
  \bibinfo{person}{Dong Yu}.} \bibinfo{year}{2014}\natexlab{}.
\newblock \showarticletitle{{1-Bit Stochastic Gradient Descent and Application
  to Data-Parallel Distributed Training of Speech DNNs}}. In
  \bibinfo{booktitle}{\emph{Conference of the International Speech
  Communication Association}}.
\newblock


\bibitem[\protect\citeauthoryear{Sharif~Razavian, Azizpour, Sullivan, and
  Carlsson}{Sharif~Razavian et~al\mbox{.}}{2014}]%
        {sharif2014cnn}
\bibfield{author}{\bibinfo{person}{Ali Sharif~Razavian},
  \bibinfo{person}{Hossein Azizpour}, \bibinfo{person}{Josephine Sullivan},
  {and} \bibinfo{person}{Stefan Carlsson}.} \bibinfo{year}{2014}\natexlab{}.
\newblock \showarticletitle{{CNN Features Off-The-Shelf: An Astounding Baseline
  for Recognition}}.
\newblock \bibinfo{journal}{\emph{arXiv preprint arXiv:1403.6382}}
  (\bibinfo{year}{2014}).
\newblock


\bibitem[\protect\citeauthoryear{Sharify, Lascorz, Siu, Judd, and
  Moshovos}{Sharify et~al\mbox{.}}{2018}]%
        {sharify2018loom}
\bibfield{author}{\bibinfo{person}{Sayeh Sharify},
  \bibinfo{person}{Alberto~Delmas Lascorz}, \bibinfo{person}{Kevin Siu},
  \bibinfo{person}{Patrick Judd}, {and} \bibinfo{person}{Andreas Moshovos}.}
  \bibinfo{year}{2018}\natexlab{}.
\newblock \showarticletitle{{Loom: Exploiting Weight and Activation Precisions
  to Accelerate Convolutional Neural Networks}}. In
  \bibinfo{booktitle}{\emph{Design Automation Conference}}.
\newblock


\bibitem[\protect\citeauthoryear{Sharma, Park, Suda, Lai, Chau, Chandra, and
  Esmaeilzadeh}{Sharma et~al\mbox{.}}{2018}]%
        {sharma2018bit}
\bibfield{author}{\bibinfo{person}{Hardik Sharma}, \bibinfo{person}{Jongse
  Park}, \bibinfo{person}{Naveen Suda}, \bibinfo{person}{Liangzhen Lai},
  \bibinfo{person}{Benson Chau}, \bibinfo{person}{Vikas Chandra}, {and}
  \bibinfo{person}{Hadi Esmaeilzadeh}.} \bibinfo{year}{2018}\natexlab{}.
\newblock \showarticletitle{{Bit Fusion: Bit-Level Dynamically Composable
  Architecture for Accelerating Deep Neural Networks}}. In
  \bibinfo{booktitle}{\emph{International Symposium on Computer Architecture}}.
\newblock


\bibitem[\protect\citeauthoryear{Shen, Baevski, Morcos, Keutzer, Auli, and
  Kiela}{Shen et~al\mbox{.}}{2021}]%
        {shen2021reservoir}
\bibfield{author}{\bibinfo{person}{Sheng Shen}, \bibinfo{person}{Alexei
  Baevski}, \bibinfo{person}{Ari~S Morcos}, \bibinfo{person}{Kurt Keutzer},
  \bibinfo{person}{Michael Auli}, {and} \bibinfo{person}{Douwe Kiela}.}
  \bibinfo{year}{2021}\natexlab{}.
\newblock \showarticletitle{{Reservoir Transformer}}. In
  \bibinfo{booktitle}{\emph{Conference of the Association for Computational
  Linguistics}}.
\newblock


\bibitem[\protect\citeauthoryear{Shen, Dong, Ye, Ma, Yao, Gholami, Mahoney, and
  Keutzer}{Shen et~al\mbox{.}}{2020}]%
        {shen2020q}
\bibfield{author}{\bibinfo{person}{Sheng Shen}, \bibinfo{person}{Zhen Dong},
  \bibinfo{person}{Jiayu Ye}, \bibinfo{person}{Linjian Ma},
  \bibinfo{person}{Zhewei Yao}, \bibinfo{person}{Amir Gholami},
  \bibinfo{person}{Michael~W Mahoney}, {and} \bibinfo{person}{Kurt Keutzer}.}
  \bibinfo{year}{2020}\natexlab{}.
\newblock \showarticletitle{{Q-BERT: Hessian Based Ultra Low Precision
  Quantization of BERT}}. In \bibinfo{booktitle}{\emph{AAAI Conference on
  Artificial Intelligence}}.
\newblock


\bibitem[\protect\citeauthoryear{Shi, Guo, Jiang, Wang, Shi, Wang, and Li}{Shi
  et~al\mbox{.}}{2020a}]%
        {shi2019pvrcnn}
\bibfield{author}{\bibinfo{person}{Shaoshuai Shi}, \bibinfo{person}{Chaoxu
  Guo}, \bibinfo{person}{Li Jiang}, \bibinfo{person}{Zhe Wang},
  \bibinfo{person}{Jianping Shi}, \bibinfo{person}{Xiaogang Wang}, {and}
  \bibinfo{person}{Hongsheng Li}.} \bibinfo{year}{2020}\natexlab{a}.
\newblock \showarticletitle{{PV-RCNN: Point-Voxel Feature Set Abstraction for
  3D Object Detection}}. In \bibinfo{booktitle}{\emph{IEEE Conference on
  Computer Vision and Pattern Recognition}}.
\newblock


\bibitem[\protect\citeauthoryear{Shi, Wang, and Li}{Shi et~al\mbox{.}}{2019}]%
        {shi2019pointrcnn}
\bibfield{author}{\bibinfo{person}{Shaoshuai Shi}, \bibinfo{person}{Xiaogang
  Wang}, {and} \bibinfo{person}{Hongsheng Li}.}
  \bibinfo{year}{2019}\natexlab{}.
\newblock \showarticletitle{{PointRCNN: 3D Object Proposal Generation and
  Detection from Point Cloud}}. In \bibinfo{booktitle}{\emph{IEEE Conference on
  Computer Vision and Pattern Recognition}}.
\newblock


\bibitem[\protect\citeauthoryear{Shi, Wang, Shi, Wang, and Li}{Shi
  et~al\mbox{.}}{2020b}]%
        {shi2019parta2}
\bibfield{author}{\bibinfo{person}{Shaoshuai Shi}, \bibinfo{person}{Zhe Wang},
  \bibinfo{person}{Jianping Shi}, \bibinfo{person}{Xiaogang Wang}, {and}
  \bibinfo{person}{Hongsheng Li}.} \bibinfo{year}{2020}\natexlab{b}.
\newblock \showarticletitle{{From Points to Parts: 3D Object Detection from
  Point Cloud with Part-Aware and Part-Aggregation Network}}.
\newblock \bibinfo{journal}{\emph{IEEE Transactions on Pattern Analysis and
  Machine Intelligence}} \bibinfo{volume}{43}, \bibinfo{number}{08}
  (\bibinfo{year}{2020}), \bibinfo{pages}{2647--2664}.
\newblock


\bibitem[\protect\citeauthoryear{Simonyan and Zisserman}{Simonyan and
  Zisserman}{2014}]%
        {simonyan2014two}
\bibfield{author}{\bibinfo{person}{Karen Simonyan} {and}
  \bibinfo{person}{Andrew Zisserman}.} \bibinfo{year}{2014}\natexlab{}.
\newblock \showarticletitle{{Two-Stream Convolutional Networks for Action
  Recognition in Videos}}. In \bibinfo{booktitle}{\emph{Conference on Neural
  Information Processing Systems}}.
\newblock


\bibitem[\protect\citeauthoryear{Simonyan and Zisserman}{Simonyan and
  Zisserman}{2015}]%
        {simonyan2015very}
\bibfield{author}{\bibinfo{person}{Karen Simonyan} {and}
  \bibinfo{person}{Andrew Zisserman}.} \bibinfo{year}{2015}\natexlab{}.
\newblock \showarticletitle{{Very Deep Convolutional Networks for Large-Scale
  Image Recognition}}. In \bibinfo{booktitle}{\emph{International Conference on
  Learning Representations}}.
\newblock


\bibitem[\protect\citeauthoryear{So, Liang, and Le}{So et~al\mbox{.}}{2019}]%
        {so2019evolved}
\bibfield{author}{\bibinfo{person}{David~R So}, \bibinfo{person}{Chen Liang},
  {and} \bibinfo{person}{Quoc~V Le}.} \bibinfo{year}{2019}\natexlab{}.
\newblock \showarticletitle{{The Evolved Transformer}}. In
  \bibinfo{booktitle}{\emph{International Conference on Machine Learning}}.
\newblock


\bibitem[\protect\citeauthoryear{Srinivas and Babu}{Srinivas and Babu}{2015}]%
        {srinivas2015data}
\bibfield{author}{\bibinfo{person}{Suraj Srinivas} {and}
  \bibinfo{person}{R~Venkatesh Babu}.} \bibinfo{year}{2015}\natexlab{}.
\newblock \showarticletitle{{Data-Free Parameter Pruning for Deep Neural
  Networks}}. In \bibinfo{booktitle}{\emph{British Machine Vision Conference}}.
\newblock


\bibitem[\protect\citeauthoryear{Sriram, Cox, Tsoi, and Luk}{Sriram
  et~al\mbox{.}}{2010}]%
        {sriram2010towards}
\bibfield{author}{\bibinfo{person}{Vinay Sriram}, \bibinfo{person}{David Cox},
  \bibinfo{person}{Kuen~Hung Tsoi}, {and} \bibinfo{person}{Wayne Luk}.}
  \bibinfo{year}{2010}\natexlab{}.
\newblock \showarticletitle{{Towards an Embedded Biologically-Inspired Machine
  Vision Processor}}. In \bibinfo{booktitle}{\emph{International Conference on
  Field-Programmable Technology}}.
\newblock


\bibitem[\protect\citeauthoryear{Stroud, Ross, Sun, Deng, and
  Sukthankar}{Stroud et~al\mbox{.}}{2020}]%
        {stroud2020d3d}
\bibfield{author}{\bibinfo{person}{Jonathan Stroud}, \bibinfo{person}{David
  Ross}, \bibinfo{person}{Chen Sun}, \bibinfo{person}{Jia Deng}, {and}
  \bibinfo{person}{Rahul Sukthankar}.} \bibinfo{year}{2020}\natexlab{}.
\newblock \showarticletitle{{D3D: Distilled 3D Networks for Video Action
  Recognition}}. In \bibinfo{booktitle}{\emph{IEEE/CVF Winter Conference on
  Applications of Computer Vision}}.
\newblock


\bibitem[\protect\citeauthoryear{Strubell, Ganesh, and McCallum}{Strubell
  et~al\mbox{.}}{2019}]%
        {strubell2019energy}
\bibfield{author}{\bibinfo{person}{Emma Strubell}, \bibinfo{person}{Ananya
  Ganesh}, {and} \bibinfo{person}{Andrew McCallum}.}
  \bibinfo{year}{2019}\natexlab{}.
\newblock \showarticletitle{{Energy and Policy Considerations for Deep Learning
  in NLP}}. In \bibinfo{booktitle}{\emph{Conference of the Association for
  Computational Linguistics}}.
\newblock


\bibitem[\protect\citeauthoryear{Sudhakaran, Escalera, and Lanz}{Sudhakaran
  et~al\mbox{.}}{2020}]%
        {sudhakaran2020gate}
\bibfield{author}{\bibinfo{person}{Swathikiran Sudhakaran},
  \bibinfo{person}{Sergio Escalera}, {and} \bibinfo{person}{Oswald Lanz}.}
  \bibinfo{year}{2020}\natexlab{}.
\newblock \showarticletitle{{Gate-Shift Networks for Video Action
  Recognition}}. In \bibinfo{booktitle}{\emph{IEEE Conference on Computer
  Vision and Pattern Recognition}}.
\newblock


\bibitem[\protect\citeauthoryear{Sun, Feng, Han, Yan, and Wen}{Sun
  et~al\mbox{.}}{2019}]%
        {sun2019optimizing}
\bibfield{author}{\bibinfo{person}{Peng Sun}, \bibinfo{person}{Wansen Feng},
  \bibinfo{person}{Ruobing Han}, \bibinfo{person}{Shengen Yan}, {and}
  \bibinfo{person}{Yonggang Wen}.} \bibinfo{year}{2019}\natexlab{}.
\newblock \showarticletitle{{Optimizing Network Performance for Distributed DNN
  Training on GPU Clusters: ImageNet/AlexNet Training in 1.5 Minutes}}.
\newblock \bibinfo{journal}{\emph{arXiv preprint arXiv:1902.06855}}
  (\bibinfo{year}{2019}).
\newblock


\bibitem[\protect\citeauthoryear{Sutskever, Vinyals, and Le}{Sutskever
  et~al\mbox{.}}{2014}]%
        {sutskever2014sequence}
\bibfield{author}{\bibinfo{person}{Ilya Sutskever}, \bibinfo{person}{Oriol
  Vinyals}, {and} \bibinfo{person}{Quoc~V Le}.}
  \bibinfo{year}{2014}\natexlab{}.
\newblock \showarticletitle{{Sequence to Sequence Learning with Neural
  Networks}}. In \bibinfo{booktitle}{\emph{Conference on Neural Information
  Processing Systems}}.
\newblock


\bibitem[\protect\citeauthoryear{Sze, Chen, Yang, and Emer}{Sze
  et~al\mbox{.}}{2017}]%
        {sze2017efficient}
\bibfield{author}{\bibinfo{person}{Vivienne Sze}, \bibinfo{person}{Yu-Hsin
  Chen}, \bibinfo{person}{Tien-Ju Yang}, {and} \bibinfo{person}{Joel~S Emer}.}
  \bibinfo{year}{2017}\natexlab{}.
\newblock \showarticletitle{{Efficient Processing of Deep Neural Networks: A
  Tutorial and Survey}}.
\newblock \bibinfo{journal}{\emph{Proceedings of the IEEE}}
  \bibinfo{volume}{105}, \bibinfo{number}{12} (\bibinfo{year}{2017}),
  \bibinfo{pages}{2295--2329}.
\newblock


\bibitem[\protect\citeauthoryear{Szegedy, Liu, Jia, Sermanet, Reed, Anguelov,
  Erhan, Vanhoucke, and Rabinovich}{Szegedy et~al\mbox{.}}{2015}]%
        {szegedy2015going}
\bibfield{author}{\bibinfo{person}{Christian Szegedy}, \bibinfo{person}{Wei
  Liu}, \bibinfo{person}{Yangqing Jia}, \bibinfo{person}{Pierre Sermanet},
  \bibinfo{person}{Scott Reed}, \bibinfo{person}{Dragomir Anguelov},
  \bibinfo{person}{Dumitru Erhan}, \bibinfo{person}{Vincent Vanhoucke}, {and}
  \bibinfo{person}{Andrew Rabinovich}.} \bibinfo{year}{2015}\natexlab{}.
\newblock \showarticletitle{{Going Deeper with Convolutions}}. In
  \bibinfo{booktitle}{\emph{IEEE Conference on Computer Vision and Pattern
  Recognition}}.
\newblock


\bibitem[\protect\citeauthoryear{Szegedy, Vanhoucke, Ioffe, Shlens, and
  Wojna}{Szegedy et~al\mbox{.}}{2016}]%
        {szegedy2016rethinking}
\bibfield{author}{\bibinfo{person}{Christian Szegedy}, \bibinfo{person}{Vincent
  Vanhoucke}, \bibinfo{person}{Sergey Ioffe}, \bibinfo{person}{Jon Shlens},
  {and} \bibinfo{person}{Zbigniew Wojna}.} \bibinfo{year}{2016}\natexlab{}.
\newblock \showarticletitle{{Rethinking the Inception Architecture for Computer
  Vision}}. In \bibinfo{booktitle}{\emph{IEEE Conference on Computer Vision and
  Pattern Recognition}}.
\newblock


\bibitem[\protect\citeauthoryear{Tambe, Hooper, Pentecost, Yang, Donato, Sanh,
  Rush, Brooks, and Wei}{Tambe et~al\mbox{.}}{2020a}]%
        {tambe2020edgebert}
\bibfield{author}{\bibinfo{person}{Thierry Tambe}, \bibinfo{person}{Coleman
  Hooper}, \bibinfo{person}{Lillian Pentecost}, \bibinfo{person}{En-Yu Yang},
  \bibinfo{person}{Marco Donato}, \bibinfo{person}{Victor Sanh},
  \bibinfo{person}{Alexander~M Rush}, \bibinfo{person}{David Brooks}, {and}
  \bibinfo{person}{Gu-Yeon Wei}.} \bibinfo{year}{2020}\natexlab{a}.
\newblock \showarticletitle{{EdgeBERT: Optimizing On-Chip Inference for
  Multi-Task NLP}}.
\newblock \bibinfo{journal}{\emph{arXiv preprint arXiv:2011.14203}}
  (\bibinfo{year}{2020}).
\newblock


\bibitem[\protect\citeauthoryear{Tambe, Yang, Wan, Deng, Reddi, Rush, Brooks,
  and Wei}{Tambe et~al\mbox{.}}{2020b}]%
        {tambe2020algorithm}
\bibfield{author}{\bibinfo{person}{Thierry Tambe}, \bibinfo{person}{En-Yu
  Yang}, \bibinfo{person}{Zishen Wan}, \bibinfo{person}{Yuntian Deng},
  \bibinfo{person}{Vijay~Janapa Reddi}, \bibinfo{person}{Alexander Rush},
  \bibinfo{person}{David Brooks}, {and} \bibinfo{person}{Gu-Yeon Wei}.}
  \bibinfo{year}{2020}\natexlab{b}.
\newblock \showarticletitle{{Algorithm-Hardware Co-Design of Adaptive
  Floating-Point Encodings for Resilient Deep Learning Inference}}. In
  \bibinfo{booktitle}{\emph{Design Automation Conference}}.
\newblock


\bibitem[\protect\citeauthoryear{Tan, Chen, Pang, Vasudevan, Sandler, Howard,
  and Le}{Tan et~al\mbox{.}}{2019}]%
        {tan2019mnasnet}
\bibfield{author}{\bibinfo{person}{Mingxing Tan}, \bibinfo{person}{Bo Chen},
  \bibinfo{person}{Ruoming Pang}, \bibinfo{person}{Vijay Vasudevan},
  \bibinfo{person}{Mark Sandler}, \bibinfo{person}{Andrew Howard}, {and}
  \bibinfo{person}{Quoc~V Le}.} \bibinfo{year}{2019}\natexlab{}.
\newblock \showarticletitle{{MnasNet: Platform-Aware Neural Architecture Search
  for Mobile}}. In \bibinfo{booktitle}{\emph{IEEE Conference on Computer Vision
  and Pattern Recognition}}.
\newblock


\bibitem[\protect\citeauthoryear{Tan, Pang, and Le}{Tan et~al\mbox{.}}{2020a}]%
        {tan2020efficientdet}
\bibfield{author}{\bibinfo{person}{Mingxing Tan}, \bibinfo{person}{Ruoming
  Pang}, {and} \bibinfo{person}{Quoc~V Le}.} \bibinfo{year}{2020}\natexlab{a}.
\newblock \showarticletitle{{EfficientDet: Scalable and Efficient Object
  Detection}}. In \bibinfo{booktitle}{\emph{IEEE Conference on Computer Vision
  and Pattern Recognition}}.
\newblock


\bibitem[\protect\citeauthoryear{Tan, Song, Ma, Tan, Chen, Miao, Wu, Ye, Wang,
  Li, and Ma}{Tan et~al\mbox{.}}{2020b}]%
        {tan2020pcnn}
\bibfield{author}{\bibinfo{person}{Zhanhong Tan}, \bibinfo{person}{Jiebo Song},
  \bibinfo{person}{Xiaolong Ma}, \bibinfo{person}{Sia-Huat Tan},
  \bibinfo{person}{Hongyang Chen}, \bibinfo{person}{Yuanqing Miao},
  \bibinfo{person}{Yifu Wu}, \bibinfo{person}{Shaokai Ye},
  \bibinfo{person}{Yanzhi Wang}, \bibinfo{person}{Dehui Li}, {and}
  \bibinfo{person}{Kaisheng Ma}.} \bibinfo{year}{2020}\natexlab{b}.
\newblock \showarticletitle{{PCNN: Pattern-Based Fine-Grained Regular Pruning
  Towards Optimizing CNN Accelerators}}.
\newblock \bibinfo{journal}{\emph{arXiv preprint arXiv:2002.04997}}
  (\bibinfo{year}{2020}).
\newblock


\bibitem[\protect\citeauthoryear{Tang, Liu, Li, Lin, and Han}{Tang
  et~al\mbox{.}}{2022}]%
        {tang2022torchsparse}
\bibfield{author}{\bibinfo{person}{Haotian Tang}, \bibinfo{person}{Zhijian
  Liu}, \bibinfo{person}{Xiuyu Li}, \bibinfo{person}{Yujun Lin}, {and}
  \bibinfo{person}{Song Han}.} \bibinfo{year}{2022}\natexlab{}.
\newblock \showarticletitle{{TorchSparse: Efficient Point Cloud Inference
  Engine}}. In \bibinfo{booktitle}{\emph{Proceedings of Machine Learning and
  Systems 2022, MLSys 2022, Santa Clara, CA, USA, August 29-September 1,
  2022}}.
\newblock


\bibitem[\protect\citeauthoryear{Tang, Liu, Zhao, Lin, Lin, Wang, and Han}{Tang
  et~al\mbox{.}}{2020}]%
        {tang2020searching}
\bibfield{author}{\bibinfo{person}{Haotian Tang}, \bibinfo{person}{Zhijian
  Liu}, \bibinfo{person}{Shengyu Zhao}, \bibinfo{person}{Yujun Lin},
  \bibinfo{person}{Ji Lin}, \bibinfo{person}{Hanrui Wang}, {and}
  \bibinfo{person}{Song Han}.} \bibinfo{year}{2020}\natexlab{}.
\newblock \showarticletitle{{Searching Efficient 3D Architectures with Sparse
  Point-Voxel Convolution}}. In \bibinfo{booktitle}{\emph{European Conference
  on Computer Vision}}.
\newblock


\bibitem[\protect\citeauthoryear{Tatarchenko, Park, Koltun, and
  Zhou}{Tatarchenko et~al\mbox{.}}{2018}]%
        {tatarchenko2018tangent}
\bibfield{author}{\bibinfo{person}{Maxim Tatarchenko}, \bibinfo{person}{Jaesik
  Park}, \bibinfo{person}{Vladlen Koltun}, {and} \bibinfo{person}{Qian-Yi
  Zhou}.} \bibinfo{year}{2018}\natexlab{}.
\newblock \showarticletitle{{Tangent Convolutions for Dense Prediction in 3D}}.
  In \bibinfo{booktitle}{\emph{IEEE Conference on Computer Vision and Pattern
  Recognition}}.
\newblock


\bibitem[\protect\citeauthoryear{Tay, Dehghani, Bahri, and Metzler}{Tay
  et~al\mbox{.}}{2020}]%
        {tay2020efficient}
\bibfield{author}{\bibinfo{person}{Yi Tay}, \bibinfo{person}{Mostafa Dehghani},
  \bibinfo{person}{Dara Bahri}, {and} \bibinfo{person}{Donald Metzler}.}
  \bibinfo{year}{2020}\natexlab{}.
\newblock \showarticletitle{{Efficient Transformers: A Survey}}.
\newblock \bibinfo{journal}{\emph{arXiv preprint arXiv:2009.06732}}
  (\bibinfo{year}{2020}).
\newblock


\bibitem[\protect\citeauthoryear{Tchapmi, Choy, Armeni, Gwak, and
  Savarese}{Tchapmi et~al\mbox{.}}{2017}]%
        {tchapmi2017segcloud}
\bibfield{author}{\bibinfo{person}{Lyne~P Tchapmi},
  \bibinfo{person}{Christopher~B Choy}, \bibinfo{person}{Iro Armeni},
  \bibinfo{person}{JunYoung Gwak}, {and} \bibinfo{person}{Silvio Savarese}.}
  \bibinfo{year}{2017}\natexlab{}.
\newblock \showarticletitle{{SEGCloud: Semantic Segmentation of 3D Point
  Clouds}}. In \bibinfo{booktitle}{\emph{International Conference on 3D
  Vision}}.
\newblock


\bibitem[\protect\citeauthoryear{Thomas, Qi, Deschaud, Marcotegui, Goulette,
  and Guibas}{Thomas et~al\mbox{.}}{2019}]%
        {thomas2019kpconv}
\bibfield{author}{\bibinfo{person}{Hugues Thomas}, \bibinfo{person}{Charles~R
  Qi}, \bibinfo{person}{Jean-Emmanuel Deschaud}, \bibinfo{person}{Beatriz
  Marcotegui}, \bibinfo{person}{Fran\c{c}ois Goulette}, {and}
  \bibinfo{person}{Leonidas~J Guibas}.} \bibinfo{year}{2019}\natexlab{}.
\newblock \showarticletitle{{KPConv: Flexible and Deformable Convolution for
  Point Clouds}}. In \bibinfo{booktitle}{\emph{International Conference on
  Computer Vision}}.
\newblock


\bibitem[\protect\citeauthoryear{Tran, Bourdev, Fergus, Torresani, and
  Paluri}{Tran et~al\mbox{.}}{2015}]%
        {tran2015learning}
\bibfield{author}{\bibinfo{person}{Du Tran}, \bibinfo{person}{Lubomir Bourdev},
  \bibinfo{person}{Rob Fergus}, \bibinfo{person}{Lorenzo Torresani}, {and}
  \bibinfo{person}{Manohar Paluri}.} \bibinfo{year}{2015}\natexlab{}.
\newblock \showarticletitle{{Learning Spatiotemporal Features with 3D
  Convolutional Networks}}. In \bibinfo{booktitle}{\emph{International
  Conference on Computer Vision}}.
\newblock


\bibitem[\protect\citeauthoryear{Tran, Wang, Torresani, and Feiszli}{Tran
  et~al\mbox{.}}{2019}]%
        {tran2019video}
\bibfield{author}{\bibinfo{person}{Du Tran}, \bibinfo{person}{Heng Wang},
  \bibinfo{person}{Lorenzo Torresani}, {and} \bibinfo{person}{Matt Feiszli}.}
  \bibinfo{year}{2019}\natexlab{}.
\newblock \showarticletitle{{Video Classification with Channel-Separated
  Convolutional Networks}}. In \bibinfo{booktitle}{\emph{International
  Conference on Computer Vision}}.
\newblock


\bibitem[\protect\citeauthoryear{Tran, Wang, Torresani, Ray, LeCun, and
  Paluri}{Tran et~al\mbox{.}}{2018}]%
        {tran2018closer}
\bibfield{author}{\bibinfo{person}{Du Tran}, \bibinfo{person}{Heng Wang},
  \bibinfo{person}{Lorenzo Torresani}, \bibinfo{person}{Jamie Ray},
  \bibinfo{person}{Yann LeCun}, {and} \bibinfo{person}{Manohar Paluri}.}
  \bibinfo{year}{2018}\natexlab{}.
\newblock \showarticletitle{{A Closer Look at Spatiotemporal Convolutions for
  Action Recognition}}. In \bibinfo{booktitle}{\emph{IEEE Conference on
  Computer Vision and Pattern Recognition}}.
\newblock


\bibitem[\protect\citeauthoryear{Umuroglu, Fraser, Gambardella, Blott, Leong,
  Jahre, and Vissers}{Umuroglu et~al\mbox{.}}{2017}]%
        {umuroglu2017finn}
\bibfield{author}{\bibinfo{person}{Yaman Umuroglu}, \bibinfo{person}{Nicholas~J
  Fraser}, \bibinfo{person}{Giulio Gambardella}, \bibinfo{person}{Michaela
  Blott}, \bibinfo{person}{Philip Leong}, \bibinfo{person}{Magnus Jahre}, {and}
  \bibinfo{person}{Kees Vissers}.} \bibinfo{year}{2017}\natexlab{}.
\newblock \showarticletitle{{FINN: A Framework for Fast, Scalable Binarized
  Neural Network Inference}}. In \bibinfo{booktitle}{\emph{International
  Symposium on Field-Programmable Gate Arrays}}.
\newblock


\bibitem[\protect\citeauthoryear{Umuroglu, Rasnayake, and
  Sj{\"a}lander}{Umuroglu et~al\mbox{.}}{2018}]%
        {umuroglu2018bismo}
\bibfield{author}{\bibinfo{person}{Yaman Umuroglu}, \bibinfo{person}{Lahiru
  Rasnayake}, {and} \bibinfo{person}{Magnus Sj{\"a}lander}.}
  \bibinfo{year}{2018}\natexlab{}.
\newblock \showarticletitle{{BISMO: A Scalable Bit-Serial Matrix Multiplication
  Overlay for Reconfigurable Computing}}. In
  \bibinfo{booktitle}{\emph{International Conference on Field-Programmable
  Logic and Applications}}.
\newblock


\bibitem[\protect\citeauthoryear{Vaswani, Shazeer, Parmar, Uszkoreit, Jones,
  Gomez, Kaiser, and Polosukhin}{Vaswani et~al\mbox{.}}{2017}]%
        {vaswani2017attention}
\bibfield{author}{\bibinfo{person}{Ashish Vaswani}, \bibinfo{person}{Noam
  Shazeer}, \bibinfo{person}{Niki Parmar}, \bibinfo{person}{Jakob Uszkoreit},
  \bibinfo{person}{Llion Jones}, \bibinfo{person}{Aidan~N Gomez},
  \bibinfo{person}{Lukasz Kaiser}, {and} \bibinfo{person}{Illia Polosukhin}.}
  \bibinfo{year}{2017}\natexlab{}.
\newblock \showarticletitle{{Attention Is All You Need}}. In
  \bibinfo{booktitle}{\emph{Conference on Neural Information Processing
  Systems}}.
\newblock


\bibitem[\protect\citeauthoryear{Voita, Talbot, Moiseev, Sennrich, and
  Titov}{Voita et~al\mbox{.}}{2019}]%
        {voita2019analyzing}
\bibfield{author}{\bibinfo{person}{Elena Voita}, \bibinfo{person}{David
  Talbot}, \bibinfo{person}{Fedor Moiseev}, \bibinfo{person}{Rico Sennrich},
  {and} \bibinfo{person}{Ivan Titov}.} \bibinfo{year}{2019}\natexlab{}.
\newblock \showarticletitle{{Analyzing Multi-Head Self-Attention: Specialized
  Heads Do the Heavy Lifting, the Rest Can Be Pruned}}. In
  \bibinfo{booktitle}{\emph{Conference of the Association for Computational
  Linguistics}}.
\newblock


\bibitem[\protect\citeauthoryear{Wang, Singh, Michael, Hill, Levy, and
  Bowman}{Wang et~al\mbox{.}}{2018a}]%
        {wang2018glue}
\bibfield{author}{\bibinfo{person}{Alex Wang}, \bibinfo{person}{Amanpreet
  Singh}, \bibinfo{person}{Julian Michael}, \bibinfo{person}{Felix Hill},
  \bibinfo{person}{Omer Levy}, {and} \bibinfo{person}{Samuel~R Bowman}.}
  \bibinfo{year}{2018}\natexlab{a}.
\newblock \showarticletitle{{GLUE: A Multi-Task Benchmark and Analysis Platform
  for Natural Language Understanding}}. In
  \bibinfo{booktitle}{\emph{International Conference on Learning
  Representations}}.
\newblock


\bibitem[\protect\citeauthoryear{Wang}{Wang}{2020}]%
        {wang2020efficient}
\bibfield{author}{\bibinfo{person}{Hanrui Wang}.}
  \bibinfo{year}{2020}\natexlab{}.
\newblock \emph{\bibinfo{title}{{Efficient Algorithms and Hardware for Natural
  Language Processing}}}.
\newblock \bibinfo{thesistype}{Master's\ Thesis}.
  \bibinfo{school}{Massachusetts Institute of Technology}.
\newblock


\bibitem[\protect\citeauthoryear{Wang, Wang, Yang, Shen, Sun, Lee, and
  Han}{Wang et~al\mbox{.}}{2020e}]%
        {wang2020gcn}
\bibfield{author}{\bibinfo{person}{Hanrui Wang}, \bibinfo{person}{Kuan Wang},
  \bibinfo{person}{Jiacheng Yang}, \bibinfo{person}{Linxiao Shen},
  \bibinfo{person}{Nan Sun}, \bibinfo{person}{Hae-Seung Lee}, {and}
  \bibinfo{person}{Song Han}.} \bibinfo{year}{2020}\natexlab{e}.
\newblock \showarticletitle{{GCN-RL Circuit Designer: Transferable Transistor
  Sizing with Graph Neural Networks and Reinforcement Learning}}. In
  \bibinfo{booktitle}{\emph{Design Automation Conference}}.
\newblock


\bibitem[\protect\citeauthoryear{Wang, Wu, Liu, Cai, Zhu, Gan, and Han}{Wang
  et~al\mbox{.}}{2020f}]%
        {wang2020hat}
\bibfield{author}{\bibinfo{person}{Hanrui Wang}, \bibinfo{person}{Zhanghao Wu},
  \bibinfo{person}{Zhijian Liu}, \bibinfo{person}{Han Cai},
  \bibinfo{person}{Ligeng Zhu}, \bibinfo{person}{Chuang Gan}, {and}
  \bibinfo{person}{Song Han}.} \bibinfo{year}{2020}\natexlab{f}.
\newblock \showarticletitle{{HAT: Hardware-Aware Transformers for Efficient
  Natural Language Processing}}. In \bibinfo{booktitle}{\emph{Conference of the
  Association for Computational Linguistics}}.
\newblock


\bibitem[\protect\citeauthoryear{Wang, Yang, Lee, and Han}{Wang
  et~al\mbox{.}}{2018c}]%
        {wang2018learning}
\bibfield{author}{\bibinfo{person}{Hanrui Wang}, \bibinfo{person}{Jiacheng
  Yang}, \bibinfo{person}{Hae-Seung Lee}, {and} \bibinfo{person}{Song Han}.}
  \bibinfo{year}{2018}\natexlab{c}.
\newblock \showarticletitle{{Learning to Design Circuits}}. In
  \bibinfo{booktitle}{\emph{Workshop on ML for Systems at NeurIPS}}.
\newblock


\bibitem[\protect\citeauthoryear{Wang, Zhang, and Han}{Wang
  et~al\mbox{.}}{2021}]%
        {wang2020spatten}
\bibfield{author}{\bibinfo{person}{Hanrui Wang}, \bibinfo{person}{Zhekai
  Zhang}, {and} \bibinfo{person}{Song Han}.} \bibinfo{year}{2021}\natexlab{}.
\newblock \showarticletitle{{SpAtten: Efficient Sparse Attention Architecture
  with Cascade Token and Head Pruning}}. In \bibinfo{booktitle}{\emph{IEEE
  International Symposium on High-Performance Computer Architecture}}.
\newblock


\bibitem[\protect\citeauthoryear{Wang, Liu, Lin, Lin, and Han}{Wang
  et~al\mbox{.}}{2019b}]%
        {wang2019haq}
\bibfield{author}{\bibinfo{person}{Kuan Wang}, \bibinfo{person}{Zhijian Liu},
  \bibinfo{person}{Yujun Lin}, \bibinfo{person}{Ji Lin}, {and}
  \bibinfo{person}{Song Han}.} \bibinfo{year}{2019}\natexlab{b}.
\newblock \showarticletitle{{HAQ: Hardware-Aware Automated Quantization with
  Mixed Precision}}. In \bibinfo{booktitle}{\emph{IEEE Conference on Computer
  Vision and Pattern Recognition}}.
\newblock


\bibitem[\protect\citeauthoryear{Wang, Liu, Lin, Lin, and Han}{Wang
  et~al\mbox{.}}{2020c}]%
        {wang2020hardware}
\bibfield{author}{\bibinfo{person}{Kuan Wang}, \bibinfo{person}{Zhijian Liu},
  \bibinfo{person}{Yujun Lin}, \bibinfo{person}{Ji Lin}, {and}
  \bibinfo{person}{Song Han}.} \bibinfo{year}{2020}\natexlab{c}.
\newblock \showarticletitle{{Hardware-Centric AutoML for Mixed-Precision
  Quantization}}.
\newblock \bibinfo{journal}{\emph{International Journal of Computer Vision}}
  \bibinfo{volume}{128}, \bibinfo{number}{8} (\bibinfo{year}{2020}),
  \bibinfo{pages}{2035--2048}.
\newblock


\bibitem[\protect\citeauthoryear{Wang, Xiong, Wang, Qiao, Lin, Tang, and
  Van~Gool}{Wang et~al\mbox{.}}{2016}]%
        {wang2016temporal}
\bibfield{author}{\bibinfo{person}{Limin Wang}, \bibinfo{person}{Yuanjun
  Xiong}, \bibinfo{person}{Zhe Wang}, \bibinfo{person}{Yu Qiao},
  \bibinfo{person}{Dahua Lin}, \bibinfo{person}{Xiaoou Tang}, {and}
  \bibinfo{person}{Luc Van~Gool}.} \bibinfo{year}{2016}\natexlab{}.
\newblock \showarticletitle{{Temporal Segment Networks: Towards Good Practices
  for Deep Action Recognition}}. In \bibinfo{booktitle}{\emph{European
  Conference on Computer Vision}}.
\newblock


\bibitem[\protect\citeauthoryear{Wang, Liu, Guo, Sun, and Tong}{Wang
  et~al\mbox{.}}{2017}]%
        {wang2017ocnn}
\bibfield{author}{\bibinfo{person}{Peng-Shuai Wang}, \bibinfo{person}{Yang
  Liu}, \bibinfo{person}{Yu-Xiao Guo}, \bibinfo{person}{Chun-Yu Sun}, {and}
  \bibinfo{person}{Xin Tong}.} \bibinfo{year}{2017}\natexlab{}.
\newblock \showarticletitle{{O-CNN: Octree-Based Convolutional Neural Networks
  for 3D Shape Analysis}}. In \bibinfo{booktitle}{\emph{SIGGRAPH}}.
\newblock


\bibitem[\protect\citeauthoryear{Wang, Sun, Liu, and Tong}{Wang
  et~al\mbox{.}}{2018b}]%
        {wang2018adaptive}
\bibfield{author}{\bibinfo{person}{Peng-Shuai Wang}, \bibinfo{person}{Chun-Yu
  Sun}, \bibinfo{person}{Yang Liu}, {and} \bibinfo{person}{Xin Tong}.}
  \bibinfo{year}{2018}\natexlab{b}.
\newblock \showarticletitle{{Adaptive O-CNN: A Patch-Based Deep Representation
  of 3D Shapes}}. In \bibinfo{booktitle}{\emph{SIGGRAPH Asia}}.
\newblock


\bibitem[\protect\citeauthoryear{Wang, Li, Xiao, Zhu, Li, Wong, and Chao}{Wang
  et~al\mbox{.}}{2019a}]%
        {wang-etal-2019-learning-deep}
\bibfield{author}{\bibinfo{person}{Qiang Wang}, \bibinfo{person}{Bei Li},
  \bibinfo{person}{Tong Xiao}, \bibinfo{person}{Jingbo Zhu},
  \bibinfo{person}{Changliang Li}, \bibinfo{person}{Derek~F. Wong}, {and}
  \bibinfo{person}{Lidia~S. Chao}.} \bibinfo{year}{2019}\natexlab{a}.
\newblock \showarticletitle{{Learning Deep Transformer Models for Machine
  Translation}}. In \bibinfo{booktitle}{\emph{Conference of the Association for
  Computational Linguistics}}.
\newblock


\bibitem[\protect\citeauthoryear{Wang, Li, Khabsa, Fang, and Ma}{Wang
  et~al\mbox{.}}{2020a}]%
        {wang2020linformer}
\bibfield{author}{\bibinfo{person}{Sinong Wang}, \bibinfo{person}{Belinda Li},
  \bibinfo{person}{Madian Khabsa}, \bibinfo{person}{Han Fang}, {and}
  \bibinfo{person}{Hao Ma}.} \bibinfo{year}{2020}\natexlab{a}.
\newblock \showarticletitle{{Linformer: Self-Attention with Linear
  Complexity}}.
\newblock \bibinfo{journal}{\emph{arXiv preprint arXiv:2006.04768}}
  (\bibinfo{year}{2020}).
\newblock


\bibitem[\protect\citeauthoryear{Wang, Wang, Cai, Lin, Liu, Wang, Lin, and
  Han}{Wang et~al\mbox{.}}{2020d}]%
        {wang2020apq}
\bibfield{author}{\bibinfo{person}{Tianzhe Wang}, \bibinfo{person}{Kuan Wang},
  \bibinfo{person}{Han Cai}, \bibinfo{person}{Ji Lin}, \bibinfo{person}{Zhijian
  Liu}, \bibinfo{person}{Hanrui Wang}, \bibinfo{person}{Yujun Lin}, {and}
  \bibinfo{person}{Song Han}.} \bibinfo{year}{2020}\natexlab{d}.
\newblock \showarticletitle{{APQ: Joint Search for Network Architecture,
  Pruning and Quantization Policy}}. In \bibinfo{booktitle}{\emph{IEEE
  Conference on Computer Vision and Pattern Recognition}}.
\newblock


\bibitem[\protect\citeauthoryear{Wang, Yu, Huang, and Neumann}{Wang
  et~al\mbox{.}}{2018e}]%
        {wang2018sgpn}
\bibfield{author}{\bibinfo{person}{Weiyue Wang}, \bibinfo{person}{Ronald Yu},
  \bibinfo{person}{Qiangui Huang}, {and} \bibinfo{person}{Ulrich Neumann}.}
  \bibinfo{year}{2018}\natexlab{e}.
\newblock \showarticletitle{{SGPN: Similarity Group Proposal Network for 3D
  Point Cloud Instance Segmentation}}. In \bibinfo{booktitle}{\emph{IEEE
  Conference on Computer Vision and Pattern Recognition}}.
\newblock


\bibitem[\protect\citeauthoryear{Wang, Yu, Dou, Darrell, and Gonzalez}{Wang
  et~al\mbox{.}}{2018d}]%
        {wang2018skipnet}
\bibfield{author}{\bibinfo{person}{Xin Wang}, \bibinfo{person}{Fisher Yu},
  \bibinfo{person}{Zi-Yi Dou}, \bibinfo{person}{Trevor Darrell}, {and}
  \bibinfo{person}{Joseph~E Gonzalez}.} \bibinfo{year}{2018}\natexlab{d}.
\newblock \showarticletitle{{SkipNet: Learning Dynamic Routing in Convolutional
  Networks}}. In \bibinfo{booktitle}{\emph{European Conference on Computer
  Vision}}.
\newblock


\bibitem[\protect\citeauthoryear{Wang, Sun, Liu, Sarma, Bronstein, and
  Solomon}{Wang et~al\mbox{.}}{2019c}]%
        {wang2018dynamic}
\bibfield{author}{\bibinfo{person}{Yue Wang}, \bibinfo{person}{Yongbin Sun},
  \bibinfo{person}{Ziwei Liu}, \bibinfo{person}{Sanjay~E Sarma},
  \bibinfo{person}{Michael~M Bronstein}, {and} \bibinfo{person}{Justin~M
  Solomon}.} \bibinfo{year}{2019}\natexlab{c}.
\newblock \showarticletitle{{Dynamic Graph CNN for Learning on Point Clouds}}.
  In \bibinfo{booktitle}{\emph{SIGGRAPH}}.
\newblock


\bibitem[\protect\citeauthoryear{Wang}{Wang}{2021}]%
        {wang2021sparsednn}
\bibfield{author}{\bibinfo{person}{Ziheng Wang}.}
  \bibinfo{year}{2021}\natexlab{}.
\newblock \showarticletitle{{SparseDNN: Fast Sparse Deep Learning Inference on
  CPUs}}.
\newblock \bibinfo{journal}{\emph{arXiv preprint arXiv:2101.07948}}
  (\bibinfo{year}{2021}).
\newblock


\bibitem[\protect\citeauthoryear{Wang, Lin, Sheng, Yan, and Shao}{Wang
  et~al\mbox{.}}{2020b}]%
        {wang2020pv}
\bibfield{author}{\bibinfo{person}{Zihao Wang}, \bibinfo{person}{Chen Lin},
  \bibinfo{person}{Lu Sheng}, \bibinfo{person}{Junjie Yan}, {and}
  \bibinfo{person}{Jing Shao}.} \bibinfo{year}{2020}\natexlab{b}.
\newblock \showarticletitle{{PV-NAS: Practical Neural Architecture Search for
  Video Recognition}}.
\newblock \bibinfo{journal}{\emph{arXiv preprint arXiv:2011.00826}}
  (\bibinfo{year}{2020}).
\newblock


\bibitem[\protect\citeauthoryear{Wang and Lu}{Wang and Lu}{2019}]%
        {wang2019voxsegnet}
\bibfield{author}{\bibinfo{person}{Zongji Wang} {and} \bibinfo{person}{Feng
  Lu}.} \bibinfo{year}{2019}\natexlab{}.
\newblock \showarticletitle{{VoxSegNet: Volumetric CNNs for Semantic Part
  Segmentation of 3D Shapes}}.
\newblock \bibinfo{journal}{\emph{IEEE Transactions on Visualization and
  Computer Graphics}} \bibinfo{volume}{26}, \bibinfo{number}{9}
  (\bibinfo{year}{2019}), \bibinfo{pages}{2919--2930}.
\newblock


\bibitem[\protect\citeauthoryear{Wangni, Wang, Liu, and Zhang}{Wangni
  et~al\mbox{.}}{2018}]%
        {wangni2018gradient}
\bibfield{author}{\bibinfo{person}{Jianqiao Wangni}, \bibinfo{person}{Jialei
  Wang}, \bibinfo{person}{Ji Liu}, {and} \bibinfo{person}{Tong Zhang}.}
  \bibinfo{year}{2018}\natexlab{}.
\newblock \showarticletitle{{Gradient Sparsification for
  Communication-Efficient Distributed Optimization}}. In
  \bibinfo{booktitle}{\emph{Conference on Neural Information Processing
  Systems}}.
\newblock


\bibitem[\protect\citeauthoryear{Wei, Wang, Zhou, Lin, and Sun}{Wei
  et~al\mbox{.}}{2019}]%
        {wei-etal-2019-imitation}
\bibfield{author}{\bibinfo{person}{Bingzhen Wei}, \bibinfo{person}{Mingxuan
  Wang}, \bibinfo{person}{Hao Zhou}, \bibinfo{person}{Junyang Lin}, {and}
  \bibinfo{person}{Xu Sun}.} \bibinfo{year}{2019}\natexlab{}.
\newblock \showarticletitle{{Imitation Learning for Non-Autoregressive Neural
  Machine Translation}}. In \bibinfo{booktitle}{\emph{Conference of the
  Association for Computational Linguistics}}.
\newblock


\bibitem[\protect\citeauthoryear{Wen, Wu, Wang, Chen, and Li}{Wen
  et~al\mbox{.}}{2016}]%
        {wen2016learning}
\bibfield{author}{\bibinfo{person}{Wei Wen}, \bibinfo{person}{Chunpeng Wu},
  \bibinfo{person}{Yandan Wang}, \bibinfo{person}{Yiran Chen}, {and}
  \bibinfo{person}{Hai Li}.} \bibinfo{year}{2016}\natexlab{}.
\newblock \showarticletitle{{Learning Structured Sparsity in Deep Neural
  Networks}}. In \bibinfo{booktitle}{\emph{Conference on Neural Information
  Processing Systems}}.
\newblock


\bibitem[\protect\citeauthoryear{Wen, Xu, Yan, Wu, Wang, Chen, and Li}{Wen
  et~al\mbox{.}}{2017}]%
        {terngrad}
\bibfield{author}{\bibinfo{person}{Wei Wen}, \bibinfo{person}{Cong Xu},
  \bibinfo{person}{Feng Yan}, \bibinfo{person}{Chunpeng Wu},
  \bibinfo{person}{Yandan Wang}, \bibinfo{person}{Yiran Chen}, {and}
  \bibinfo{person}{Hai Li}.} \bibinfo{year}{2017}\natexlab{}.
\newblock \showarticletitle{{TernGrad: Ternary Gradients to Reduce
  Communication in Distributed Deep Learning}}. In
  \bibinfo{booktitle}{\emph{Conference on Neural Information Processing
  Systems}}.
\newblock


\bibitem[\protect\citeauthoryear{Wistuba, Rawat, and Pedapati}{Wistuba
  et~al\mbox{.}}{2019}]%
        {wistuba2019survey}
\bibfield{author}{\bibinfo{person}{Martin Wistuba}, \bibinfo{person}{Ambrish
  Rawat}, {and} \bibinfo{person}{Tejaswini Pedapati}.}
  \bibinfo{year}{2019}\natexlab{}.
\newblock \showarticletitle{{A Survey on Neural Architecture Search}}.
\newblock \bibinfo{journal}{\emph{arXiv preprint arXiv:1905.01392}}
  (\bibinfo{year}{2019}).
\newblock


\bibitem[\protect\citeauthoryear{Wu, Dai, Zhang, Wang, Sun, Wu, Tian, Vajda,
  Jia, and Keutzer}{Wu et~al\mbox{.}}{2019a}]%
        {wu2019fbnet}
\bibfield{author}{\bibinfo{person}{Bichen Wu}, \bibinfo{person}{Xiaoliang Dai},
  \bibinfo{person}{Peizhao Zhang}, \bibinfo{person}{Yanghan Wang},
  \bibinfo{person}{Fei Sun}, \bibinfo{person}{Yiming Wu},
  \bibinfo{person}{Yuandong Tian}, \bibinfo{person}{Peter Vajda},
  \bibinfo{person}{Yangqing Jia}, {and} \bibinfo{person}{Kurt Keutzer}.}
  \bibinfo{year}{2019}\natexlab{a}.
\newblock \showarticletitle{{FBNet: Hardware-Aware Efficient ConvNet Design via
  Differentiable Neural Architecture Search}}. In
  \bibinfo{booktitle}{\emph{IEEE Conference on Computer Vision and Pattern
  Recognition}}.
\newblock


\bibitem[\protect\citeauthoryear{Wu, Fan, Baevski, Dauphin, and Auli}{Wu
  et~al\mbox{.}}{2019b}]%
        {wu2019pay}
\bibfield{author}{\bibinfo{person}{Felix Wu}, \bibinfo{person}{Angela Fan},
  \bibinfo{person}{Alexei Baevski}, \bibinfo{person}{Yann~N Dauphin}, {and}
  \bibinfo{person}{Michael Auli}.} \bibinfo{year}{2019}\natexlab{b}.
\newblock \showarticletitle{{Pay Less Attention with Lightweight and Dynamic
  Convolutions}}. In \bibinfo{booktitle}{\emph{International Conference on
  Learning Representations}}.
\newblock


\bibitem[\protect\citeauthoryear{Wu, Leng, Wang, Hu, and Cheng}{Wu
  et~al\mbox{.}}{2016}]%
        {wu2016quantized}
\bibfield{author}{\bibinfo{person}{Jiaxiang Wu}, \bibinfo{person}{Cong Leng},
  \bibinfo{person}{Yuhang Wang}, \bibinfo{person}{Qinghao Hu}, {and}
  \bibinfo{person}{Jian Cheng}.} \bibinfo{year}{2016}\natexlab{}.
\newblock \showarticletitle{{Quantized Convolutional Neural Networks for Mobile
  Devices}}. In \bibinfo{booktitle}{\emph{IEEE Conference on Computer Vision
  and Pattern Recognition}}.
\newblock


\bibitem[\protect\citeauthoryear{Wu, Qi, and Fuxin}{Wu et~al\mbox{.}}{2019c}]%
        {wu2019pointconv}
\bibfield{author}{\bibinfo{person}{Wenxuan Wu}, \bibinfo{person}{Zhongang Qi},
  {and} \bibinfo{person}{Li Fuxin}.} \bibinfo{year}{2019}\natexlab{c}.
\newblock \showarticletitle{{PointConv: Deep Convolutional Networks on 3D Point
  Clouds}}. In \bibinfo{booktitle}{\emph{IEEE Conference on Computer Vision and
  Pattern Recognition}}.
\newblock


\bibitem[\protect\citeauthoryear{Wu, Liu, Lin, Lin, and Han}{Wu
  et~al\mbox{.}}{2020}]%
        {wu2020lite}
\bibfield{author}{\bibinfo{person}{Zhanghao Wu}, \bibinfo{person}{Zhijian Liu},
  \bibinfo{person}{Ji Lin}, \bibinfo{person}{Yujun Lin}, {and}
  \bibinfo{person}{Song Han}.} \bibinfo{year}{2020}\natexlab{}.
\newblock \showarticletitle{{Lite Transformer with Long-Short Range
  Attention}}. In \bibinfo{booktitle}{\emph{International Conference on
  Learning Representations}}.
\newblock


\bibitem[\protect\citeauthoryear{Wu, Nagarajan, Kumar, Rennie, Davis, Grauman,
  and Feris}{Wu et~al\mbox{.}}{2018}]%
        {wu2018blockdrop}
\bibfield{author}{\bibinfo{person}{Zuxuan Wu}, \bibinfo{person}{Tushar
  Nagarajan}, \bibinfo{person}{Abhishek Kumar}, \bibinfo{person}{Steven
  Rennie}, \bibinfo{person}{Larry~S Davis}, \bibinfo{person}{Kristen Grauman},
  {and} \bibinfo{person}{Rogerio Feris}.} \bibinfo{year}{2018}\natexlab{}.
\newblock \showarticletitle{{BlockDrop: Dynamic Inference Paths in Residual
  Networks}}. In \bibinfo{booktitle}{\emph{IEEE Conference on Computer Vision
  and Pattern Recognition}}.
\newblock


\bibitem[\protect\citeauthoryear{Wu, Song, Khosla, Yu, Zhang, Tang, and
  Xiao}{Wu et~al\mbox{.}}{2015}]%
        {wu20153d}
\bibfield{author}{\bibinfo{person}{Zhirong Wu}, \bibinfo{person}{Shuran Song},
  \bibinfo{person}{Aditya Khosla}, \bibinfo{person}{Fisher Yu},
  \bibinfo{person}{Linguang Zhang}, \bibinfo{person}{Xiaoou Tang}, {and}
  \bibinfo{person}{Jianxiong Xiao}.} \bibinfo{year}{2015}\natexlab{}.
\newblock \showarticletitle{{3D ShapeNets: A Deep Representation for Volumetric
  Shapes}}. In \bibinfo{booktitle}{\emph{IEEE Conference on Computer Vision and
  Pattern Recognition}}.
\newblock


\bibitem[\protect\citeauthoryear{Xie, Gu, Guo, Qi, Guibas, and Litany}{Xie
  et~al\mbox{.}}{2020}]%
        {xie2020pointcontrast}
\bibfield{author}{\bibinfo{person}{Saining Xie}, \bibinfo{person}{Jiatao Gu},
  \bibinfo{person}{Demi Guo}, \bibinfo{person}{Charles~R Qi},
  \bibinfo{person}{Leonidas~J Guibas}, {and} \bibinfo{person}{Or Litany}.}
  \bibinfo{year}{2020}\natexlab{}.
\newblock \showarticletitle{{PointContrast: Unsupervised Pre-Training for 3D
  Point Cloud Understanding}}. In \bibinfo{booktitle}{\emph{European Conference
  on Computer Vision}}.
\newblock


\bibitem[\protect\citeauthoryear{Xie, Sun, Huang, Tu, and Murphy}{Xie
  et~al\mbox{.}}{2018}]%
        {xie2018rethinking}
\bibfield{author}{\bibinfo{person}{Saining Xie}, \bibinfo{person}{Chen Sun},
  \bibinfo{person}{Jonathan Huang}, \bibinfo{person}{Zhuowen Tu}, {and}
  \bibinfo{person}{Kevin Murphy}.} \bibinfo{year}{2018}\natexlab{}.
\newblock \showarticletitle{{Rethinking Spatiotemporal Feature Learning:
  Speed-Accuracy Trade-Offs in Video Classification}}. In
  \bibinfo{booktitle}{\emph{European Conference on Computer Vision}}.
\newblock


\bibitem[\protect\citeauthoryear{Xin, Tang, Lee, Yu, and Lin}{Xin
  et~al\mbox{.}}{2020}]%
        {xin2020deebert}
\bibfield{author}{\bibinfo{person}{Ji Xin}, \bibinfo{person}{Raphael Tang},
  \bibinfo{person}{Jaejun Lee}, \bibinfo{person}{Yaoliang Yu}, {and}
  \bibinfo{person}{Jimmy Lin}.} \bibinfo{year}{2020}\natexlab{}.
\newblock \showarticletitle{{DeeBERT: Dynamic Early Exiting for Accelerating
  BERT Inference}}. In \bibinfo{booktitle}{\emph{Conference of the Association
  for Computational Linguistics}}.
\newblock


\bibitem[\protect\citeauthoryear{Xiong, Wu, Alleva, Droppo, Huang, and
  Stolcke}{Xiong et~al\mbox{.}}{2018}]%
        {xiong2018microsoft}
\bibfield{author}{\bibinfo{person}{Wayne Xiong}, \bibinfo{person}{Lingfeng Wu},
  \bibinfo{person}{Fil Alleva}, \bibinfo{person}{Jasha Droppo},
  \bibinfo{person}{Xuedong Huang}, {and} \bibinfo{person}{Andreas Stolcke}.}
  \bibinfo{year}{2018}\natexlab{}.
\newblock \showarticletitle{{The Microsoft 2017 Conversational Speech
  Recognition System}}. In \bibinfo{booktitle}{\emph{IEEE International
  Conference on Acoustics, Speech and Signal Processing}}.
\newblock


\bibitem[\protect\citeauthoryear{Xu, Liu, Liu, Lin, Liu, and Liu}{Xu
  et~al\mbox{.}}{2019}]%
        {xu2019first}
\bibfield{author}{\bibinfo{person}{Mengwei Xu}, \bibinfo{person}{Jiawei Liu},
  \bibinfo{person}{Yuanqiang Liu}, \bibinfo{person}{Felix~Xiaozhu Lin},
  \bibinfo{person}{Yunxin Liu}, {and} \bibinfo{person}{Xuanzhe Liu}.}
  \bibinfo{year}{2019}\natexlab{}.
\newblock \showarticletitle{{A First Look at Deep Learning Apps on
  Smartphones}}. In \bibinfo{booktitle}{\emph{International World Wide Web
  Conference}}.
\newblock


\bibitem[\protect\citeauthoryear{Xu, Sun, Wu, Wang, and Neumann}{Xu
  et~al\mbox{.}}{2020}]%
        {xu2020grid}
\bibfield{author}{\bibinfo{person}{Qiangeng Xu}, \bibinfo{person}{Xudong Sun},
  \bibinfo{person}{Cho-Ying Wu}, \bibinfo{person}{Panqu Wang}, {and}
  \bibinfo{person}{Ulrich Neumann}.} \bibinfo{year}{2020}\natexlab{}.
\newblock \showarticletitle{{Grid-GCN for Fast and Scalable Point Cloud
  Learning}}. In \bibinfo{booktitle}{\emph{IEEE Conference on Computer Vision
  and Pattern Recognition}}.
\newblock


\bibitem[\protect\citeauthoryear{Xu, Fan, Xu, Zeng, and Qiao}{Xu
  et~al\mbox{.}}{2018}]%
        {xu2018spidercnn}
\bibfield{author}{\bibinfo{person}{Yifan Xu}, \bibinfo{person}{Tianqi Fan},
  \bibinfo{person}{Mingye Xu}, \bibinfo{person}{Long Zeng}, {and}
  \bibinfo{person}{Yu Qiao}.} \bibinfo{year}{2018}\natexlab{}.
\newblock \showarticletitle{{SpiderCNN: Deep Learning on Point Sets with
  Parameterized Convolutional Filters}}. In \bibinfo{booktitle}{\emph{European
  Conference on Computer Vision}}.
\newblock


\bibitem[\protect\citeauthoryear{Xue, Li, and Gong}{Xue et~al\mbox{.}}{2013}]%
        {xue2013restructuring}
\bibfield{author}{\bibinfo{person}{Jian Xue}, \bibinfo{person}{Jinyu Li}, {and}
  \bibinfo{person}{Yifan Gong}.} \bibinfo{year}{2013}\natexlab{}.
\newblock \showarticletitle{{Restructuring of Deep Neural Network Acoustic
  Models with Singular Value Decomposition}}. In
  \bibinfo{booktitle}{\emph{Conference of the International Speech
  Communication Association}}.
\newblock


\bibitem[\protect\citeauthoryear{Yan, Mao, and Li}{Yan et~al\mbox{.}}{2018}]%
        {yan2018second}
\bibfield{author}{\bibinfo{person}{Yan Yan}, \bibinfo{person}{Yuxing Mao},
  {and} \bibinfo{person}{Bo Li}.} \bibinfo{year}{2018}\natexlab{}.
\newblock \showarticletitle{{SECOND: Sparsely Embedded Convolutional
  Detection}}.
\newblock \bibinfo{journal}{\emph{Sensors}} \bibinfo{volume}{18},
  \bibinfo{number}{10} (\bibinfo{year}{2018}).
\newblock


\bibitem[\protect\citeauthoryear{Yan, Wang, Guo, and Han}{Yan
  et~al\mbox{.}}{2020}]%
        {yan2020micronet}
\bibfield{author}{\bibinfo{person}{Zhongxia Yan}, \bibinfo{person}{Hanrui
  Wang}, \bibinfo{person}{Demi Guo}, {and} \bibinfo{person}{Song Han}.}
  \bibinfo{year}{2020}\natexlab{}.
\newblock \showarticletitle{{MicroNet for Efficient Language Modeling}}.
\newblock \bibinfo{journal}{\emph{Journal of Machine Learning Research}}
  \bibinfo{volume}{123}, \bibinfo{number}{20} (\bibinfo{year}{2020}),
  \bibinfo{pages}{215 -- 231}.
\newblock


\bibitem[\protect\citeauthoryear{Yang, Yan, Li, Kwon, Lai, Krishna, Chandra,
  Jiang, and Shi}{Yang et~al\mbox{.}}{2020}]%
        {yang2020co}
\bibfield{author}{\bibinfo{person}{Lei Yang}, \bibinfo{person}{Zheyu Yan},
  \bibinfo{person}{Meng Li}, \bibinfo{person}{Hyoukjun Kwon},
  \bibinfo{person}{Liangzhen Lai}, \bibinfo{person}{Tushar Krishna},
  \bibinfo{person}{Vikas Chandra}, \bibinfo{person}{Weiwen Jiang}, {and}
  \bibinfo{person}{Yiyu Shi}.} \bibinfo{year}{2020}\natexlab{}.
\newblock \showarticletitle{{Co-Exploration of Neural Architectures and
  Heterogeneous ASIC Accelerator Designs Targeting Multiple Tasks}}. In
  \bibinfo{booktitle}{\emph{Design Automation Conference}}.
\newblock


\bibitem[\protect\citeauthoryear{Yang, Howard, Chen, Zhang, Go, Sandler, Sze,
  and Adam}{Yang et~al\mbox{.}}{2018}]%
        {yang2018netadapt}
\bibfield{author}{\bibinfo{person}{Tien-Ju Yang}, \bibinfo{person}{Andrew
  Howard}, \bibinfo{person}{Bo Chen}, \bibinfo{person}{Xiao Zhang},
  \bibinfo{person}{Alec Go}, \bibinfo{person}{Mark Sandler},
  \bibinfo{person}{Vivienne Sze}, {and} \bibinfo{person}{Hartwig Adam}.}
  \bibinfo{year}{2018}\natexlab{}.
\newblock \showarticletitle{{NetAdapt: Platform-Aware Neural Network Adaptation
  for Mobile Applications}}. In \bibinfo{booktitle}{\emph{European Conference
  on Computer Vision}}.
\newblock


\bibitem[\protect\citeauthoryear{Yang, Sun, Liu, Shen, and Jia}{Yang
  et~al\mbox{.}}{2019}]%
        {yang2019std}
\bibfield{author}{\bibinfo{person}{Zetong Yang}, \bibinfo{person}{Yanan Sun},
  \bibinfo{person}{Shu Liu}, \bibinfo{person}{Xiaoyong Shen}, {and}
  \bibinfo{person}{Jiaya Jia}.} \bibinfo{year}{2019}\natexlab{}.
\newblock \showarticletitle{{STD: Sparse-to-Dense 3D Object Detector for Point
  Cloud}}. In \bibinfo{booktitle}{\emph{International Conference on Computer
  Vision}}.
\newblock


\bibitem[\protect\citeauthoryear{Yeung, Downing, Fei-Fei, and Milstein}{Yeung
  et~al\mbox{.}}{2018}]%
        {yeung2018bedside}
\bibfield{author}{\bibinfo{person}{Serena Yeung}, \bibinfo{person}{N~Lance
  Downing}, \bibinfo{person}{Li Fei-Fei}, {and} \bibinfo{person}{Arnold
  Milstein}.} \bibinfo{year}{2018}\natexlab{}.
\newblock \showarticletitle{{Bedside Computer Vision — Moving Artificial
  Intelligence from Driver Assistance to Patient Safety}}.
\newblock \bibinfo{journal}{\emph{New England Journal of Medicine}}
  \bibinfo{number}{14} (\bibinfo{year}{2018}), \bibinfo{pages}{1271--1273}.
\newblock
Issue 378.


\bibitem[\protect\citeauthoryear{Yu and Huang}{Yu and Huang}{2019a}]%
        {yu2019autoslim}
\bibfield{author}{\bibinfo{person}{Jiahui Yu} {and} \bibinfo{person}{Thomas
  Huang}.} \bibinfo{year}{2019}\natexlab{a}.
\newblock \showarticletitle{{AutoSlim: Towards One-Shot Architecture Search for
  Channel Numbers}}.
\newblock \bibinfo{journal}{\emph{arXiv preprint arXiv:1903.11728}}
  (\bibinfo{year}{2019}).
\newblock


\bibitem[\protect\citeauthoryear{Yu and Huang}{Yu and Huang}{2019b}]%
        {yu2019universally}
\bibfield{author}{\bibinfo{person}{Jiahui Yu} {and} \bibinfo{person}{Thomas~S
  Huang}.} \bibinfo{year}{2019}\natexlab{b}.
\newblock \showarticletitle{{Universally Slimmable Networks and Improved
  Training Techniques}}. In \bibinfo{booktitle}{\emph{International Conference
  on Computer Vision}}.
\newblock


\bibitem[\protect\citeauthoryear{Yu, Lukefahr, Palframan, Dasika, Das, and
  Mahlke}{Yu et~al\mbox{.}}{2017}]%
        {yu2017scalpel}
\bibfield{author}{\bibinfo{person}{Jiecao Yu}, \bibinfo{person}{Andrew
  Lukefahr}, \bibinfo{person}{David Palframan}, \bibinfo{person}{Ganesh
  Dasika}, \bibinfo{person}{Reetuparna Das}, {and} \bibinfo{person}{Scott
  Mahlke}.} \bibinfo{year}{2017}\natexlab{}.
\newblock \showarticletitle{{Scalpel: Customizing DNN Pruning to the Underlying
  Hardware Parallelism}}. In \bibinfo{booktitle}{\emph{International Symposium
  on Computer Architecture}}.
\newblock


\bibitem[\protect\citeauthoryear{Yu, Yang, Xu, Yang, and Huang}{Yu
  et~al\mbox{.}}{2019}]%
        {yu2018slimmable}
\bibfield{author}{\bibinfo{person}{Jiahui Yu}, \bibinfo{person}{Linjie Yang},
  \bibinfo{person}{Ning Xu}, \bibinfo{person}{Jianchao Yang}, {and}
  \bibinfo{person}{Thomas Huang}.} \bibinfo{year}{2019}\natexlab{}.
\newblock \showarticletitle{{Slimmable Neural Networks}}. In
  \bibinfo{booktitle}{\emph{International Conference on Learning
  Representations}}.
\newblock


\bibitem[\protect\citeauthoryear{Zach, Pock, and Bischof}{Zach
  et~al\mbox{.}}{2007}]%
        {zach2007duality}
\bibfield{author}{\bibinfo{person}{Christopher Zach}, \bibinfo{person}{Thomas
  Pock}, {and} \bibinfo{person}{Horst Bischof}.}
  \bibinfo{year}{2007}\natexlab{}.
\newblock \showarticletitle{{A Duality Based Approach for Realtime TV-L1
  Optical Flow}}. In \bibinfo{booktitle}{\emph{Joint Pattern Recognition
  Symposium}}.
\newblock


\bibitem[\protect\citeauthoryear{Zadeh, Edo, Awad, and Moshovos}{Zadeh
  et~al\mbox{.}}{2020}]%
        {zadeh2020gobo}
\bibfield{author}{\bibinfo{person}{Ali~Hadi Zadeh}, \bibinfo{person}{Isak Edo},
  \bibinfo{person}{Omar~Mohamed Awad}, {and} \bibinfo{person}{Andreas
  Moshovos}.} \bibinfo{year}{2020}\natexlab{}.
\newblock \showarticletitle{{GOBO: Quantizing Attention-Based NLP Models for
  Low Latency and Energy Efficient Inference}}. In
  \bibinfo{booktitle}{\emph{IEEE/ACM International Symposium on
  Microarchitecture}}.
\newblock


\bibitem[\protect\citeauthoryear{Zagoruyko and Komodakis}{Zagoruyko and
  Komodakis}{2017}]%
        {zagoruyko2017paying}
\bibfield{author}{\bibinfo{person}{Sergey Zagoruyko} {and}
  \bibinfo{person}{Nikos Komodakis}.} \bibinfo{year}{2017}\natexlab{}.
\newblock \showarticletitle{{Paying More Attention to Attention: Improving the
  Performance of Convolutional Neural Networks via Attention Transfer}}. In
  \bibinfo{booktitle}{\emph{International Conference on Learning
  Representations}}.
\newblock


\bibitem[\protect\citeauthoryear{Zhang, Xiong, and Su}{Zhang
  et~al\mbox{.}}{2018a}]%
        {zhang2018accelerating}
\bibfield{author}{\bibinfo{person}{Biao Zhang}, \bibinfo{person}{Deyi Xiong},
  {and} \bibinfo{person}{Jinsong Su}.} \bibinfo{year}{2018}\natexlab{a}.
\newblock \showarticletitle{{Accelerating Neural Transformer via an Average
  Attention Network}}. In \bibinfo{booktitle}{\emph{Conference of the
  Association for Computational Linguistics}}.
\newblock


\bibitem[\protect\citeauthoryear{Zhang, Li, Sun, Guan, Xiao, and Cong}{Zhang
  et~al\mbox{.}}{2015a}]%
        {zhang2015optimizing}
\bibfield{author}{\bibinfo{person}{Chen Zhang}, \bibinfo{person}{Peng Li},
  \bibinfo{person}{Guangyu Sun}, \bibinfo{person}{Yijin Guan},
  \bibinfo{person}{Bingjun Xiao}, {and} \bibinfo{person}{Jason Cong}.}
  \bibinfo{year}{2015}\natexlab{a}.
\newblock \showarticletitle{{Optimizing FPGA-Based Accelerator Design for Deep
  Convolutional Neural Networks}}. In \bibinfo{booktitle}{\emph{International
  Symposium on Field-Programmable Gate Arrays}}.
\newblock


\bibitem[\protect\citeauthoryear{Zhang, Du, Zhang, Lan, Liu, Li, Guo, Chen, and
  Chen}{Zhang et~al\mbox{.}}{2016a}]%
        {zhang2016cambricon}
\bibfield{author}{\bibinfo{person}{Shijin Zhang}, \bibinfo{person}{Zidong Du},
  \bibinfo{person}{Lei Zhang}, \bibinfo{person}{Huiying Lan},
  \bibinfo{person}{Shaoli Liu}, \bibinfo{person}{Ling Li}, \bibinfo{person}{Qi
  Guo}, \bibinfo{person}{Tianshi Chen}, {and} \bibinfo{person}{Yunji Chen}.}
  \bibinfo{year}{2016}\natexlab{a}.
\newblock \showarticletitle{{Cambricon-X: An Accelerator for Sparse Neural
  Networks}}. In \bibinfo{booktitle}{\emph{IEEE/ACM International Symposium on
  Microarchitecture}}.
\newblock


\bibitem[\protect\citeauthoryear{Zhang, Hou, Yin, Shang, Chen, Jiang, and
  Liu}{Zhang et~al\mbox{.}}{2020a}]%
        {zhang2020ternarybert}
\bibfield{author}{\bibinfo{person}{Wei Zhang}, \bibinfo{person}{Lu Hou},
  \bibinfo{person}{Yichun Yin}, \bibinfo{person}{Lifeng Shang},
  \bibinfo{person}{Xiao Chen}, \bibinfo{person}{Xin Jiang}, {and}
  \bibinfo{person}{Qun Liu}.} \bibinfo{year}{2020}\natexlab{a}.
\newblock \showarticletitle{{TernaryBERT: Distillation-Aware Ultra-Low Bit
  BERT}}. In \bibinfo{booktitle}{\emph{Conference on Empirical Methods in
  Natural Language Processing}}.
\newblock


\bibitem[\protect\citeauthoryear{Zhang, Zhao, and LeCun}{Zhang
  et~al\mbox{.}}{2015b}]%
        {NIPS2015_5782}
\bibfield{author}{\bibinfo{person}{Xiang Zhang}, \bibinfo{person}{Junbo Zhao},
  {and} \bibinfo{person}{Yann LeCun}.} \bibinfo{year}{2015}\natexlab{b}.
\newblock \showarticletitle{{Character-Level Convolutional Networks for Text
  Classification}}. In \bibinfo{booktitle}{\emph{Conference on Neural
  Information Processing Systems}}.
\newblock


\bibitem[\protect\citeauthoryear{Zhang, Zhou, Lin, and Sun}{Zhang
  et~al\mbox{.}}{2018b}]%
        {zhang2018shufflenet}
\bibfield{author}{\bibinfo{person}{Xiangyu Zhang}, \bibinfo{person}{Xinyu
  Zhou}, \bibinfo{person}{Mengxiao Lin}, {and} \bibinfo{person}{Jian Sun}.}
  \bibinfo{year}{2018}\natexlab{b}.
\newblock \showarticletitle{{ShuffleNet: An Extremely Efficient Convolutional
  Neural Network for Mobile Devices}}. In \bibinfo{booktitle}{\emph{IEEE
  Conference on Computer Vision and Pattern Recognition}}.
\newblock


\bibitem[\protect\citeauthoryear{Zhang, Zou, He, and Sun}{Zhang
  et~al\mbox{.}}{2016b}]%
        {zhang2016accelerating}
\bibfield{author}{\bibinfo{person}{Xiangyu Zhang}, \bibinfo{person}{Jianhua
  Zou}, \bibinfo{person}{Kaiming He}, {and} \bibinfo{person}{Jian Sun}.}
  \bibinfo{year}{2016}\natexlab{b}.
\newblock \showarticletitle{{Accelerating Very Deep Convolutional Networks for
  Classification and Detection}}.
\newblock \bibinfo{journal}{\emph{IEEE Transactions on Pattern Analysis and
  Machine Intelligence}} \bibinfo{volume}{38}, \bibinfo{number}{10}
  (\bibinfo{year}{2016}), \bibinfo{pages}{1943--1955}.
\newblock


\bibitem[\protect\citeauthoryear{Zhang, Wang, Han, and Dally}{Zhang
  et~al\mbox{.}}{2020b}]%
        {sparch}
\bibfield{author}{\bibinfo{person}{Zhekai Zhang}, \bibinfo{person}{Hanrui
  Wang}, \bibinfo{person}{Song Han}, {and} \bibinfo{person}{William~J Dally}.}
  \bibinfo{year}{2020}\natexlab{b}.
\newblock \showarticletitle{{SpArch: Efficient Architecture for Sparse Matrix
  Multiplication}}. In \bibinfo{booktitle}{\emph{IEEE International Symposium
  on High-Performance Computer Architecture}}.
\newblock


\bibitem[\protect\citeauthoryear{Zhao, Wang, Chen, Zheng, Liu, and Tang}{Zhao
  et~al\mbox{.}}{2020}]%
        {zhao2020autoemb}
\bibfield{author}{\bibinfo{person}{Xiangyu Zhao}, \bibinfo{person}{Chong Wang},
  \bibinfo{person}{Ming Chen}, \bibinfo{person}{Xudong Zheng},
  \bibinfo{person}{Xiaobing Liu}, {and} \bibinfo{person}{Jiliang Tang}.}
  \bibinfo{year}{2020}\natexlab{}.
\newblock \showarticletitle{{AutoEmb: Automated Embedding Dimensionality Search
  in Streaming Recommendations}}.
\newblock \bibinfo{journal}{\emph{arXiv preprint arXiv:2002.11252}}
  (\bibinfo{year}{2020}).
\newblock


\bibitem[\protect\citeauthoryear{Zhong, Yan, Wu, Shao, and Liu}{Zhong
  et~al\mbox{.}}{2018}]%
        {zhong2018practical}
\bibfield{author}{\bibinfo{person}{Zhao Zhong}, \bibinfo{person}{Junjie Yan},
  \bibinfo{person}{Wei Wu}, \bibinfo{person}{Jing Shao}, {and}
  \bibinfo{person}{Cheng-Lin Liu}.} \bibinfo{year}{2018}\natexlab{}.
\newblock \showarticletitle{{Practical Block-Wise Neural Network Architecture
  Generation}}. In \bibinfo{booktitle}{\emph{IEEE Conference on Computer Vision
  and Pattern Recognition}}.
\newblock


\bibitem[\protect\citeauthoryear{Zhou, Wu, Ni, Zhou, Wen, and Zou}{Zhou
  et~al\mbox{.}}{2018}]%
        {zhou2018dorefa}
\bibfield{author}{\bibinfo{person}{Shuchang Zhou}, \bibinfo{person}{Yuxin Wu},
  \bibinfo{person}{Zekun Ni}, \bibinfo{person}{Xinyu Zhou}, \bibinfo{person}{He
  Wen}, {and} \bibinfo{person}{Yuheng Zou}.} \bibinfo{year}{2018}\natexlab{}.
\newblock \showarticletitle{{DoReFa-Net: Training Low Bitwidth Convolutional
  Neural Networks with Low Bitwidth Gradients}}. In
  \bibinfo{booktitle}{\emph{IEEE Conference on Computer Vision and Pattern
  Recognition}}.
\newblock


\bibitem[\protect\citeauthoryear{Zhou and Tuzel}{Zhou and Tuzel}{2018}]%
        {zhou2018voxelnet}
\bibfield{author}{\bibinfo{person}{Yin Zhou} {and} \bibinfo{person}{Oncel
  Tuzel}.} \bibinfo{year}{2018}\natexlab{}.
\newblock \showarticletitle{{VoxelNet: End-to-End Learning for Point Cloud
  Based 3D Object Detection}}. In \bibinfo{booktitle}{\emph{IEEE Conference on
  Computer Vision and Pattern Recognition}}.
\newblock


\bibitem[\protect\citeauthoryear{Zhu, Han, Mao, and Dally}{Zhu
  et~al\mbox{.}}{2017a}]%
        {zhu2017trained}
\bibfield{author}{\bibinfo{person}{Chenzhuo Zhu}, \bibinfo{person}{Song Han},
  \bibinfo{person}{Huizi Mao}, {and} \bibinfo{person}{William Dally}.}
  \bibinfo{year}{2017}\natexlab{a}.
\newblock \showarticletitle{{Trained Ternary Quantization}}. In
  \bibinfo{booktitle}{\emph{International Conference on Learning
  Representations}}.
\newblock


\bibitem[\protect\citeauthoryear{Zhu, Lin, Lu, Lin, and Han}{Zhu
  et~al\mbox{.}}{2021}]%
        {zhu2021dga}
\bibfield{author}{\bibinfo{person}{Ligeng Zhu}, \bibinfo{person}{Hongzhou Lin},
  \bibinfo{person}{Yao Lu}, \bibinfo{person}{Yujun Lin}, {and}
  \bibinfo{person}{Song Han}.} \bibinfo{year}{2021}\natexlab{}.
\newblock \showarticletitle{Delayed Gradient Averaging: Tolerate the
  Communication Latency in Federated Learning}. In
  \bibinfo{booktitle}{\emph{Conference on Neural Information Processing
  Systems}}.
\newblock


\bibitem[\protect\citeauthoryear{Zhu, Liu, and Song}{Zhu
  et~al\mbox{.}}{2019a}]%
        {zhu2019deep}
\bibfield{author}{\bibinfo{person}{Ligeng Zhu}, \bibinfo{person}{Zhijian Liu},
  {and} \bibinfo{person}{Han Song}.} \bibinfo{year}{2019}\natexlab{a}.
\newblock \showarticletitle{{Deep Leakage for Gradient}}. In
  \bibinfo{booktitle}{\emph{Conference on Neural Information Processing
  Systems}}.
\newblock


\bibitem[\protect\citeauthoryear{Zhu, Lu, Lin, and Han}{Zhu
  et~al\mbox{.}}{2019c}]%
        {zhu2019distributed}
\bibfield{author}{\bibinfo{person}{Ligeng Zhu}, \bibinfo{person}{Yao Lu},
  \bibinfo{person}{Yujun Lin}, {and} \bibinfo{person}{Song Han}.}
  \bibinfo{year}{2019}\natexlab{c}.
\newblock \showarticletitle{{Distributed Training Across the World}}. In
  \bibinfo{booktitle}{\emph{NeurIPS Workshop on Systems for ML}}.
\newblock


\bibitem[\protect\citeauthoryear{Zhu, Yang, Mendieta, and Chen}{Zhu
  et~al\mbox{.}}{2020}]%
        {zhu2020a3d}
\bibfield{author}{\bibinfo{person}{Sijie Zhu}, \bibinfo{person}{Taojiannan
  Yang}, \bibinfo{person}{Matias Mendieta}, {and} \bibinfo{person}{Chen Chen}.}
  \bibinfo{year}{2020}\natexlab{}.
\newblock \showarticletitle{{A3D: Adaptive 3D Networks for Video Action
  Recognition}}.
\newblock \bibinfo{journal}{\emph{arXiv preprint arXiv:2011.12384}}
  (\bibinfo{year}{2020}).
\newblock


\bibitem[\protect\citeauthoryear{Zhu, Wang, Dai, Yuan, and Wei}{Zhu
  et~al\mbox{.}}{2017b}]%
        {zhu2017flow}
\bibfield{author}{\bibinfo{person}{Xizhou Zhu}, \bibinfo{person}{Yujie Wang},
  \bibinfo{person}{Jifeng Dai}, \bibinfo{person}{Lu Yuan}, {and}
  \bibinfo{person}{Yichen Wei}.} \bibinfo{year}{2017}\natexlab{b}.
\newblock \showarticletitle{{Flow-Guided Feature Aggregation for Video Object
  Detection}}. In \bibinfo{booktitle}{\emph{International Conference on
  Computer Vision}}.
\newblock


\bibitem[\protect\citeauthoryear{Zhu, Liu, Yang, Yuille, and Xu}{Zhu
  et~al\mbox{.}}{2019b}]%
        {zhu2019vnas}
\bibfield{author}{\bibinfo{person}{Zhuotun Zhu}, \bibinfo{person}{Chenxi Liu},
  \bibinfo{person}{Dong Yang}, \bibinfo{person}{Alan Yuille}, {and}
  \bibinfo{person}{Daguang Xu}.} \bibinfo{year}{2019}\natexlab{b}.
\newblock \showarticletitle{{V-NAS: Neural Architecture Search for Volumetric
  Medical Image Segmentation}}. In \bibinfo{booktitle}{\emph{International
  Conference on 3D Vision}}.
\newblock


\bibitem[\protect\citeauthoryear{Zhuo and Prasanna}{Zhuo and Prasanna}{2005}]%
        {zhuo2005sparse}
\bibfield{author}{\bibinfo{person}{Ling Zhuo} {and} \bibinfo{person}{Viktor~K
  Prasanna}.} \bibinfo{year}{2005}\natexlab{}.
\newblock \showarticletitle{{Sparse Matrix-Vector Multiplication on FPGAs}}. In
  \bibinfo{booktitle}{\emph{International Symposium on Field-Programmable Gate
  Arrays}}.
\newblock


\bibitem[\protect\citeauthoryear{Zoph and Le}{Zoph and Le}{2017}]%
        {zoph2017neural}
\bibfield{author}{\bibinfo{person}{Barret Zoph} {and} \bibinfo{person}{Quoc~V
  Le}.} \bibinfo{year}{2017}\natexlab{}.
\newblock \showarticletitle{{Neural Architecture Search with Reinforcement
  Learning}}. In \bibinfo{booktitle}{\emph{International Conference on Learning
  Representations}}.
\newblock


\bibitem[\protect\citeauthoryear{Zoph, Vasudevan, Shlens, and Le}{Zoph
  et~al\mbox{.}}{2018}]%
        {zoph2018learning}
\bibfield{author}{\bibinfo{person}{Barret Zoph}, \bibinfo{person}{Vijay
  Vasudevan}, \bibinfo{person}{Jonathon Shlens}, {and} \bibinfo{person}{Quoc~V
  Le}.} \bibinfo{year}{2018}\natexlab{}.
\newblock \showarticletitle{{Learning Transferable Architectures for Scalable
  Image Recognition}}. In \bibinfo{booktitle}{\emph{IEEE Conference on Computer
  Vision and Pattern Recognition}}.
\newblock


\bibitem[\protect\citeauthoryear{Zou, Dou, Guo, and Ni}{Zou
  et~al\mbox{.}}{2013}]%
        {zou2013high}
\bibfield{author}{\bibinfo{person}{Dan Zou}, \bibinfo{person}{Yong Dou},
  \bibinfo{person}{Song Guo}, {and} \bibinfo{person}{Shice Ni}.}
  \bibinfo{year}{2013}\natexlab{}.
\newblock \showarticletitle{{High Performance Sparse Matrix-Vector
  Multiplication on FPGA}}.
\newblock \bibinfo{journal}{\emph{IEICE Electronics Express}}
  (\bibinfo{year}{2013}).
\newblock


\end{thebibliography}
